\documentclass[12pt,openany,oneside]{book}

\usepackage[utf8]{inputenc}
\usepackage[
	backend=bibtex,
	style=numeric,
]{biblatex}
\usepackage{csquotes}
\usepackage[french]{babel}
\usepackage[T1]{fontenc}
\usepackage[a4paper,left=3cm,right=2.5cm,top=3cm,bottom=3cm, twoside]{geometry}
\usepackage{xcolor}
\usepackage{stmaryrd}
\usepackage{amssymb}
\usepackage{amsmath}
\usepackage{libertine}
\usepackage[pdftex]{graphicx}
\usepackage{caption}
\usepackage{subcaption}
\usepackage{float}
\usepackage{multirow}
\usepackage{apalike}
\usepackage{minitoc}
\usepackage{tikz}
\usepackage{enumitem}
\usepackage{epigraph}
\usepackage{lscape}
\usepackage{titletoc}
\usepackage{booktabs}
\usepackage{arydshln}
\usepackage[nonumberlist]{glossaries}
\usepackage{calc}
\usepackage[]{titlesec}
\definecolor{linkColor}{HTML}{32a852} 
\definecolor{headerColor}{HTML}{EEEEEE}
\usepackage{stackengine}
\usepackage[colorlinks=true,citecolor=linkColor,linkcolor=black]{hyperref}
\usepackage{fancyhdr}
\usepackage{lipsum}
\usepackage{tabularx}
\usepackage{booktabs,makecell}
\usepackage{afterpage}
\usepackage{color, colortbl}
\usepackage{dblfloatfix}
\usepackage{rotating}
\usepackage{pdfpages}
\usepackage{tabularray}

\usepackage{arabtex} 
\usepackage{utf8}
\usepackage{pifont}  
\usepackage{setspace}
\usepackage{fancyhdr}

\definecolor{yourcolor}{HTML}{8a0e19}

\titleformat{\chapter}[display]
{\normalfont\color{yourcolor}}
{\filleft\huge\color{black}\textsc\chaptertitlename\hspace*{2mm}%
	\begin{tikzpicture}[baseline={([yshift=-.6ex]current bounding box.center)}]
	\node[fill=yourcolor,circle,text=white] {\thechapter};
	\end{tikzpicture}
}
{1ex}
{\titlerule[1.5pt]\vspace*{1.5ex}\Huge\color{black}\textsc}
[]

\titleformat{name=\chapter,numberless}[display]
{\normalfont\color{black}}
{}
{1ex}
{\vspace*{-5cm}\Huge\textsc}
[]

\newcommand{\printmyminitoc}[1]{%
	\noindent\hspace{1cm}%
	\colorlet{chpnumbercolor}{black}%
	\begin{tikzpicture}
	\node(s){
		\begin{minipage}{.9\linewidth}
		\printcontents[chapters]{}{1}{}
		\end{minipage}
	};
	{
		\color{yourcolor}
		\draw(s.north west)--(s.north east) (s.south west)--(s.south east);
	}
	\end{tikzpicture}
	\vspace*{3ex}
	
	#1
	\vfill
	\pagebreak
}

\newcommand{\HRule}{\rule{\linewidth}{0.7mm}}
\newcommand{\Hrule}{\rule{\linewidth}{0.3mm}}

\begin{document}

 \let\cleardoublepage=\clearpage
\includepdf[pages={1}]{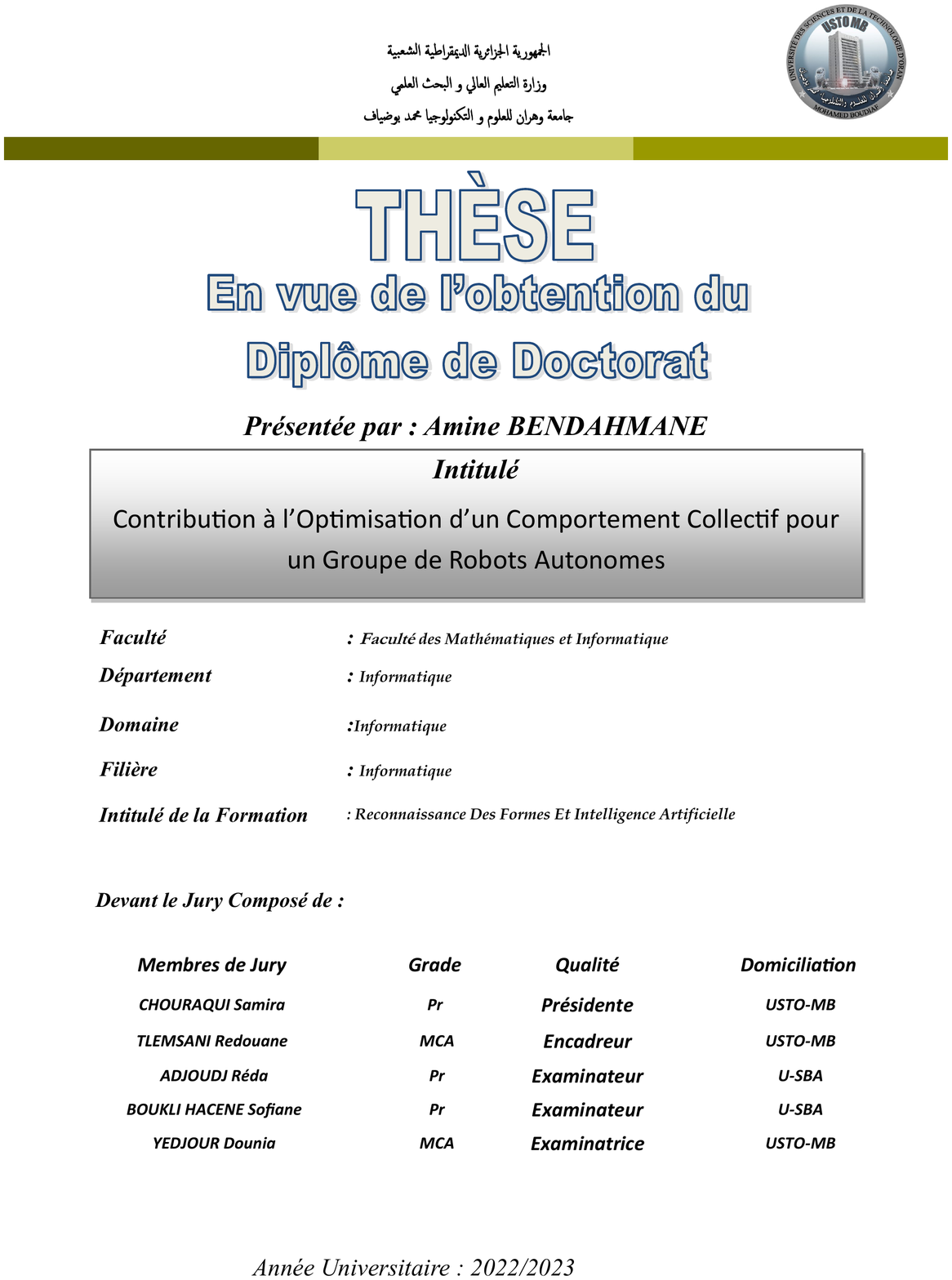}

\begin{titlepage}
 
 \newgeometry{top=0.3in}

    \begin{center}
	\begin{tabular}{c@{\hskip 9cm}c}
		\includegraphics[height=2.5cm]{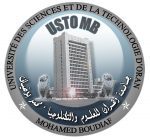}
             &
		\includegraphics[height=2cm]{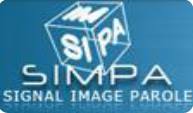}
	\end{tabular}
    \end{center}
    
    \begin{center}
    République Algérienne Démocratique et Populaire\\
    Ministère de l'Enseignement Supérieur et de la Recherche Scientifique\\
    Université des Sciences et de la Technologie d'Oran Mohamed Boudiaf\\
    Faculté des Mathématiques et Informatique \\
    Laboratoire Signal-Image-Parole (SIMPA)

    \vfill
    
    \HRule \\[0.1cm]
    { \Large \bfseries Contribution à l’Optimisation d’un Comportement Collectif pour un Groupe de Robots Autonomes }
    \Hrule \\
    \end{center}
		
    \vfill

    \begin{center}
    
    Thèse en vue de l'obtention de diplôme de\\
    doctorat en \textsc{\large Informatique}\\[1cm]
    Présentée par \textsc{\Large Amine BENDAHMANE}\\[1cm]

    \begin{center}
    Devant le jury composé de \\
    \begin{tabular}{llll}
	\textsc{CHOURAQUI Samira}  &  \textsc{Pr.} & Présidente & USTOMB \\
	\textsc{TLEMSANI Redouane}  &  \textsc{MCA} & Directeur de thèse & USTOMB  \\
	\textsc{ADJOUDJ Réda}  &  \textsc{Pr.} & Examinateur & U-SBA \\
	\textsc{BOUKLI HACENE Sofiane}  &  \textsc{Pr.} & Examinateur & U-SBA \\
	\textsc{YEDJOUR Dounia}  &  \textsc{MCA} & Examinateur & USTOMB 
				
    \end{tabular}\\[1cm]
    \end{center}

    \vspace{1cm}
    2023
    \end{center}
		
    \vspace{1cm}
    
    
    
\end{titlepage}

\restoregeometry


    \pagestyle{fancy}
    \fancyhead{}
    \renewcommand{\chaptermark}[1]{\markboth{\textsc{#1}}{}}
    \frontmatter
    \pagenumbering{arabic}

    
\setcounter{page}{2}%

\vspace*{\fill}
\null\hfill A ma famille, ma femme \\
\null\hfill et mes amis.

\vspace*{\fill}
    \chapter{Remerciements}
\setstretch{1.5}

Je tiens à remercier mes deux directeurs de thèse, le Pr. BENYETTOU Abdelkader et le Dr. TLEMSANI Redouane, pour leurs conseils inestimables et pour avoir accepté d'encadrer ce travail dont j'espère qu'il sera à la hauteur de leurs espérances.

Je remercie également Mme la présidente du jury, Pr. CHOURAQUI Samira, pour sa disponibilité, ainsi que l'ensemble des membres du jury, à savoir:
le Pr. ADJOUDJ Réda, Pr. BOUKLI HACENE Sofiane et Dr. YEDJOUR Dounia, pour avoir accepté de donner de leurs temps pour examiner ce travail.

Je remercie le Pr. BERACHED Nasreddine pour m'avoir toujours ouvert son laboratoire et m'avoir permis de faire des expériences sur des robots réels sans poser de conditions.

Je remercie aussi l'ensemble de mes enseignants qui m'ont appris la méthodologie scientifique et les connaissances techniques requises pour aboutir à ce travail, dont j'espère qu'il sera à la hauteur de leurs espérances. Sans oublier le personnel administratif de la faculté ainsi que le service de poste-graduation.

Pour finir, je remercie tous ceux qui ont contribué de près ou de loin pour la réalisation de ce travail. Notamment ma famille qui m'a accompagné tout au long de la réalisation de travail, ainsi que mes amis et mes collègues.

\setstretch{1}
    \chapter{Résumé}


\setstretch{1.5}
Cette thèse étudie le domaine de la robotique collective, et plus particulièrement les problèmes d'optimisation des systèmes multirobots dans le cadre de l'exploration, de la planification de trajectoires et de la coordination. 
Elle inclut deux contributions. La première est l'utilisation de l'algorithme d'optimisation des papillon (BOA : Butterfly Optimization Algorithm) pour résoudre le problème d'exploration de zone inconnue avec des contraintes d'énergie dans des environnements dynamiques. A notre connaissance, cet algorithme n'a jamais été utilisé pour résoudre des problèmes de robotique auparavant. Nous avons également proposé une nouvelle version de cet algorithme appelée xBOA basée sur l'opérateur de croisement pour améliorer la diversité des solutions candidates et accélérer la convergence de l'algorithme.
La deuxième contribution présenté dans cette thèse est le développement d'une nouvelle plateforme de simulation pour l'analyse comparative de problèmes incrémentaux en robotique tels que les tâches d'exploration. La plateforme est conçue de manière à être générique pour comparer rapidement différentes métaheuristiques avec un minimum de modifications, et pour s'adapter facilement aux scénarios mono et multirobots. De plus, elle offre aux chercheurs des outils pour automatiser leurs expériences et générer des visuels, ce qui leur permettra de se concentrer sur des tâches plus importantes telles que la modélisation de nouveaux algorithmes.
Nous avons mené une série d'expériences qui ont montré des résultats prometteurs et nous ont permis de valider notre approche et notre modélisation.

\subsubsection{Mots clés:} Systèmes multirobots, comportements intelligents, métaheuristiques, exploration robotique, benchmarking, Butterfly Optimization Algorithm, xBOA

\pagebreak

\text{\LARGE ABSTRACT} 

\vspace{0.5cm}

\setstretch{1.7}
This thesis studies the domain of collective robotics, and more particularly the optimization problems of multirobot systems in the context of exploration, path planning and coordination.
It includes two contributions. The first one is the use of the Butterfly Optimization Algorithm (BOA) to solve the Unknown Area Exploration problem with energy constraints in dynamic environments. This algorithm was never used for solving robotics problems before, as far as we know. We proposed a new version of this algorithm called xBOA based on the crossover operator to improve the diversity of the candidate solutions and speed up the convergence of the algorithm.
The second contribution is the development of a new simulation framework for benchmarking dynamic incremental problems in robotics such as exploration tasks. The framework is made in such a manner to be generic to quickly compare different metaheuristics with minimum modifications, and to adapt easily to single and multi-robot scenarios. Also, it provides researchers with tools to automate their experiments and generate visuals, which will allow them to focus on more important tasks such as modeling new algorithms.
We conducted a series of experiments that showed promising results and allowed us to validate our approach and model.

\subsubsection{Keywords:}
Multirobot systems, intelligent behaviors, metaheuristics, robotics exploration, benchmarking, Butterfly Optimization Algorithm, xBOA

\pagebreak

 
\begin{flushright}
\setcode{utf8}

\vspace{0.2cm}
 \RL{
 \textbf{\large ملخص }
 }

\setstretch{1.7}

 \RL{
 تدرس هذه الأطروحة مجال الروبوتات الجماعية، وبالتحديد مسائل تحسين أنظمة الروبوتات المتعددة في سياق الاستكشاف وتخطيط المسارات والتنسيق الأوتوماتيكي.
 }

 \RL{
تشمل هذه الأطروحة على مساهمتين في المجال العلمي، المساهمة الأولى تتمثل في استخدام خوارزمية جديدة \LR{(Butterfly Optimization Algorithm)} لحل مشكلة استكشاف المناطق غير المعروفة مع مراعات شروط الطاقة المحدودة والتغيرات في بيئة الروبوت. على حد علمنا لم يتم استخدام هذه الخوارزمية من قبل لحل مسائل الروبوتات. في نفس السياق، سنقدم إصدار جديد من هذه الخوارزمية يسمى \LR{xBOA} بهدف تحسين أداءها من جانب الدقة وكذا تنوع الحلول التي يتم إنشاؤها. 
}

\RL{
أما المساهمة الثانية، فهي تتمثل في تطوير برنامج هدفه تقييم آداء مختلف الخوارزميات في حل مشاكل الاستكشاف، وتخطيط المسارات، ونقل البضائع، والمراقبة باستعمال مجموعة من الروبوتات. تم تصميم البرنامج بطريقة تجعله عامًا وسهل الاستعمال لمقارنة الخصائص الوصفية بسرعة، والتكيف أوتوماتيكيا مع سيناريوهات الروبوتات الفردية والمتعددة. كما أنه يوفر للباحثين أدوات لأتمتة تجاربهم مما سيسمح لهم بالتركيز على مهام أكثر أهمية مثل نمذجة خوارزميات جديدة ومقارنتها مع الخوارزميات القديمة.
}

\vspace{0.5cm}

\RL{
\textbf{\large الكلمات المفتاحية}
}

\RL{
أنظمة الروبوتات الجماعية، السلوكيات الذكية، الاستكشاف الآلي، التحسين العشوائي ، قياس آداء الخوارزميات، خوارزمية تحسين الفراشات، خوارزمية \LR{xBOA}
}

\end{flushright}

\setstretch{1}

\tableofcontents
\addcontentsline{toc}{chapter}{Table des matières}

\renewcommand{\listfigurename}{Liste des figures}
\listoffigures
\addcontentsline{toc}{chapter}{Liste des figures}

\listoftables
\addcontentsline{toc}{chapter}{Liste des tableaux}



    \let\mainmatterorig\mainmatter
    \renewcommand\mainmatter
     {\edef\temppagenumber{\arabic{page}}%
      \mainmatterorig
      \setcounter{page}{\temppagenumber+1}%
     }
     \renewcommand{\arraystretch}{1.8}
    
    \mainmatter
    \setlength{\parskip}{.7em}
    \titlespacing*{\section}{0pt}{.9em}{.8em}
    \renewcommand{\baselinestretch}{1.1}
    \fancyhead[RO]{\leftmark}
    \fancyhead[LE]{\textsc{\chaptername~\thechapter}}

    
    \chapter*{Introduction générale}
\addcontentsline{toc}{chapter}{Introduction Générale}  
\rhead{Introduction Générale}

\setstretch{1.5}
La robotique collective est un domaine actif de recherche et développement qui a le potentiel de révolutionner de nombreux secteurs. Elle prend une place de plus en plus importante dans l’agriculture, l’industrie, le transport, la sécurité, le sauvetage, l'exploration et le divertissement. Elle se base sur le principe d’utilisation de plusieurs robots qui travaillent ensemble pour atteindre un objectif commun. Cette collectivité implique la collaboration et/ou la coopération des robots entre eux, ce qui la lie intimement avec le domaine de l’intelligence artificielle distribuée.

Ces deux domaines combinés peuvent offrir de nombreux avantages, notamment : la réalisation de tâches plus rapidement et plus efficacement, et de manière plus sécurisée, même dans des environnements dangereux qui pourraient présenter des risques pour les êtres humains. Par ailleurs, l’utilisation d’un groupe de robots au lieu d’un seul permet d’augmenter la polyvalence et la flexibilité du système, ainsi que la robustesse aux pannes.

Notre thèse s’inscrit dans le cadre d’optimisation d’un comportement collectif pour un groupe de robots autonomes, nous nous intéressons aux scénarios de planification de trajectoires et d’exploration de zones en vue de détection d'incidents par exemple, ou de la localisation d'une personne en détresse.

La thèse est organisée en deux parties. La partie théorique – divisée en deux chapitres – vise à introduire tous les fondements théoriques nécessaires à la compréhension de notre travail. La deuxième partie – elle aussi divisée en deux chapitres – inclut tous les détails relatifs à la modélisation de notre solution et l’analyse des résultats expérimentaux.

Nous introduirons le lecteur dans le premier chapitre aux différents types de systèmes multirobots ainsi que les mécanismes de synchronisation existants, tout en détaillant les différentes problématiques de recherche engendrées par ce type de systèmes. Nous aborderons également un état de l’art sur les techniques d’intelligence artificielle utilisées pour les résoudre.

Le deuxième chapitre sera consacré à l’étude des métaheuristiques et leurs mécanismes de fonctionnement. Nous présenterons les fondements théoriques de quelques métaheuristiques populaires que nous utiliserons durant nos expériences. Nous présenterons par la suite notre contribution à l’amélioration du Butterfly Optimization Algorithm, en expliquant tous les détails mathématiques nécessaires à son bon fonctionnement.

Dans le troisième chapitre, nous introduirons le lecteur à la deuxième contribution de notre thèse concrétisée dans le développement d’une plateforme de simulation spécialisée dans le benchmarking et l’évaluation des algorithmes d’optimisation pour les problématiques de navigation, de planification de trajectoires, d'exploration, de balayage et de surveillance. Nous présenterons ensuite la méthodologie utilisée et notre modélisation du problème d’exploration de zones inconnues avec des contraintes d'énergie.

Le dernier chapitre regroupera l’ensemble des résultats expérimentaux obtenus durant notre thèse. Nous présenterons aussi notre interprétation de ces résultats à travers une analyse dédiée pour chaque expérience. Ce chapitre conclura nos travaux en présentant une critique des techniques utilisées ainsi que des pistes potentielles pour les améliorer.

\setstretch{1}
\pagebreak
\rhead{}
    
    \part{Etude théorique}
    \chapter{Les systèmes multirobots}

\startcontents[chapters]
\printmyminitoc{
	
}

\section{Introduction}

L'étude de la robotique collective dans un contexte informatique concerne l'étude des systèmes dits "multirobots".

L’objectif de ce chapitre est de donner une idée d'ensemble sur ce sujet en essayant d'englober chacun de ses aspects. Nous commencerons par présenter les systèmes multirobots, leurs types et leurs domaines d'applications. Nous détaillerons également les axes de recherche relatifs à ce sujet avec les différentes problématiques qui en découlent.

Nous aborderons ensuite un état de l’art des techniques d'intelligence artificielle en lien avec les problématiques traitées dans cette thèse, à savoir les problématiques d'optimisation de trajectoires et d'exploration de zones inconnues en utilisant plusieurs robots. 

\section{Les systèmes multirobots}
    \subsection{Présentation des systèmes multirobots}

\noindent
\begin{figure}[ht]
    \setlength{\abovecaptionskip}{0.4cm} 
    \setlength{\belowcaptionskip}{-0.4cm} 
    \centering 
    \includegraphics[width=0.9\textwidth]{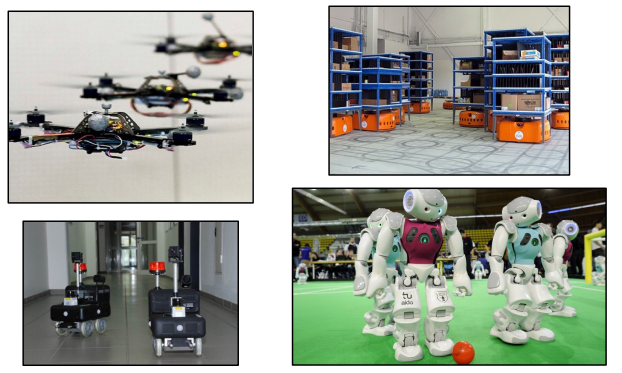}
\caption{Images de systèmes multirobots dans le monde réel}
    \label{fig:c1.1}
\end{figure}
    
Les systèmes multirobots (\textit{Multirobots Systems},en anglais) sont des systèmes composés de plusieurs robots programmés de sorte à travailler ensemble afin d’accomplir une certaine mission. Cette mission est souvent décomposée en un ensemble de tâches afin de réduire sa complexité. Ces tâches peuvent être effectuées une à une séquentiellement, ou en parallèle, selon l’interdépendance entre elles et le type d’architecture du système.
    
Les robots sont des machines avec des propriétés physiques et logiques, ils peuvent donc être analysés d’un point de vue mécanique et électronique pour étudier la manière dont ils détectent et impactent l'environnement qui les entoure. Comme ils peuvent aussi être étudiés d’un point de vue algorithmique pour spécifier la manière dont cet environnement doit être représenté en interne dans le but de prendre des décisions ou d’effectuer des actions. L’intelligence artificielle est donc au cœur de ce processus décisionnel, afin de s’assurer que ces machines s’adaptent de manière efficace à leur environnement.

Dans le domaine de l'informatique, un robot peut être vu comme un agent qui interagit selon certaines règles. L’étude d’un système multirobots peut être vue comme l’étude d’un système multiagents avec certaines limitations physiques telles que leur degré d’autonomie et leur capacité de stockage et de traitement de l’information.

D’un autre côté, ils peuvent être apparentés aux systèmes embarqués en les considérant comme un réseau d’objets connectés qui communiquent entre eux pour échanger des informations. Le terme Réseaux Multirobots (\textit{Multirobots Network} en anglais) est parfois utilisé dans la littérature pour décrire ce concept.

En parallèle, ils peuvent être considérés dans le domaine de la logistique comme des flottes de véhicules autonomes (\textit{Autonomous Vehicules Fleet}). L’objet d’étude se concentrera donc sur les techniques de gestion de la flotte (\textit{fleet management}), suivi de la position et état des robots (\textit{tracking}), ainsi que les techniques de commande à distance (\textit{remote control}). Il est à noter que le terme « \textit{Véhicule autonome} » dans ce domaine doit être compris au sens large, il inclut les robots de type: voitures, tracteurs, drones et sous-marins.

\noindent
\begin{figure}[ht]
    \vspace{-1cm}
    \setlength{\abovecaptionskip}{0.4cm} 
    \setlength{\belowcaptionskip}{-0.4cm} 
    \centering 
        
    \begin{subfigure}[b]{0.3\textwidth}
         \centering
         \includegraphics[width=\textwidth]{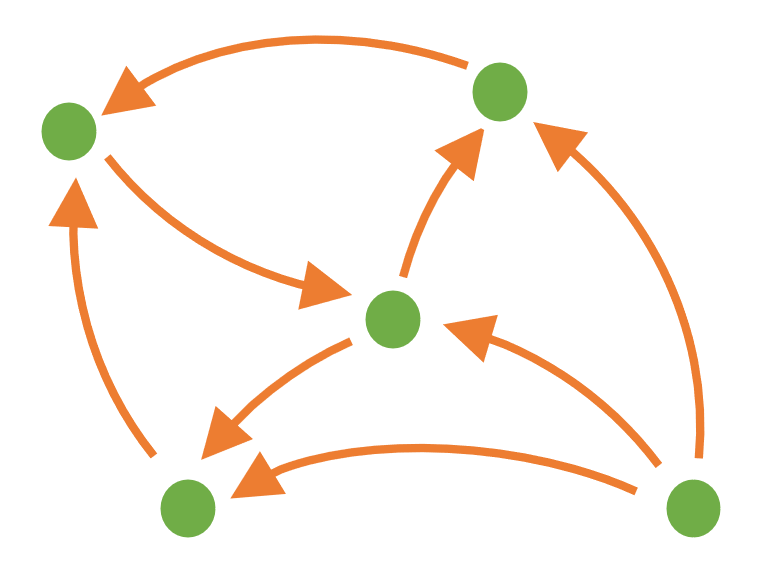}
         \caption{\textit{Intelligence Artificielle : Étude des décisions et interactions entre les agents}}
     \end{subfigure}\quad
      \begin{subfigure}[b]{0.3\textwidth}
         \centering
         \includegraphics[width=\textwidth]{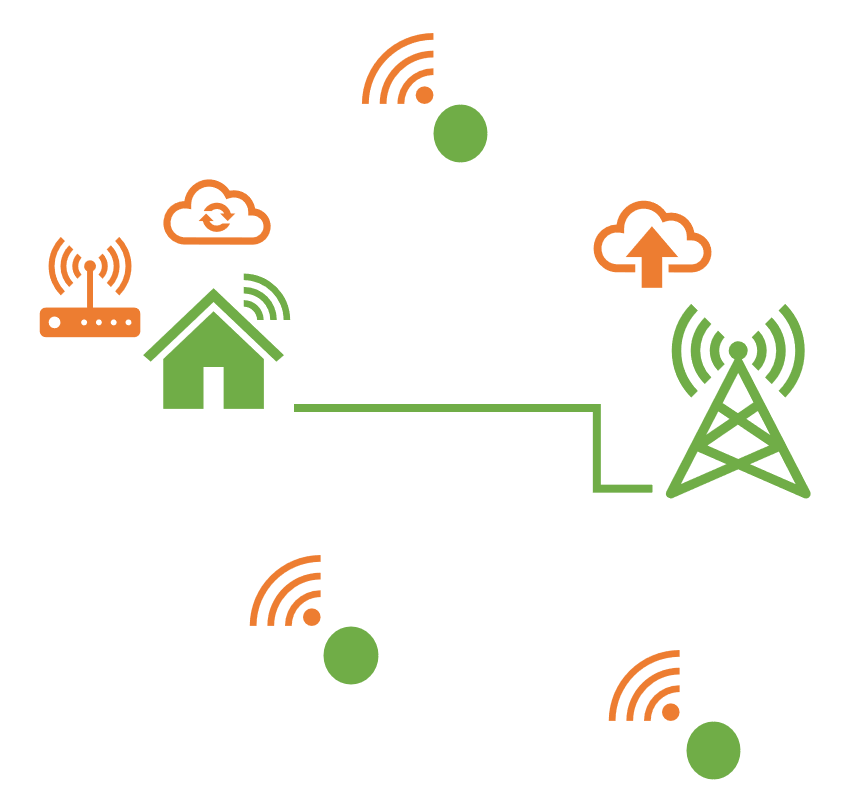}
         \caption{\textit{Réseaux et systèmes embarqués : Connectivité, topologie réseau et protocoles}}
     \end{subfigure}\quad
      \begin{subfigure}[b]{0.3\textwidth}
         \centering
         \includegraphics[width=\textwidth]{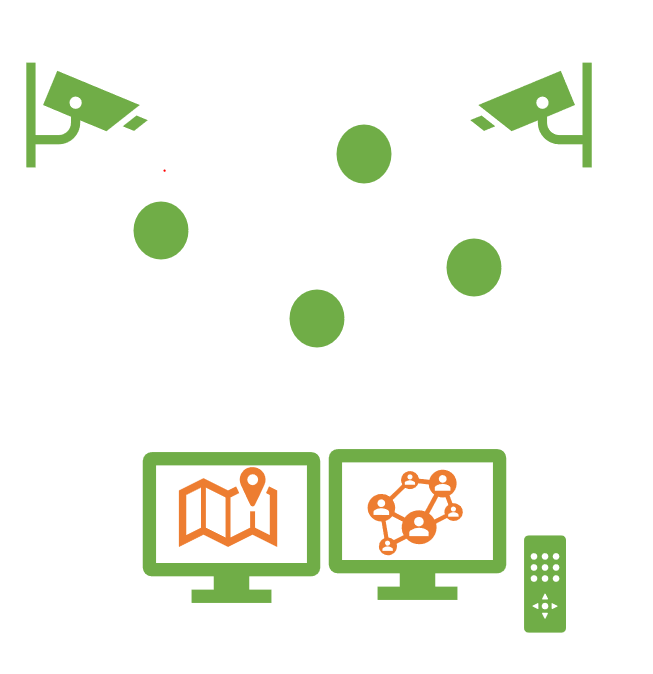}
         \caption{\textit{Logistique autonome :  Suivi, visualisation et commande à distance}}
     \end{subfigure}
     \vspace{0.5\baselineskip}
    \caption{Différents aspects étudiés selon la nature du domaine appliqué aux systèmes multirobots}
    \label{fig:c1.2}
\end{figure}

    \subsection{Les domaines d'applications}
    Les systèmes multirobots sont souvent utilisés dans des domaines industriels, sécuritaires et ludiques. Parmi les applications notables qui ont permis d’exploiter les pleines capacités qu’offrent ces systèmes, nous pouvons citer : 

    \begin{itemize}
  \item La détection de feux de forêt ;
  \item Le nettoyage de zones contaminées ou dangereuses ;
  \item L’accomplissement d’opérations agricoles dans les fermes intelligentes ;
  \item La recherche et localisation de survivants lors des catastrophes naturelles ;
  \item La gestion des entrepôts de marchandises ;
  \item Le déplacement de conteneurs dans les ports ;
  \item L’inspection des conduits souterrains, pipe-lines, canalisations et fonds marins ;
  \item Le divertissement tel que la dance coordonnée de robots, les spectacles de lumière et les compétitions de football robotique ;
  \item La surveillance et détection d’intrus ;
  \item Le déploiement de capteurs lors des expéditions scientifiques, tel que les études climatologiques, environnementales et écologiques ;
  \item Ainsi que d’autres applications militaires tel que l’espionnage et la télésurveillance.

\end{itemize}

\noindent
\begin{figure}[ht]
    \setlength{\abovecaptionskip}{0.4cm} 
    \setlength{\belowcaptionskip}{-0.4cm} 
    \centering 
    \includegraphics[width=0.9\textwidth]{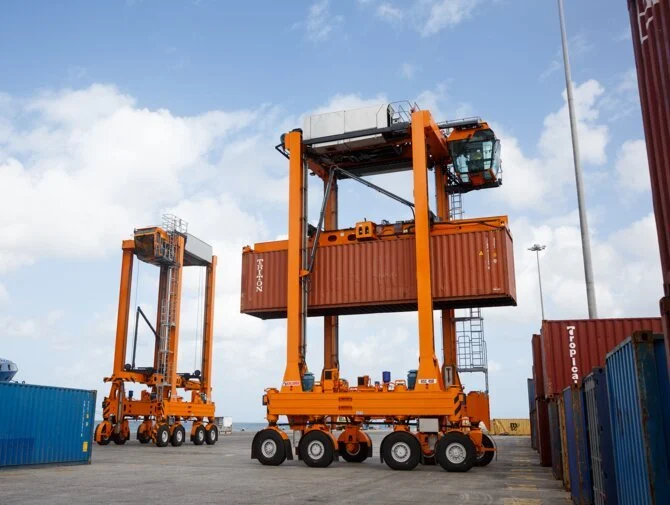}
\caption{Utilisation de robots pour la gestion portuaire (\textit{système Kalmar AutoStrad} \cite{durrant07})}
    \label{fig:c1.3}
\end{figure}

L’utilisation de ces systèmes offre plusieurs avantages tels que l’augmentation de la productivité industrielle et agricole, l’accélération de tâches de sauvetage et lutte contre les incendies, le renforcement de la fiabilité des systèmes sécuritaires, l’élargissement de la surface de couverture dans les systèmes de surveillance, ainsi que la possibilité à effectuer des tâches impossibles à faire en utilisant un seul robot.

De plus, l’utilisation de ces systèmes permet parfois aussi de diminuer les coûts de production par rapport à l’utilisation d’un seul robot, bien que cette idée puisse paraître contre-intuitive. En effet, il est parfois moins coûteux de fabriquer plusieurs petits robots bon marché avec des capacités limitées au lieu d’un seul robot complexe et sophistiqué. D’autant plus que la consommation énergétique d’un robot peut augmenter rapidement avec l’augmentation de sa puissance de calcul et poids maximal à déplacer. Afin de palier à ce type de problèmes, certains chercheurs se sont basés sur la distribution de charge entre plusieurs petits robots qui travaillent ensemble au lieu d’investir sur l’augmentation de la capacité d’un seul robot.

D'un autre côté, les systèmes multirobots permettent d'augmenter la robustesse du système en évitant l’inconvénient du point de défaillance unique (\textit\{Single point of failure\}). Par exemple, dans un scénario de sauvetage ou d’exploration d’une zone dangereuse, il est beaucoup moins grave qu’un ou plusieurs petits robots tombent en panne tant que les autres robots restent opérationnels, comparé à la situation où un seul robot complexe est déployé risquant d’entraîner l’échec automatique de la mission s’il tombe en panne ou se retrouve bloqué dans les débris. 

\noindent
\begin{figure}[ht]
    \setlength{\abovecaptionskip}{0.4cm} 
    \setlength{\belowcaptionskip}{-0.4cm} 
    \centering 
    \includegraphics[width=0.8\textwidth]{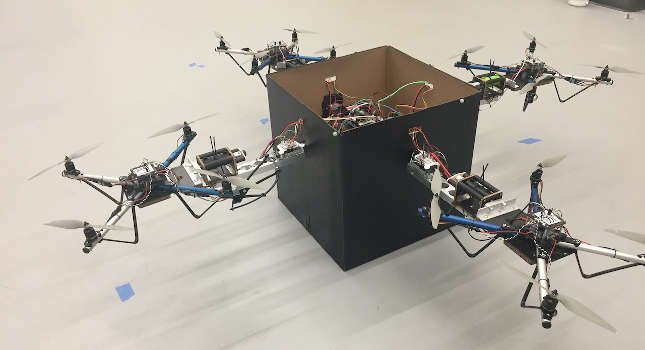}
\caption{Exemple d'un scénario où un groupe de robots s'entraident pour déplacer un objet lourd (\textit{Université Georgia Tech})}
    \label{fig:c1.4}
\end{figure}

Il existe toutefois des inconvénients à utiliser ce genre de systèmes : la nécessité à intégrer des moyens de communication entre robots entraîne l’augmentation de la complexité des logiciels et la nécessité à intégrer des dispositifs électroniques supplémentaires (Wifi, 4G/LTE, GPS…) qui peuvent être lourds en consommation énergétique. L’ajout de ces éléments engendre l’augmentation des coûts de production ainsi que la consommation énergétique des robots, notamment dans les situations où le rayon de communication est large.

Un autre problème est la nécessité d’inclure des mécanismes de coordination et/ou collaboration entre les robots pour éviter le chevauchement entre les tâches effectuées, ou encore l’implémentation de processus de prise de décision collective. Ces fonctionnalités additionnelles entraînent souvent le recourt à l’augmentation de la puissance de calcul des robots ainsi que la difficulté à détecter les bogues informatiques à cause de la nature hautement parallèle d’un tel système.

En outre, l’utilisation des systèmes multirobots dans certains domaines d’application requiert l’installation d’infrastructures supplémentaires (antennes relais, serveurs de récolte de données, balises de localisation, bornes de recharge...etc), ce qui accroît la complexité du système.

\noindent
\begin{figure}[ht]
    \setlength{\abovecaptionskip}{0.4cm} 
    \setlength{\belowcaptionskip}{-0.4cm} 
    \centering 
    \includegraphics[width=0.95\textwidth]{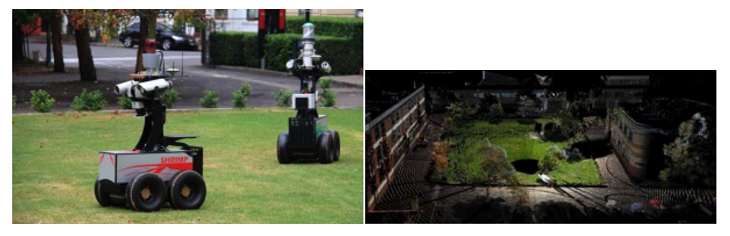}
\caption{Exemple d'un scénario de modélisation 3D d’un parc urbain en utilisant plusieurs plusieurs robots (\textit{Australian Centre for Field Robotics})}
    \label{fig:c1.5}
\end{figure}

\section{Les types de systèmes multirobots}
Les systèmes multirobots peuvent être catégorisés selon plusieurs critères : architecture du système, type d’interactions, type de synchronisation, autonomie ou encore le type des robots utilisés.

Nous allons détailler ci-dessous chaque classification en commençant de la plus abstraite vers la plus spécifique.

    \subsection{Classification par type de comportements}
    Un système multirobot doit intégrer des mécanismes d’interaction entre les robots, ces interactions forment des comportements de travail collectif qui peuvent être classés en plusieurs catégories :

     \subsubsection{Coordination}
La coordination est le fait de synchroniser ses efforts pour réaliser une tâche en équipe. Sachant que cette tâche pouvait être réalisée par un seul agent, mais de manière moins efficace que lorsque celle-ci est réalisée par une équipe d’agents.

L’efficacité augmente donc avec l’augmentation du nombre d’agents dans le groupe, jusqu’à une certaine limite où l’ajout de nouveaux membres n’ajoute aucun apport.
Un exemple de missions qui peuvent être réalisées seules ou en coordination avec d’autres robots comprend les tâches d’exploration, la recherche de personnes et le déploiement de capteurs.

     \subsubsection{Coopération}
     La coopération entre plusieurs agents est semblable à la coordination, à la différence que la tâche en question ne peut d’être réalisée par un seul agent, comme l’action de déplacer un objet très lourd par exemple, ou jouer à un match dans un sport collectif.
  
     L’efficacité passe donc de 0\% en utilisant un seul agent à 100\% en utilisant un groupe d’agents.

     \subsubsection{Collaboration}
     La collaboration est l’utilisation d’un groupe d’agents différents pour effectuer une tâche ensemble qui ne pouvait pas être réalisée en utilisant un seul type d’agents.

     Ce genre de comportement nécessite souvent la distribution de rôles selon les capacités de chaque agent. On retrouve souvent ce type de tâches dans les chaines de production industrielles où chaque type de robots est responsable de la réalisation d’une sous-tâche.

     L’efficacité passe dans ce cas de figure de 0\% en utilisant un type d’agents à 100\% en utilisant plusieurs types d’agents.

     \subsubsection{Compétition}
     Un comportement compétitif est l’opposé de la coordination ou de la coopération dans le sens ou chaque agent agit de manière à maximiser son gain au profit du gain des autres agents. Un exemple d’un comportement compétitif peut être trouvé dans les compétitions de robotiques pour collecter le maximum nombre d’objets par exemple.
     
     \subsubsection{Opposition}
     Dans le comportement d’opposition, chaque agent agit de manière adverse contre les autres agents de manière à annuler leurs actions. Un exemple de ce type de comportement peut être trouvé dans les systèmes anti-intrusion ou un groupe de robots essaient de neutraliser un autre robot intrus.
     
     \subsubsection{Emergence de comportement}
     C’est le fruit d’accumulation de plusieurs actions effectuées par un ensemble d’agents agissant de manière séparée en suivant des règles simples, mais qui résulte en la réalisation d’une tâche commune et souvent complexe. La coordination émerge donc à partir d’un travail collectif même si elle n’était pas intentionnelle à l’échelle individuelle. 
     
     On retrouve ce genre de comportement dans les essaims de robots inspirés à partir de colonies d’insectes et d’animaux sociaux, pour réaliser des tâches de collecte de nourriture par exemple.

    \subsection{Classification par type de synchronisation}\label{section:sync_architectures}
Il est possible de classer la synchronisation des systèmes multirobots en deux types :

         \subsubsection{Synchronisation implicite}
            Ce mode se base sur l’échange de données de manière indirecte en utilisant une mémoire partagée stockée sur un serveur central. Elle est souvent utilisée dans le but d’éviter le chevauchement entre les actions réalisées par plusieurs robots sans nécessiter une communication directe entre eux. Les robots dans ce type de scénarios ne sont pas forcément conscients de l’existence des autres robots, ils peuvent tout de même profiter des informations récoltées par ceux-ci en consultant les données stockées sur le serveur.

            Ce type de synchronisations est largement utilisé dans des travaux d’exploration et de recherche de survivants où le serveur combine les données des zones explorées par chaque robot, puis les redistribuent aux robots pour leur permettre d’éviter de réexplorer une zone qui a déjà été visitée par un autre robot auparavant.

         \subsubsection{Synchronisation  explicite}
            Ce type concerne l’échange de messages directs entre les robots pour se partager des informations et diviser le travail. Cette communication permet d’effectuer une coordination proactive en permettant aux robots de décider d’une stratégie commune qui maximise les gains du groupe. 

            Un cas concret de ce type de synchronisation est l’échange d’actions futures planifiées par les robots telles que les trajectoires de déplacement afin d’éviter les collisions, ou l’échange d’informations sur leurs états internes (comme le niveau de batterie disponible) afin de négocier la distribution de tâches de manière optimale.

    \subsection{Classification par type d'architecture du système}
    On peut classer les architectures des systèmes multirobots en 3 catégories principales :
    
        \subsubsection{Système centralisé}
        Il s’agit d’un ensemble de robots communiquant avec un nœud central dont le rôle est de gérer le groupe en envoyant des ordres à chaque robot, ou en combinant les données collectées par ceux-ci. Ce nœud central peut être un serveur distant, ou un robot leader qui évolue dans le même environnement que les autres robots, et possède généralement des capacités de calcul et de stockage largement supérieur aux robots exécutants.

        Il constitue le point le plus fragile du système puisque ce dernier ne sera plus opérationnel si la communication avec ce nœud est interrompue ou si celui-ci tombe en panne. Une pratique courante pour diminuer ce genre de risque dans les systèmes critiques est de miser sur la redondance de ce nœud. Ceci permet de renforcer la robustesse du système tout en gardant l’avantage de cette topologie qui se caractérise par sa simplicité et la disponibilité de toutes les informations au même endroit ; le nœud central possède une vue d’ensemble sur l’état de tous les robots, des actions effectuées, et de l’avancement de la mission, et pourra planifier de manière efficace la distribution des tâches.

\noindent
\begin{figure}[ht]
    \setlength{\abovecaptionskip}{0.8cm} 
    \setlength{\belowcaptionskip}{-0.4cm} 
    \centering 

     \begin{subfigure}[b]{0.24\textwidth}
         \centering
         \includegraphics[width=\textwidth]{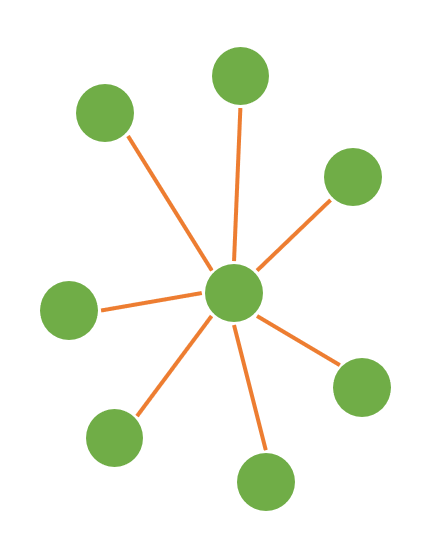}
         \caption{\textit{Centralisé}}
     \end{subfigure}\quad
      \begin{subfigure}[b]{0.33\textwidth}
         \centering
         \includegraphics[width=\textwidth]{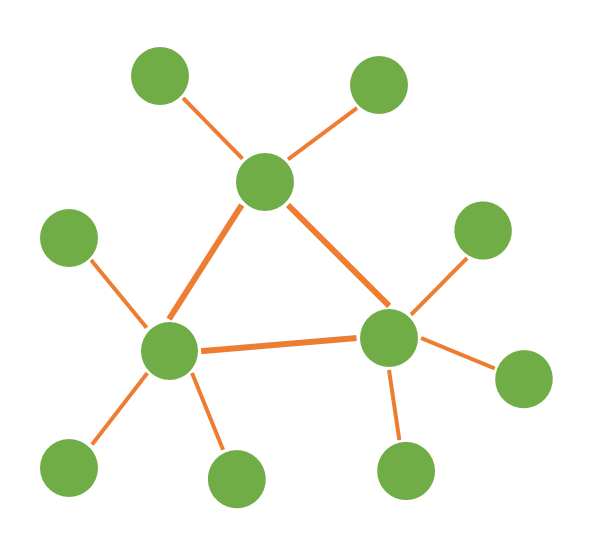}
         \caption{\textit{Décentralisé}}
     \end{subfigure}\quad
      \begin{subfigure}[b]{0.28\textwidth}
         \centering
         \includegraphics[width=\textwidth]{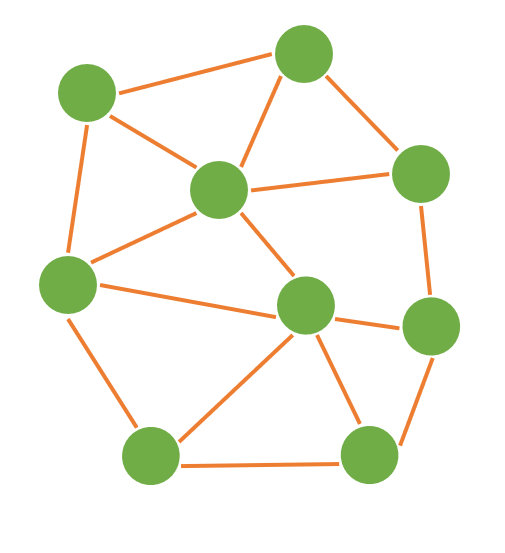}
         \caption{\textit{Distribué}}
     \end{subfigure}

\caption{Types d'architectures de systèmes multirobots}
    \label{fig:c1.6}
\end{figure}

        \subsubsection{Système décentralisé}
        Dans un système décentralisé, il n’y a pas un seul nœud central, mais plutôt plusieurs petits nœuds centraux dont chacun est responsable de gérer un groupe de robots. Ces nœuds centraux communiquent entre eux pour se synchroniser, mais restent indépendants dans les décisions et choix des tâches à affecter aux robots de leurs équipes respectives.
        
        L’avantage d’un tel système est de ne pas posséder un seul point de défaillance (\textit{Single point of failure}) puisque chaque équipe est indépendante des autres. Par ailleurs, il est possible de réaffecter les équipes de sorte à ne pas pénaliser une équipe entière si son nœud central n’est plus opérationnel.

        Ces points centraux jouent souvent le rôle de relais de communication dans les scénarios nécessitant le déploiement de robots dans une zone à large surface, ce qui permet d’éviter l’isolement d’un robot si son rayon de communication est trop loin du nœud central.
        
        Ce genre de systèmes nécessitent toutefois l’utilisation d’un grand nombre d’agents puisqu’il y aura très peu d’intérêts à diviser un petit groupe en plusieurs petites équipes. Une complexité de coordination s’ajoute à cela puisque chaque nœud ne possède qu’une partie de l’état global du système, il est donc impératif que les nœuds centraux échangent entre eux les informations de manière régulière afin d’éviter le chevauchement entre les tâches réalisées par chaque équipe.
        
        Une autre utilité de ce genre d’architecture est de les utiliser dans des scénarios de collaboration entre plusieurs types de robots. Chaque équipe sera constituée par un type de robots et sera coordonnée par un leader qui prend en considération les capacités des membres de son équipe.

        \subsubsection{Système distribué}
        Dans un système distribué, il n’existe pas de notion d’agent central ou de leader. Chaque nœud communique avec ses voisins et décide des tâches qu’il devra accomplir, il est donc totalement indépendant des autres nœuds, ce qui renforce la robustesse du système puisque celui-ci reste opérationnel tant qu’il y a au moins un robot en service. De plus, il n’y a pas l’obligation de disposer d’un serveur central pour combiner les données.

        Cette architecture se caractérise par une communication limitée et une prise de décision au niveau local, ce qui permet d’agrandir la taille du groupe sans avoir besoin d’augmenter les performances des robots ou le débit du réseau. Elle est souvent utilisée dans le cas des essaims de robots (\textit{swarm of robots}) où l’objectif est de déployer un grand nombre de petits robots avec des capacités limitées, dans le but de faire émerger un comportement plus intelligent pour réaliser des tâches complexes. L’idée de base tourne donc autour de la mise à l’échelle (\textit{scalability}) du système multirobots tout en augmentant sa robustesse.

        Le principal inconvénient d’un système distribué est la dispersion de l’information. En effet, chaque robot ne possède qu’une vue très limitée de l’état de la mission. Un robot peut communiquer avec d’autres robots qui ne sont pas dans son voisinage en passant par d’autres robots qui joueront le rôle de relais, mais ceci induira nécessairement à une certaine latence entre l’envoi d’un message et sa réception. Afin de pallier à ces inconvénients, les chercheurs se tournent souvent vers la synchronisation implicite entre les agents et éliminent la nécessité de communiquer entre eux, un exemple peut être trouvé dans les applications de déploiement de capteurs où chaque robot calcule la distance entre les capteurs déployés par ses voisins et adapte sa propre position en conséquence. Un autre exemple se trouve dans les applications de déplacement en groupe ou les robots adaptent leurs vitesses et directions selon la vitesse moyenne et direction de leurs voisins.

\begin{table}[t]
    \centering
    
    \caption{Comparaison entre les architectures des systèmes multirobots}
    \begin{tabularx}{\textwidth}{ |p{0.3\textwidth}|X|X|X| }
    \hline
        \rowcolor{headerColor}
        \multicolumn{1}{|c|}{\textbf{Type d'architecture}} & \multicolumn{1}{c|}{\textbf{Centralisée}} & \multicolumn{1}{c|}{\textbf{Décentralisée}} & \multicolumn{1}{c|}{\textbf{Distribuée}} \\
        \hline
        \textbf{Dépendance au nœud central} & Elevée & Moyenne & Basse  \\
        \hline
        \textbf{Latence de communication} & Basse & Basse & Elevée  \\
        \hline
        \textbf{Robustesse} & Basse & Moyenne & Elevée  \\
        \hline
        \textbf{Simplicité de coordination et synchronisation} & Très facile & Relativement facile & Difficile  \\
        \hline
        \textbf{Mise à l’échelle} & Limitée & Limitée & Sans limite  \\
        \hline
        \textbf{Disponibilité de l’information} & Regroupée dans le nœud central & Répliquée dans les nœuds centraux & Dispersée  \\
        \hline
        \textbf{Adaptabilité au changement d’objectifs} & Facile & Relativement facile & Difficile  \\
        \hline
    \end{tabularx}
    \label{tab1.1}
\end{table}

Le tableau \ref{tab1.1} résume les différentes entre ces trois types d'architectures.

    \subsection{Classification par autonomie}
    Les systèmes multirobots peuvent aussi être classifiés par type d’autonomie :
        \subsubsection{Système autonome}
        Un système autonome est un système totalement indépendant de l’intervention de l’être humain. Il est responsable d’affecter les tâches, les réaliser, faire le suivi de l’avancement de la mission et regrouper les informations collectées par chaque robot.

        Le rôle de l’être humain pendant la mission se limite donc dans la mise en marche ou l’arrêt du système, et l’intervention en cas de dysfonctionnement.

        Les systèmes autonomes sont souvent utilisés dans des environnements contrôlés impliquant des risques limités. Les scénarios les plus communs sont l’automatisation des chaines de production industrielles, la gestion des entrepôts de marchandises, et le nettoyage de surfaces.

        \subsubsection{Système semi-autonome}
        Un système semi-autonome possède une certaine dépendance à l’être humain, qui jouera un rôle plus ou moins important selon la nature de la mission. L’opérateur humain pourra effectuer des tâches de suivi par exemple, de redéfinition des priorités pendant la mission, ou encore le contrôle manuel du robot leader. Les autres robots devront donc s’adapter automatiquement aux décisions de l’opérateur humain ; ils jouent un rôle de support/assistants.
        
        On retrouve ce genre d’organisation dans les scénarios où l’intervention de l’être humain est limitée, mais nécessaire pour le bon déroulement de la mission, par exemple: la surveillance, la recherche des survivants dans les catastrophes naturelles, ainsi que la détection de feux de forêt. 

        \subsubsection{Système contrôlé}
        Il s’agit d’un système contrôlé manuellement par un opérateur humain, souvent à distance à travers une interface de commande offrant une visualisation complète de l’état du système et un retour visuel sur les opérations. 

        Une problématique de base de ce type de systèmes est la manière qu’un seul opérateur humain peut contrôler plusieurs robots à la fois. Une solution serait d’offrir à l’opérateur la possibilité d’affecter les tâches pour chaque robot, et laisser les robots effectuer ces tâches de manière automatique. La différence clé avec un système semi-autonome est que dans un système contrôlé, les robots n’ont pas la liberté de choisir une tâche ou de se déplacer vers un point sauf s’ils reçoivent l’ordre explicite de la part de l’opérateur.

        Ce type de systèmes est utilisé dans les scénarios à fort risque dont la prise de décisions requiert une expertise élevée dans le domaine ou lorsqu'elle est liée à une responsabilité morale qui ne peut être automatisée, tels que les interventions à distance dans les zones contaminées, le diagnostic des pipe-lines, et les applications militaires.

        \subsection{Classification par type de robots}
    Les systèmes multirobots peuvent aussi être classifiés par type de robots qui compose le système : 
        \subsubsection{Classification par homogénéité}
        Un système homogène est un système constitué par des robots similaires du même type, et qui ont les mêmes capacités. Ceci simplifie beaucoup la distribution des tâches puisque les robots sont interchangeables et chaque tâche peut être effectuée par n’importe quel membre du groupe.

        Un système hétérogène est en revanche composé de robots de différents types. La distribution de tâches doit être adaptée selon les capacités de chaque type de robots, ce qui implique un mécanisme de prise de décisions plus spécifique.

\noindent
\begin{figure}[ht]
    \setlength{\abovecaptionskip}{0.4cm} 
    \setlength{\belowcaptionskip}{-0.4cm} 
    \centering 
    \includegraphics[width=0.9\textwidth]{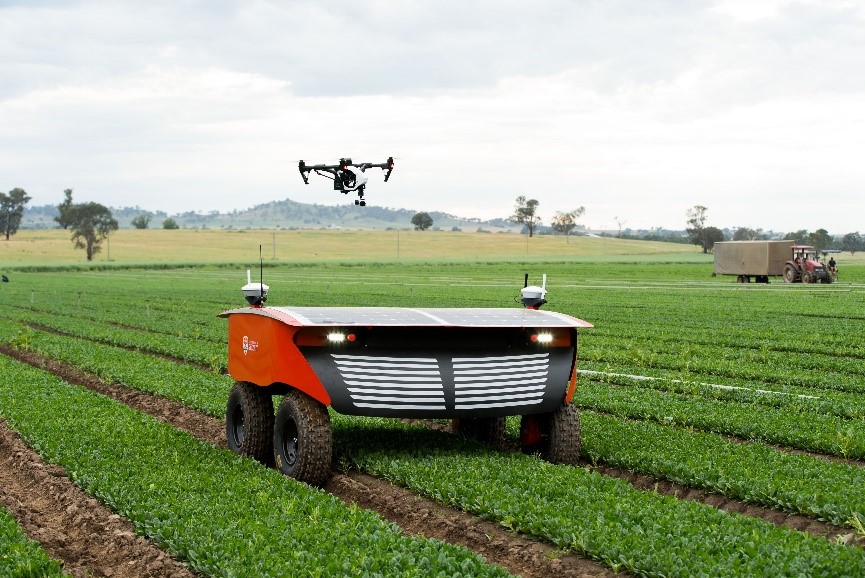}
\caption{Exemple d’un système hétérogène constitué d’un robot volant et d'un robot mobile terrestre pour la supervision d’une zone agricole (\textit{Australian Centre for Field Robotics})}
    \label{fig:c1.7}
\end{figure}

        \subsubsection{Classification par mobilité}
        Un système mobile est constitué de robots qui peuvent se déplacer dans l’espace où ils sont déployés, tels que les robots de type véhicule, les robots volants et les robots marins.
        
        Un système statique est constitué de robots fixés dans le sol, tel que les bras manipulateurs dans les chaines de production industrielles.

        À noter qu’un système hétérogène peut aussi être constitué de robots mobiles et robots statiques qui travaillent en collaboration pour effectuer une mission commune.

        \subsubsection{Classification par mode de locomotion}
        Il existe plusieurs types de locomotion pour les robots mobiles : on trouve par exemple les robots humanoïdes qui ressemblent à des êtres humains ou à des animaux et se déplacent en utilisant des pieds. Il existe aussi des robots de type véhicule, qui se déplacent en utilisant des roues ou des chaînes. Ou encore des robots à propulsion qui se déplacent dans l’air (drones) ou dans l’eau (bateaux, sous-marins). On trouve également un mode de déplacement par frottement, inspiré de certains animaux tels que les serpents.
        
\section{Catégorisation des problématiques liées aux systèmes multirobots}
Les problématiques liées à la robotique en général touchent à plusieurs disciplines dont la mécanique, l’électronique, l’informatique ou encore l’énergétique et les sciences sociales.

Nous nous intéressons dans cette thèse aux problématiques liées à l’informatique et l’intelligence artificielle en général, et plus précisément et au traitement de l’information et à la prise de décisions.

Nous pouvons distinguer dans la littérature plusieurs problématiques qui se chevauchent et qui constituent chacune un axe de recherche très actif dans la communauté :

    \subsection{La navigation}
    La navigation est l’une des problématiques de base dans le domaine de la robotique. Elle soulève la question de comment permettre à un robot de se déplacer tout en évitant les collisions avec les obstacles présents dans son environnement.

    Un obstacle peut être de nature statique comme les murs et les objets. Il peut aussi être de nature dynamique comme les êtres humains, les véhicules et les autres robots.

    L’axe de recherche de la navigation s’intéresse à la modélisation géométrique des robots et de leur mode de déplacement, les degrés de liberté d’un robot, et les techniques de détection et d’évitement d’obstacles.

    La modélisation géométrique du robot est essentielle pour pouvoir contrôler sa vitesse et sa direction de mouvement. Un robot à roues ne se déplace pas de la même manière qu’un drone volant ou qu’un robot à deux pieds. Cette modélisation peut aussi varier pour des robots de même type selon leur nombres de moteurs, leur forme et le nombre de degrés de liberté qu'ils possèdent.

    La navigation nécessite également de pouvoir mesurer l’accélération du robot, qui est souvent obtenue en calculant le nombre de rotations des roues pour les robots de type véhicule par exemple, mais elle nécessite l’utilisation de dispositifs électroniques plus complexes pour les robots volants tels qu'un gyroscope pour mesurer l’orientation, un altimètre pour mesurer l’attitude et un accéléromètre pour mesurer l’accélération.

    D’autres dispositifs doivent aussi être utilisés pour l’évitement d’obstacles, il s’agit souvent de capteurs de distance de type laser, ultrason ou infrarouge pour pouvoir construire un histogramme de distances et choisir une direction de mouvement sans danger. Mais on peut aussi utiliser des méthodes plus complexes comme des caméras 2D ou 3D pour la reconnaissance d’objets, qui s’avèrent utiles lorsqu’il est nécessaire d’interagir avec l’obstacle en question (ouverture de portes par exemple, collecte d’objets…etc.).

    Dans un système multirobots, chaque agent considère les autres robots comme des obstacles à éviter. Toutefois, dans un scénario de déplacement en groupe la navigation devient plus compliquée car elle prend en considération des critères supplémentaires dont la distance maximale autorisée entre chaque robot, la vitesse des robots à proximité, et leur direction de mouvement. Un scénario typique est le déplacement en formation où les robots se déplacent en suivant les mouvements d’un robot leader, ou bien en essayant de garder une certaine formation (ligne droite, cercle, triangle...etc.). Lorsque les robots rencontrent un obstacle, ils sont souvent obligés de rompre la formation pour l’éviter, ils doivent ensuite retourner à la formation initiale en réajustant leurs vitesses et positions.

\noindent
\begin{figure}[ht]
    \setlength{\abovecaptionskip}{0.4cm} 
    \setlength{\belowcaptionskip}{-0.4cm} 
    \centering 
    \includegraphics[width=0.6\textwidth]{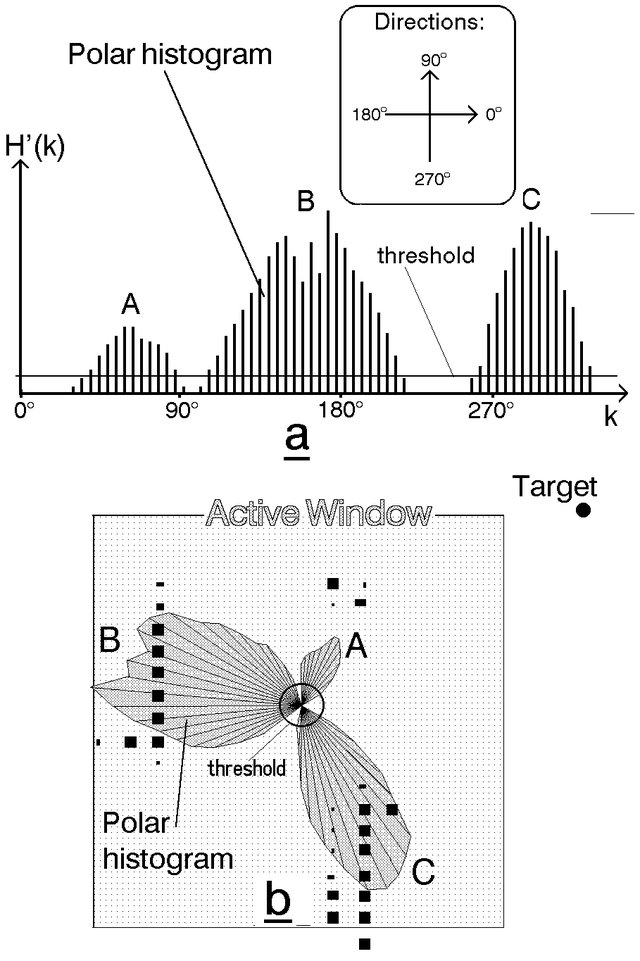}
\caption{Détection d’obstacles en utilisant la technique du \textit{Vector Field Histogram} \cite{borenstein91}}
    \label{fig:c1.8}
\end{figure}

    \subsection{La cartographie}
    La problématique de cartographie tente de répondre à la question « à quoi ressemble l’environnement ? ». Le but est de permettre à un robot de créer un modèle interne de son environnement à partir de ses observations.

\noindent
\begin{figure}[h]
    \setlength{\abovecaptionskip}{0.4cm} 
    \setlength{\belowcaptionskip}{-0.4cm} 
    \centering 
    \includegraphics[width=0.99\textwidth]{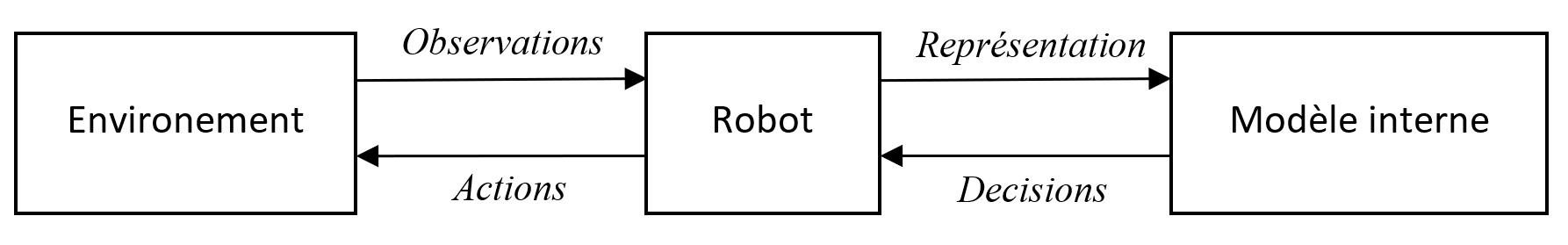}
\caption{Diagramme de relations entre l'environnement du robot et sa représentation interne}
    \label{fig:c1.9}
\end{figure}
    
    Ce processus de modélisation représente l’espace qui entoure le robot dans une structure de données qui permettra de faciliter les autres opérations tel que la planification des trajectoires, distribution de tâches, et localisation des points d’intérêts. Cela permet aussi d’optimiser la navigation puisque le robot pourra savoir à l’avance la position des obstacles qu’il faudra éviter.

    Afin de pouvoir créer une carte de l’environnement, le robot devra traduire les observations récoltées en données utiles. Ces observations sont souvent obtenues soit en utilisant des dispositifs électroniques de mesure de distances (télémètres, radars...) qui permettent de calculer rapidement et avec grande précision l’emplacement des obstacles dans un rayon allant de 180° à 360°, soit en utilisant des capteurs optiques (caméras) qui permettent d’extraire des informations plus riches tels que les couleurs, les formes ou les textures. Chaque méthode à des avantages et des inconvénients, et il n’est pas rare dans les applications réelles de combiner les deux méthodes afin de maximiser la qualité des cartes crées, au profit d’une augmentation des besoins en puissance de calcul et de la capacité de stockage. Le tableau \ref{tab1.2} présente une comparaison entre les deux méthodes.

    \begin{table}
        \centering
        
        \caption{Comparaison entre les capteurs de distance et les capteurs optiques}
        \begin{tabularx}{\textwidth}{ |p{0.25\textwidth}|X|X| }
            \hline
            \rowcolor{headerColor}
            \multicolumn{1}{|c|}{\textbf{Type}} &
            \multicolumn{1}{c|}{\textbf{Capteurs de distance}} &
            \multicolumn{1}{c|}{\textbf{Capteurs optiques}}  \\
            \hline
            \textbf{Exemples} & Télémètres (laser / infrarouge), radars (ultrason) & Caméras 2D, caméras 3D  \\
            \hline
            \textbf{Type d’information} & Distance des obstacles & Couleurs, formes, textures + distance en utilisant des caméras 3D  \\
            \hline
            \textbf{Fréquence de mesure} & 100 à 1000 observations par seconde & 15 à 30 observations par seconde pour la majorité des caméras  \\
            \hline
            \textbf{Rapidité de traitement*} & Supérieur à 100 observations par seconde & 1 à 10 observations par seconde. Peut être augmenté en relayant les calculs à une carte graphique  \\
            \hline
            \textbf{Distance maximale} & 10, 20, 40m pour le laser selon la gamme; 0.5 à 10m pour l’ultrason & Techniquement pas de limite  \\
            \hline
            \textbf{Angle d’observation} & 180° à 360° pour les télémètres laser; 30° à 60° pour les capteurs ultrason. & Varie selon l’objectif;
            100° à 180° pour la majorité des caméras. \\
            \hline
            \textbf{Précision} & 0.1° pour le laser;
            1° à 3° l’ultrason & Dépends de la distance avec l’objet \\
            \hline
            \textbf{Nature de la donnée} & Vecteur de nombre réels & Matrice de pixels  \\
            \hline
            \textbf{Taille de la donnée} & Quelques kilo-octets & > 2 méga-octets selon la résolution de l’image  \\
            \hline
            \multicolumn{3}{p{0.95\textwidth}}{\textit{*Rapidité de traitement : temps nécessaire pour effectuer les opérations d’extraction d’informations utiles à partir d’observations brutes en utilisant un processeur moyen}}
            \\
            
        \end{tabularx}
        \label{tab1.2}
    \end{table}

     Les cartes peuvent être classées en deux catégories selon le type de la structure de données utilisée pour les représenter : 

        \subsubsection{Cartes métriques}
            Elles sont le résultat de la représentation de l’environnement sous forme matricielle, afin de représenter les obstacles avec des formes géométriques simples, telles que des lignes, cercles et polygones.

            Elle s’intéresse à la représentation des périmètres des objets et leur emplacement sans forcément connaître leurs natures, puisque l’objectif est souvent de pouvoir distinguer entre l’espace « vide » où le robot peut naviguer en toute sécurité, et l’espace « occupé » où il y a un grand risque de collision.
      
\begin{figure}[ht]
    \setlength{\abovecaptionskip}{0.4cm} 
    \setlength{\belowcaptionskip}{-0.4cm} 
    \centering
    \includegraphics[width=\textwidth]{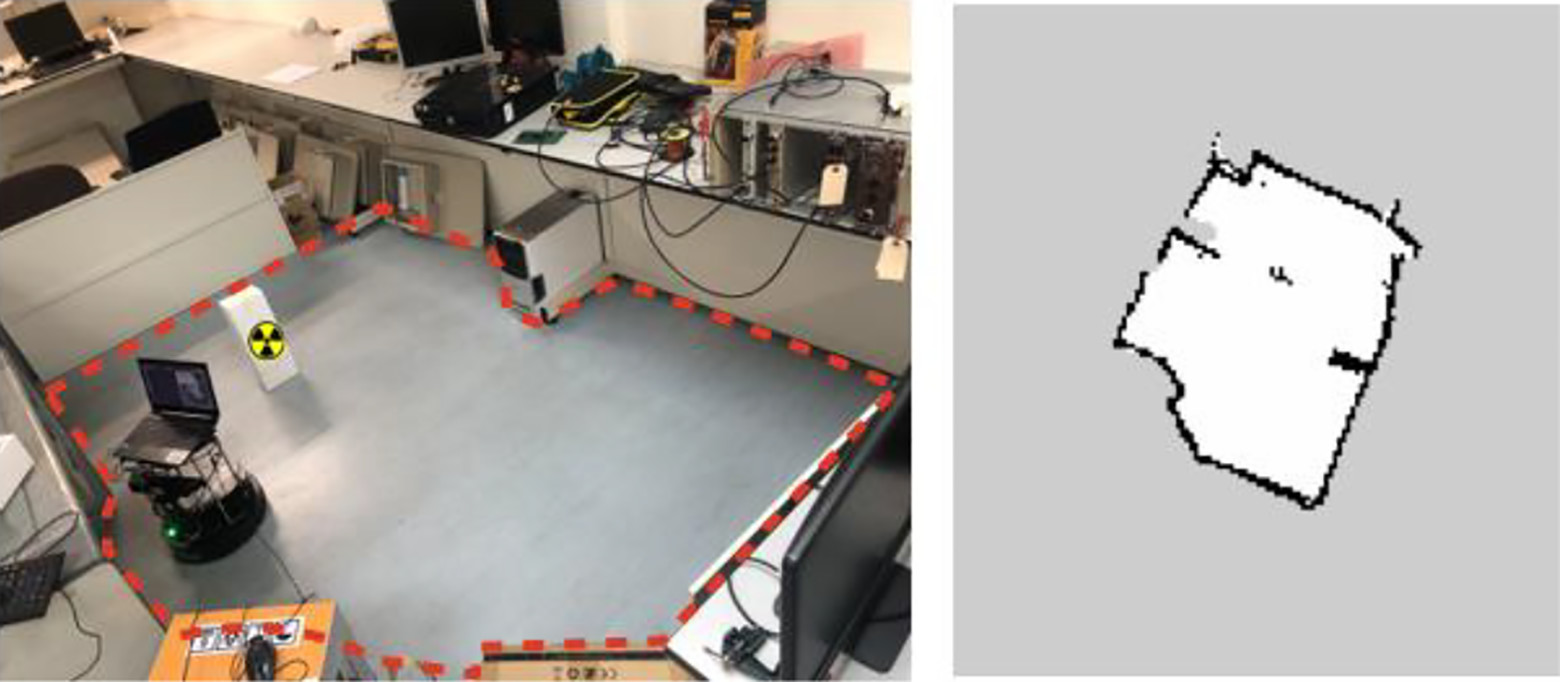}
    \caption{Exemple d’une carte métrique 2D construite en utilisant la technique des grilles d’occupations \cite{abd22}}
    \label{fig:c1.10}
\end{figure}

            Dans la figure \ref{fig:c1.10} par exemple, nous retrouvons une représentation d’une carte métrique 2D construire en utilisant la méthode des grilles d’occupations : la zone blanche représente l’espace vide de l’environnement, les lignes noires représentent la périphérie des murs et obstacles, tandis que l’espace gris représente l’espace non accessible.

\noindent

            Ce type de cartes est simple à réaliser à partir des observations, mais nécessite de connaître la position exacte des obstacles et du robot, ce qui la lie fortement à la problématique de localisation décrite un peu plus loin dans ce chapitre.
        
\begin{figure}[ht]
    \setlength{\abovecaptionskip}{0.4cm} 
    \setlength{\belowcaptionskip}{-0.4cm} 
    \centering 
    \includegraphics[width=0.95\textwidth]{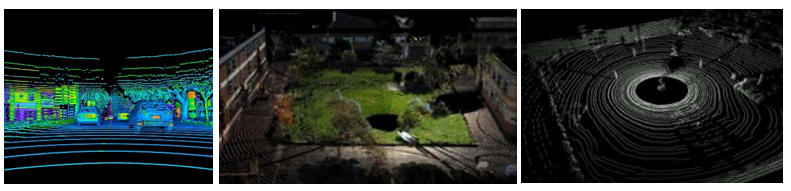}
\caption{Exemples de cartes métriques 3D}
    \label{fig:c1.11}
\end{figure}

            Un autre inconvénient est la nature limitée du modèle. En effet, ces cartes 2D représentent l’environnement par vue d’oiseau, ce qui rends difficile l’estimation de la hauteur d’un objet ou de connaître sa nature. Afin de pallier à cette limitation, une tendance récente est l’utilisation de télémètres 3D ou de caméras 3D pour représenter ces cartes sous forme de nuage de points comme on peut le voir sur la figure \ref{fig:c1.11}. L’avantage de cette représentation est d’offrir la notion de « profondeur » qui est très utile pour pouvoir distinguer entre les murs, les êtres vivants et les objets mobiles.

\noindent
   
        \subsubsection{Cartes topologiques}
            Elles sont le résultat de la représentation de l’environnement sous forme de graphe. Ce type de cartes est plus indulgent quant à l’erreur dans la position exacte du robot puisque l'environnement est plutôt schématisé sous forme de graphe. Pour y arriver, cet environnement est d'abord  divisé en plusieurs régions, qui sont par la suite représentées par des nœuds, puis reliées entre elles par des arêtes représentant les chemins possibles pour se déplacer d’une région à une autre.

            Ces cartes sont plus difficiles à construire et à mettre à jour comparé aux cartes métriques, mais elles sont plus adaptées d’un point de vue algorithmique à la planification de chemins à long terme et à la distribution de tâches entre plusieurs robots. De plus, elles ont l’avantage de consommer moins d’espace mémoire.

\noindent
\begin{figure}[ht]
    \setlength{\abovecaptionskip}{0.4cm} 
    \setlength{\belowcaptionskip}{-0.4cm} 
    \centering 
    \includegraphics[width=0.8\textwidth]{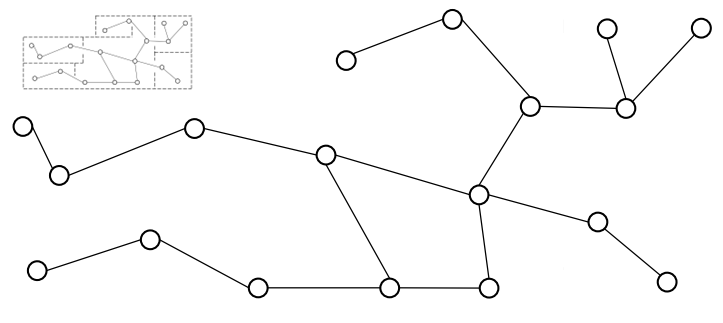}
\caption{Exemple d’une carte topologique}
    \label{fig:c1.12}
\end{figure}
            
            Les cartes topologiques peuvent être enrichies d’informations supplémentaires comme la distance entre les nœuds ou la densité d’occupation (pourcentage du nombre d’obstacles par rapport à la surface de l’espace vide). Ces informations peuvent s’avérer utiles pour la recherche du chemin optimal entre deux points.
            
            Elles peuvent être aussi être combinées avec des cartes métriques afin de profiter des avantages des deux types. La partie métrique sera utilisée pour la navigation à court terme tandis que la partie topologique sera utilisée pour la planification à long terme.
   
        \subsubsection{La cartographie dans un contexte multirobots}

        Dans un contexte multirobots, les cartes sont utilisées comme moyen de communication et de coordination. En combinant les observations partielles récoltées par chaque robot nous obtenons une carte complète de l’environnement.
        
        Cette carte est souvent construite dans une machine centrale. Au début, chaque robot construira sa propre carte locale à partir des observations qu’il enregistre. Ces cartes sont ensuite envoyées au serveur pour les combiner dans une seule carte globale. Plusieurs techniques existent pour effectuer cette combinaison comme la corrélation de scans, la superposition de cartes, ou la fusion de graphes.
        
        Cette opération peut être difficile surtout lorsque la position des robots n’est pas connue avec précision, ou lorsque les observations sont bruitées à cause de capteurs de mauvaise qualité, ce qui peut induire à des inconsistances dans la carte globale.
        
        Le facteur temps est également important, deux robots peuvent passer par un même endroit, mais générer quand même deux cartes différentes si la position des obstacles a changé entre temps. Il sera donc encore plus difficile de les superposer sans avoir recours à des informations supplémentaires comme le temps de passage de chaque robot ou la position de points de repères.
        
        Un autre défi concerne la manière de fusionner des cartes construites par une équipe de robots hétérogènes. Puisque des robots différents peuvent avoir des capteurs différents ou des points de vue différents (robot aérien et robot terrestre par exemple), il faudra veiller à transformer leurs observations vers un format uniforme qui facilitera la fusion des cartes.

\noindent
\begin{figure}[ht]
    \setlength{\abovecaptionskip}{0.4cm} 
    \setlength{\belowcaptionskip}{-0.4cm} 
    \centering 
    \includegraphics[width=0.5\textwidth]{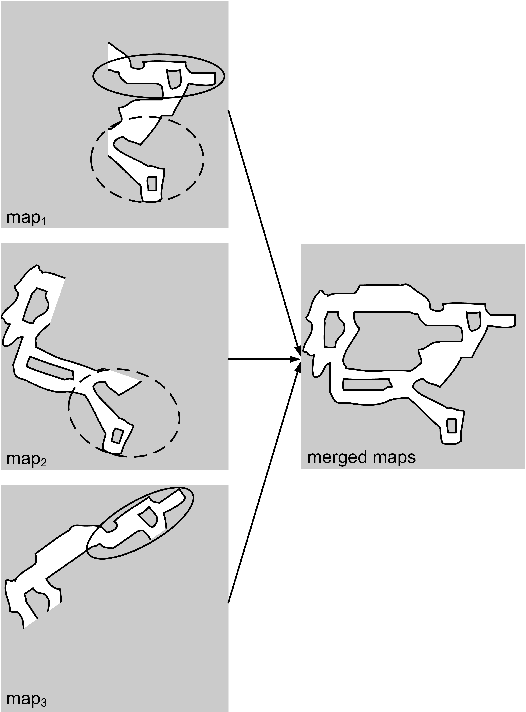}
\caption{Résultat d’un processus de fusion de cartes \cite{saeedi15}}
    \label{fig:c1.13}
\end{figure}

        Les cartes sont aussi un moyen de communication avec l’être humain, car elles permettent à un opérateur de facilement comprendre la structure de l’environnement et visualiser l’état d’avancement d’une mission. Aussi, il sera plus facile de sélectionner graphiquement des points sur une carte pour délimiter une région d’intérêt, que d’insérer un ensemble de coordonnées dans un tableau.
  
    \subsection{La localisation}
    Dans la problématique de localisation, le but est de répondre à la question « où est le robot ? ». On s’intéresse ici à connaître la position du robot par rapport à un repaire fixe. 
    
    Étant donné la nature incertaine de l’environnement en général, cette position est souvent estimée selon une certaine probabilité en se référant à des objets externes, bien qu’il soit possible d’estimer sa position en utilisant des informations internes telles que la vitesse de déplacement du robot et son orientation.
    
    La position d’un robot qui se déplace sur une surface 2D (sol par exemple) est déterminée par deux coordonnées (x, y) et une orientation définie par un angle $\theta$.
    D’un autre côté, un robot qui se déplace sur un espace 3D (robot volant par exemple), nécessite l’utilisation de trois coordonnées (x, y, z) pour déterminer sa position, et trois angles pour déterminer son orientation ($\alpha$, $\beta$, $\gamma$). Ces coordonnées définissent la position du robot par rapport à un certain repère.
    
    Dans les applications réelles de robotique, il est fréquent d’utiliser plusieurs repères pour le calcul des positions. L’un d’entre eux -appelé repère global- est un repère de référence fixe qui sert à calculer la position du robot par rapport au monde qui l’entoure, ce qui permet d'effectuer l'opération de cartographie.
    
    De manière opposée, un repère mobile dont le point d'origine est la position du robot est utilisé pour déterminer la position relative des obstacles en utilisant la distance mesurée par ses capteurs, ce qui est particulièrement utile pendant la phase de navigation pour éviter les collisions. La position de ces obstacles est ensuite recalculée par rapport au repère fixe afin de pouvoir déterminer leurs positions réelles dans la carte de l’environnement. Nous remarquons donc que les problématiques de navigation, de localisation et de cartographie sont fortement liées.
\noindent
\begin{figure}[h]
    \setlength{\abovecaptionskip}{0.4cm}
    \setlength{\belowcaptionskip}{-0.4cm}
    \centering
    \includegraphics[width=0.8\textwidth]{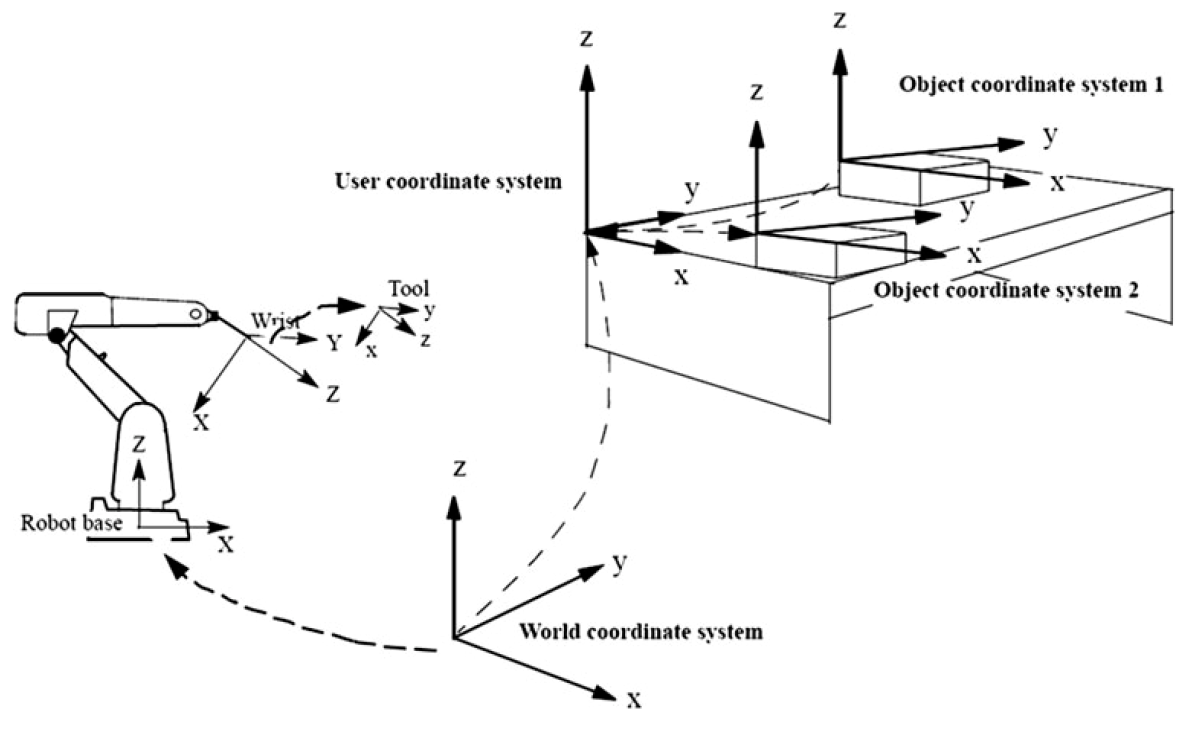}
    \caption{Schématisation des différents repères utilisés pour la localisation relative et globale \cite{deng12}}
    \label{fig:c1.14}
\end{figure}

    Le choix du point d’origine pour un repère de référence présente lui aussi son lot de difficultés. Dans un environnement contrôlé, ce point peut être défini par l’utilisateur et sera utilisé pour positionner tout objet modélisé dans l’environnement. Toutefois, ceci n'est pas toujours possible dans le cas où le robot est déployé dans un environnement inconnu au préalable. Il est donc plus judicieux de fixer la position initiale du robot comme point de référence.
    
    Dans un contexte multirobots, ceci devient plus difficile puisque chaque robot possède son propre point de référence. Il est donc important que ces robots synchronisent leurs positions par rapport à un repère commun comme la sélection d’un objet fixe comme point de référence ou en utilisant un moyen de localisation externe, telle que la géolocalisation.
    
    Certains travaux ont démontré que l’échange des positions des robots entre eux permettrait aussi de réduire les erreurs dans leur localisation. Pour cela, chaque robot détermine la position de l’autre robot lorsque celui-ci rentre dans son champ de vision, puis la lui envoie. Celui-ci comparera la valeur reçue avec celle calculée en utilisant ses propres informations. Ceci permet d’avoir un « point de vue externe » lorsqu’un robot recalcule sa position pendant un déplacement.
    
\begin{figure}
    \setlength{\abovecaptionskip}{0.4cm} 
    \setlength{\belowcaptionskip}{-0.4cm} 
    \centering 
    \includegraphics[width=0.8\textwidth]{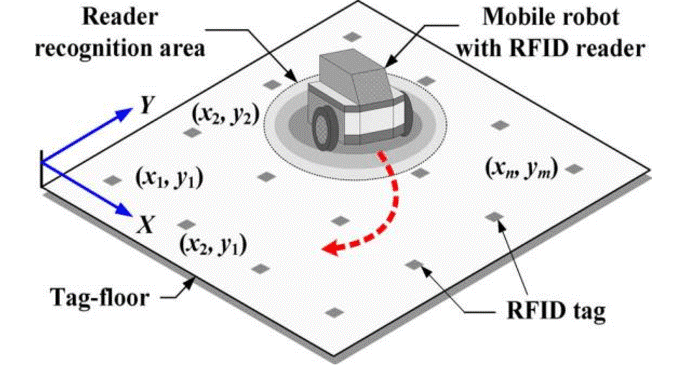}
\caption{Localisation par balises RFID \cite{choi11}}
    \label{fig:c1.15}
\end{figure}

    Une autre difficulté liée au problème de localisation est l’estimation exacte de la position du robot en utilisant des informations incomplètes ou bruitées. En effet, les capteurs du robot sont souvent limités et sujets à de petites erreurs, qui peuvent rapidement s’accumuler pour donner résultat à une localisation incorrecte. Plusieurs méthodes ont été utilisées dans la littérature pour améliorer la localisation des robots en utilisant des informations visuelles recueillies à partir de caméras comme l’identification de points d’intérêts par exemple (portes, fenêtres, objets...) ou l’utilisation de marqueurs placés préalablement dans l’environnement (bornes, balises...) tel que décrit dans la figure \ref{fig:c1.15}.
    
    Dans certains travaux, la localisation est externalisée vers un serveur central lié à des caméras placées en hauteur. C’est le cas lorsque les capacités de calcul des robots ne permettent pas de faire un traitement assez complexe pour les besoins de l’expérience.

\noindent

    \subsection{La planification}
    La planification est une problématique très importante dans le domaine de la robotique parce qu’elle est au centre du processus décisionnel. Elle répond à la question : quelle est la meilleure façon pour accomplir une certaine tâche ?
    
    Le but est de décomposer cette tâche en plusieurs actions (ou sous-tâches) afin de choisir le meilleur ordre d’actions parmi la liste des combinaisons possibles. 
    
    Dans un contexte multirobots, ce choix devient plus compliqué puisqu’il faut répartir ces sous-tâches de manière optimale sur plusieurs agents. Ceci correspond à un type de problèmes mathématiques dont la complexité est combinatoire (NP-complet) \cite{kancir18}, il n’est donc pas toujours envisageable de vérifier toutes les combinaisons possibles.
    
    Dans un système centralisé, cette répartition est généralement effectuée par le nœud central qui affecte les tâches à chaque agent. Dans le cas d’un système décentralisé ou distribué, un consensus doit être trouvé par les robots pour se diviser les tâches de sorte à maximiser le profit cumulé du groupe, même si cela implique que les actions effectuées séparément par chaque robot ne maximisent pas ses profits au plan individuel.
    
    Lorsque le groupe est constitué de robots hétérogènes, l’affectation de tâches doit aussi prendre en considération leur capacités. En effet, certaines tâches ne sont pas effectuées de la même manière par tous les robots et il se pourrait que certaines tâches ne puissent être réalisées que par un type spécifique de robots. Il faudra donc veiller à inclure ces contraintes au processus d’affectation.
    
    Un type particulier de planification concerne le calcul de chemins (\textit{path planning} ou \textit{path finding} en anglais). Il s’agit d’un axe de recherche très actif qui vise à trouver le meilleur ordre d’actions pour se déplacer d’un point A vers un point B. Ces actions prennent la forme de mouvements, d’où l’appellation « planification de mouvements » (\textit{motion planning}). Il y a là aussi des contraintes à prendre en compte telles que la présence d’obstacles, la longueur du chemin choisi, ainsi que les restrictions géométriques du robot.
    
    La planification de chemins peut aussi prendre la forme de répartitions de tâches dans les systèmes multirobots : étant donné plusieurs points de destinations à visiter, le but est de trouver la meilleure combinaison possible pour répartir ces points de destination entre les robots de sorte que chaque point ne soit visité que par un seul robot pour éviter la redondance. Ce type de problèmes est populaire dans les applications de transport de marchandises et de gestion des entrepôts.
    
    Un autre aspect à prendre en compte lors de la planification de chemins dans les systèmes multirobots est l’évitement des collisions. Le critère du temps est très important dans ce contexte puisqu’il n’est pas interdit que deux chemins se chevauchent tant que deux robots ne sont pas présents au même moment au même endroit. Il faut donc veiller à intégrer une stratégie de gestion de conflits entre les robots, en gérant les priorités de passage des robots par exemple ou en intégrant des contraintes supplémentaires lors du calcul de chemins.

\noindent
\begin{figure}[ht]
    \setlength{\abovecaptionskip}{0.4cm} 
    \setlength{\belowcaptionskip}{-0.4cm} 
    \centering 
    \includegraphics[width=0.95\textwidth]{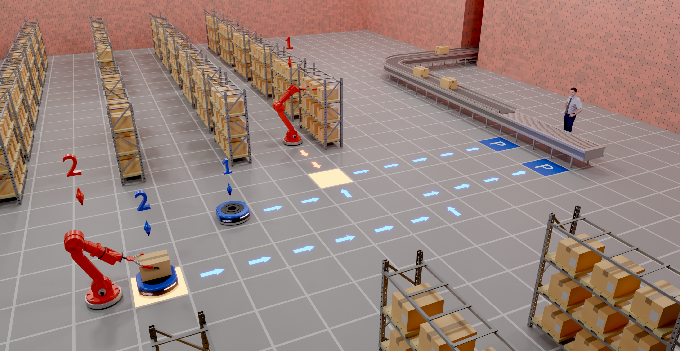}
\caption{Simulation d'un scénario industriel basé sur les algorithmes de planification de trajectoires \cite{greshler21}}
    \label{fig:c1.16}
\end{figure}

    \subsection{L'exploration}

    La tâche d'exploration consiste à parcourir une zone dont le robot n'a aucune information au préalable (ou peu d'informations). Le but est de collecter le maximum de données utiles afin de pouvoir mener à bien la mission.
    
    La problématique d'exploration peut devenir particulièrement difficile avec l'augmentation de la surface de la zone à parcourir et des contraintes de mouvement et d’énergie du robot. Ceci devient particulièrement critique lorsque le facteur temps est limité en raison de nature même de la mission, comme les opérations de sauvetage et recherches de personnes pendant les catastrophes naturelles par exemple. 
    
    La tâche d'exploration est souvent couplée avec d'autres problématiques selon la nature de la mission. Ceci ouvre la possibilité à plusieurs variantes que nous pouvons classer selon les catégories suivantes :

    \begin{itemize}
  \item \underline{Exploration :} Consiste à visiter une zone inconnue dans le but de collecter le maximum de données possibles, souvent sous forme de positions de points d’intérêts ou d’ensemble de routes et chemins possibles.

  \item \underline{Target Searching :} Consiste à explorer une zone dans le but de trouver un individu ou un certain objet. 
  \item \underline{Patroling :} Consiste à explorer une zone de manière répétitive, souvent dans le but de détecter des intrus dans un contexte de surveillance. 
  \item \underline{Consistent Surveillance (Consistent Monitoring) :} Consiste à surveiller une zone de sorte que chaque point de l'environnement soit toujours dans le champ de vision des robots. Le but est de déployer les robots de sorte qu'aucun angle mort ne soit toléré. 
  \item \underline{Total Coverage (Complete Coverage) :} Consiste à faire le balayage complet d'une zone, c'est à dire l'explorer en visitant tous les points atteignables par les robots. On ne se contente pas d'observer et détecter les obstacles à distance, mais de se déplacer et parcourir chaque petite parcelle de la surface de la zone dans le but de la nettoyer par exemple, ou détecter la présence de danger (mines, fuite de gaz...).
  
  \end{itemize}

La problématique d'exploration avec ses différentes variantes est une tâche qu'on peut décomposer en plusieurs sous-tâches. En effet, plusieurs stratégies peuvent être employées pour répartir cet ensemble de sous-tâches sur un groupe de robots afin d'accélérer sa réalisation. 

La stratégie la plus intuitive est de diviser la zone à parcourir en plusieurs régions, puis d'affecter à chaque robot une ou plusieurs régions à explorer. Le problème se transforme donc en un problème d'affectation pouvant prendre en compte un ou plusieurs critères d'optimisation, tel que la distance du robot par rapport à la région affectée, la taille de la région, ou encore l’énergie restante du robot.

Toutefois, ceci soulève aussi plusieurs questions : quel est le nombre optimal de régions ? Quelle est la forme des régions ? Quel type de trajectoire faut-il utiliser ?

\begin{figure}[ht]
    \setlength{\abovecaptionskip}{0.4cm} 
    \setlength{\belowcaptionskip}{-0.4cm} 
    \centering 
    \includegraphics[width=0.98\textwidth]{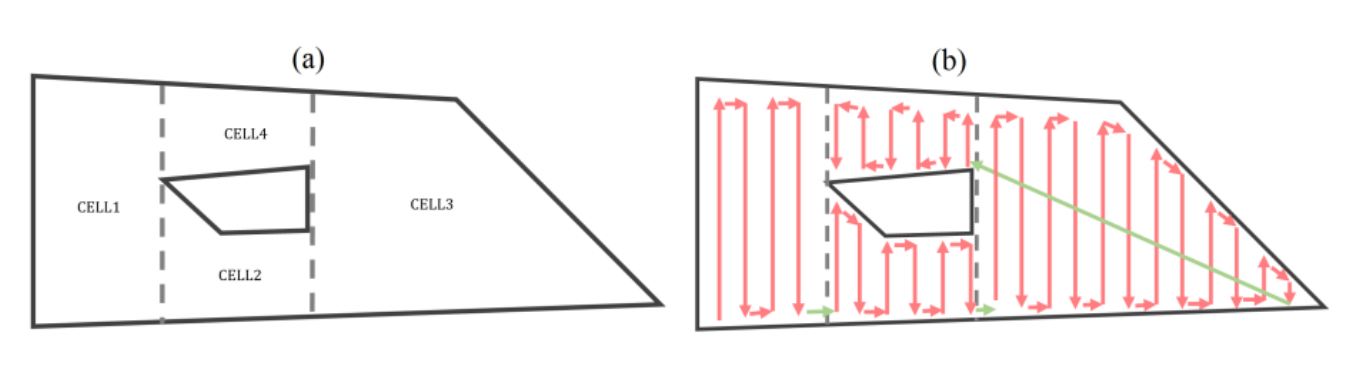}
    \caption{Exemple de décomposition d'une carte et balayage en utilisant une trajectoire en zigzag \cite{horvath18}}
    \label{fig:c1.17}
\end{figure}

\begin{figure}[ht]
    \setlength{\abovecaptionskip}{0.4cm} 
    \setlength{\belowcaptionskip}{-0.4cm} 
    \centering 
    \includegraphics[width=0.9\textwidth]{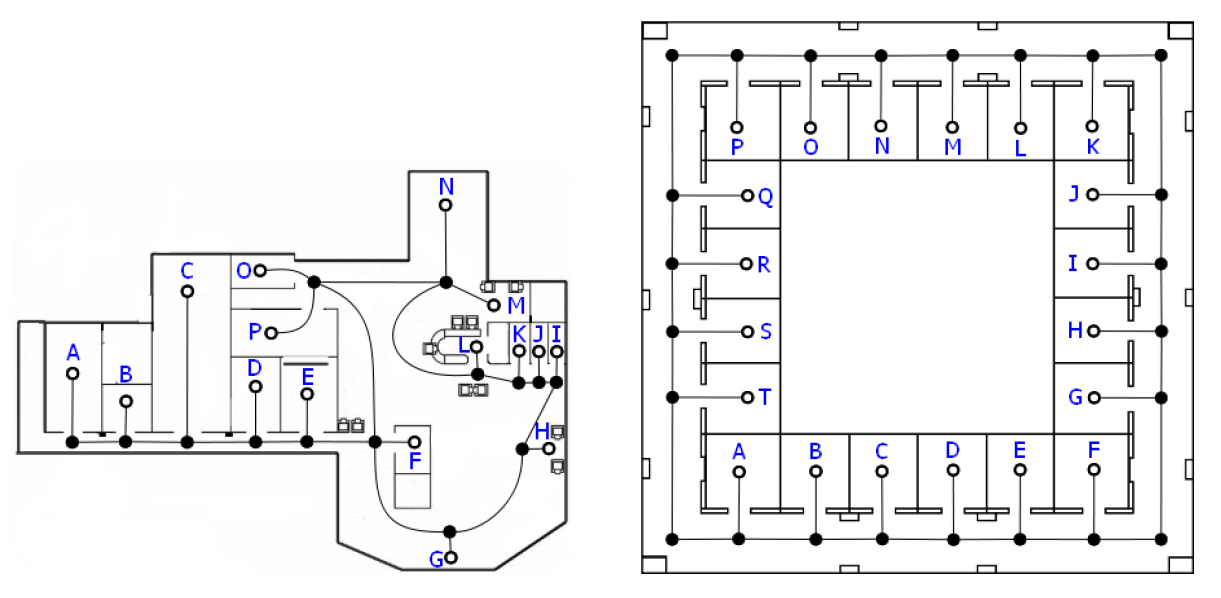}
    \caption{Exemple de sélection de régions à visiter dans un scénario de surveillance \cite{iocchi11}}
    \label{fig:c1.18}
\end{figure}

Les réponses à ces questions varient selon la nature de la mission et des techniques à utiliser. Les figures \ref{fig:c1.17} et \ref{fig:c1.18} montrent des exemples de ce genre de décisions dans deux scénarios différents.

\noindent

\noindent

Une stratégie d’exploration efficace poussera les robots à minimiser le chevauchement entre les régions visitées et éviter de retourner par le même chemin, sauf si cela est inévitable comme lorsqu’un robot atteint une route fermée et doit faire demi-tour, ou dans le cas d’une intersection de plusieurs chemins (ex : couloir, hall).

Dans les scénarios de surveillance par contre, la visite répétitive d’une même région est obligatoire puisque le but est de s’assurer qu’aucun intrus n’a pénétré la zone explorée. Les robots devront faire en sorte de toujours revenir vers les régions visitées à une fréquence régulière. Les critères d’optimisation devront donc être adaptés selon chaque type de scénario.

    \subsection{La communication}
    La problématique de communication est au cœur des systèmes multirobots. Plusieurs stratégies de communication peuvent être employées selon le type de coordination et l'architecture générale du système.
        \subsubsection{Type de communication :}
    \begin{itemize}
\item \textbf{Communication passive :} 

Il s'agit d'échanger des informations indirectement, soit en détectant les autres robots et en calculant leurs positions et trajectoires lorsqu'ils rentrent dans le champ de vision. Soit en déposant ou déplaçant des objets dans l'environnement, similairement au mode de communication de certains insectes qui dégagent des produits chimiques lorsqu'ils se déplacent (odeurs, phéromones...) et déposent des morceaux de nourriture dans les endroits de stockage (fourmilières).

Ce type de communication est souvent facile à mettre en œuvre, mais ne garantit pas la réception du message de la part des autres robots. Aussi, il est difficile de garantir la confidentialité ou de vérifier l'authenticité de l'émetteur puisqu'un robot intrus peut copier le mode d'interaction avec l'environnement pour influencer la décision du système.

\item \textbf{Communication active :} 

Il s'agit d'échanger des messages de manière intentionnelle en utilisant un protocole de communication qui garantit l'adressage unique de chaque robot. Le protocole le plus souvent utilisé est le TCP/IP à travers une liaison Wifi ou Bluetooth, ce qui introduit une augmentation de la consommation d’énergie du robot, notamment pour les dispositifs à rayon large. Mais d'autres moyens peuvent aussi être utilisés comme l'échange de messages avec des balises qui jouent le rôle de relais pour augmenter le rayon de couverture du dispositif de communication installé sur les robots.

\end{itemize}

        \subsubsection{Mode de communication :}
        Un autre point important lié à cette problématique est le choix du mode de communication. En effet dans une communication décentralisée, les robots s'échangent des messages directement lorsqu'ils sont à proximité. Ceci permet d'identifier qui est l'émetteur et le destinataire afin d'envoyer des messages de manière ciblée. Toutefois, il est difficile de combiner les informations de tous les robots afin d'avoir une vue d'ensemble générale.

        De l'autre côté, le mode de communication centralisé permet de combiner toutes les informations dans un même endroit. Les robots ne communiquent pas directement entre eux, mais passent par un serveur qui sera responsable de traiter et distribuer les informations. Ce serveur fera en sorte que chaque robot ait accès à la même information que les autres, ou au contraire de choisir quel robot a accès à quelle information.

        Dans ce dernier mode de communication, il est important de spécifier l'endroit où sera installé le nœud central. En effet, une première solution est de mettre ce nœud dans le même réseau local que le système multirobots ce qui permet d'avoir un débit de communication élevé. Une autre solution est d'accéder à ce nœud à distance — souvent en utilisant le réseau internet — afin de pouvoir profiter de fonctionnalités plus évoluées telles que l'accès aux serveurs Cloud par exemple, ou la visualisation des actions des robots à travers un centre de commandement.


\renewcommand{\arraystretch}{1.7}
\begin{table}[ht]
    \centering
    \caption{Récapitulatif des problématiques de bases dans le domaine des systèmes multirobots}
    \begin{tabularx}{\textwidth}{ |p{0.25\textwidth}|X|X| }
    \hline
        \rowcolor{headerColor}
            \multicolumn{1}{|c|}{\textbf{Problématique}} &
            \multicolumn{1}{c|}{\textbf{Objectif}} &
            \multicolumn{1}{c|}{\textbf{Types}}  \\
        \hline
        \textbf{Navigation} & Se déplacer tout en évitant les obstacles & --  \\ 
        \hline
        \textbf{Cartographie} & Dessiner la structure de l’environnement & Métrique (grille), Topologique (graphe)  \\ 
        \hline
        \textbf{Localisation} & Calculer la position du robot & Localisation relative, géolocalisation  \\ 
        \hline
        \textbf{Planification} & Calculer un chemin d’un point A vers un point B & Planification à court terme, à long terme  \\ 
        \hline
        \textbf{Exploration} & Visiter une zone inconnue pour collecter des informations & Exploration, Balayage, Surveillance  \\ 
        \hline
        \textbf{Communication} & Échanger des informations & Communication robot-server, communication inter-robot  \\ 
        \hline
        \textbf{Contrôle et gestion de flotte} & Envoi des ordres aux robots par l’opérateur humain & Contrôle de robot individuel, contrôle groupé  \\ \hline
        \textbf{Interfaces homme-machines} & Visualiser l’état des robots et l’avancement de leurs tâches & --  \\ 
        \hline
    \end{tabularx}
\end{table}
    
    \subsection{Autres problématiques}
    D’autres problématiques liées aux systèmes multirobots comprennent des sujets rattachés généralement à d’autres domaines scientifiques, citons par exemple : 

\begin{itemize}
    \item L’étude de l’assemblage de plusieurs robots en un seul (mécanique et électronique) ;
    \item Les dispositifs d’interconnectivité des robots (télécommunications) ;
\end{itemize}
	
Par ailleurs, d’autres sujets liés au domaine de la robotique en général comprennent l’étude des :

\begin{itemize}
    \item Moyens de locomotion des robots ;
    \item Dispositifs de détection (détecteurs de distances, caméras, capteurs spécifiques : gaz, température, métaux…etc) ;
    \item Dispositifs de mesures internes (vitesse, accélération, orientation, géolocalisation, force exercée, énergie consommée);
    \item Ressources énergétiques du robot (batteries, combustibles…) ; 
    \item Effets psychologies liés à l’utilisation des robots en présence des êtres humains :
    \item Sujets liés à l’éthique et réglementations dans le domaine de la robotique.
\end{itemize}
    
\section{Etat de l'art et travaux connexes}
Cette section présente un aperçu des techniques d’apprentissage machine et d’optimisation utilisées pour résoudre les problématiques liées aux systèmes multirobots, nous nous intéresserons surtout aux problématiques liées à notre présente thèse à savoir la navigation, la planification et l’exploration.

De nombreuses techniques ont été utilisées en robotique pour l'exploration de zones. Ces techniques peuvent être classées selon plusieurs critères concernant leur déterminisme, la nécessité d'utiliser des informations préalables, ou l'adaptabilité à un contexte multirobots.

    \subsection{Les méthodes déterministes}
        Une méthode déterministe populaire pour résoudre le problème d'exploration de zone inconnue a été introduite par \cite{yamauchi97} où le robot continue de se déplacer vers le point de frontière le plus proche (\emph{frontier-based exploration}). Les frontières sont les lignes séparant les régions explorées et inexplorées d'une zone. Cette technique est facile à mettre en œuvre, nécessite peu de ressources de calcul, et donne de bons résultats en pratique. Ceci a encouragé les chercheurs à développer de nombreuses variantes du même algorithme dans le but de trouver la meilleure stratégie pour sélectionner les points de frontières les plus intéressants \cite{holz10}. 

        Une version multirobots a aussi été proposée par les auteurs initiaux \cite{yamauchi98} où chaque robot se déplace vers sa frontière la plus proche tout en envoyant la mise à jour de l’opération de cartographie aux autres robots ; cependant, cette stratégie ne parvient pas à éviter la redondance puisque plusieurs robots peuvent se déplacer vers la même frontière. L'auteur de \cite{bautin12} a essayé de résoudre ce problème en utilisant la technique de propagation d'onde (\emph{wavefront propagation}) pour dispatcher les robots. Les frontières sont classées en fonction du nombre de robots à proximité. Chaque robot est alors affecté à une frontière différente, ce qui lui permet de s'éloigner le plus possible des autres robots et maximiser la zone d'exploration.

        Les auteurs de \cite{alkhawaldah15} se sont orientés vers une stratégie différente pour les environnements d’intérieurs : le premier robot explore tout le couloir et détecte les portes, puis chacun des autres robots sélectionne une porte différente et explore la pièce correspondante en utilisant la technique des frontières les plus proches. Le robot ne sort pas d'une pièce tant qu'il ne l'a pas entièrement explorée. Cette stratégie encourage les robots à explorer les pièces individuellement et à réduire le chevauchement entre les régions qui leur sont assignées, ce qui contribue à réduire le temps total de mission. Plus récemment, \cite{luperto2020} a utilisé des données incomplètes telles que des cartes d'évacuation simplifiées ou des plans d’architecture comme entrée à l’algorithme afin de choisir la meilleure frontière à explorer. Les expériences ont montré que l'exploitation de cartes approximatives peut accélérer la mission d'exploration, même si les données sont inexactes. Cependant, cela nécessite que ces cartes soient importées, prétraitées et alignées manuellement par un opérateur humain.
      
        Une autre famille d'approches déterministes repose sur la décomposition de l'environnement en plusieurs sous-régions, puis sur l'exploration de chaque région indépendamment à l'aide d'une stratégie simple telle qu'un mouvement en zigzag ou en forme circulaire. Une technique populaire de cette famille est la \emph{Boustrophédon Decomposition} proposé par \cite{choset98}. Le principe est de décomposer la carte en régions polygonales en fonction de la position des obstacles. Elle a été utilisée avec succès pour effectuer des tâches de couverture complète (\emph{Complete Coverage}). Cependant, elle suppose que l’environnement soit statique, c'est-à-dire que tous les obstacles sont fixes et ne changent pas de position. Elle nécessite aussi que la structure de l’environnement (carte métrique) soit connue à l'avance puisqu’elle figure parmi les paramètres d’entrée de l’algorithme à fournir par l’utilisateur. D'autres techniques utilisent les diagrammes de Voronoï pour décomposer la carte en régions plus flexibles \cite{garcia07, masehian04}. \cite{dakulovic15} a utilisé la propagation d’onde pour adapter l'algorithme D* au problème de couverture. Au lieu de planifier un chemin à partir d’une position de départ vers une position d’arrivée, le D* modifié génére  un chemin pour visiter tous les points d'une carte donnée. La fonction de replanification rapide de l'algorithme D* lui permet de s'adapter aux environnements dynamiques en modifiant rapidement la trajectoire en cas de changement de position d'un obstacle..

        Les auteurs de \cite{song15} ont proposé une nouvelle approche pour générer des chemins de couverture efficaces. Elle utilise une représentation de cartes multicouches appelée \emph{Exploratory Turing Machine} pour produire une trajectoire en zigzag avec une direction de déplacement réglable. Cette stratégie se traduit par des longueurs de trajectoire plus courtes par rapport aux méthodes classiques basées sur des mouvements de va-et-vient. Les auteurs de \cite{shen20} l'ont étendu récemment en ajoutant des contraintes d'énergie : le robot exécute le trajet de couverture jusqu'à ce que son énergie soit faible, avant de retourner à la borne de recharge de batteries. Après cela, il redémarre la couverture à partir de la région inexplorée la plus proche afin d’éviter de faire un long trajet pour retourner jusqu'au point précédent où il s'était arrêté. Cette approche garantit la couverture complète de l'environnement avec un chevauchement réduit.

    \subsection{Les méthodes à base d'apprentissage machine}
        Des méthodes basées sur l'apprentissage ont également été utilisées pour résoudre le problème d'exploration robotique. Dans l'approche proposée par \cite{lei16}, le robot ne s'appuie sur aucune carte pour explorer l'environnement, il utilise plutôt un réseau de neurones par renforcement de type \emph{Deep Q-Network} de bout en bout pour choisir la direction la plus appropriée à suivre en utilisant uniquement les images de la caméra comme paramètre d’entrée. Cette méthode a été testée pour naviguer dans un environnement sous forme de couloir inconnu tout en évitant les murs.

        D'autre part, l’approche de \cite{strom17} utilise une base de données de cartes déjà vues pour prédire les régions inconnues d'une carte partiellement explorée. Elle utilise une technique inspirée du sac de mots (\emph{bag of words}) pour détecter les similitudes entre des cartes sous forme de grilles et apprendre à compléter les zones manquantes. Cela avait permis  la planification des chemins au-delà de la région explorée, ce qui a réduit la distance parcourue par le robot comparé à la méthode d'exploration à base de frontières. De même, le modèle de \cite{shrestha19} prédit l'emplacement et la forme des obstacles se trouvant au-delà des frontières dans les régions inconnues. Pour celà, les auteurs ont utilisé un auto-encodeur variationnel (\emph{Variational Autoencoders}) pour la prédiction des régions à explorer, et une heuristique pour évaluer leur coûts et utilité.

        Récemment, les auteurs de \cite{yu23} ont proposé une approche où les robots utilisent l'apprentissage par renforcement dans un contexte multirobots afin d'apprendre une stratégie leur permettant d'explorer une zone efficacement. Une autre approche basée sur l'apprentissage par renforcement, proposée par \cite{zhi19}, permet à un robot d'apprendre à explorer une zone tout en utilisant les images de sa caméra afin de faire une reconnaissance visuelle et éviter les endroits déjà visités. L'approche proposée par \cite{durdu21} se base aussi sur la classification d'images à base de réseaux convolutionnels, mais cette fois dans le but de guider un robot à naviguer dans un environnement sous forme de labyrinthe afin de le cartographier. Les résultats expérimentaux ont montré que l'algorithme a appris à choisir la direction de mouvement du robot de telle façon à éviter les obstacles.

\begin{table*}
\setlength\extrarowheight{7pt}
\centering
\footnotesize
\caption{\label{table:c1.3}{Résumé comparatif des travaux cités}}
\begin{tabular}{|c|c|c|c|c|c|c|c|}
\hline
            \multicolumn{1}{|c|}{\textbf{Ref.}} &
            \multicolumn{1}{c|}{\textbf{Carte}} &
            \multicolumn{1}{c|}{\makecell[c]{\textbf{Famille}\\\textbf{d'approche}}} &
            \multicolumn{1}{c|}{\makecell[c]{\textbf{Type}\\\textbf{d'approche}}} &
            \multicolumn{1}{c|}{\makecell[c]{\textbf{Energie}\\\textbf{limitée}}} &
            \multicolumn{1}{c|}{\textbf{Expérience}} &
            \multicolumn{1}{c|}{\makecell[c]{\textbf{Nbr}\\\textbf{robots}}} &
            \multicolumn{1}{c|}{\makecell[c]{\textbf{Type}\\\textbf{exploration}}}
            \\
            \hline

\footnotesize{\cite{yamauchi97}} &
    Inconnue &
    \multicolumn{1}{c|}{\makecell[c]{Frontier-\\based}} &
    Déterministe &
    Non           &
    Robot réel               &
    Un seul    &
    Exploration       \\ \hline
\footnotesize{\cite{alkhawaldah15}} & 
    Inconnue & \multicolumn{1}{c|}{\makecell[c]{Frontier-\\based}}      & Déterministe    &
    Non             &
    Simulation               &
    Plusieurs  &
    Exploration       \\ \hline
\footnotesize{\cite{bautin12}}& 
    Inconnue & 
    \multicolumn{1}{c|}{\makecell[c]{Wavefront\\propagat.}} &
    Déterministe    &
    Non             &
    Simulation               &
    Plusieurs  &
    Exploration       \\ \hline
\footnotesize{\cite{luperto2020}} &
  \multicolumn{1}{c|}{\makecell[c]{Connue\\Partiellement}} &
  \multicolumn{1}{c|}{\makecell[c]{Frontier-\\based}} &
  Déterministe &
  Non &
  \multicolumn{1}{c|}{\makecell[c]{Simulation et\\Robot réel}}  &
  Un seul &
  Exploration \\ \hline
\footnotesize{\cite{dakulovic15}} &
    Connue   &
    D*                  &
    Déterministe    &
    Non             &
    Simulation               &
    Un seul    &
    \multicolumn{1}{c|}{\makecell[c]{Complete\\coverage}} \\ \hline
\footnotesize{\cite{song15}}        &
    Connue   & $\epsilon*$          &
    Déterministe    &
    Non             &
    \multicolumn{1}{c|}{\makecell[c]{Simulation et\\Robot réel}}  &
    Un seul    &
    \multicolumn{1}{c|}{\makecell[c]{Complete\\coverage}} \\ \hline
\footnotesize{\cite{shen20}}        &
    Inconnue & $\epsilon*$           &
    Déterministe    &
    Oui             &
    Simulation               &
    Un seul  &  
    \multicolumn{1}{c|}{\makecell[c]{Complete\\coverage}} \\ \hline
\footnotesize{\cite{ahmadi18}}      &
    Connue   &
    GA                  &
    Stochastique    &
    Oui             &
    Simulation               &
    Un seul    &
    Exploration       \\ \hline
\footnotesize{\cite{albina20}}      &
    Inconnue &
    GWO                 &
    Stochastique    &
    Non             &
    \multicolumn{1}{c|}{\makecell[c]{Simulation et\\Robot réel}}  &
    Un seul    &
    Exploration       \\ \hline
\footnotesize{\cite{albina19a}}     &
    Inconnue &
    GWO                 &
    Stochastique    &
    Non             &
    Simulation               &
    Plusieurs  & Exploration       \\ \hline
\footnotesize{\cite{lei16}}         &
    /        &
    \multicolumn{1}{c|}{\makecell[c]{Deep\\Q-Network.}}  &
    Apprentissage   &
    Non             &
    Simulation               &
    Un seul    &
    Exploration       \\ \hline
\footnotesize{\cite{strom17}}       &
    Inconnue &
    FabMap2             &
    Apprentissage   &
    Non             &
    Robot réel               &
    Un seul    &
    Exploration       \\ \hline
\footnotesize{\cite{shrestha19}} &
  \multicolumn{1}{c|}{\makecell[c]{Connue\\Partiellement}} &
  \multicolumn{1}{c|}{\makecell[c]{Variational\\Autoencod.}}  &
  Apprentissage &
  Non &
  Simulation &
  Un seul &
  Exploration \\ \hline
\footnotesize{\cite{bendahmane22}}* &
  Inconnue &
  BOA/xBOA &
  Stochastique &
  Oui &
  Simulation &
  \multicolumn{1}{c|}{\makecell[c]{Un seul et\\Plusieurs}}  &
  Exploration \\ \hline

  \multicolumn{8}{l}{\textit{* Notre approche}} \\  
\end{tabular}

\end{table*}

    \subsection{Les méthodes stochastiques}
        Les métaheuristiques ont été largement utilisées dans différents domaines de la robotique \cite{fong15} et sont encore largement utilisées pour les robots terrestres et aériens.
        
        \cite{ahmadi18} a utilisé un algorithme génétique (\textit{Genetic Algorithms}) pour surveiller une zone connue à l'aide d'un robot aérien, tout en satisfaisant certaines contraintes telles que la longueur et la régularité du chemin.

        Les auteurs de \cite{zhou13} ont utilisé l’algorithme des particules en essaim (\emph{Particle Swarm Optimization}) afin de distribuer un groupe de robots sur plusieurs régions différentes de l'environnement. Chaque robot explore la région où il se trouve puis utilise l'algorithme des particules en essaim afin de se diriger vers la prochaine région à explorer en se basant sur l'optimisation des frontières. 
        
        \cite{xiao13} ont utilisé l’algorithme de colonies de fourmis (\emph{Ant Colony Optimization}) pour effectuer une tâche d’exploration dans un contexte multirobots. L'approche est modélisée sous forme de problème d'affectation, où l'algorithme de colonies de fourmis affecte à chaque robot une frontière à explorer tout en minimisant certaines contraintes de distance et de densité des robots dans une même région.

        Récemment, \cite{albina20} ont utilisé l’algorithme d'optimisation des loups (\emph{Grey Wolf Optimizer}) pour affecter au robot un point de frontière à explorer. Une fois ce point atteint, le robot répète l'opération afin de sélectionner le prochain point, et ainsi de suite. Les mêmes auteurs ont également proposé une version multiobjectifs de cet algorithme dans le but de maximiser la surface de la zone explorée et la précision de la carte produite \cite{albina19a}.

        Quant à l'approche proposée dans \cite{gul22}, elle utilise la méthode stochastique de l'optimisation arithmétique (Arithmetic Optimizer) pour renforcer  les capacités d'une méthode déterministe à maximiser l'utilité lors d'une tâche d'exploration.

        Bien que les métaheuristiques les plus utilisées dans le domaine de la robotique soient généralement des techniques classiques utilisées pour résoudre des problèmes d'optimisation globale, d'autres métaheuristiques peuvent également être appliquées conformément au théorème No-Free-Lunch \cite{wolpert97} qui stipule qu'aucun algorithme n’est meilleur qu'un autre algorithme dans tous les types de problèmes. Cela signifie que si une technique montre des résultats supérieurs dans certaines classes de problèmes, elle ne peut pas montrer des résultats optimaux pour toutes les autres classes.

        Ce théorème a motivé les chercheurs à inventer de nouvelles métaheuristiques et à les appliquer à différents domaines, dont la robotique \cite{romeh23}. Cependant, il existe de nombreuses nouvelles métaheuristiques qui n'ont pas encore été utilisées dans le contexte de l'exploration de zones. Quelques exemples de ces techniques récemment développées incluent les algorithmes suivants : \emph{Butterfly Optimization Algorithm} (BOA) \cite{arora19}, \emph{Atomic Orbital Search} \cite{azizi21}, \emph{Dwarf Mongoose Optimization Algorithm} \cite{agushaka22}, \emph{Arithmetic Optimization Algorithm} \cite{abualigah21a}, \emph{Tuna Swarm Optimization} \cite{ xie21}, et \emph{Reptile Search Algorithm} \cite{ abualigah22}.

\section{Conclusion}
Nous avons présenté dans ce chapitre les types de systèmes multirobots, leurs domaines d’applications, ainsi que les problématiques clés formant les principaux axes de recherche de cette discipline.

Nous avons aussi présenté un état de l’art sur la problématique d’exploration, avec un état récapitulatif des principales approches citées, ceci nous a permis de positionner notre travail et donner au lecteur un aperçu de nos objectifs.

Le prochain chapitre sera dédié aux fondements théoriques des métaheuristiques. Nous y présenterons leur structure et mécanismes internes. Nous aborderons également le mode de fonctionnement de quelques métaheuristiques populaires, ainsi que notre contribution à améliorer l'algorithme d'optimisation des papillons (\textit{Butterfly Optimization Algorithm}).

    \chapter{Les métaheuristiques}

\startcontents[chapters]
\printmyminitoc{
}

\section{Introduction}
        L'optimisation dans le domaine mathématique est un terme utilisé pour désigner la recherche de la meilleure solution à un problème donné. L'optimisation peut être considérée comme un processus essayant de répondre à la question suivante : « existe-t-il ou non une meilleure solution au problème ? ». 
 
        Nous allons présenter dans ce chapitre une famille de méthodes utilisées dans le domaine de l’optimisation numérique qui ont connu un grand succès durant le demi-siècle dernier en raison de leur capacité à s’adapter à différents types de problèmes et fournir des résultats satisfaisants. 

        Nous introduisons aussi dans ce chapitre les fondements théoriques d’une nouvelle technique appelée xBOA \cite{bendahmane22} et qui constitue l’une des contributions principales de cette thèse.

\section{Les métaheuristiques}
        Avec l'augmentation rapide des ressources de calcul introduites par les ordinateurs, le besoin de résoudre des problèmes plus complexes est devenu de plus en plus important. Des techniques d’optimisation stochastique ont été développées pour fournir des solutions efficaces à ce type de problèmes.

        Les heuristiques sont une classe d'algorithmes visant à trouver des solutions approximatives à des problèmes d'optimisation dont la complexité est combinatoire. L'optimisation stochastique est un type de techniques d'optimisation se basant sur une recherche aléatoire afin d'explorer plus efficacement l'espace de solutions possibles.
        
        Les métaheuristiques combinent le concept d’heuristique avec l'optimisation stochastique dans le but de trouver des solutions acceptables à des problèmes non linéaires dans un intervalle de temps raisonnable comparé aux techniques de programmation mathématique traditionnelles qui garantissent l'optimalité, mais peuvent entraîner une explosion du temps d'exécution \cite{kaveh19}.

        Le succès des métaheuristiques est causé par leur capacité à résoudre des problèmes à très haute complexité ainsi que des problèmes d’optimisation ayant des objectifs contradictoires. Elles peuvent également gérer des contraintes non linéaires et être appliquées à une variété de problèmes du monde réel sans nécessiter de gros changement du point de vue de la programmation. Un autre avantage important des métaheuristiques est leur capacité à résoudre des problèmes dont le formalisme mathématique n’est pas connu avec précision.
        
        Toutefois, elles peuvent être coûteuses en temps de calcul dans certains cas et ne garantissent pas toujours de trouver la solution optimale. De plus, elles sont difficiles à paramétrer à cause de leur sensibilité aux valeurs initiales, ce qui peut amener à une convergence prématurée.

        Elles fournissent un moyen efficace pour trouver des solutions optimales lorsque l'espace de recherche est trop grand ou lorsque les données sont incomplètes. Elles sont facilement adaptables et peuvent être utilisées dans une variété de domaines, tels que l'ingénierie, l'informatique, l'économie ou les finances.

\section{Les types de métaheuristiques}
        Les métaheuristiques peuvent être classées en deux grandes catégories : les algorithmes à trajectoire et les algorithmes à base de population.
        
        Les algorithmes à trajectoire, aussi appelés algorithmes à solution unique (\emph{Single-Solution Based}) proposent une seule solution et la modifient afin de la faire déplacer dans l’espace de recherche, tandis que les algorithmes basés sur une population \emph{(Population Based)} maintiennent un groupe de solutions potentielles et les font évoluer itérativement, puis sélectionnent la meilleure.

\noindent
\begin{figure}[ht]
    \setlength{\abovecaptionskip}{0.4cm} 
    \setlength{\belowcaptionskip}{-0.4cm} 
    \centering 
    \includegraphics[width=0.9\textwidth]{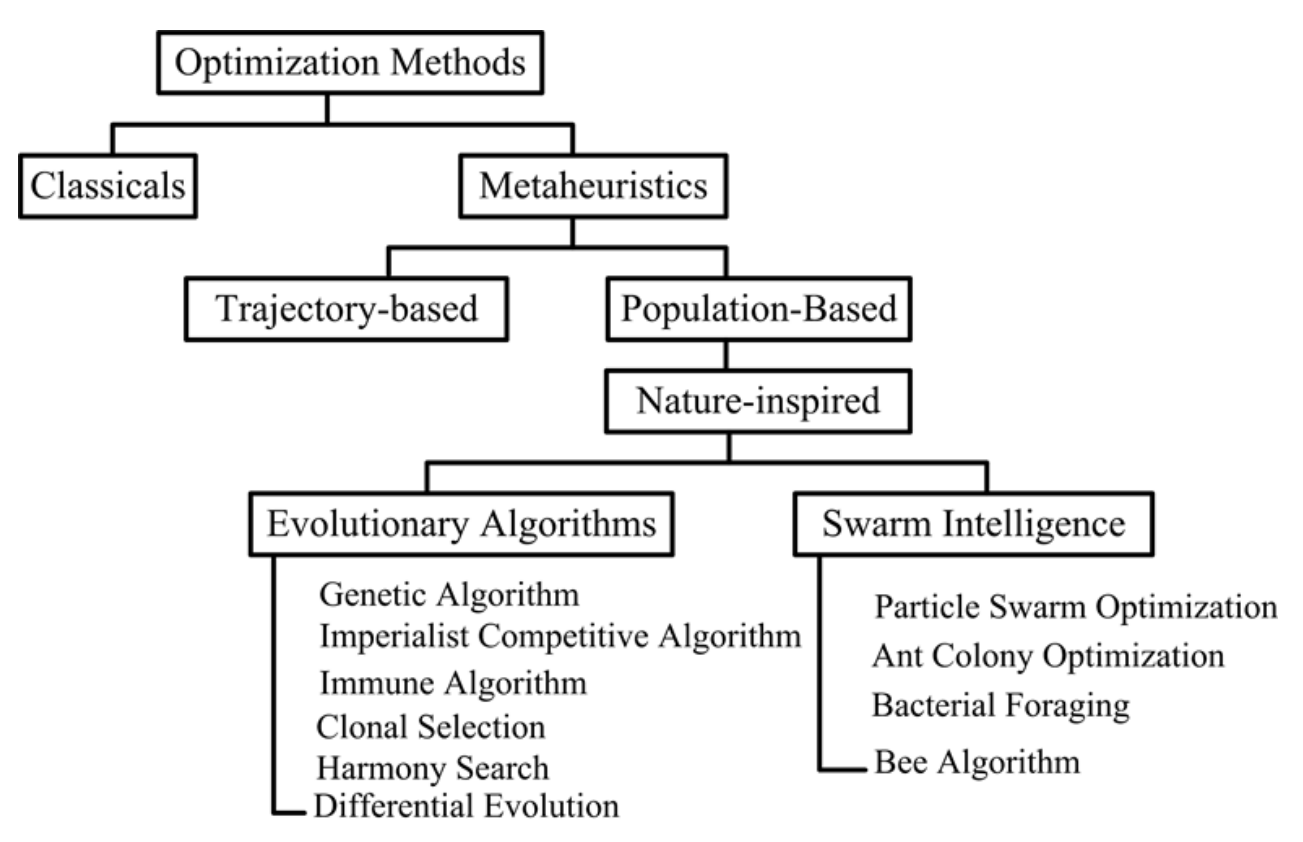}
\caption{Classification des métaheuristiques \cite{aldaco15}}
    \label{fig:c2.1}
\end{figure}

    \subsection{Les méthodes à base de trajectoires}
        Ce type de méthodes se concentre sur l'amélioration d'une solution en la faisant déplacer dans l'espace de recherche formant ainsi une trajectoire.

        Les trajectoires qui améliorent la solution seront automatiquement acceptées, tandis qu'une trajectoire qui décroît la qualité de la solution n'est pas forcément refusée.

\begin{figure}[ht]
    \setlength{\abovecaptionskip}{0.4cm} 
    \setlength{\belowcaptionskip}{-0.4cm} 
    \centering 
    \includegraphics[width=0.9\textwidth]{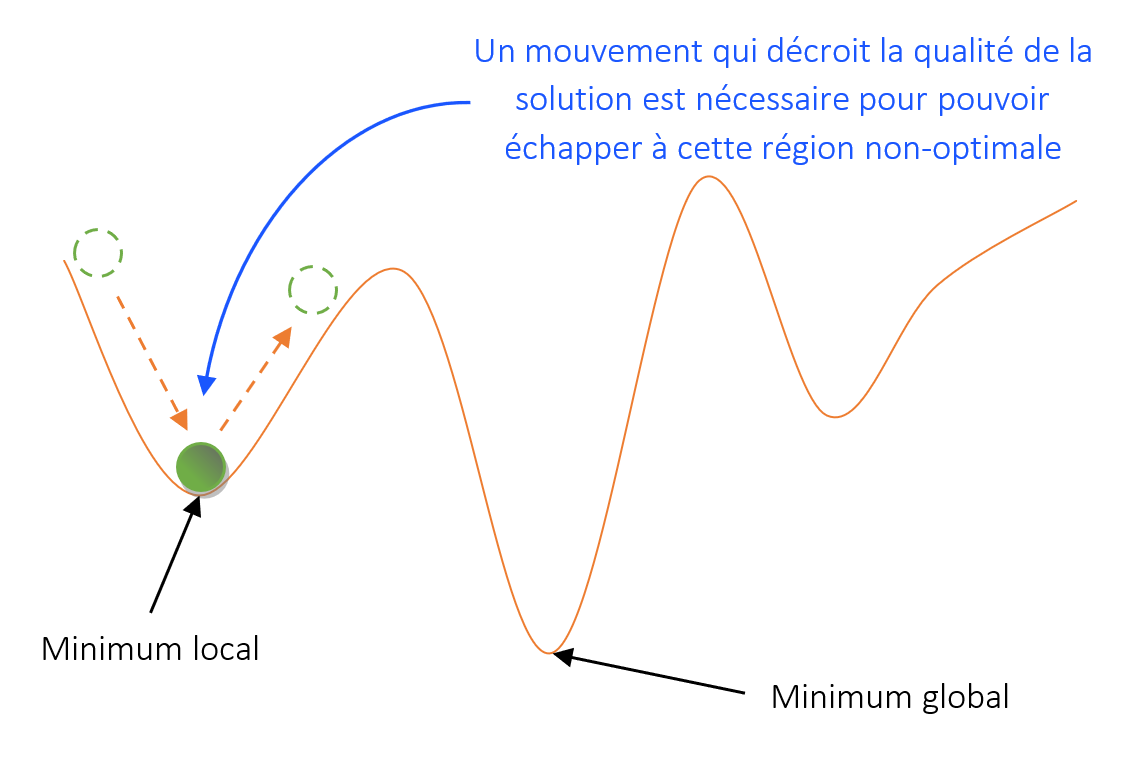}
\caption{Exemple d'une recherche à solution unique}
    \label{fig:c2.2}
\end{figure}

        Elle pourra être acceptée avec une certaine probabilité afin de permettre à l'algorithme d'éviter un blocage dans une région non optimale, tel que l'on peut voir dans l'exemple présenté dans la figure \ref{fig:c2.2}.

\noindent

        Ces méthodes se basent généralement sur des stratégies de recherche locale et recherche de voisinage couplée avec des mécanismes de mémorisation ou de sauts \cite{memari17}

        Parmi les méthodes les plus connues de cette catégorie, nous pouvons citer les suivantes :

        \begin{itemize}
            \item \textbf{La méthode du recuit simulé (\textit{Simulated Annealing})} \cite{kirkpatrick83}
            
            Inspirée de la technique de traitement des métaux, cet algorithme choisit à chaque étape s'il faut accepter de déplacer la solution vers un état voisin qui risque de décroître sa qualité ou s’il faut maintenir le même état. 
            
            Ce choix se fait sur la base d’un calcul de probabilité dont la valeur décroît régulièrement après chaque itération jusqu’à atteindre zéro. En d’autres termes, l’algorithme aura de moins en moins de chances d’accepter les mauvaises solutions au fur et à mesure que le nombre d’itérations augmente. Il finira par converger vers une solution de meilleure qualité.
            
            \item \textbf{La recherche tabou (\textit{Tabu Search})} \cite{glover86}
            
            Cet algorithme se base sur une recherche locale, tout en évitant activement les points de l'espace de recherche déjà visités. Ceci se fait en gardant en mémoire ces points visités dans le but d’éviter les boucles dans les trajectoires de recherche qui conduiront vers le blocage dans un minimum local.

            \item \textbf{La recherche guidée (\textit{Guided Local Search})} \cite{davenport94}
            
            Cet algorithme se base sur une recherche locale classique, à la différence que lorsqu’un minimum local est détecté (aucune amélioration de la qualité de la solution n’est possible), une pénalité est rajoutée à la fonction objective de sorte à encourager la solution à sortir du voisinage courant et passer vers un nouveau voisinage.
            
        \end{itemize}

    \subsection{Les méthodes à base de population}
\begin{figure}[ht]
    \setlength{\abovecaptionskip}{0.4cm} 
    \setlength{\belowcaptionskip}{-0.4cm} 
    \centering 
    \includegraphics[width=0.9\textwidth]{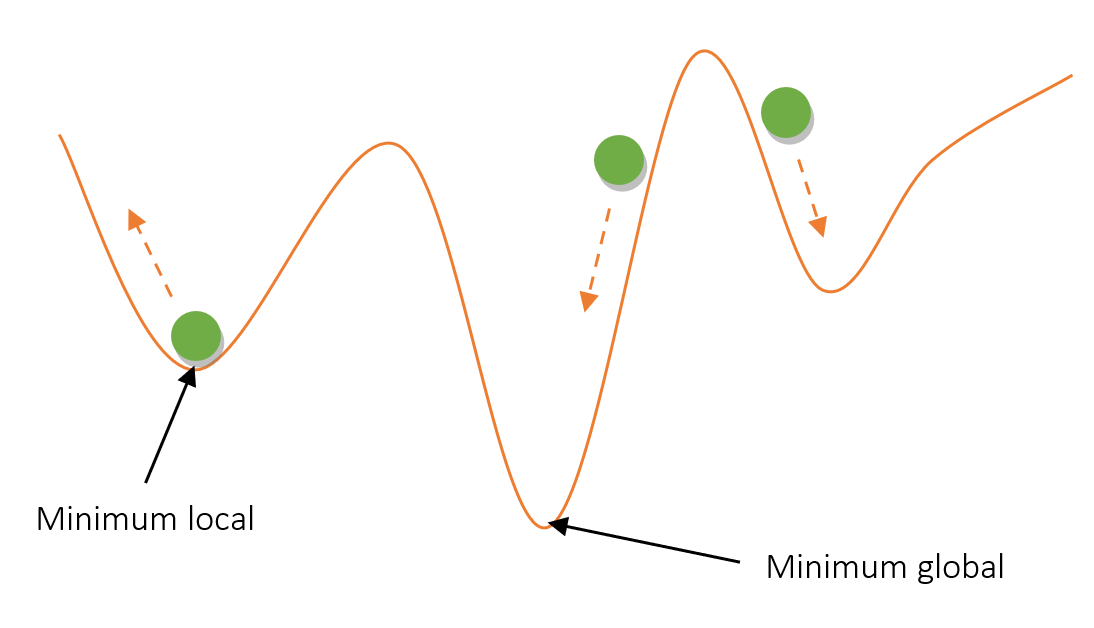}
\caption{Exemple d'une recherche à base de population de solutions}
    \label{fig:c2.3}
\end{figure}

    Ce type de méthodes démarre avec une population de solutions et les améliore toutes en même temps afin d’explorer plusieurs régions de l’espace de recherche en parallèle. Ceci permet une plus grande flexibilité et une robustesse plus élevée par rapport aux minimas locaux (voir figure \ref{fig:c2.3}).

\noindent

    Le choix de la population initiale est crucial pour la réussite du processus d’optimisation. En effet, si toutes les solutions de la population initiale sont presque identiques, il n'y aura pas suffisamment de diversité et elles convergeront prématurément vers la même solution.
    
    Généralement les solutions sont initialisées aléatoirement, cependant, des heuristiques peuvent être utilisées pour influencer cette initialisation de façon à avoir une population suffisamment diversifiée pour pouvoir explorer la plus grande partie de l'espace de recherche.
    
    Les méthodes à base de population peuvent être classées en plusieurs catégories selon le type d’inspiration :
        \subsubsection{Algorithmes évolutionnaires}
        Les algorithmes évolutionnaires ont vu le jour au milieu des années 70 avec les travaux de John Holland \cite{holland76} puis ceux de David Goldberg \cite{goldberg89} sur les algorithmes génétiques \emph{(GA: Genetic Algorithms)}, 
        
        Ces algorithmes, inspirés des principes de l'évolution biologique des êtres vivants, se basent sur le processus d’autoadaptation des espèces dont l’idée générale est que les individus d’une population qui s’adaptent le mieux aux conditions de leur environnement survivent le plus longtemps, ce qui résulte en une évolution progressive de cette population au cours du temps vers des générations meilleures. 
        
        D’autres algorithmes évolutionnaires ont vu le jour depuis la fin des années 80, tous utilisent des populations d’individus représentant un ensemble de solutions potentielles. Ces individus sont mélangés et modifiés en utilisant certains opérateurs spécifiquement créés pour ce type d’algorithmes inspirés des principes de reproduction et de mutation biologique. Seuls les meilleurs individus seront gardés pour la prochaine itération en analogie au processus de la sélection naturelle.

\noindent
\begin{figure}[ht]
    \setlength{\abovecaptionskip}{0.4cm} 
    \setlength{\belowcaptionskip}{-0.4cm} 
    \centering 
    \includegraphics[width=0.9\textwidth]{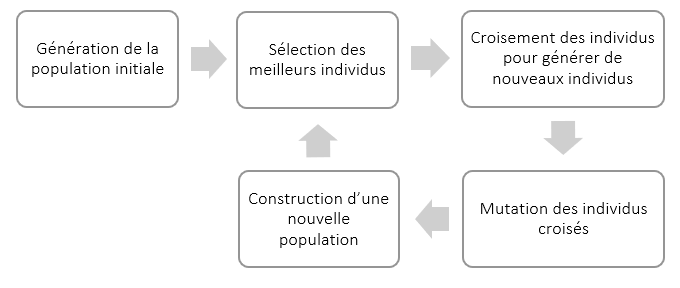}
\caption{Processus de base d'un algorithme génétique}
    \label{fig:c2.4}
\end{figure}

        Après les succès des algorithmes génétiques, d'autres algorithmes bio-inspirés ont été proposés. Les systèmes immunitaires artificiels \emph{(AIS : Artificial Immune Systems)} \cite{farmer86} comptent parmi les premiers et utilisent le principe de fonctionnement des systèmes immunitaires des vertébrés basés sur l’adaptation et la mémorisation.
        
        D’autres types d’algorithmes sont aussi apparus tels que la programmation génétique \emph{(Genetic Programming)} \cite{cramer85} où le but est de faire évoluer un programme sous forme d’arbre en adaptant les opérateurs des algorithmes génétiques à ce type de modèles, ou l’évolution différentielle \emph{(Differential Evolution)} \cite{storn97} qui se basent sur l’évolution d’individus codés sous forme de vecteurs de nombres réels, et dont l’opérateur de croisement se base sur un calcul de distances.

        \subsubsection{Algorithmes d’intelligence en essaim (Swarm Intelligence)}
        
        Le terme d’intelligence en essaim, ou intelligence collective, désigne un phénomène où une population d'agents simples et réactifs interagissent les uns avec les autres de manière à ce qu’un comportement intelligent émerge à la suite de ces interactions.
        
        Ce phénomène est souvent observé dans les essaims de créatures sociales comme les fourmis, les abeilles et les oiseaux.
        
        L’analogie la plus facile pour expliquer ce phénomène est la manière dont les fourmis travaillent. À l’échelle individuelle, une fourmi n’est pas intelligente et son comportement obéit à un ensemble de règles simples, toutefois, en combinant le comportement de toutes les fourmis d’une colonie, nous remarquons l’émergence d’un comportement complexe et auto-organisé. Nous disons donc que le groupe est intelligent, mais que l’individu ne l’est pas.

\noindent
\begin{figure}[thb!]
    \setlength{\abovecaptionskip}{0.4cm} 
    \setlength{\belowcaptionskip}{-0.4cm} 
    \centering 
    \includegraphics[width=0.9\textwidth]{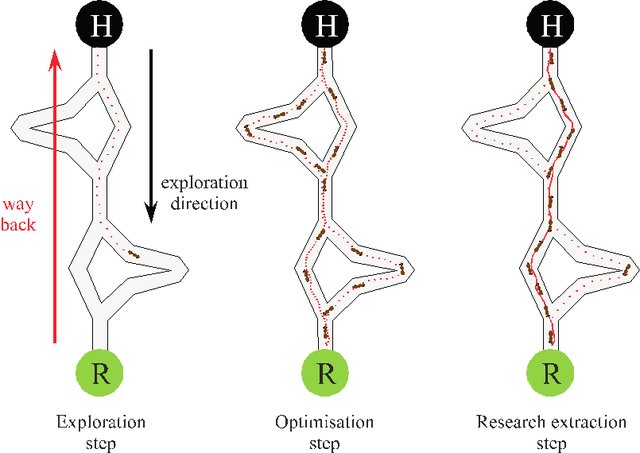}
\caption{Optimisation de chemins par un ensemble de fourmis \cite{katona19}}
    \label{fig:c2.5}
\end{figure}
        
        Des métaheuristiques à base de population sont apparues au début des années 90 s’inspirant de ce phénomène. Elles se basent généralement sur le principe de modéliser les solutions sous forme de vecteurs de nombres réels. Ces vecteurs évolueront à travers les itérations en se rapprochant ou s’éloignant des autres individus selon certaines probabilités et en obéissant à certaines contraintes.
        L'une des méthodes les plus populaires dans ce contexte est l’algorithme d'Optimisation par Colonie de Fourmis (\textit{ACO : Ant Colony Optimization}) \cite{colorni91} qui fut créée pour la recherche de chemins optimaux dans les graphes en modélisant le comportement des fourmis lorsqu’elles se dirigent vers la nourriture.

        Une autre méthode populaire dans cette catégorie est appelée Optimisation par Essaim de Particules (\textit{PSO : Particle Swarm Optimization}) \cite{kennedy95}. Elle imite le comportement social de groupe de poissons et d’oiseaux en vol où chaque individu modifie sa direction de mouvement en réponse à ses propres expériences et à celles de ses voisins. Elle a été appliquée avec succès dans de nombreux domaines et a donné des résultats satisfaisants, en particulier pour les problèmes non linéaires.
        
        Le nombre de métaheuristiques qui rentre dans la catégorie de l’intelligence en essaim ne cesse de croître, et il est difficile de toutes les citer. Il s’agit d’un axe de recherche très actif puisant sa source d'inspiration de la nature. Citons par exemple les algorithmes d’optimisation inspirés du comportement des abeilles par exemple (\textit{ABC: Artificial Bee Colony}) \cite{karaboga05}, les loups (\textit{GWO: Grey Wolf Optimizer}) \cite{mirjalili14}, les chauves-souris (\textit{Bat Algorithm}) \cite{yang10a}, les lucioles (\textit{Firefly Algorithm})\cite{yang10b}...etc.

        \subsubsection{Algorithmes inspirés de la physique }
        
        Il s’agit d’un type d’algorithmes à base de populations -toujours inspiré de la nature- mais cette fois des phénomènes physiques. Ces techniques suivent les mêmes principes que les algorithmes d’intelligence en essaim et sont parfois considérées comme une sous-catégorie de ceux-ci.
        
        Parmi ces algorithmes nous retrouvons ceux basés sur les principes d’interaction entre les masses telles que l’algorithme de recherche à gravitation (\textit{GSA : Gravitational Search Algorithm}) \cite{rashedi09} dont les équations sont inspirées de la loi de la gravité de Newton.

        La population de solution dans cet algorithme est représentée par un ensemble d'atomes qui interagissent entre eux proportionnellement à leurs masses. Plus une solution sera de meilleure qualité, plus sa masse augmentera et attirera donc les autres atomes vers elle.
        
        Nous retrouvons aussi des méthodes basées sur les principes de la mécanique des fluides, telles que le \textit{Vortex Search Algorithm} \cite{dougan15}. Ainsi que d’autres inspirées des principes d’électromagnétisme tel que l’algorithme des champs électriques artificiels (\textit{Artificial Electric Field Algorithm}) \cite{yadav19}.

\noindent
\begin{figure}[bht!]
    \setlength{\abovecaptionskip}{0.4cm} 
    \setlength{\belowcaptionskip}{-0.4cm} 
    \centering 
    \includegraphics[width=0.65\textwidth]{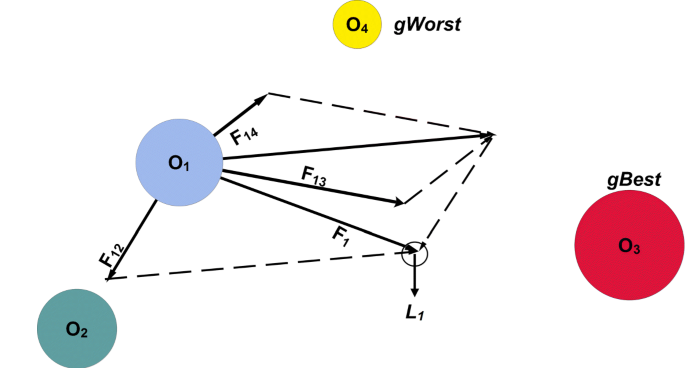}
\caption{Optimisation à base d'interactions physiques entre les atomes \cite{mittal21}}
    \label{fig:c2.6}
\end{figure}
        
\section{Les mécanisme des métaheuristiques}
    \subsection{La fonction objectif}
    Une fonction d’évaluation – communément appelée \textit{fonction objectif} ou \textit{fonction fitness} – est une fonction mathématique qui évalue la qualité d’une solution dans un problème d'optimisation spécifique. 
    
    Elle prend une solution candidate en entrée et retourne un nombre scalaire (voir équation \ref{eq:2.0}). Ce scalaire sera utilisé par l’algorithme pour comparer les solutions trouvées et sélectionner la meilleure. De ce fait, elle dirige le processus de recherche de l'algorithme.
    
    En d’autres termes, pour chaque solution $ s \in S(\Omega) $, une valeur de fitness $f(s)$ existe définie par la formulation suivante :
    
    \begin{align}\label{eq:2.0}
        f : S \longrightarrow R_0^+
    \end{align}    
    \begin{align*}
        s = \{X_i / i = 1..n \}
    \end{align*}

    Sachant que $S$ est l’espace de recherche à $n$ dimensions lié à notre problème, et $\Omega$ est l’ensemble des contraintes du problème.

    \subsection{L'exploration et l'exploitation}
    L’exploration et l’exploitation sont deux concepts clés des métaheuristiques qui doivent constamment être balancées afin de garantir une bonne convergence vers la solution optimale.
    
    L’exploration, aussi appelée \textit{diversification}, est l’habilité de l’algorithme à diversifier les solutions en couvrant plusieurs régions de l’espace de recherche. Tandis que l’exploitation, aussi appelée \textit{intensification}, est l’aptitude à se concentre sur une seule région de l’espace de recherche dans le but d’améliorer une solution en cherchant une meilleure solution dans son voisinage. 

\begin{figure}[ht]
    \setlength{\abovecaptionskip}{0.4cm} 
    \setlength{\belowcaptionskip}{-0.4cm} 
    \centering 
    \includegraphics[width=0.7\textwidth]{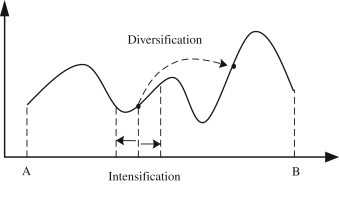}
\caption{Difference entre les stratégies d'exploration et d'exploitation d'une métaheuristique \cite{purnomo15}}
    \label{fig:c2.7}
\end{figure}

    La figure \ref{fig:c2.7} schématise cette différence. La phase d'intensification revient donc à faire une recherche locale. Alors que la phase de diversification à pour but d'explorer l'espace de manière générale.

\noindent

    \subsection{La convergence}
\begin{figure}[ht!]
    \setlength{\abovecaptionskip}{0.4cm} 
    \setlength{\belowcaptionskip}{-0.4cm} 
    \centering 
    \includegraphics[width=0.9\textwidth]{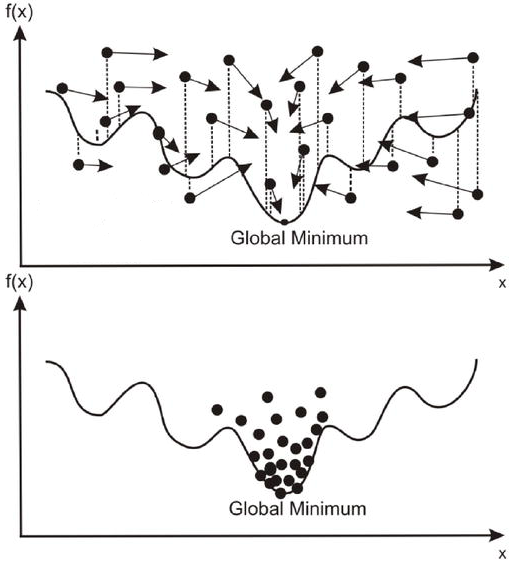}
\caption{Exemple de convergence d'une population de solutions \cite{buyuk21}}
    \label{fig:c2.7.2}
\end{figure}

    On dit qu’une métaheuristique converge, lorsque la majorité des individus constituant sa population aboutissent à la même solution (voir figure \ref{fig:c2.7.2}).
    
    Cette convergence est le résultat d’une comparaison avec différentes solutions trouvées de l’espace de recherche, puis l’élimination des solutions de mauvaise qualité, de sorte à ne garder à la fin que la solution optimale.
    
    Dans le cas idéal, la convergence vers la meilleure solution se fait après que l’algorithme ait exploré toutes les régions de l’espace de recherche afin de garantir de tomber sur la solution optimale. On appelle cette solution \textit{Optimum Global} ou \textit{Minima Global}. Toutefois, il se pourrait dans certains cas que l’algorithme converge vers une certaine région sans avoir exploré d’autres, et pourra donc rater la solution optimale en convergeant vers une solution de moindre qualité. On appelle cette solution \textit{Optimum Local} ou \textit{Minima Local}. 
    
    L’objectif d’une métaheuristique est donc de trouver un compromis entre les phases d’exploration et d’exploitation afin d’éviter de converger trop rapidement vers un optimum local, mais sans pour autant perdre trop de temps à explorer chaque région en détail. L’idée est d’éliminer les mauvaises régions rapidement afin de pouvoir se concentrer sur les régions les plus prometteuses.

    D’un point de vue mathématique, une solution $s^\ast$ est appelée optimum global si et seulement si elle respecte l'équation \ref{eq:2.0.5}.

    \begin{align}\label{eq:2.0.5}
        f(s^\ast) \leqslant f(s) \forall s	\in S.
    \end{align}
    
    Sachant que $S$ est l’espace de recherche et $ f:S \longleftarrow R^+_0 $ est la fonction objectif à minimiser.

\noindent

    \subsection{Les hyperparamètres}
    Ce sont les paramètres spéciaux que l'utilisateur doit définir manuellement pour réguler le comportement de l'algorithme. Ces hyperparamètres sont définis avant le début de l’exécution, souvent de manière expérimentale en essayant plusieurs combinaisons.
    
    Parmi les hyperparamètres les plus utilisés dans les métaheuristiques, nous retrouvons :

    \begin{itemize}
    \item La taille de la population : qui représente le nombre d’individus (ou nombre de solutions candidates).
    \item La taille de l’individu : nombre de variables constituant une solution.
    \item Nombre d'itérations : nombre maximal d’itérations avant l’arrêt de l’algorithme.
    \item Le taux de mutation : probabilité à laquelle une solution est modifiée.
    \item …etc.
    \end{itemize}

    \subsection{Les contraintes}

    Des contraintes peuvent être ajoutées à la modélisation d’un problème d’optimisation. Ces contraintes divisent l’espace de recherche en deux catégories : d’une part l’ensemble des solutions valides satisfaisant toutes les contraintes du problème, d’autre part l’ensemble des solutions non valides où au moins une contrainte est violée.
    
    Le moyen le plus simple pour intégrer une contrainte à la fonction objectif d’un problème est d’ajouter un facteur qui pénalise les solutions non conformes. Ceci prend généralement la forme d’une multiplication entre le nombre de contraintes violées et un paramètre de pénalité fixé manuellement par l’utilisateur. 
    
    \subsection{Les générations}

    Les métaheuristiques se basent sur un principe de répétition d’un ensemble d’opérations. Suivant l’analogie des algorithmes évolutionnaires, une \textit{itération} dans le contexte des métaheuristiques à base de population est souvent appelée \textit{génération} ou \textit{évolution}.
    
    Le principe est qu’à chaque génération, la population subit plusieurs transformations de sorte qu’elle devienne meilleure que ce qu'elle n'était durant la génération précédente. 
    
    Le meilleur individu de la dernière génération constitue donc la meilleure solution trouvée au cours du processus d’optimisation. Il est parfois appelé le \textit{champion}.

    \subsection{Les critères d'arrêt}
    Les métaheuristiques étant un processus itératif, des critères d’arrêts doivent être mis en place afin d’éviter le prolongement de l’exécution de l’algorithme indéfiniment.
    
    Les critères d’arrêts les plus populaires utilisés dans le domaine de l'optimisation combinatoire sont les suivants :
    \begin{itemize}

    \item Atteindre un nombre maximal d’itérations.
    \item Atteindre un nombre maximal de solutions évaluées en utilisant la fonction objectif.
    \item Atteindre un nombre maximal d’itérations où aucun changement n’a été mesuré sur les valeurs de la fonction objectif (\textit{early stopping}).
    \item Atteindre une certaine valeur de la fonction fitness.

    \end{itemize}
   
    \subsection{L'optimalité et la dominance}

\noindent
\begin{figure}[bht!]
    \setlength{\abovecaptionskip}{0.4cm} 
    \setlength{\belowcaptionskip}{-0.4cm} 
    \centering 
    \includegraphics[width=0.9\textwidth]{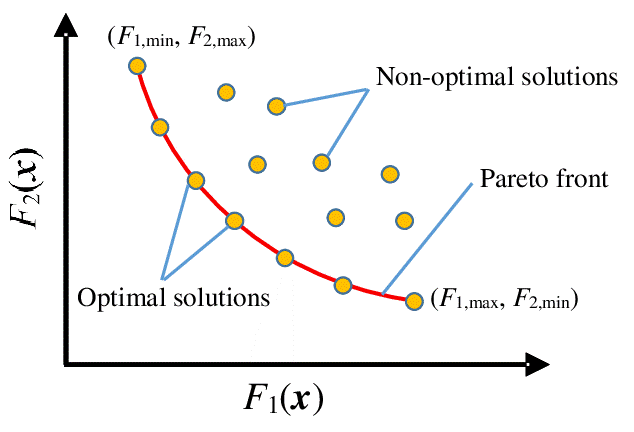}
\caption{Visualisation d'un front optimal (Pareto Front) séparant les solutions dominées des solutions non dominées \cite{mergos18}}
    \label{fig:c2.8}
\end{figure}

    Les métaheuristiques peuvent résoudre des problèmes d’optimisation ayant plusieurs objectifs contradictoires.
    
    Une stratégie est de réduire ces objectifs à un seul facteur d’optimisation en utilisant une somme pondérée. Une autre stratégie consiste à garder ces objectifs contradictoires et explorer plusieurs espaces de recherches en parallèle, selon le nombre d’objectifs à optimiser. 
    
    Lorsque l’algorithme optimise un problème mono-objectif, la meilleure solution est celle qui obtient la meilleure valeur de fitness.
    
    Dans le cas d’une optimisation multi-objectifs nous pouvons avoir deux cas de figure : soit une solution obtient la valeur optimale dans tous les espaces de recherches, nous disons donc que cette solution domine les autres solutions. Soit, nous nous retrouvons avec plusieurs solutions, chacun domine l’autre dans un objectif en particulier, mais pas dans l’autre. Ces solutions sont donc toutes les deux optimales et on les appelle solutions non dominées. L’ensemble des solutions non dominées est appelé Pareto-front. La figure \ref{fig:c2.8} illustre ceci d'une manière graphique.

\section{Structure de base d'une métaheuristique}
Malgré la diversité des métaheuristiques et leurs types, elles partagent toutes la même structure de base. L’algorithme de base d’une métaheuristique se découpe en 3 parties : 

    \subsection{La phase d'initialisation}
    Cette phase est exécutée une seule fois au début de l’algorithme, son but est de définir toutes les variables et les paramètres nécessaires pour son bon fonctionnement. Ceci implique généralement d'effectuer les actions suivantes : 

    \begin{itemize}
        \item Définir le problème : taille de la population, dimensions des individus, contraintes.
        \item Définir la fonction objectif et ses paramètres.
        \item Définir les autres hyperparamètres : nombre d’itérations, critère d’arrêt…etc.
        \item Générer la solution – ou la population de solutions – initiale.
        \item Evaluer les solutions initiales en utilisant la fonction objectif.

    \end{itemize}

    \subsection{Le corps de l'algorithme}
    Il s’agit d’exécuter les opérations constituant le cœur du processus d’optimisation. Cette phase est répétée jusqu’à ce que le critère d’arrêt soit atteint : 
     \begin{itemize}
        \item Exécuter l’ensemble des opérations constituant la logique de l’algorithme.
        \item Evaluer la qualité des solutions.
        \item Mémoriser la meilleure solution trouvée jusqu’ici.
        \item Répéter les opérations jusqu’à ce que le critère d’arrêt soit atteint.

    \end{itemize}
    \subsection{La phase finale}
    C’est la dernière phase qui sera exécutée lorsque le processus d’optimisation est terminé. Elle comprend généralement des actions d’aide à l’analyse des résultats :
   \begin{itemize}
   
        \item Sauvegarder les résultats : meilleure solution, qualité de la solution finale, temps d’exécution…etc.
       
        \item Tracer les graphes de performances, historique d’évolution de la fonction objectif…etc.

    \end{itemize}

\begin{figure}[pht]
    \setlength{\abovecaptionskip}{0.4cm} 
    \setlength{\belowcaptionskip}{-0.4cm} 
    \centering 
    \includegraphics[width=\textwidth]{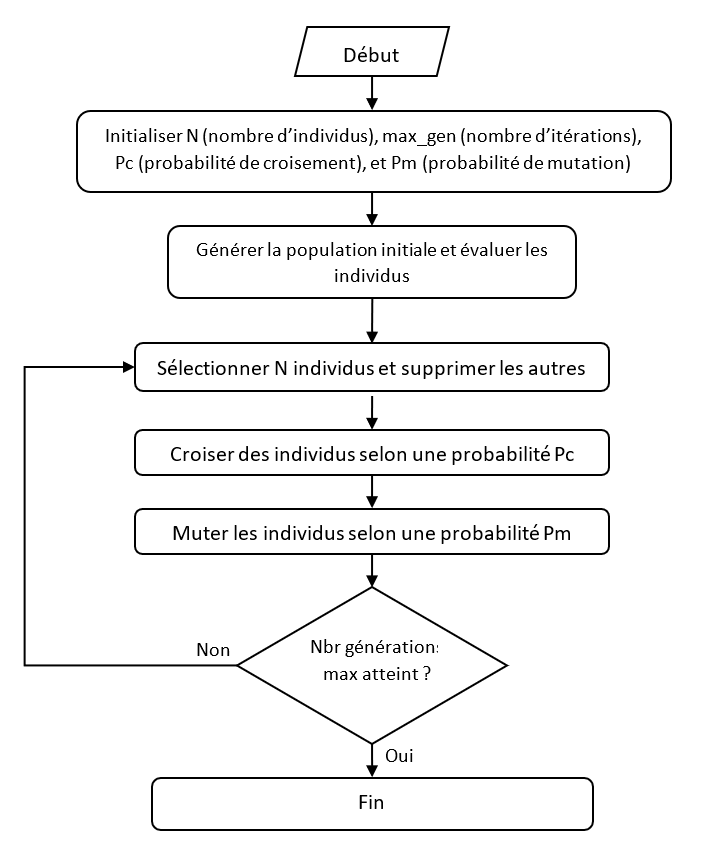}
\caption{Structure de base d'une métaheuristique}
    \label{fig:c2.9}
\end{figure}

    La figure \ref{fig:c2.9} schématise la structure de base d'une métaheuristique.

\noindent

\section{Fondements théoriques de métaheuristiques populaires}
Dans cette section, nous allons présenter les fondements théoriques et le principe de fonctionnement de quelques métaheuristiques populaires. Ces métaheuristiques seront utilisées dans les chapitres suivants lors des expériences de benchmarking et de résolution du problème d’exploration multirobots.

Les algorithmes seront présentés en suivant l’ordre chronologique de leur apparition.

    \subsection{Les algorithmes génétiques (GA)}
    Les Algorithmes Génétiques utilisent des concepts issus de la génétique et de la sélection naturelle. Ils ont été introduits par les travaux de John Holland \cite{holland76} et D. Goldberg \cite{goldberg89}.
    
    La population dans un algorithme génétique est constituée d’un certain nombre d’individus, eux-mêmes constitués d’un ensemble de gènes (regroupés en chromosomes). Un gène représente une variable spécifique au problème qu’on veut optimiser et est généralement codé par un nombre binaire ou réel.
    
    Les individus dans cet algorithme sont soumis aux 3 opérations suivantes : 
    \subsubsection{Sélection}
    Durant cette opération, les meilleurs candidats sont choisis pour servir de parents à la génération suivante. Ce processus est similaire au processus de sélection naturelle : les individus les plus adaptés à leur environnement sont conservés tandis que les moins adaptés meurent avant la reproduction.
    
    Toutefois, des stratégies sont parfois mises en place pour conserver certains individus de mauvaise qualité afin de diversifier la population et éviter une convergence trop rapide vers l’optimum local.
    
    L’adaptation d’un individu est calculée en fonction de sa valeur de fitness. L’opération de sélection nécessite donc d’évaluer l’ensemble des individus de la population à chaque itération afin de pouvoir les comparer, ce qui a un impact sur la complexité globale de l’algorithme. Pour une population de \emph{N} individus et \emph{M} itérations, la complexité sera égale à \emph{N} x \emph{M}.

\noindent
\begin{figure}[pht!]
    \setlength{\abovecaptionskip}{0.4cm} 
    \setlength{\belowcaptionskip}{-0.4cm} 
    \centering 
    \includegraphics[width=0.9\textwidth]{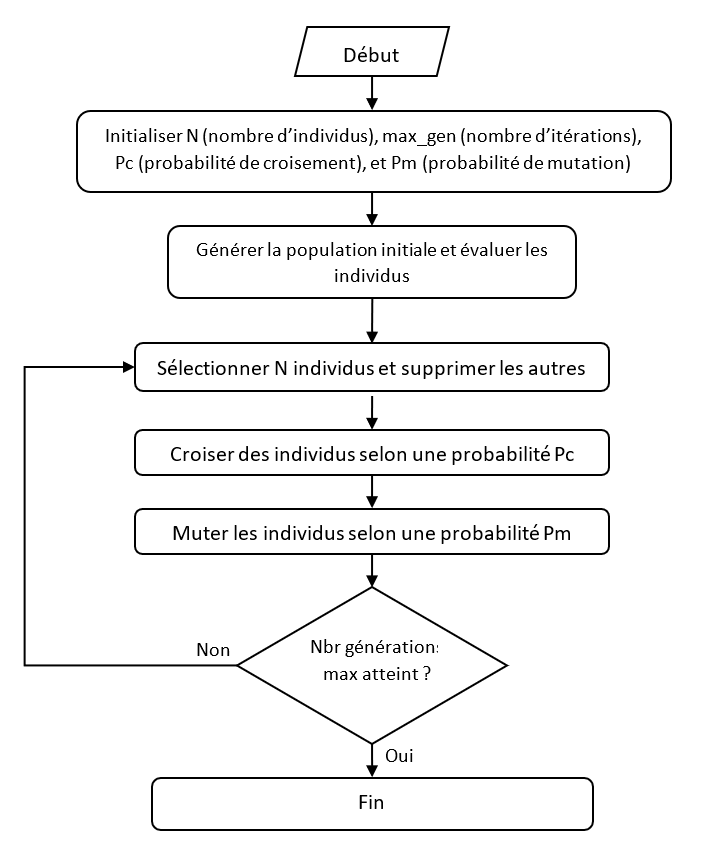}
\caption{Diagramme de l'algorithme génétique}
    \label{fig:c2.10}
\end{figure}
    
    \subsubsection{Croisement (crossover)}
    Cette opération est parfois appelée \textit{reproduction}, elle vise à combiner deux individus au hasard afin d’en produire d’autres. Pour ce faire, l’algorithme sélectionne quelques gènes du premier individu et les mélange avec des gènes du deuxième individu de sorte à créer un tout nouvel individu différent des deux autres. Cette opération est répétée plusieurs fois jusqu’à ce que le nombre de nouveaux individus créés (appelées \textit{enfants}) soit égal au nombre des individus déjà présent dans la population (\textit{parents}).
    
    Bien que la taille de la population augmente après l’exécution de cette opération, elle sera réduite à sa taille originale lors de la prochaine itération après avoir appliqué l’opérateur de sélection.
    
    Il est important à noter que l’opérateur de croisement n’est pas forcément appliqué systématiquement à tous les individus. Un facteur de probabilité est utilisé pour contrôler le taux de croisement dans la population. Ce facteur est souvent choisi dans un intervalle entre 0.7 à 1.0.

    \subsubsection{Mutation }
    Cette opération apporte des changements mineurs aux individus en modifiant aléatoirement la valeur de certains gènes. Ceci permet à l'algorithme d'examiner une plus grande variété de solutions potentielles et l'empêche de rester coincé dans des optimums locaux. Cependant, il ne faut pas que cette opération soit effectuée trop souvent pour éviter de tomber dans une recherche aléatoire. Elle est donc appliquée avec une probabilité relativement faible (souvent inférieure à 5\%).

    Le diagramme présenté dans la figure \ref{fig:c2.10} montre les étapes de l'algorithme génétique.

    \subsection{L'optimisation par Essaim de Particules (PSO)}
    Cette méthode a été publiée en 1995 \cite{kennedy95}. Elle se base sur le comportement social des essaims observés dans la nature.
    
    Dans cet algorithme, une population de particules représentant des solutions potentielles est générée aléatoirement. Ces particules se déplacent dans l’espace de recherche en modifiant leurs emplacements actuels en fonction de leurs emplacements précédents et de ceux de leurs voisins.
    
    L’équation qui gouverne le déplacement d’une particule met à jour à chaque itération deux informations : la vitesse de la particule et sa position (voir les équations \ref{eq:2.1.1} et \ref{eq:2.1.2}). Ces deux informations sont calculées en fonction de 3 composants :
    
    \begin{itemize}
        \item Composant social : attire la particule à se diriger vers la position de la meilleure solution connue par la population (dénotée $g*$).
        \item Composant cognitif : attire la particule à se diriger vers la position de la meilleure solution qu'elle a visité dans le passé (dénotée $x_i*$).
        \item Composant inertiel : pousse la particule à garder sa direction courante.

    \end{itemize}
 
L’utilisateur attribue au début de l’algorithme un facteur de pondération pour chaque composant pour contrôler le degré d’influence de ces composants sur le mouvement des particules. Plus le facteur du composant social est grand, plus l’algorithme s’oriente vers la diversification de la population. Plus le facteur du composant cognitif est grand, l’algorithme s’oriente vers une stratégie d’intensification.

\begin{align} \label{eq:2.1.1}
\text{Vitesse: }
v_i^{t+1} &= \omega v_i^t + c_1r_1(p_i^{t\_best} x_i^t) + c_2r_2(g^{t\_best}  x_i^t)
\end{align}
\begin{align} \label{eq:2.1.2}
\text{Position: } 
x_i^{t+1} &= x_i^t + v_i^{t+1}
\end{align}

Où :
\begin{itemize}
    \item $\omega$ est le poids d’inertie
    \item $r_{1}$ et $r_{2}$ sont des vecteurs aléatoires distribués uniformément dans l’interval [0,1] 
    \item $c_{1}$ et $c_{2}$ sont des hyperparamètres appelés facteurs d’accélération, ou facteur cognitif et facteur social respectivement.
\end{itemize}


    \subsection{L'otimisation par Colonie de Fourmies (ACO)}
   
    Cet algorithme a été introduit par Colorni et Dorigo en 1991 \cite{colorni91}. Il simule la façon dont les fourmis naviguent pour trouver la source de nourriture la plus proche de leur nid.
    
    Étant donné que les fourmis dans le monde réel dégagent des odeurs (ou phéromones) volatiles lorsqu'elles trouvent une source de nourriture, et que les autres fourmis ont tendance à suivre les chemins qui contiennent ces phéromones. Les itinéraires les plus courts entre la source de nourriture et le nid seront favorisés puisque la quantité de phéromones non volatilisée sera plus grande en raison de la courte distance. Ces chemins optimaux seront donc de plus en plus empruntés par les fourmis. Ce qui résulte en un effet de renforcement.

    De la même manière, les agents de l'ACO utilisent des phéromones virtuelles pour marquer les meilleures solutions trouvées. Au début de l’algorithme, les fourmis se déplacent de manière aléatoire tout en mettant à jour une matrice de phéromones. Cette matrice contient la quantité de phéromones pour chaque solution trouvée, qui est incrémentée avec une valeur proportionnelle à la qualité de la solution.

    Cette valeur décroit avec le temps suivant l’équation \ref{eq:2.3} afin de simuler l’effet d’évaporation :

 \begin{align} \label{eq:2.3}
      \tau_{i}\longleftarrow \left( 1-\rho \right)\tau_{i}
\end{align}

Où $\rho$ est une constante d’évaporation à définir par l’utilisateur et $\tau_{i}$ est la quantité de phéromones de la solution i

\noindent
\begin{figure}[pht!]
    \setlength{\abovecaptionskip}{0.4cm} 
    \setlength{\belowcaptionskip}{-0.4cm} 
    \centering 
    \includegraphics[width=0.9\textwidth]{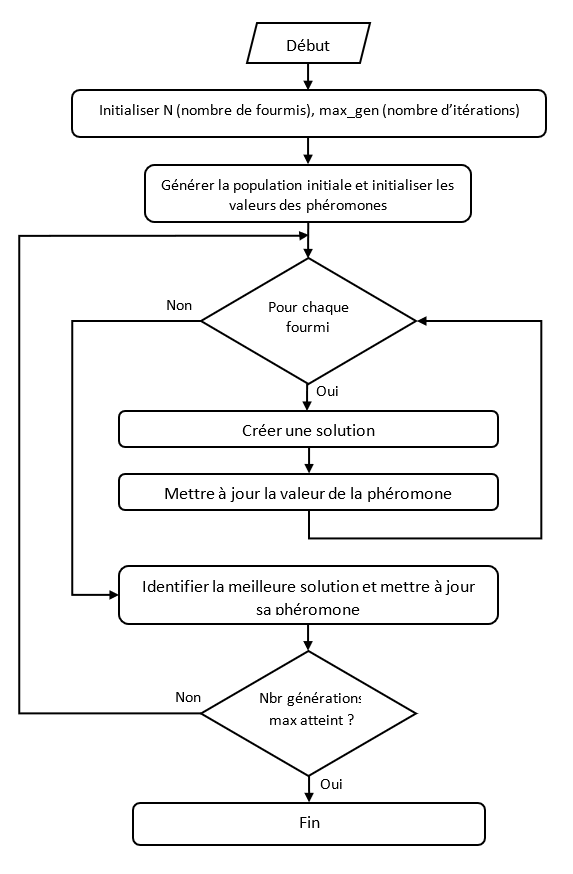}
\caption{Pseudo-code de l'algorithme ACO}
    \label{fig:c2.12}
\end{figure}
     

Après un certain nombre d’itérations, les solutions les plus visitées auront une quantité de phéromones plus grande, ce qui augmente la probabilité que les fourmis les choisissent.
La baisse régulière de la quantité de phéromones permet aux fourmis d'explorer des solutions différentes pendant le processus de recherche. C’est un mécanisme de diversification sans lequel l’algorithme risque de converger rapidement vers un optimum local. Un autre mécanisme est de garder en mémoire la liste des solutions que la fourmi a déjà visitées, de sorte à les éliminer dans ses futurs mouvements.

    

\section{Les Fondements théoriques de l'Algorithme d'Optimisation des Papillons (BOA)}

\noindent
\begin{figure}[pt!]
    \setlength{\abovecaptionskip}{0.4cm} 
    \setlength{\belowcaptionskip}{-0.4cm} 
    \centering 
    \includegraphics[width=0.9\textwidth]{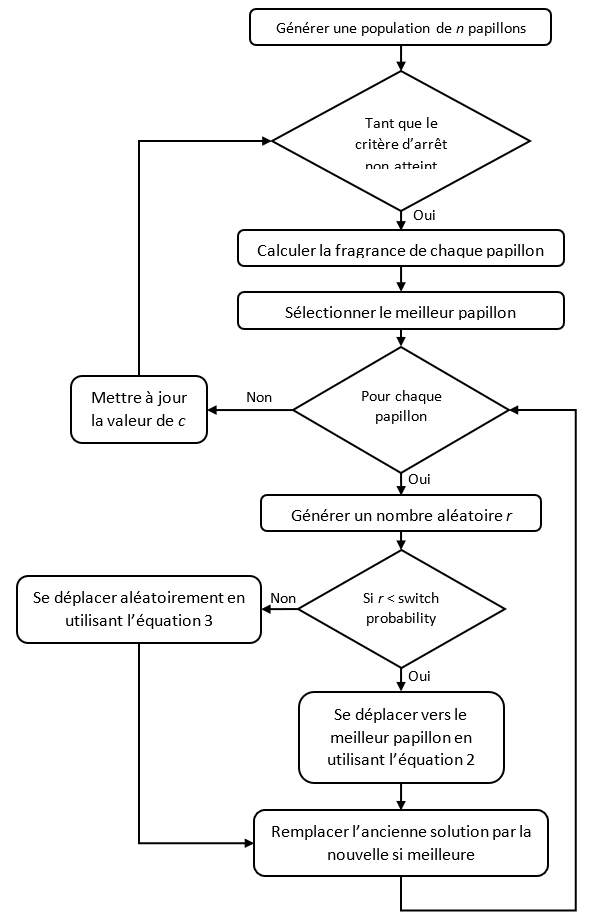}
\caption{Diagramme de l'algorithme BOA}
    \label{fig:c2.13}
\end{figure}

\emph{Butterfly Optimization Algorithm} (BOA) est une métaheuristique à base de populations s’inspirant du comportement des papillons, elle a été proposée par \emph{S. Arora} et \emph{S. Singh} \cite{arora19} en tant qu’algorithme d’optimisation globale.

Au départ, l’algorithme génère une population de solutions aléatoires (les papillons), puis les modifie à travers plusieurs itérations jusqu’à ce qu’un critère d’arrêt soit atteint.

Dans la nature, les papillons utilisent l’odorat pour trouver des sources de nourriture et des partenaires de reproduction, pour y arriver, ils utilisent des cellules dans leurs corps pour percevoir les odeurs (\textit{fragrance}) et mesurer leurs intensités \cite{arora19}. Plus la fragrance est intense, plus le papillon est attiré vers la source de cette odeur.

L’algorithme BOA modélise ce comportement en calculant une valeur de fragrance proportionnelle à la valeur de fitness de l’individu. Plus la valeur de fitness d’un papillon est de meilleure qualité, plus sa fragrance est grande, et plus les autres papillons y sont attirés. Toutefois, la fragrance émise par un papillon dans la nature est souvent altérée par les conditions météorologiques. Deux paramètres sont donc introduits dans l’algorithme pour simuler ce phénomène et modifier l’intensité de la fragrance qui sera captée par les autres papillons. 

Ces paramètres sont les suivants :
\begin{itemize}
    \item \textit{Sensor modality (c)} : Un facteur de multiplication contrôlant la proportion de la fragrance qui sera perçue.
    \item \textit{Power exponent (a)} : Un exposant qui contrôle la puissance à laquelle l’intensité originale de la fragrance est amplifiée.
    
\end{itemize}


L’équation \ref{eq:2.4} décrit la règle de calcul de la fragrance : 
\begin{align}\label{eq:2.4}
    F=c * I^a 
\end{align}
Où : 	
\begin{itemize}
    \item c = sensor modality ;
    \item a = power exponent ; et
    \item I = intensité originale de la fragrance, qui est égale à la valeur de fitness du papillon. 
\end{itemize}

Une fois les fragrances calculées, les papillons vont se déplacer graduellement vers le meilleur papillon qui représente la meilleure solution trouvée jusqu’à présent dans l’espace de recherche. Cette étape est appelée « recherche globale » \cite{arora19}

Afin d’éviter une convergence prématurée, une phase de recherche locale a été introduite par les auteurs de l’algorithme pendant laquelle les papillons se déplacent aléatoirement dans la région où ils se trouvent.

À chaque itération, l’algorithme exécutera soit la phase de recherche globale, soit celle de la recherche locale, selon la valeur d’une probabilité d’alternance (\textit{switching probability}).

Les équations suivantes décrivent respectivement comment les individus sont mis à jour pendant les phases de recherche globale et locale :

\begin{align}\label{eq:2.5}
x_{i}^{t+1}=x_{i}^{t}+\left ( r^2 * g^* - x_{i}^{t}\right )* f_{i} 
\end{align}
\begin{align}\label{eq:2.6}
x_{i}^{t+1}=x_{i}^{t}+\left ( r^2 * x_{j}^{t} - x_{k}^{t}\right )* f_{i} 
\end{align}


Où : 	

\begin{itemize}
    \item $x_{i}$ est le papillon n° $i$ ;
    \item $r$ est un nombre aléatoire appartenant à l’interval [0, 1] ;
    \item $g*$ est le meilleure papillon dans la population ;
    \item $f_{i}$ est la fragrance du papillon n° $i$ ;
    \item $x_{j}$ et $x_{k}$ sont deux papillons sélectionnés aléatoirement parmi la population.
\end{itemize}

En exécutant l’équation \ref{eq:2.5} plusieurs fois à travers les itérations, les papillons convergeront vers l’individu ayant la meilleure valeur de fitness qui représente la meilleure solution connue. Toutefois, une convergence trop rapide pourrait piéger les individus dans le voisinage d’un optimum local alors qu’une meilleure solution se trouve ailleurs dans l’espace de recherche. L’équation \ref{eq:2.6} évite à ce problème en poussant les papillons à se déplacer aléatoirement pour explorer de nouvelles solutions. 

À chaque itération, les valeurs de fitness des papillons augmenteront en qualité, et leurs fragrances seront de plus en plus grandes. Afin d’éviter une convergence trop rapide engendrée par une trop grande augmentation de la fragrance, les auteurs de l’algorithme ont introduit une règle pour diminuer le facteur de multiplication \textit{c} (\textit{sensor modality}) à la fin de chaque itération \cite{arora16}. L’ajout de cette règle – décrite par l’équation \ref{eq:2.7} – leur a permis d’améliorer les résultats de l’algorithme.

\begin{align}\label{eq:2.7}
c^{t+1}=c^{t}+\left ( \frac{0.025}{c^{t}*nbr\_iterations}\right )  
\end{align}

\begin{figure}[ht!]
    \setlength{\abovecaptionskip}{0.4cm} 
    \setlength{\belowcaptionskip}{-0.4cm} 
    \centering 
    \includegraphics[width=0.99\textwidth]{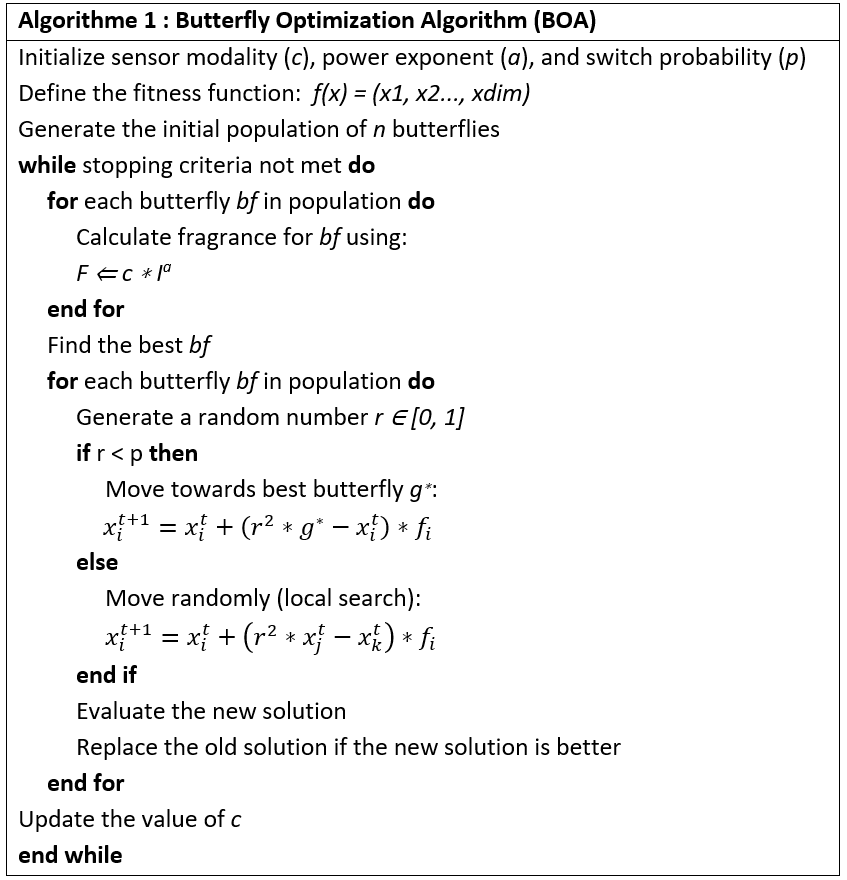}
\caption{Pseudo-code de l'algorithme BOA}
    \label{fig:c2.14}
\end{figure}

La simplicité de l’approche produit un pseudo-code facile à comprendre, ce qui constitue un avantage considérable lors de l’implémentation et débogage de l’algorithme (voir la figure \ref{fig:c2.14}). Par ailleurs, l’utilisation de formules simples et rapides à calculer permet de l’utiliser dans des machines à faible puissance de calcul tels que les petits robots ou les ordinateurs d’ancienne génération.

Le diagramme présenté dans la figure \ref{fig:c2.13} résume toutes les étapes de l’algorithme.

\noindent

\section{Les variantes de l'algorithme BOA}
\label{section:BOA_variants}

Bien que le pseudo-code tel que décrit dans la section précédente a été publié en 2019 en tant que version officielle du \emph{Butterfly Optimization Algorithm} \cite{arora19}, une version préliminaire avait déjà été présentée dans un article de conférence en 2015 \cite{arora15} avec un pseudo-code assez similaire du \emph{Flower Pollination Algorithm} (FPA) \cite{yang12} utilisant une distribution Lèvy pour les règles de mises à jour des positions des papillons. Dans les articles ultérieurs, les auteurs du BOA ont changé les équations et ont ajouté la règle de mise à jour automatique du paramètre \textit{c} (\textit{sensor modality}), ce qui a permis d’améliorer le taux de convergence de l’algorithme \cite{arora16} \cite{arora19}. Dans un autre article paru la même année, ils ont proposé une variante binaire de la méthode pour résoudre le problème de sélection d’attributs \cite{arora19b}.

D’autres auteurs ont proposé l’hybridation de l’algorithme avec différentes méthodes. Les auteurs de \cite{jalali19} ont utilisé un modèle de Perceptron Multicouches \emph{(MLP : Multi-Layer Perceptron)} hybridé avec BOA pour résoudre le problème de classification supervisé. Les résultats des expériences ont démontré que l’introduction du MLP a permis de garder une bonne balance entre les phases d’exploration et d’exploitation de l’algorithme BOA et améliorer les résultats. Les auteurs de \cite{zhang20} ont proposé une approche hybride entre PSO \emph{(Particle Swarm Optimization)} et BOA couplée avec la théorie des systèmes chaotiques afin d’obtenir de meilleurs résultats pour les problèmes à grande dimension. Ils ont aussi proposé une règle de mise à jour non-linéaire des paramètres qui a amélioré les résultats par rapport à l’utilisation de la règle classique linéaire de l’équation \ref{eq:2.7}. Les auteurs de \cite{wang21} ont utilisé une hybridation basée sur le mécanisme de mutualisme avec le \emph{Flower Pollination Algorithm} (FPA). L’approche a donné de bons résultats mais nécessite un long temps d’exécution. Le principe de mutualisme a aussi été utilisé par \cite{sharma21} pour une hybridation avec l’algorithme \emph{Symbiosis Organisms Search}. Les auteurs de \cite{zounemat21} ont combiné l’algorithme BOA avec la méthode ANFIS \emph{(Adaptive Neuro-Fuzzy Inference System)} et ont obtenu de meilleures performances par rapport à l’approche ANFIS classique ou de l’approche hybride entre ANFIS et \emph{FireFly Algorithm}.

\noindent
\begin{figure}[hb!]
    \setlength{\abovecaptionskip}{0.4cm} 
    \setlength{\belowcaptionskip}{-0.4cm} 
    \centering 
    \includegraphics[width=0.95\textwidth]{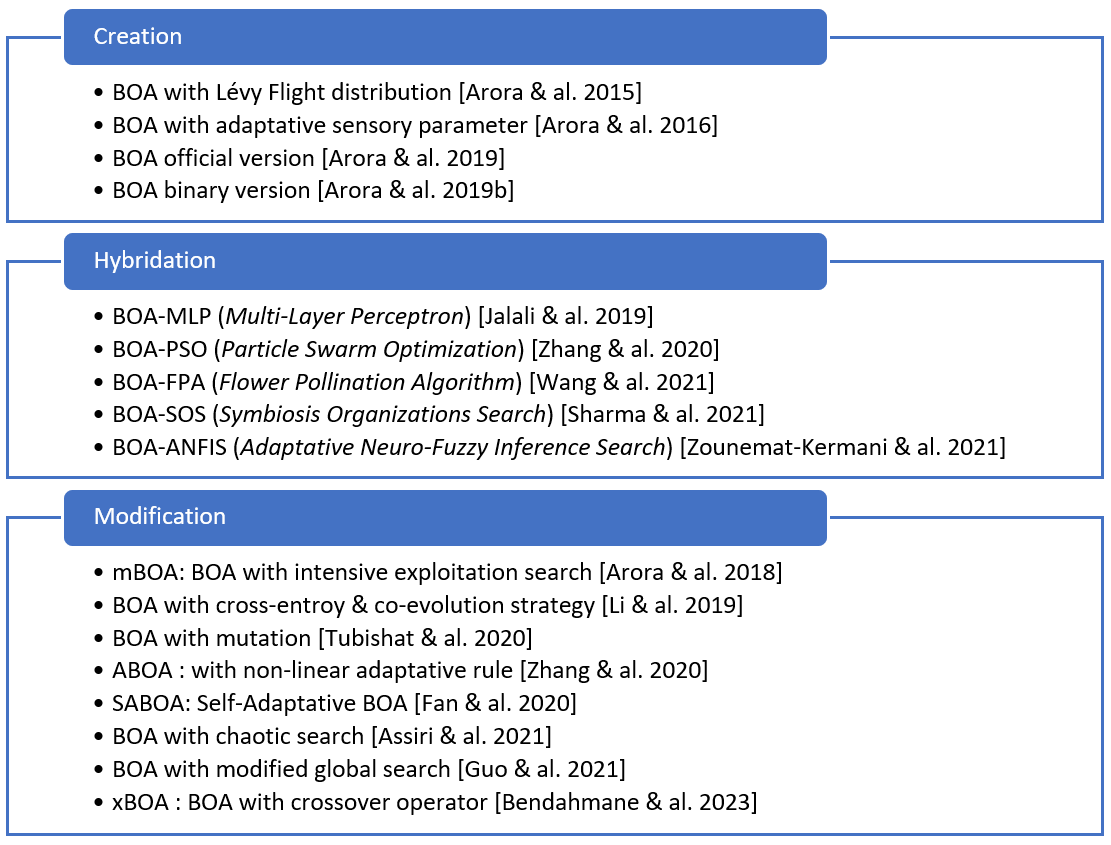}
\caption{Résumé de l'état de l'art de l'algorithme BOA et ses différentes variantes}
    \label{fig:c2.15}
\end{figure}

Une autre direction pour l’amélioration de la méthode consiste à modifier la logique de l’algorithme. Dans cette perspective, les auteurs du BOA ont proposé une variante appelée mBOA \cite{arora18} qui introduit une nouvelle étape d’exploitation intensive juste après les phases de recherches locale et globale de l’algorithme. Le but de cette nouvelle étape est d’éviter à la population de solutions d’être bloquée dans un optimum local. Les résultats des expériences ont montré une convergence plus rapide par rapport à la version classique du BOA. Les auteurs de \cite{tubishat20} ont inclus une étape de recherche locale basée sur l’opération de mutation afin d’améliorer les performances de la méthode, ce qui a donné de meilleurs résultats dans 15 parmi les 20 corpus de tests utilisés par les auteurs, mais a nécessité un temps d’exécution plus long. Les auteurs de \cite{assiri21} ont utilisé la théorie des systèmes chaotiques comme phase de recherche locale pour augmenter les traits d’exploitation de l’algorithme. Ils l’ont utilisé pour résoudre le problème de sélection d’attributs. Les auteurs de \cite{li19} ont utilisé la méthode de \emph{cross-entroy} et une technique de co-évolution pour résoudre 3 problèmes classiques d’engineering. Les auteurs de \cite{fan20} ont opté pour la réduction du nombre de paramètres en remplaçant la formule de calcul de la fragrance par une nouvelle règle autoadaptative qui ne nécessite plus l’utilisation des deux paramètres c (\emph{sensory modality}) et a (\emph{power exponent}). Cette nouvelle variante est appelée SABOA (\textit{Self-Adaptative BOA}). D’un autre côté, les auteurs de \cite{guo21} ont modifié l’équation de la recherche globale dans le but d’améliorer la convergence de l’algorithme, couplé avec une stratégie de réinitialisation périodique de la population afin d’éviter le blocage dans un minima local.

Les variantes citées ci-haut ont montré des résultats prometteurs pour la résolution des problèmes d’optimisation globaux à grandes dimensions, ce qui constituait une des faiblesses de l’approche BOA classique. Une autre faiblesse consiste en la lenteur de la convergence de l’algorithme causée par le faible taux de diversité des solutions dans la population.

Pour résoudre ces deux problèmes, nous proposons de modifier la méthode en introduisant l’opérateur de croisement et modifiant la stratégie de mouvement des papillons dans les phases de recherches globale et locale. Ces changements ont permis l’introduction d’une nouvelle variante de l’algorithme appelée xBOA (\emph{crossover BOA}) \cite{bendahmane22}.

La figure \ref{fig:c2.15} résume l’état de l’art des variantes de BOA citées ci-haut.

\section{Amélioration de l'Algorithme d'Optimisation des Papillons en utilisation l'opérateur de croisement (xBOA)}

\noindent
\begin{figure}[b!]
    \setlength{\abovecaptionskip}{0.4cm} 
    \setlength{\belowcaptionskip}{-0.4cm} 
    \centering 
    \begin{subfigure}[b]{\textwidth}
         \centering
         \includegraphics[width=0.5\textwidth]{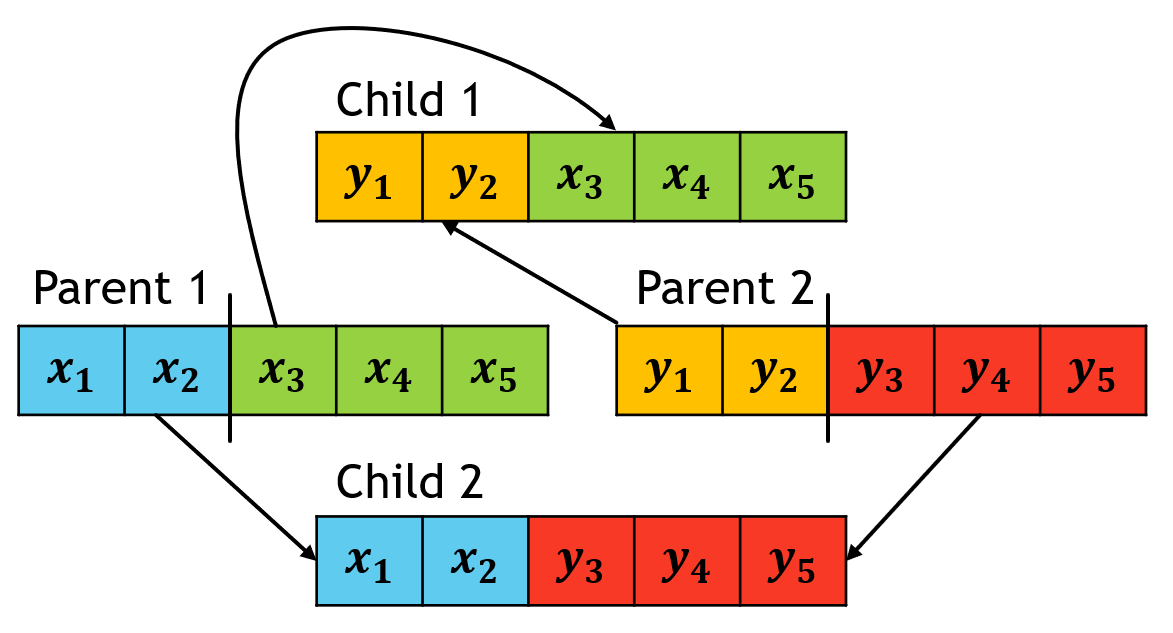}
         \vspace{10mm}%
     \end{subfigure}
     \begin{subfigure}[b]{\textwidth}
         \centering
         \includegraphics[width=0.55\textwidth]{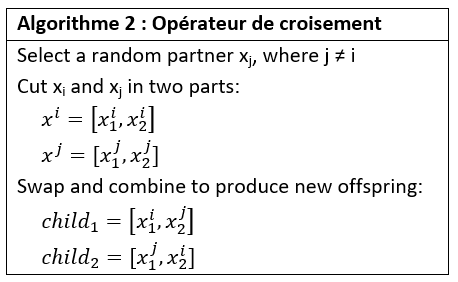}
     \end{subfigure}
\caption{Exemple et pseudo-code de l'opérateur de croisement}
    \label{fig:c2.16}
\end{figure}

L’équation \ref{eq:2.5} déplace tous les papillons vers la meilleure solution de la population, ce qui ignore les autres solutions qui ont la même valeur de fitness ou qui ont le potentiel de devenir de meilleures solutions après quelques itérations. Afin de dépasser cette limitation, nous proposons de modifier l’algorithme BOA en remplaçant cette équation avec l’opérateur de croisement durant la phase de recherche globale.

L’opérateur de croisement a été introduit dans l’Algorithme Génétique \cite{goldberg89}. Il consiste à combiner deux individus parents pour créer de nouveaux individus appelés enfants (ou \emph{offsprings} en anglais). L’idée de base inspirée de la nature simule la manière dont les enfants héritent une partie des caractéristiques de chaque parent.

Plusieurs stratégies de combinaisons ont été proposées dans la littérature \cite{pavai16}. Pour des raisons de simplicité, nous allons utiliser la stratégie de croisement à un point (\emph{Single-point Crossover}). Elle consiste à diviser le vecteur de données du premier parent en deux sous-vecteurs, puis les permuter avec les sous-vecteurs du deuxième parent afin de produire deux nouveaux individus.

L’exemple présenté dans la figure \ref{fig:c2.16} montre le résultat de cette opération pour deux individus de taille 5 ainsi que le pseudo-code de cette opération.

Vu que l’insertion de nouveaux individus dans la population peut rapidement faire croître sa taille, nous avons choisi de remplacer un individu parent par un de ses enfants ; le meilleur enfant est choisi dans ce cas. Ceci nous permet de garder une taille de population fixe pendant toute la durée du processus d’optimisation.

\noindent
\begin{figure}[ht!]
    \setlength{\abovecaptionskip}{0.4cm} 
    \setlength{\belowcaptionskip}{-0.4cm} 
    \centering 
    \includegraphics[width=0.9\textwidth]{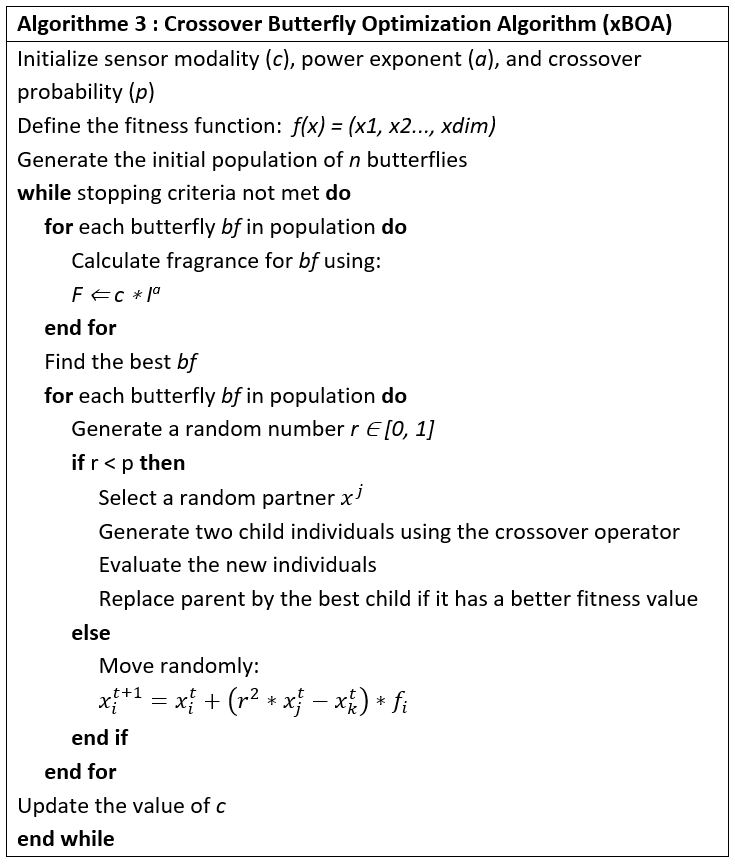}
\caption{Pseudo-code de l'algorithme xBOA}
    \label{fig:c2.17}
\end{figure}

En utilisant l’opérateur de croisement, nous encourageons les papillons à se déplacer vers plusieurs potentielles solutions au lieu de converger tous vers la meilleure solution connue. Ceci crée une diversité dans la population et permet à l’algorithme d’échapper au piège de converger prématurément vers un minima local. En d’autres termes, la population de papillons va inspecter plusieurs régions de l’espace de recherche simultanément afin de trouver la solution globale plus rapidement.

L’utilisation de l’opérateur de croisement à chaque itération pourrait toutefois déséquilibrer la balance entre les propriétés d’exploration et d’exploitation de l’algorithme xBOA, nous utilisons donc un paramètre qui déterminera à quelle fréquence cet opérateur est utilisé. Ce paramètre est appelé probabilité de croisement (\emph{crossover probability}). Il remplacera le paramètre \textit{Switch Probability} utilisé dans l’algorithme BOA original.

Une autre différence entre les deux algorithmes se situe au niveau de la recherche locale. En effet, dans xBOA l’équation \ref{eq:2.6} est utilisée même si elle engendre une dégradation de la qualité des solutions, contrairement à l’algorithme original. Ceci semble contre-productif, néanmoins cela permet à l’algorithme d’augmenter la diversité des solutions en leur permettant de se déplacer aléatoirement dans l’espace de recherche et d’explorer de nouvelles régions qui pourraient cacher la solution globale. La stratégie est d’accepter une perte en qualité à court terme pour pouvoir découvrir des solutions encore meilleures que celles trouvées jusqu’ici.

Les résultats des expériences effectuées durant notre thèse ont montré que les modifications proposées ont permis à xBOA de trouver la solution globale plus rapidement que l’algorithme BOA original, et d’être plus robuste aux minima locaux.

\begin{figure}[ht!]
    \setlength{\abovecaptionskip}{0.4cm} 
    \setlength{\belowcaptionskip}{-0.4cm} 
    \centering 
    \includegraphics[width=\textwidth]{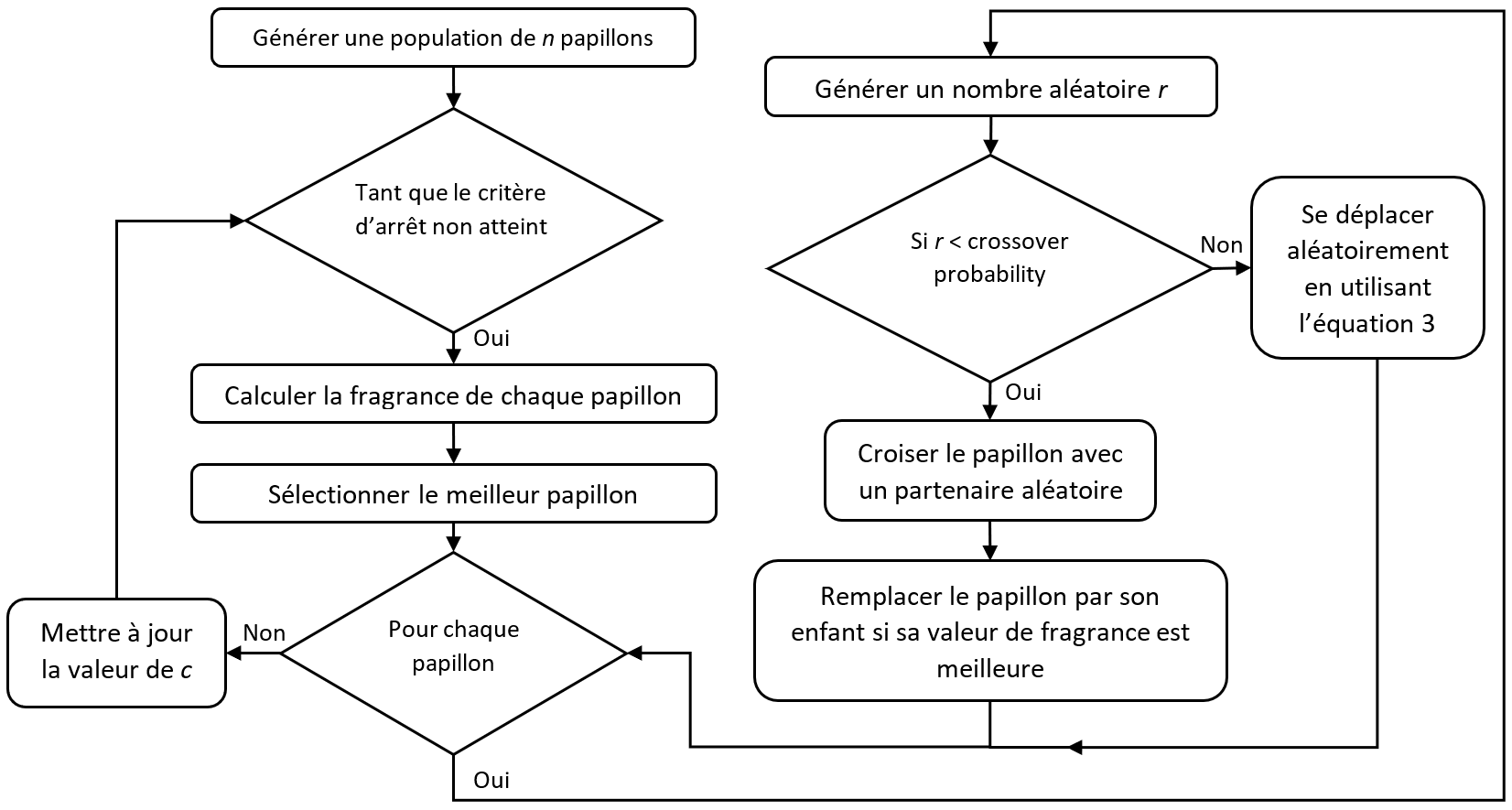}
\caption{Diagramme de l'algorithme xBOA}
    \label{fig:c2.19}
\end{figure}

Le diagramme de la figure \ref{fig:c2.19} décrit toutes les opérations de l'algorithme xBOA, son pseudo-code est présenté dans la figure \ref{fig:c2.17}.

\noindent

\section{Conclusion}

Nous avons présenté dans ce chapitre le principe de fonctionnement des métaheuristiques et leurs mécanismes internes. Nous avons également présenté les fondements théoriques de plusieurs métaheuristiques populaires utilisées dans la littérature pour résoudre divers problèmes d'optimisation globale, dont les problèmes de robotique.

Ce chapitre a aussi présenté les fondements mathématiques de la méthode BOA (\textit{Butterfly Optimization Algorithm}) ainsi que ses avantages et limitations. Nous avons proposé des modifications visant à l'améliorer ce qui a donné lieu à une nouvelle variante de l'algorithme que nous avons appelé xBOA (\textit{crossover Butterfly Optimization Algorithm}).

Cette variante se base sur l'intégration de l'opérateur de croisement durant la recherche globale afin de lui permettre de diversifier les solutions et échapper aux optimums locaux. Elle intègre aussi une modification dans la stratégie de la recherche locale pour éviter de converger trop rapidement vers un optimum local.

Ce chapitre conclut la partie théorique de notre thèse. Les deux autres chapitres restants seront consacrés à l'étude expérimentale.

    \part{Etude expérimentale}
    \chapter{Méthodologie, modélisation et paramétrage}

\startcontents[chapters]
\printmyminitoc{
}

\section{Introduction}
Nous allons présenter dans ce chapitre la méthodologie utilisée, ainsi que la modélisation de la problématique d’exploration d’une zone inconnue avec des contraintes d’énergie. Nous présenterons également les outils, l’environnement de test, ainsi que les critères de performance utilisés pour la validation de l’approche proposée.
Étant donné la difficulté rencontrée pour la comparaison de notre approche avec des cas d’études similaires, nous présenterons un nouvel outil créé afin de faciliter la comparaison et l’évaluation des algorithmes de navigation, d’exploration et de planification de trajectoires pour les robots mobiles, ainsi que les métaheuristiques appliquées au domaine de la robotique.

Cette plateforme de benchmarking a été créée d’une telle manière à nécessiter peu de ressources de calcul, et être facile à utiliser et à automatiser afin d’offrir aux chercheurs du domaine un moyen de prototypage rapide et une suite d’expérimentation unifiée. Elle fait partie des principales contributions de cette thèse et sera accessible en open source pour l’utilisation générale de la communauté scientifique.

\section{PyRoboticsLab: un nouvel outil de simulation et de benchmarking}
Le teste et l'évaluation comparative de nouveaux algorithmes constitue une étape importante de la validation des idées de recherche. 
Étant donné que les expériences réelles en robotique exigent des investissements en temps et en équipements, de nombreux chercheurs préfèrent utiliser des environnements virtuels pour tester et valider leurs algorithmes avant de les déployer sur des robots réels.
Même si les simulateurs de robotique ont été largement critiqués au début des années 90 \cite{jakobi95}, nous pouvons aujourd'hui identifier de nombreux avantages à les utiliser :

\begin{itemize}
\item Expérimentation plus rapide, en accélérant l'horloge de simulation par rapport à l'horloge du monde réel.
\item Automatisation des expériences à l’aide de scripts
\item Mise en pause, enregistrement et reproduction d'une expérience particulière pour une analyse approfondie.
\item Réduction des coûts en évitant l'endommagement du matériel en cas d’erreurs de programmation.
\item Évitement des problèmes techniques liés aux défaillances matérielles.
\item Expérimentation de plusieurs configurations matérielles sans devoir acheter des pièces de rechange ou de modifier la partie électronique du robot.
\item Évitement des inconvénients liés à la limitation des batteries et au temps de recharge pendant les expériences longues et/ou répétitives.
\item Éliminer les artefacts causés par les interférences et la faible qualité du matériel.
\item Accélération du processus de validation et de débogage des logiciels, ce qui réduit les délais de livraison.
\end{itemize}

Cette section présente PyRoboticsLab : un nouvel outil de simulation dédié au benchmarking des algorithmes de navigation, de planification de chemins, ainsi que l'exploration de zones en utilisant un ou plusieurs robots. Ce simulateur est disponible en open source sur le lien cité en bas de page\footnote{https://github.com/amineHorseman/PyRoboticsLab}.

\begin{figure}[b!]
    \centering    
    \vspace{0.3cm}
    \includegraphics[width=0.9\textwidth]{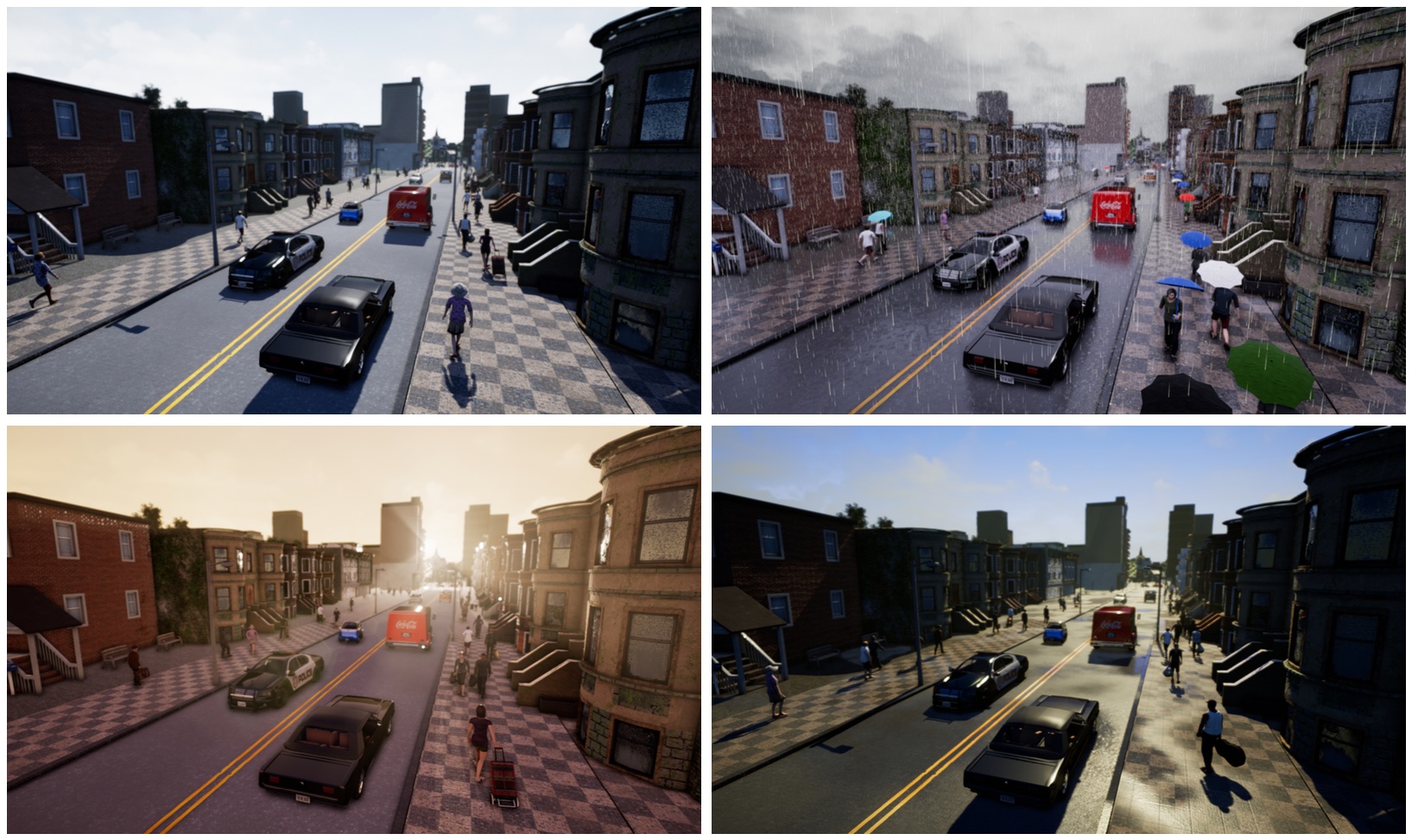}
    \caption{Exemple d'un simulateur moderne basé sur un moteur de jeux vidéos pour l'entraînement des voitures autonomes \cite{dosovitskiy17}}
    \label{fig:0.2}
\end{figure}

\subsection{Motivations pour la création d’une nouvelle plateforme de simulation}
    
De nombreux simulateurs modernes ont été introduits par la communauté robotique ces dernières années \cite{rohmer13} \cite{ferigo20} \cite{dosovitskiy17} \cite{koenig04}. Ces simulateurs offrent un grand degré de fidélité au monde réel qui leur permet d’entrainer des algorithmes d’intelligence artificielle avec une grande précision avant de les déployer en production \cite{dosovitskiy17}.
    
Toutefois, pour atteindre ce degré élevé de fidélité, il faut effectuer des calculs complexes, ce qui entraine des charges élevées en termes de mémoire et de puissance du processeur, même pour les petites expériences. Ceci est un facteur de frein pour la portabilité et le passage à l’échelle des applications.

Certains chercheurs tentent de surpasser ces limites en utilisant des services Cloud \cite{yin21}. Néanmoins, cette mise à l’échelle implique des frais supplémentaires et affecte négativement la reproductibilité de l'ensemble du projet.

D’autre part, les simulateurs modernes ont tendance à être de plus en plus polyvalents, ce qui rend leurs bases de code difficiles à comprendre et à maintenir. De nombreux projets open source souffrent de la difficulté de trouver de nouveaux contributeurs en raison de leur courbe d'apprentissage croissante \cite{afzal20}. Cela crée une barrière psychologique pour les étudiants et les jeunes développeurs, tout en décourageant les chercheurs trop occupés à les utiliser pour prototyper de nouvelles idées \cite{afzal20}. Souvent, ces chercheurs finissent par mettre en œuvre des environnements simplifiés et personnalisés pour leurs expériences afin d'obtenir des résultats rapides, ce qui rend difficile la comparaison de leurs algorithmes avec d'autres approches puisque chaque article utilise des données, une structure d'environnement ou des critères d'évaluation différents.

Afin de standardiser l'évaluation des algorithmes dans certaines problématiques de robotique, nous avons développé un nouveau simulateur spécialement conçu pour évaluer les performances des algorithmes de navigation, de planification de chemins, d'exploration, et d’optimisation dans des environnements de type grille.

Notre simulateur est dédié à la simulation d'un type spécifique de scénarios et ne vise pas à remplacer les simulateurs modernes polyvalents. Il nécessite moins de ressources CPU et de mémoire ; peut simuler une centaine de robots dans un ordinateur portable ordinaire sans carte graphique ; et est conçu pour être facile à utiliser et à porter vers différents systèmes d’exploitation, ce qui convient plus à un usage éducatif ou pour accélérer les expériences de recherche dans les axes cités ci-haut.

\subsection{Les objectifs de conception}
\label{section:simulator_goals}

Avant de pouvoir établir l’architecture du système, nous devons d’abord fixer les objectifs de conception de notre simulateur. La liste ci-dessous décrit les caractéristiques souhaitées pour notre cas d’études :

\subsubsection{Modularité}
Le simulateur doit offrir plusieurs modules agencés de façon à pouvoir facilement ajouter, modifier ou étendre les fonctionnalités d’un module sans affecter les autres. Ceci permet par exemple d’ajouter de nouveaux types de capteurs, de nouveaux algorithmes de planification, ou modifier le fonctionnement de base du système de navigation.

Le but est d’augmenter l’extensibilité du projet, et faciliter la maintenance et la mise à jour de la plateforme au fil du temps.

\subsubsection{Flexibilité}
Le simulateur doit être suffisamment flexible pour s'adapter à différents types de problèmes d’optimisation et d'algorithmes de contrôle. Cela inclut la possibilité de :
\begin{itemize}
    \item Modéliser différentes formes de fonctions objectif.
    \item Définir différents scénarios d’expérimentation.
    \item Paramétrer la position des obstacles.
    \item Personnaliser les caractéristiques de robots telles que les distances et rayons de détection ; niveau d’énergie disponible ; mouvements autorisés…etc.
\end{itemize}

\subsubsection{Passage à l’échelle}
Le simulateur doit permettre de simuler un grand nombre de robots et s’adapter à l’évolution des ressources de la machine sans nécessiter des retouches de la part de l’utilisateur sur le code source. Il doit intégrer des fonctionnalités permettant à l’utilisateur de configurer des expériences à large échelle et offrir la possibilité de paralléliser instantanément ses expériences en déterminant le nombre de processeurs à utiliser.

Ceci a pour but de permettre aux chercheurs de porter facilement leur code sur des machines plus puissantes afin d’accélérer leurs expériences sans devoir maîtriser les techniques de la programmation parallèle.

\subsubsection{Efficacité}
Le simulateur doit minimiser l’utilisation des ressources de la machine telles que le temps de calcul et la mémoire virtuelle, il doit permettre aux utilisateurs d’exécuter les expériences sans nécessiter l’acquisition de cartes graphiques couteuses ou de serveurs de calcul.

Ceci a pour but d’améliorer l’accessibilité à la recherche pour les étudiants et les jeunes chercheurs n’ayant pas forcément les moyens d’investir sur un matériel plus coûteux.

\subsubsection{Automatisation}
Le simulateur doit automatiser tout le processus d’expérimentation, allant du paramétrage automatique, l’exécution répétitive des expériences, la génération des graphes et tableaux de statistiques, ainsi que la sauvegarde des résultats pour analyse ultérieure.

Ceci a pour but d’alléger le temps dédié à la gestion des expériences et permettre aux chercheurs de se concentrer sur des tâches plus importantes, telles que l’amélioration des algorithmes d’optimisation et le développement de nouveaux modèles.

\subsubsection{Facilité d’utilisation}
Le simulateur doit permettre aux utilisateurs de facilement apprendre à l’utiliser sans nécessiter de fortes compétences en programmation ou en mathématique.

Ceci a pour but de réduire la barrière d’entrée pour les développeurs et les chercheurs intéressés par le domaine de la robotique sans vouloir s’aventurer sur certains détails mathématiques tels que la modélisation géométrique du déplacement des robots ou du calcul de probabilité des grilles d’occupation. 

L’objectif est de leur permettre de tester et de valider rapidement et facilement de nouveaux algorithmes, et positionner leur travail en comparant facilement leurs résultats avec ceux obtenus par d’autres chercheurs.

\subsubsection{Visualisation}
Le simulateur doit fournir des outils de visualisation afin de permettre de dresser une représentation visuelle de l’état des robots ; les mesures des capteurs ; les obstacles détectés ; et les signaux de commandes. Il doit aussi permettre de visualiser l’objectif de chaque robot, les chemins planifiés, ainsi que les différentes statistiques utiles pour l’utilisateur.

Un autre point important dans notre contexte est la génération de graphes pour l’analyse et l’évaluation des performances des métaheuristiques, telles que la courbe de convergence ; le taux d’erreur ; ou le temps d’exécution.

\subsubsection{Abstraction}
Le simulateur doit fournir une interface de programmation (API) à haut niveau pour simplifier et abstraire les fonctionnalités du simulateur. Cette interface doit être conçue pour faciliter l'utilisation du simulateur en supprimant les détails de bas niveau du fonctionnement du simulateur tels que les modèles géométriques du robot, la simulation des faisceaux laser, la gestion du parallélisme et la visualisation de l’environnement.

Ceci a pour but d’augmenter la productivité des utilisateurs en leur permettant de se concentrer sur ce qui est essentiel pour leurs travaux de recherche.

\begin{figure}[ht!]
    \centering    
    \vspace{0.3cm}
    \includegraphics[width=0.9\textwidth]{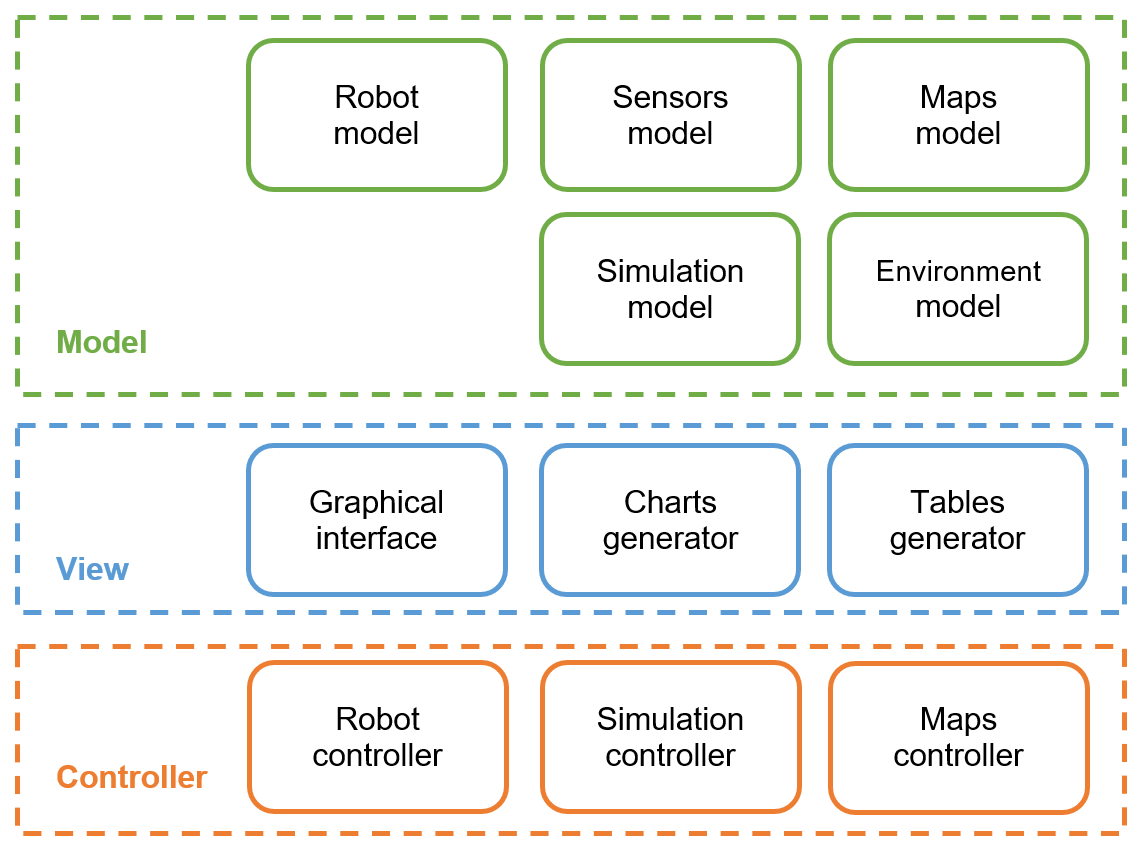}
    \caption{Architecture générale du simulateur PyRoboticsLab}
    \label{fig:0}
\end{figure}

\subsection{L'architecture du système}

Le simulateur PyRoboticsLab se base sur une architecture MVC (Modèle-Vue-Contrôleur). Ce concept vise à séparer le programme en plusieurs couches :
\begin{itemize}
    \item Le modèle représente les caractéristiques physiques des robots, des leurs capteurs et de l’environnement dans lequel ils opèrent.
    \item La vue permet de visualiser la simulation et afficher les données telles que les cartes, les obstacles, la position des robots…etc.
    \item Le contrôleur définit les algorithmes de contrôle des robots pour les différentes tâches de navigation, planification et exploration…etc.
\end{itemize}

\vspace{2mm}
La figure \ref{fig:0} schématise cette architecture.

Séparer la simulation en composants distincts nous permet de créer une architecture plus modulaire et maintenable. Chaque composant est divisé en plusieurs modules qui sont responsables de gérer un aspect spécifique de la simulation :

\begin{itemize}
    \item Modèle de robot : définit les caractéristiques physiques du robot, telles que son mode de déplacement et sa consommation d’énergie, ainsi que toutes les routines pour lui permettre de se déplacer dans l’environnement et se localiser.
    \item Modèle de capteur : définit les caractéristiques des capteurs montés sur les robots, tels le LIDAR ainsi que toutes les routines pour la simulation des données produites par ces capteurs, en incluant l’intégration du bruit artificiel dans les mesures afin de reproduire les conditions imparfaites du monde réel ainsi que les routines de raytracing.
    \item Modèle de carte : définit les caractéristiques des cartes utilisées par le robot, y compris les cartes locales et la carte globale. Chaque carte est modélisée en plusieurs couches pour séparer les données relatives aux probabilités d'occupation (position des obstacles) des données relatives à l'exploration (marquage des cellules visitées...).
    \item Modèle d'environnement : définit les caractéristiques de l'environnement dans lequel les robots se déplacent dans la surface opérationnelle ainsi que les routines nécessaires pour le chargement, copie et sauvegarde des modèles d’environnement.
    \item Modèle de simulation : permet de sauvegarder l’historique de toutes les opérations réalisées et les mesures de performances instantanées, dans le but de pouvoir retracer l’expérience étape par étape, ou de visualiser les courbes de convergence, de consommation d’énergie ou du temps d’exécution des algorithmes.
    \item Contrôleur du robot : inclut les routines pour contrôler le robot en lui permettant de naviguer, planifier les trajectoires, et choisir des points de destination. Ce module contient un sous-module définissant toutes les routines nécessaires pour le fonctionnement des algorithmes d’optimisation et métaheuristiques. L’utilisateur peut étendre ce module en ajoutant ces algorithmes afin de les intégrer facilement dans le scénario de simulation sans modifier les autres modules.
    \item Contrôleur des cartes: Ce module définit les routines nécessaires à la cartographie telles que la mise à jour des probabilités d'occupation, la fusion de cartes ainsi que le calcul des statistiques d'exploration.
    \item Contrôleur de la simulation : inclut les routines pour définir le scénario de simulation, et charger les modèles d’environnement et des robots. Ce module est responsable d’initialiser les expériences, de mesurer les performances des robots, et d’arrêter la simulation lorsque les critères d’arrêts sont atteints.
    \item Interface visuelle : inclut toutes les routines qui permettent l'affichage des aspects relatifs à la simulation et à l’état des robots.
    \item Générateur de graphes : inclut toutes les routines permettant de visualiser et manipuler les courbes de statistiques.
\end{itemize}

\begin{figure}[phbt!]
    \centering    
    \vspace{0.3cm}
    \includegraphics[width=\textwidth]{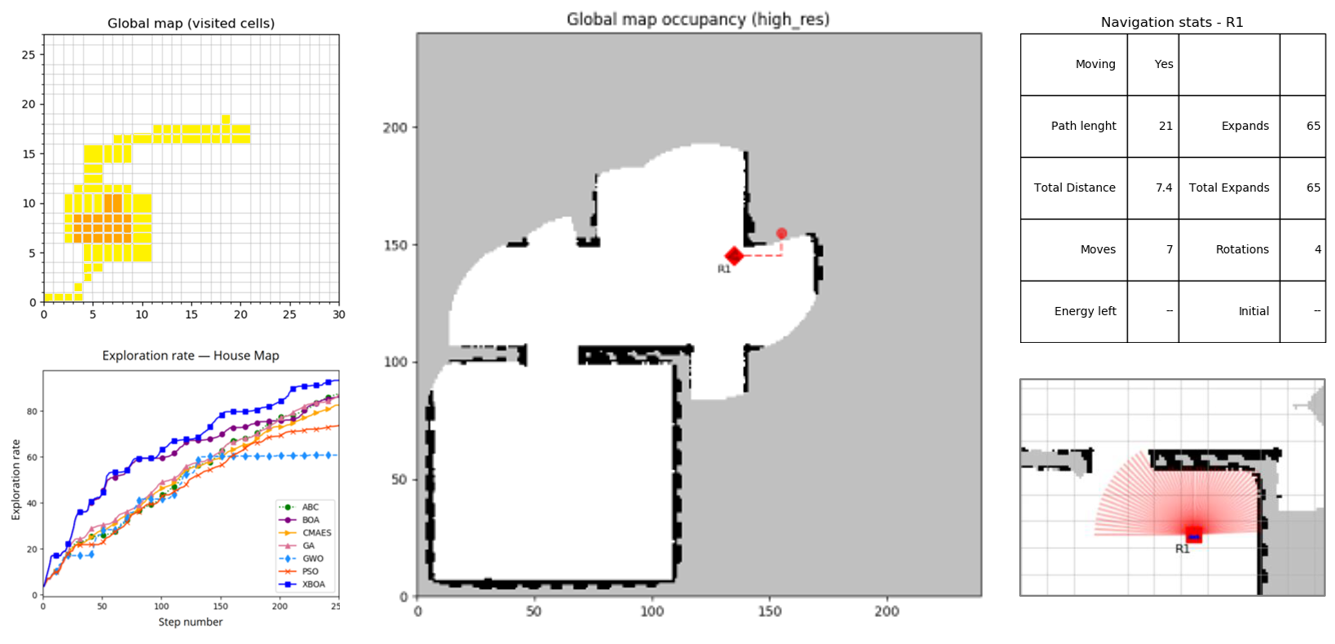}
    \caption{Exemple de quelques types de visualisations générées par le simulateur PyRoboticsLab}
    \label{fig:3.15}
\end{figure}

La figure \ref{fig:3.15} montre quelques exemples de visualisations générées automatiquement par le simulateur pendant les expériences. Elles peuvent être facilement paramétrables pour afficher différents types de cartes, statistiques, graphes et autres informations utiles.

    \subsection{Les outils et technologies}
Notre plateforme de simulation est entièrement développée en Python, avec une attention particulière portée à la réduction des dépendances afin de permettre une portabilité facile vers tous les systèmes d’exploitation (Linux, Windows, OS X...).

Python est un langage de programmation interprété, créé avec l’intention d’offrir une syntaxe simple et facile à apprendre. Toutefois, ceci ne réduit en rien sa puissance puisqu’il est versatile et peut être utilisé pour créer des plateformes web, logicielles, applications mobiles, et systèmes embarqués. Son interpréteur permet d’exécuter le même code sur des systèmes d’exploitation différents sans nécessiter une modification dans le code source ou sa recompilation.

L’utilisation de Python pour développer un outil de recherche nous permet de tirer profit de la puissance des librairies disponibles pour le calcul scientifique. Afin de respecter les objectifs définis par l’architecture globale du système, nous avons réduit l’utilisation des dépendances externes aux trois librairies suivantes :

\begin{itemize}
    \item Numpy : Est la librairie Python de référence lorsqu’il s’agit de calcul scientifique et manipulation de tableaux multidimensionnels. Elle offre une multitude de routines pour les calculs d’algèbre linaires, opérations de tri, statistiques, nombres aléatoires, manipulation de dates...etc.
    Nous utilisons Numpy dans notre plateforme pour gérer toutes les opérations de calcul matriciel nécessaires pour les opérations de cartographie et la manipulation des populations.
    \item Matplotlib : Est une librairie destinée à la création de visualisations scientifiques, telles que les graphes, les histogrammes, les nuages de points…etc. Elle permet de générer des figures interactives hautement personnalisables, et exporter les résultats sous différents formats.
    Nous utilisons Matplotlib dans notre projet pour la génération de l’interface utilisateur permettant de visualiser la position des robots, les obstacles détectés, les chemins planifiés, ainsi que les différentes statistiques et cartes d’exploration générées par les robots.
    \item Pygmo2 \cite{biscani20} : 
    Est une librairie réalisée par l’Agence Spatiale Européenne offrant une interface pour implémenter des algorithmes d’optimisation massivement parallèles. Elle supporte des fonctionnalités avancées telles que la parallélisation des métaheuristiques, le tri de populations, la visualisation des solutions non dominées (Pareto front)...etc.
    Nous utilisons cette librairie pour uniformiser l’implémentation des métaheuristiques que nous utilisons pour résoudre les problèmes d’exploration et de planification de trajectoires. Cette uniformisation est un aspect important pour notre projet puisqu’elle permet de s’assurer que la différence entre les résultats expérimentaux des algorithmes n’est pas causée par une différence entre les techniques d’implémentation utilisées à bas niveau par les librairies.
\end{itemize}

L’intégration d'un nombre réduit de libraires Python permet de faciliter la portabilité du code et minimiser les dépendances. Ceci offre les avantages suivants :

\begin{itemize}
    \item Facilité à comprendre, déboguer ou modifier le code source de la plateforme d’afin d’y intégrer les changements souhaités.
    \item Facilité d’installation et de configuration, à la différence de nombreux simulateurs qui sont plus concentrés sur des environnements Linux et les systèmes de type UNIX.
    \item Possibilité d’intégration avec les librairies d’apprentissage machine populaires, pour l’ajout de fonctionnalités additionnelles tel que le Deep Learning par exemple.
    \item Possibilité de déployer rapidement le projet en tant que service web dans un serveur Cloud pour une utilisation ouverte au public.
    \item Facilité à ajouter un nouvel algorithme et comparer ses performances avec les autres algorithmes déjà implémentés sur la plateforme.
\end{itemize}

Nous pensons que ces caractéristiques sont un atout pour permettre son adoption par les étudiants et les enseignants comme outil pédagogique et de recherche.
    
    \subsection{Les scénarios de bechmarking}

La plateforme de benchmarking proposée permet de simuler les scénarios suivants:

\begin{itemize}
    \item Path Planning: Planification de trajectoires.
    \item Exploration : Découverte, balayage, surveillance, détection d'intrus.
    \item Target Searching : Rechercher une personne ou un objet.
    \item Foraging : Déplacement et tri d'objets.
    \item Map Decomposition: décomposition d’environnement en plusieurs régions.
    \item Formation Control :  Coordination pour le déplacement en formation de plusieurs robots.
\end{itemize}

\section{Modélisation géométrique des robots}


\begin{figure*}[b!]
    \vspace{1mm}%
    \centering
    \setlength{\abovecaptionskip}{0.3cm} 

    \hspace{3mm}%
     \begin{subfigure}[b]{0.37\textwidth}
         \centering
         \includegraphics[width=\textwidth]{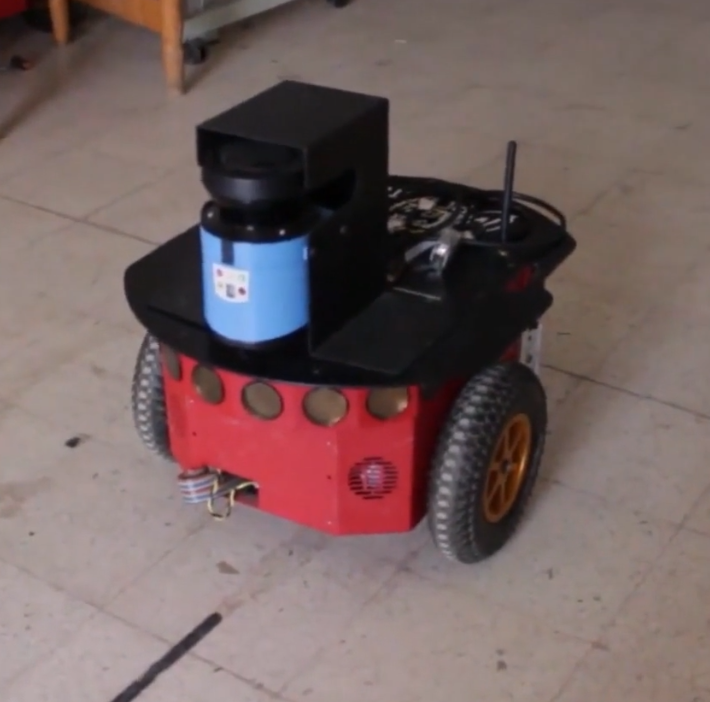}
         \caption{Robot P3DX utilisé}
     \end{subfigure}
     \hspace{3mm}%
    \begin{subfigure}[b]{0.57\textwidth}
         \centering
         \includegraphics[width=\textwidth]{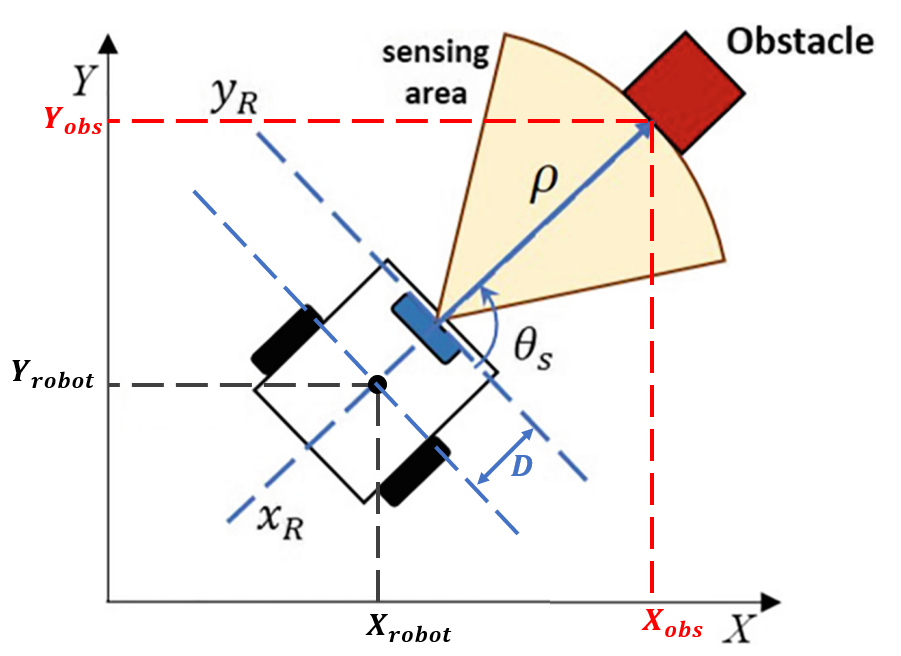}
         \caption{Schéma géométrique \cite{ng19}}
     \end{subfigure}
     
    \caption{Modélisation géométrique du robot et des différents repères utilisés pour la détection d'obstacles}
    \label{fig:c3.16}
\end{figure*}

Le simulateur PyRoboticsLab se base sur une représentation géométrique en 2 dimensions des robots mobiles. Il simule les robots terrestres à commande différentielle ainsi que les robots aériens de type quadrirotors. Les moteurs de ces robots ont la particularité d'être commandés séparément, ce qui permet de faire des rotations en 360° sur place sans nécessiter des manœuvres tel que les véhicules de type voiture par exemple.

La commande différentielle permet de contrôler la direction du robot en variant les vitesses de rotation de chaque moteur. Si tous les moteurs tournent à la même vitesse, le robot se dirigera tout droit. Si un des moteurs tourne à une vitesse plus élevée que les autres, le robot fera une rotation dans une direction opposée à l'emplacement de ce moteur. L'angle de cette rotation est influencé par plusieurs paramètres dont la distance du moteur par rapport au centre de gravité du robot, ainsi que le diamètre des roues pour les robots terrestres par exemple.

Afin de faciliter la commande des robots dans notre simulateur, nous mettons à la disposition de l'utilisateur des routines permettant de faire abstraction des détails de calcul pour ne se focaliser que sur l'action souhaitée. Nous présenterons un exemple de ce type de routine dans la section suivante (voir section \ref{nav_model}).

Notre modélisation des robots terrestres se base sur un robot de type Pioneer P3DX dont nous avons eu l'occasion d'utiliser pour faire des expériences dans le laboratoire LARESI situé au département d'électronique à l'USTOMB. Ce robot très populaire dans le domaine de la recherche possède un télémètre laser qui lui permet de détecter les obstacles aux alentours dans un rayon de 180° jusqu'à 270° selon les modèles. La distance de détection maximale du télémètre est de 10 mètres; toutefois, nous avons choisi de la limiter à 4 mètres pour des raisons pratiques (adaptation aux endroits étroits).

\begin{figure}[htb!]
    \setlength{\abovecaptionskip}{0.5cm} 
    \setlength{\belowcaptionskip}{-0.4cm} 
    \centering 

    \begin{subfigure}[b]{0.43\textwidth}
         \centering
         \includegraphics[width=\textwidth]{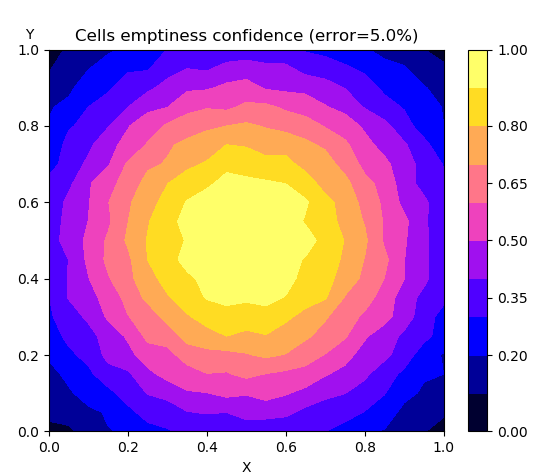}
         \caption{Le degré de confiance que la cellule soit vide est plus élevé lorsque le rayon traverse cette cellule près de son centre.}
         \vspace{3mm}%
     \end{subfigure}
     \hspace{5mm}%
     \begin{subfigure}[b]{0.48\textwidth}
         \centering
         \includegraphics[width=\textwidth]{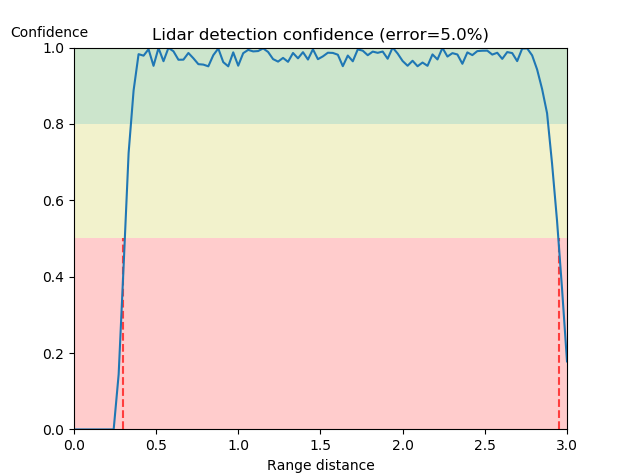}
         \caption{Le degré de confiance de la distance mesurée  est faible lorsque l'obstacle est trop proche ou trop loin du capteur.}
         
         \vspace{4mm}%
     \end{subfigure}

\caption{Modèle du bruit ajouté au capteur LIDAR et calcul du degré de confiance de la mesure}
    \label{fig:3.9}
\end{figure}

Étant donné que le Lidar du robot était sujet à un taux d'erreur conséquent lors des expériences réelles, nous avons choisi de modéliser cette erreur estimée à 5\% en intégrant dans notre simulateur un bruit gaussien. Nous avons aussi intégré la distance aveugle du Lidar comprise entre 0 et 30cm, c'est-à-dire que si l'obstacle est trop proche du Lidar il ne sera pas détecté, et ceci à cause de la limite physique de ce type de capteurs. La figure \ref{fig:3.9} schématise de ce modèle d'erreurs.

L'intégration de ces imperfections dans le modèle géométrique de PyRoboticsLab permet d'effectuer des simulations plus réalistes. En effet, l'intégration de ces imperfections permettra de tester la validité des algorithmes d'évitement d'obstacles et de cartographie à s'adapter plus facilement aux conditions du monde réel lorsqu'ils sont déployés en production, contrairement aux tests effectués dans des simulateurs qui assument l'hypothèse du monde parfait.

La modélisation des rayons laser du Lidar a été implémentée en utilisant la technique du \textit{Ray Casting}, qui consiste à tracer des rayons virtuels et suivre leur trajectoire pixel par pixel jusqu'à trouver le plus proche objet bloquant le chemin de ce rayon. Ceci nous permet de calculer la distance entre le robot et cet objet.

Le Lidar modélisé possède une résolution de 1°, nous projetons donc 181 rayons virtuels avec chacun un angle différent allant de 0° à 180°, ce qui nous permet de connaître la distance et l'angle des objets détectés. Si le rayon tracé n'est bloqué par aucun obstacle, nous concluons que l'espace traversé par ce rayon est vide et peut être parcouru par le robot en toute sécurité.

Étant donné que l'environnement dans notre simulateur est échantillonné sous forme de grille dont chaque cellule représente un espace de 10cm² (voir section \ref{map_model}) il peut arriver qu'un rayon projeté ne traverse qu'une partie infime de cette cellule, ce qui ne permet pas d'être sûr que cette cellule soit réellement vide. Ceci nous a poussés à modéliser une fonction gaussienne pour mesurer le degré de confiance de la mesure du rayon laser : plus le rayon est proche du centre de la cellule, plus la confiance de mesure est élevée, tel qu'est modélisé par la figure \ref{fig:3.9}. Ce degré de confiance de la mesure est linéairement proportionnel à la probabilité de détection. Lorsque plusieurs rayons laser traversent une même cellule, cette probabilité de détection est cumulée et augmente le taux de confiance de la mesure. Le calcul des probabilités sera détaillé dans la section \ref{map_model} dédiée à la cartographie.

\noindent

Une fois un obstacle détecté avec un degré de confiance suffisamment élevé, nous effectuons une transformation géométrique pour calculer la position des obstacles par rapport au repère fixe à partir de sa position relative au robot. Ce repère fixe est défini au point (0,0) de la carte globale, et ceci afin de faciliter les calculs par rapport au repère relatif du Lidar puisque celui-ci est attaché à un robot qui se déplace. En d'autres termes : les distances des obstacles mesurées par le Lidar sont toutes relatives au repère du robot mobile et il faut donc les transformer vers des positions globales relatives au repère fixe de l'environnement afin de pouvoir les dessiner sur la carte. Ce processus est illustré par la figure \ref{fig:c3.16}

\section{Modélisation du processus de navigation, planification et évitement d'obstacles} \label{nav_model}


\begin{figure}[t!]
    \centering    
    \vspace{0.3cm}
    \includegraphics[width=0.7\textwidth]{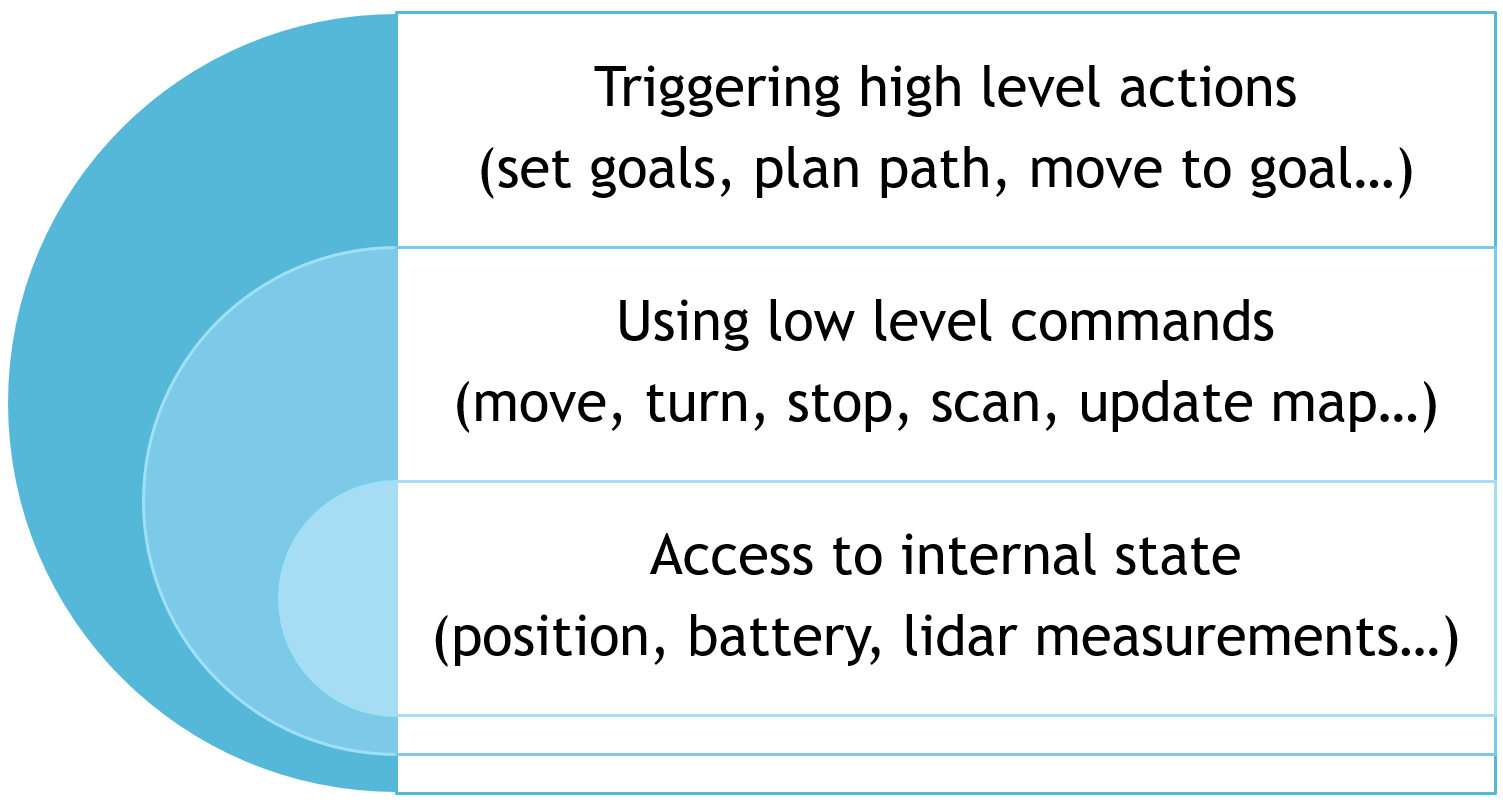}
    \caption{Niveaux d'abstraction du processus de navigation, planification et évitement d'obstacles du simulateur PyRoboticsLab}
    \label{fig:3.17}
\end{figure}

Afin de faciliter la commande des robots dans notre simulateur, nous mettons à la disposition de l'utilisateur des routines permettant de faire abstraction des détails de calcul et ne se focaliser que sur l'action souhaitée. Il suffit d'utiliser par exemple la commande 
\textit{robot.move(forward)} pour le déplacer un mètre en avant, et \textit{robot.turn(45)} pour faire une rotation de 45° dans le sens des aiguilles d'une montre. D'autres routines permettent d'effectuer des tâches relatives à la détection d'obstacles et de cartographie telle que \textit{lidar.scan()} pour récupérer la liste des obstacles détectés, \textit{lidar.get\_measurement\_confidence()} pour calculer le degré de confiance dans la mesure du lidar selon le modèle du bruit, et \textit{robot.update\_map()} pour mettre à jour la carte.

Afin de simplifier encore plus le processus de développement de nouveaux algorithmes, le simulateur offre un autre niveau d'abstraction permettant à l'utilisateur de définir directement les coordonnées (x, y) du point auquel le robot devra se diriger sans se soucier des commandes de direction (move) et de rotations (turn). Ceci peut se faire en utilisant donc la commande \textit{robot.move\_toward\_point(x,y)}. De même il suffit d'utiliser les commandes \textit{robot.set\_goals()} et \textit{robot.plan\_path()} pour définir un ou plusieurs points à visiter et planifier une trajectoire vers ces points.

L'idée est donc d'offrir plusieurs niveaux d'abstraction permettant aux utilisateurs de choisir différentes commandes à utiliser selon leurs objectifs. Un utilisateur qui souhaite utiliser le simulateur pour développer un nouvel algorithme d'exploration choisira de préférence un niveau d'abstraction simple afin de ne se focaliser que sur la stratégie du robot, alors qu'un utilisateur souhaitant tester de nouvel algorithme de navigation ou de cartographie choisira sûrement l'utilisation de routines de bas niveau pour contrôler les chaque action individuellement.

Nous laissons aussi la flexibilité aux utilisateurs de pouvoir redéfinir n'importe quelle fonction et apporter des modifications dans l'implémentation à bas niveau afin de leur permettre d'adapter le simulateur à leurs besoins spécifiques. Il pourront pour cela utiliser les variables de classe pour avoir accès aux informations internes du robot telles que le niveau de batterie restant, l'état du robot (en mouvement, en attente, bloqué par un obstacle), la distance parcourue, le nombre de mouvements effectués...etc.

\begin{figure}[phbt!]
    \centering    
    \vspace{0.3cm}
    \includegraphics[width=1.4\textwidth, angle=90]{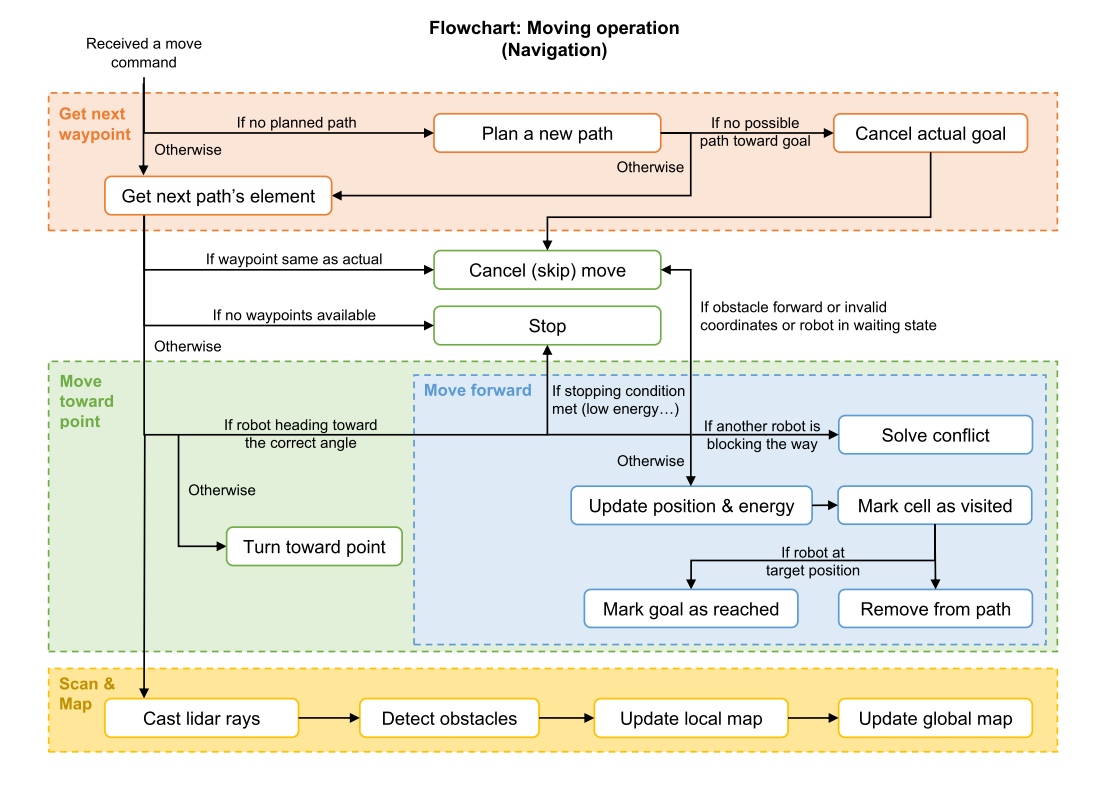}
    \caption{Schéma général du modèle de navigation, planification et évitement d'obstacles du simulateur PyRoboticsLab}
    \label{fig:3.8.1}
\end{figure}

La figure \ref{fig:3.8.1} présente un schéma général de notre modélisation du processus de navigation, planification et évitement d'obstacles. Cette modélisation se base sur l'implémentation en interne d'un automate à états finis. 

\begin{figure}[phbt!]
    \centering    
    \vspace{0.3cm}
    \includegraphics[width=\textwidth]{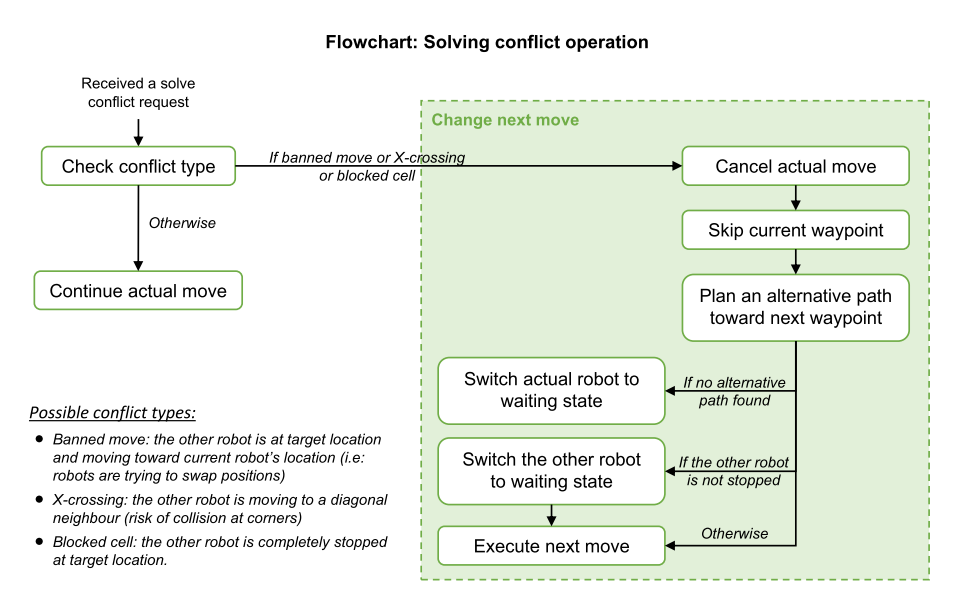}
    \caption{Diagramme de la routine "\textit{solve conflict}" pour résoudre un problème de blocage entre deux robots}
    \label{fig:3.8.2}
\end{figure}

Chaque état présenté dans ce schéma possède une routine définissant l'ensemble des opérations à effectuer. La figure \ref{fig:3.8.2} par exemple décrit les détails de l'opération "\textit{solve conflict}" dont l'objectif est de trouver un compromis entre deux robots qui se bloquent mutuellement le chemin. Cette routine poussera un des robots à changer de trajectoire, ou de se mettre dans un état d'attente pour laisser l'autre robot passer.


\section{Modélisation des grilles d'occupation} \label{map_model}

L'exploration de zones inconnues est étroitement liée au problème de navigation et de cartographie. Le robot doit se déplacer dans l'environnement et le découvrir progressivement. Lors de cette opération, il est fort probable de rencontrer des impasses et autres obstacles bloquant le chemin. Le robot mémorisera alors la position des obstacles et les utilisera pour planifier des chemins alternatifs et découvrir de nouvelles régions.

Les robots utilisent des capteurs pour détecter les murs et les obstacles. Étant donné que ces capteurs ont une portée limitée, il n'est pas possible d'observer l'ensemble de l'environnement à la fois. Dans ce cas, nous devons sauvegarder les positions des obstacles détectés dans une structure de données qui permet au robot d'agréger facilement de nouvelles observations et de les combiner de manière à simplifier le calcul des trajectoires.

\begin{figure}[phbt!]
    \centering    
    \vspace{0.3cm}
    \includegraphics[width=\textwidth]{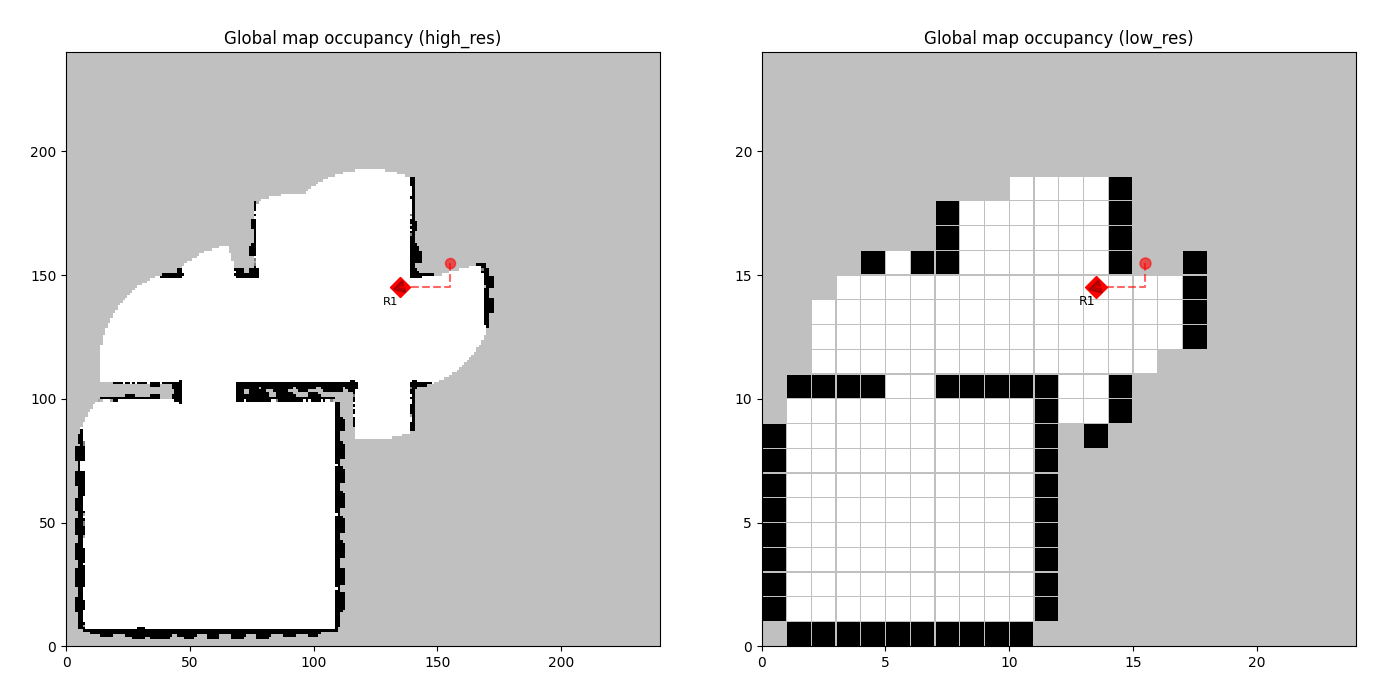}

    \begin{description}
    \item[] \textit{A gauche: carte à haute résolution (0.1m²/px). A droite: carte à basse résolution (1m²/px).}
    \item[] \textit{Les pixels blancs représentent la zone explorée ; les pixels gris représentent la zone inconnue ; les pixels noirs représentent les obstacles détectés.}
    \end{description}
     \vspace{0.5mm}
    
    \caption{Exemple d'exécution d'un scénario d'exploration et décomposition de la carte de l'environnement sous forme de grille d'occupation}
    \label{fig:3.10}
\end{figure}

La Grille d'Occupation (Occupancy Grid Map \cite{elfes89}) est la structure de données la plus utilisée pour représenter l'environnement en robotique. C'est une matrice 2D où chaque cellule représente une partie de l'environnement. La taille des cellules influence le degré de détails affichés sur la grille, elle est généralement définie sur une taille égale ou inférieure à la circonférence du robot. La figure \ref{fig:3.10} montre un exemple de carte quadrillée.

La valeur de chaque cellule de la grille représente la probabilité que la région correspondante de l'environnement soit vide ou occupée par un obstacle. Puisque le robot n'a aucune information à priori sur la région à explorer, toutes les cellules ont une probabilité d'occupation initiale de 0,5. Cette probabilité sera mise à jour à l'aide de la règle de Bayes (équation \ref{eq:5}) chaque fois que les capteurs du robot détectent une cellule vide ou occupée.

\begin{equation} \label{eq:5}
    p(A/B) = \frac{p(B/A)*p(A)}{p(B)}
\end{equation}
\begin{align*}
    &\text{A est la valeur d'occupation, et} \\
    &\text{B est l'observation}
\end{align*}

Une pratique courante dans le domaine de cartographie vise à utiliser les valeurs logarithmiques d'occupation au lieu des probabilités, et ceci afin de convertir les opérations de multiplication en additions tel que décrit par les équations \ref{eq:6} et \ref{eq:7}.

\begin{align} \label{eq:6}
    odds(A) &= \frac{p(A)}{P(\neg A)} \nonumber \\
    odds(A/B) &= \frac{p(A/B)}{P(\neg A/B)}
\end{align}

Après avoir appliqué l'équation \ref{eq:6} dans la règle de Bayes, nous obtenons l'équation \ref{eq:7}. La valeur logarithmique varie entre [$-\infty$,$+\infty$], ce qui est utile pour éviter la multiplication de petits nombres lors de l'implémentation qui pourront causer des problèmes en raison de la précision limitée des valeurs flottantes dans l'ordinateur.

\begin{equation} \label{eq:7}
    log odds(A/B) = log \frac{p(B/A)}{P(B/\neg A)} + log odds(A)
\end{equation}

En conséquence, nous pouvons classer chaque cellule $ C_{ij} $ de la grille dans l'une des trois catégories suivantes : cellule vide, cellule occupée par un obstacle et cellule inconnue. L'équation \ref{eq:8} formalise ceci.

\begin{equation} \label{eq:8}
C_{ij} \ is \  \left\{\begin{matrix}
Occupied \ &if \ Occ(C_{ij}) > 0 \\
Empty \ &if \ Occ(C_{ij}) < 0 \\
Unknown \ &if \ Occ(C_{ij}) = 0
\end{matrix}\right.
\end{equation}

Où $ Occ(C_{ij}) $ est la valeur logarithmique de l'occupation calculée en utilisant l'équation \ref{eq:7}. 

Dans l'équation \ref{eq:8}, la valeur $0$ représente le seuil pour savoir si une cellule contient un obstacle où non. Ce seuil a été défini à cette valeur parce qu'il correspond à la probabilité d'occupation initiale fixée à 0.5 ($logodds(0.5)=0 $)

Dans un scénario de navigation, le robot aura donc pour objectif d'éviter toutes les cellules occupées dont la valeur logarithmique d'occupation est supérieure à ce seuil. De même, l'objectif d'une mission d'exploration serait de  diriger le robot vers toutes les cellules ayant une valeur égale à ce seuil puisqu'elles représentent les cellules inconnues qui n'ont pas encore été balayées par les capteurs du robot. Ceci nous permet de modéliser notre problème sous forme de problème d'optimisation dont le but est de minimiser le nombre de ces cellules. La figure \ref{fig:3.10} montre un exemple d'exécution d'un tel scénario.

La section suivante présentera les détails relatifs à la modélisation de ce type de scénarios.


\section{Modélisation du problème d'exploration}

    \subsection{Modélisation mono et multirobots}   

\begin{figure*}[phb!]
    \centering

     \begin{subfigure}[b]{0.48\textwidth}
         \centering
         \includegraphics[width=\textwidth]{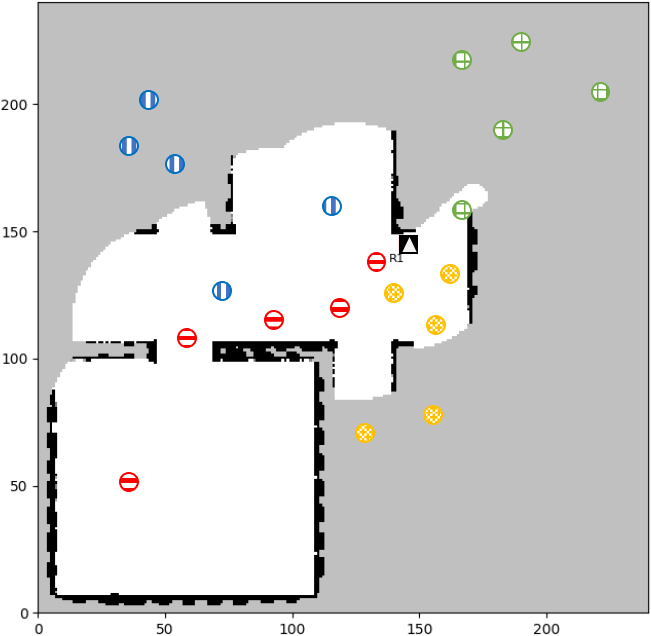}
         \caption{Générer une population de points de destinations \\}
     \end{subfigure}
     \hspace{5mm}%
    \begin{subfigure}[b]{0.48\textwidth}
         \centering
         \includegraphics[width=\textwidth]{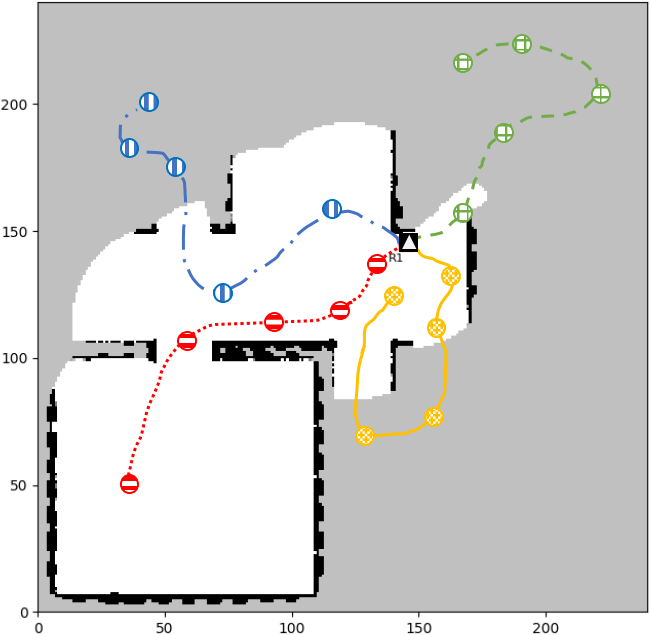}
         \caption{Calculer les plus courts chemins pour visiter les points de destinations de chaque solution candidate}
     \end{subfigure}
     \vspace{2mm}%

     \begin{subfigure}[b]{0.48\textwidth}
         \centering
         \includegraphics[width=\textwidth]{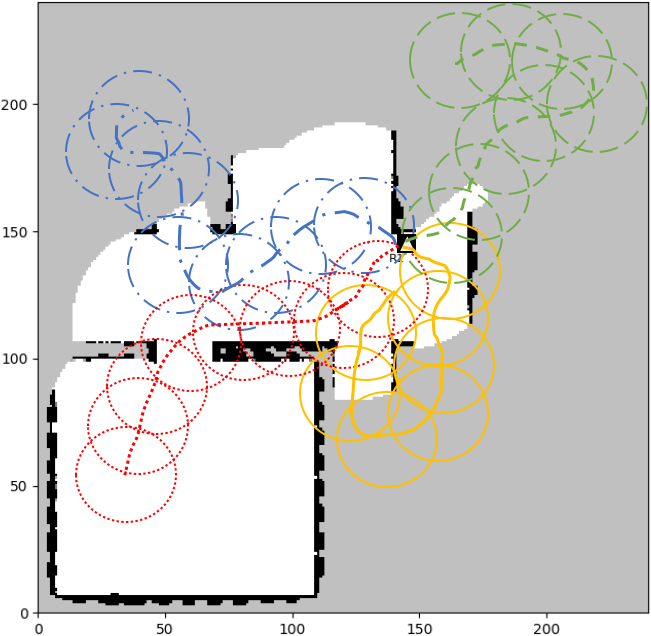}
         \caption{Estimer le gain en terme de surface de la zone explorée pour chaque chemin \\}
     \end{subfigure}
     \hspace{5mm}%
    \begin{subfigure}[b]{0.48\textwidth}
         \centering
         \includegraphics[width=\textwidth]{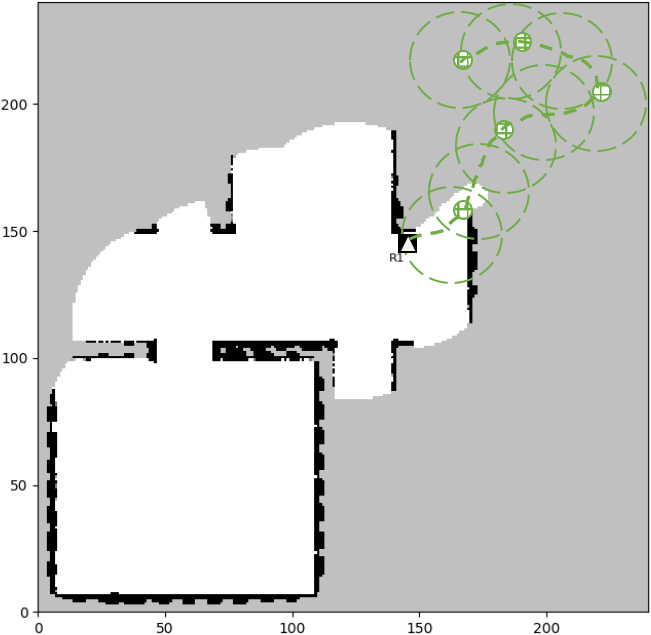}
         \caption{La meilleure solution est celle qui maximize la surface d'exploration de régions pas encore explorées}
     \end{subfigure}

     \vspace{1mm}
    \caption{Un exemple du processus d'évaluation de la fonction fitness pour une population de 4 solutions candidates}
    \label{fig:3}
\end{figure*}

La tâche d'exploration de zones inconnues est souvent modélisée comme un problème d'optimisation. Le but de ce processus est d'attribuer une probabilité d'occupation à chaque cellule de la carte. Afin d'atteindre cet objectif, le robot doit maximiser la surface de la zone explorée tout en minimisant l'énergie utilisée.

Le rôle des métaheuristiques dans notre modélisation est au cœur de ce processus. Elles commencent par générer une population aléatoire de points de destination à visiter, puis améliorent la position de ces points cibles à travers une succession d'opérations d'optimisation.

Mathématiquement, chaque solution candidate $X_k$ dans la population représente un ensemble d'emplacements de cellules cibles $ C_{ij} $, où $(i,j)$ sont les coordonnées $(x,y)$ à l'intérieur des limites de la grille.

\vspace{-0.3cm}
\begin{equation}
    X = {C_{ij}} \nonumber
\end{equation}

Par conséquent, la fonction fitness peut être modélisée comme une maximisation du nombre de cellules qui ont une valeur d'occupation logarithmique égale à 0 (cellules inexplorées). L'équation \ref{eq:9} définit la formulation mathématique de cette fonction.

\vspace{-0.5cm}
\begin{align} \label{eq:9}
    F &= max(Observed \ Cells) \nonumber \\
    &= min(\sum_{i,j}\delta(C_{ij,0})) \\
    Where\ &\delta(C_{ij,0}) = \left\{\begin{matrix}
        1 \ &if \ Occ(C_{ij})\neq 0 \\
        0 \ &otherwise
    \end{matrix} \right. \nonumber
\end{align}

\vspace{-0.8mm}
Avec la contrainte suivante:
\begin{equation}
\setlength{\abovedisplayskip}{6pt}
\setlength{\belowdisplayskip}{3pt}
    \sum_{i,j} E(C_{ij}) < current\ battery\ level \nonumber
\end{equation}

Où $ E(C_{ij}) $ est l'énergie nécessaire pour déplacer le robot de la position actuelle à la cellule $C_{ij}$.

La figure \ref{fig:3} montre un exemple d'application de cette opération pour sélectionner le meilleur ensemble d'emplacements cibles à visiter parmi 4 solutions candidates.

\noindent
\begin{figure}[pht!]
    \setlength{\abovecaptionskip}{0.4cm} 
    \setlength{\belowcaptionskip}{-0.4cm} 
    \centering 
    \includegraphics[width=0.9\textwidth]{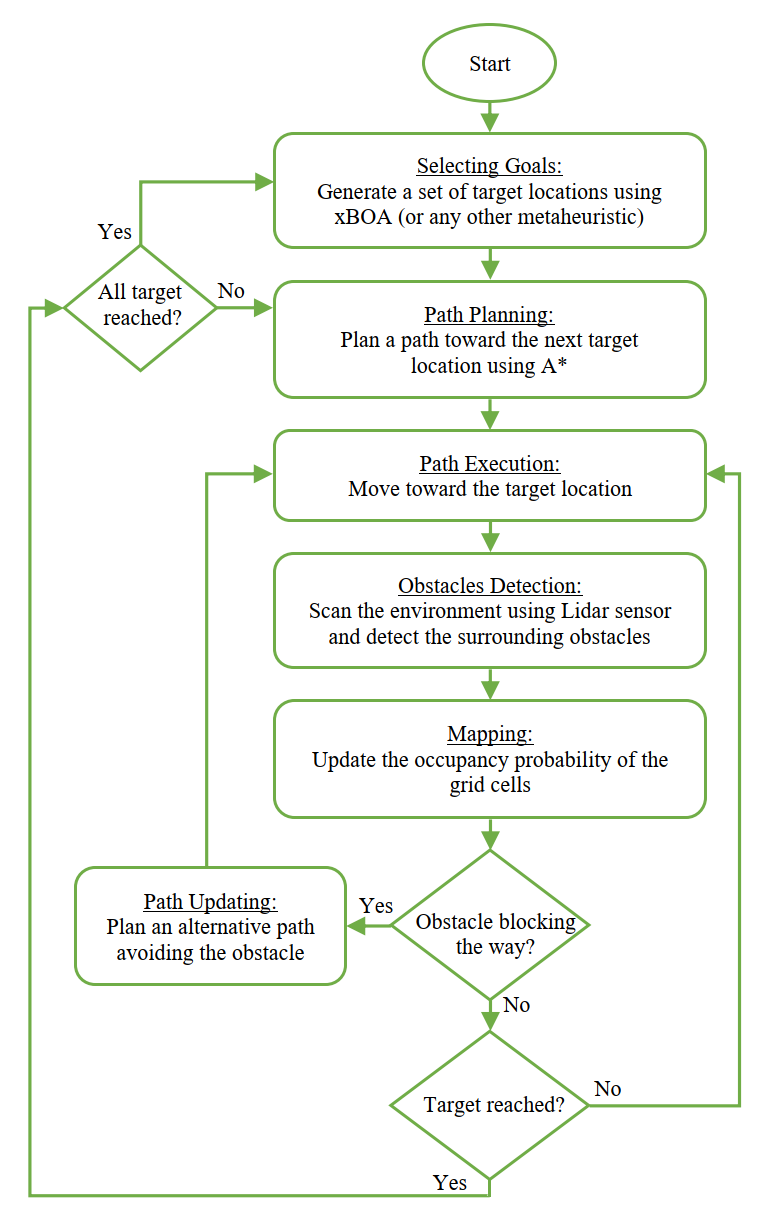}
\caption{Diagramme du processus d'exploration d'une zone inconnue}
    \label{fig:2}
\end{figure}

Une fois que le meilleur ensemble d'emplacements cibles qui satisfait la contrainte d'énergie est trouvé, le robot calcule le chemin le plus court qui relie ces emplacements cibles en utilisant l'algorithme $A^*$ \cite{hart68}, puis il exécute ce chemin jusqu'à visiter tous les emplacements cibles. Après cela, il répète l'algorithme d'optimisation pour générer un nouvel ensemble d'emplacements à visiter et continue le processus jusqu'à ce que toutes les cellules de la carte aient été observées (c'est-à-dire que le robot a exploré toute la zone).

Il est important de rappeler que la trajectoire prévue n'est pas nécessairement optimale, puisque le robot ne peut pas détecter les obstacles hors de portée de ses capteurs. De plus, il n'a aucune information préalable sur la région à explorer. Ainsi, il est fort probable qu'il rencontre des impasses et autres obstacles barrant le chemin lors de la navigation. Il sera alors obligé de trouver un chemin alternatif pour sortir de l'impasse, même si cela nécessite de retourner dans une région précédemment visitée et de consommer plus d'énergie.

Par conséquent, la solution est construite de manière incrémentale, en commençant par un chemin sous-optimal, puis en le mettant à jour régulièrement au fur et à mesure que de nouveaux obstacles sont observés, ce qui garantit également que le robot s'adapte aux environnements dynamiques et évite les obstacles mobiles.

\begin{figure}[pht!]
    \setlength{\abovecaptionskip}{0.4cm} 
    \setlength{\belowcaptionskip}{-0.4cm} 
    \centering 
    \includegraphics[width=\textwidth]{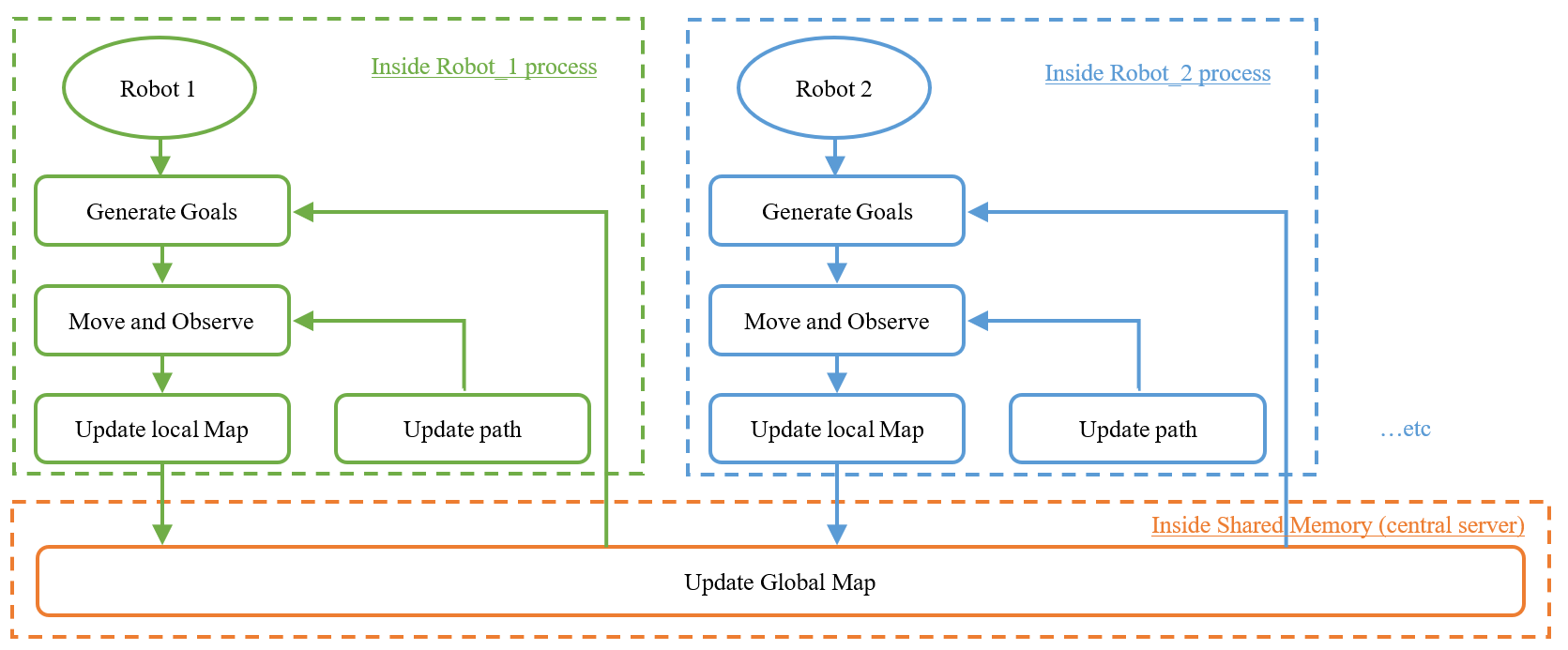}
\caption{Schéma du modèle utilisé pour assurer la coordination entre les robots}
    \label{fig:3.7}
\end{figure}

La figure \ref{fig:2} schématise le processus modélisé pour résoudre le problème d'exploration, et la figure \ref{fig:3.7} présente le schéma de coordination utilisé pour produire un comportement collectif dans les scénarios multirobots.


\noindent

    \subsection{Les critères d'évaluation}

Afin de pouvoir analyser les performances des métaheuristiques utilisées durant les expériences et comparer leurs résultats, nous allons utiliser les critères d’évaluation suivants :
\begin{itemize}
\item Nombre de pas (\textit{step number}) : Nombre de mouvement et rotations effectué par le robot. Ce nombre est proportionnel à la quantité d’énergie consommée.
\item Temps d’exécution (\textit{execution time}) : Durée totale de la mission, mesurée en secondes. 
\item Taux d’exploration (\textit{exploration rate}) : Surface de la zone observée par le robot pendant la mission, mesurée en pourcentage par rapport à la surface totale.
\item Nombre d’évaluation de la fonction de fitness (\textit{Fitevals}) : Nombre de solutions évaluées par la métaheuristique pendant le processus d’optimisation.
\item Temps de calcul (\textit{computation time}) : Durée d’exécution de l’ensemble des opérations de calculs requis par la métaheuristique pour effectuer toutes les itérations et sélectionner la solution optimale, mesurée en secondes.
\end{itemize}

    \subsection{La complexité du modèle} 

L'utilisation de l'équation \ref{eq:9} comme fonction de fitness signifie que pour chaque solution candidate dans la population, nous allons planifier un chemin vers les emplacements cibles définis par cette solution, puis estimer combien de nouvelles cellules seront observées si ce chemin est exécuté par le robot. La complexité de cette opération est $O(M²)$ où $M$ est la longueur du chemin.

Étant donné que la valeur de fitness est évaluée pour chaque solution candidate à chaque génération, la complexité globale devient $O(N x K x M²)$ où $N$ est la taille de la population, $K$ est le nombre de générations et $M$ est la longueur du chemin.

Si la taille de la population ou le nombre de générations est défini sur une grande valeur, le processus deviendra coûteux en calcul. Une approche alternative consisterait à calculer une estimation approximative des cellules observées (en utilisant un vecteur de distance par exemple) comme critère de fitness, mais cela diminuerait la qualité des solutions générées. Afin d'éviter cette perte d'informations, nous garderons notre modélisation tout en essayant de réduire la taille de la population au minimum afin d'obtenir un compromis acceptable entre la qualité des solutions générées et le temps d'exécution.

\section{Paramétrage et configuration}

\subsection{Paramétrage des expériences}

Afin d'évaluer les performances des algorithmes BOA et xBOA, nous allons les comparer à cinq autres métaheuristiques fréquemment utilisées dans la littérature, à savoir : 
\begin{itemize}
    \item Artificial Bees Colony (ABC) \cite{karaboga05}
    \item Covariance Matrix Analysis Evolution Strategy (CMAES) \cite{hansen03}
    \item Genetic Algorithm (GA) \cite{goldberg89}
    \item Grey Wolf Optimizer (GWO) \cite{mirjalili14}
    \item Particle Swarm Optimization (PSO) \cite{kennedy95}
\end{itemize}

Pour ces cinq méthodes, nous avons utilisé les implémentations fournies par \textit{Pygmo2} \cite{biscani20} qui est une bibliothèque Python offrant une interface unifiée pour implémenter des algorithmes d’optimisation parallèles.

Étant donné que ces méthodes sont basées sur des opérations stochastiques, nous allons utiliser la même population initiale pour chacune d’elles et répéter l’exécution 10 fois afin d’obtenir une comparaison plus objective. Le critère d’arrêt sera défini tel que suit : "\textit{arrêter l’expérience si l’énergie du robot atteint zéro, ou la surface de la zone explorée atteint un taux supérieur ou égal à 99\%}".

Nous allons effectuer une autre série d’expériences pour comparer les performances de l'algorithme xBOA par rapport aux autres variantes de l’algorithme BOA citées ci-dessous :

\begin{itemize}
    \item Self-adaptative BOA (SABOA) \cite{fan20} 
    \item BOA with intensive search (mBOA) \cite{arora18}
    \item BOA with non-linear adaptative rule (ABOA) \cite{zhang20}
\end{itemize}

Nous avons réimplémenté BOA et chacune de ces variantes en langage Python, et les avons adaptés à la bibliothèque \textit{Pygmo2} pour s’assurer que la différence des performances n’est pas causée par l’utilisation de techniques d’implémentation différentes, ou par des outils différents.

Finalement, nous allons effectuer un test en modifiant le nombre de robots pour valider l’adaptabilité de notre modélisation à des scénarios mono et multirobots.

\subsection{Paramétrage de l'environnement de simulation}

\begin{figure*}[pht!]
    \vspace{1mm}%
    \centering
    \setlength{\abovecaptionskip}{0.3cm} 

     \begin{subfigure}[b]{0.45\textwidth}
         \centering
         \includegraphics[width=\textwidth]{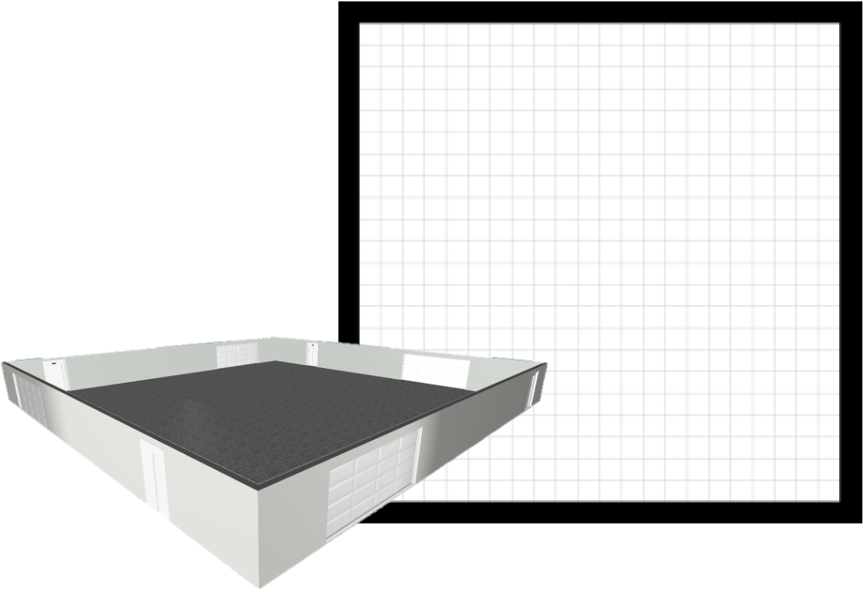}
         .\hspace{27mm}\textit{Empty Map (24x24m)}
     \end{subfigure}
     \hspace{5mm}%
    \begin{subfigure}[b]{0.425\textwidth}
         \centering
         \includegraphics[width=\textwidth]{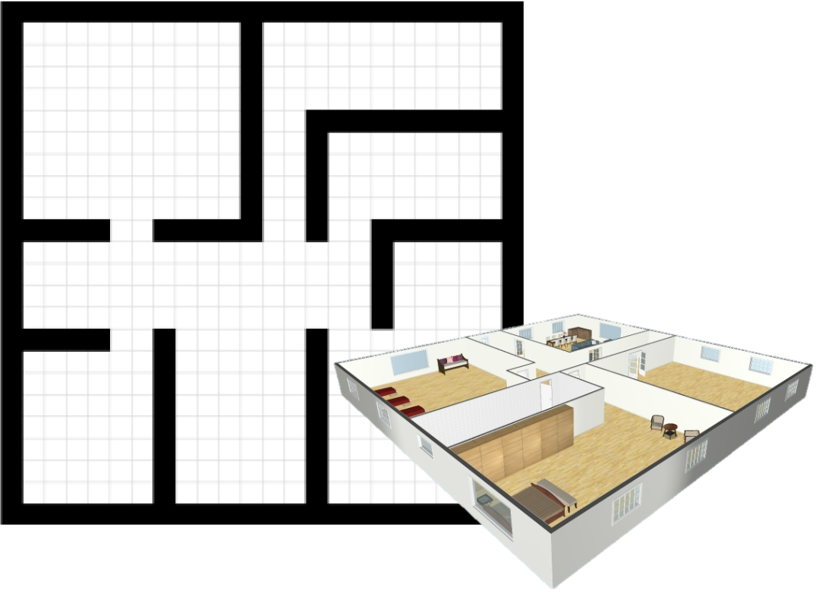}
         \textit{House Map (24x24m)}\hspace{24mm}
     \end{subfigure}
     
    \caption{Cartes de l'environnement de simulation utilisé durant les expériences}
    \label{fig:c3.8}
\end{figure*}

La figure \ref{fig:c3.8} montre la cartographie en 2D des deux environnements qui seront utilisés pour effectuer les expériences. La première (dénommée \textit{Empty map}) représente une zone vide ne contenant aucun obstacle à l’exception du mur d’entourage qui délimite son périmètre. La deuxième zone (dénommée \textit{House map}) est inspirée de l’architecture d’une maison et est partiellement couverte d’obstacles à un taux d'occupation de 27\% après suppression des portes et fenêtres. Ces deux zones ont une taille de 24x24m soit une surface totale de 576m² chacune.

Pour chaque zone, deux types d’expériences seront effectuées en variant le nombre d’emplacements cibles (points de destinations) générés par le processus d’optimisation. Ce paramètre influencera la stratégie d’exploration des robots : un nombre de destinations réduit poussera le robot à effectuer une planification à court terme, alors qu’un nombre plus élevé le poussera à effectuer une planification à long terme. Pour chaque expérience, nous enregistrerons le temps d’exécution, le taux d’exploration ainsi que les performances de chaque métaheuristique.

\begin{figure*}[pht!]
    \vspace{1mm}%
    \centering
    \setlength{\abovecaptionskip}{0.3cm} 

    \includegraphics[width=0.7\textwidth]{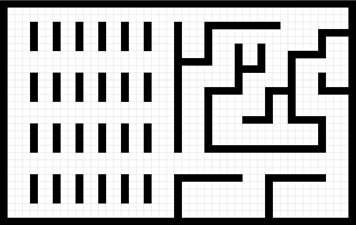}
    \begin{description}
        \centering
        \vspace{1mm}%
        \item[] \textit{Factory Map (30x50m)}
    \end{description}
     
    \caption{Carte d'environnement utilisé pour simuler les scénarios à large échelle}
    \label{fig:c3.19}
\end{figure*}

Nous utiliserons également une carte d'environnement inspirée d'une usine avec une superficie de 1500m² afin d'effectuer des expériences de validation de l'approche sur des zones à plus large échelle. Cet environnement contient un nombre plus important d'obstacles et plusieurs couloirs étroits et des chemins dont l'extrémité finale conduit à une impasse, ce qui sera utile pour tester les capacités des robots à naviguer efficacement sans se bloquer mutuellement le chemin. La figure \ref{fig:c3.19} montre la cartographie de cette usine.

Dans le but de réduire la variabilité des expériences causées par différentes initialisations, nous allons fixer les mêmes conditions de départ pour chaque session de test :

\begin{itemize}
\item Position de départ du robot :  dans les coordonnées (1,1) avec un angle de 0° vers le nord.
\item Distance de mesure du capteur LIDAR : 4m, avec un taux de 5\% d’erreurs.
\item Plage de mesure angulaire du capteur LIDAR : 180°, avec une résolution de 1°.
\item Taille de la population : 20 individus.
\item Nombre maximal de générations : 30 itérations.
\item Condition d’arrêt prématuré : arrêt si aucune amélioration de la valeur de fitness pendant 10 générations consécutives.
\item Population initiale : même population pour toutes les méthodes.
\item SEED du générateur de nombre aléatoires : 25 (choisi arbitrairement).
\end{itemize}

Le nombre de simulations prévues pour pouvoir réaliser les objectifs fixés peut être calculé en multipliant le \textit{nombre des expériences} x \textit{nombre de méthodes} x \textit{nombre de répétitions}. Ce chiffre s'élève donc à 380 exécutions en total. 
Nous diviserons ces exécutions sur plusieurs machines en utilisant la configuration suivante:  

\begin{itemize}
    \item Python 3.9.5
    \item Numpy 1.21
    \item Matplotlib 3.4.2
    \item Pygmo 2.16.1
\end{itemize}

Ces machines utilisent des systèmes d'exploitation différents:
\begin{itemize}
    \item Ubuntu 20.04
    \item CentOS 7.7
    \item Windows 10
\end{itemize}

À noter que la possibilité de porter notre simulateur sur des  environnements de type Linux et Windows sans nécessiter l'apport de changements dans le code source, ou de son encapsulation dans des \textit{containers virtuels}, est un point positif augmentant la facilité d'utilisation de l'outil.

\section{Conclusion}

Nous avons présenté dans ce chapitre la méthodologie utilisée pour la résolution du problème d’exploration d’une zone inconnue avec des contraintes d’énergie. Nous avons expliqué notre modèle en formalisant la fonction objectif sur la base des informations récoltées à partir d'une grille d'occupation.

Nous avons également présenté une nouvelle plateforme de simulation dédiée à l'évaluation et le benchmarking de métaheuristiques ainsi que d'autres algorithmes d'optimisation. Cet outil a été implémenté de façon à abstraire les détails de bas niveau des tâches de navigation, de cartographie et de planification de chemins, et ceci afin de permettre aux chercheurs de se concentrer sur des tâches plus importantes telles que la modélisation de nouveaux algorithmes pour résoudre les problèmes d'exploration ou d'optimisation de trajectoires. Une attention particulière a été donnée à l'intégration de fonctionnalités dédiées à l'automatisation du processus d'expérimentation (paramétrage, répétition des expériences, génération de graphes) ainsi que la facilité du passage à l'échelle en la déployant sans difficulté sur des serveurs Cloud pour simuler des environnements à très large échelle.

Dans le chapitre suivant, nous allons effectuer une série d'expériences afin de valider notre modélisation, et évaluer les performances de l'algorithme xBOA. Nous présenterons également une analyse des résultats et une comparaison entre plusieurs variantes de l'algorithme BOA récemment introduit dans la littérature.

    \chapter{Expériences et analyse}

\startcontents[chapters]
\printmyminitoc{
}

\section{Introduction}





Ce chapitre présente la série d'expériences effectuées afin de valider notre modélisation du problème d'exploration et évaluer les performances de l'algorithme xBOA.

Nous utiliserons notre plateforme de benchmarking afin de comparer les performances de cet algorithme avec les différentes métaheuristiques incluses dans le simulateur (GA, ACO, PSO, CMAES, GWO, ABC) pour la résolution du problème d'exploration de zones inconnues. Nous comparerons également les performances du xBOA avec l'algorithme BOA original et ses autres variantes (MBOA, SABOA, ABOA).

Nous analyserons ensuite les résultats d'une série d'expériences visant à accélérer la vitesse d'exécution de l'algorithme tout en analysant la robustesse des méthodes citées face à la réduction de la taille de la population.

Nous présenterons aussi une expérience pour évaluer l’adaptabilité de notre approche dans un contexte multirobots. Le but étant de valider les capacités du modèle à pouvoir générer un comportement collectif pour les robots sans besoin d'intégrer des mécanismes de synchronisation explicite. Pour finir, nous testerons l'approche sur un robot réel.

\section{Configuration matérielle}

Dans le but d'effectuer les expériences souhaitées, nous avons utilisé trois  machines. 

La première consiste en un ordinateur portable de moyenne gamme avec une mémoire vive de 8Go et un processeur \textit{Intel i7} de 4ème génération. Ce processeur possède 4 cores ayant chacun une fréquence d'horloge de 2.8Ghz. Bien que cet ordinateur inclut aussi une carte graphique, celle-ci n'a pas été utilisée pour les raisons citées dans la section \ref{section:simulator_goals}, ainsi que le fait que les deux autres machines utilisées pour effectuer les expériences ne possédaient pas de capacités de calcul graphique. 

La deuxième machine utilisée consiste en un service cloud de \textit{Google} appelé \textit{Collaboratory} (ou \textit{Google Colab}) offrant une machine virtuelle utilisant un processeur \textit{Intel Xeon} de 2 cores cadencés à 2.3Ghz avec une mémoire vive de 12Go. L'utilisation de ce service est gratuite pour des sessions de calcul dont la durée est inférieure à 12h.

La troisième machine utilisée consiste en un cluster disponible au \textit{Plateau Technique de Calcul Intensif IBN-BAJA} à l'\textit{Université des Sciences et de la Technologie d'Oran Mohamed Boudiaf - USTOMB}. Ce cluster destiné à effectuer des calculs à hautes performances (\textit{High Performance Computing}) possède une mémoire vive de 32Go et 24 processeurs \textit{Intel Xeon} de 6 cores chacun cadencés à 2Ghz. Ce cluster nous a permis d'effectuer des sessions de simulation de longue durée en répétant la même expérience plusieurs fois.

Chaque machine utilise un système d'exploitation différent. Le code a été déployé sur ces machine sans nécessiter de changement ou de compilation, ce qui a démontré la portabilité de notre simulateur sur des systèmes de type Linux et Windows. Le code n'a pas encore été testé sur un système de type iOS mais ne devrait présenter aucun problème de compatibilité. 

\section{Expérience 1: Évaluation de l'algorithme xBOA dans un contexte multirobots}


\begin{figure*}[b!]
    \vspace{1mm}%
    \centering
    \setlength{\abovecaptionskip}{0.3cm} 

     \begin{subfigure}[b]{\textwidth}
         \centering
         \includegraphics[width=\textwidth]{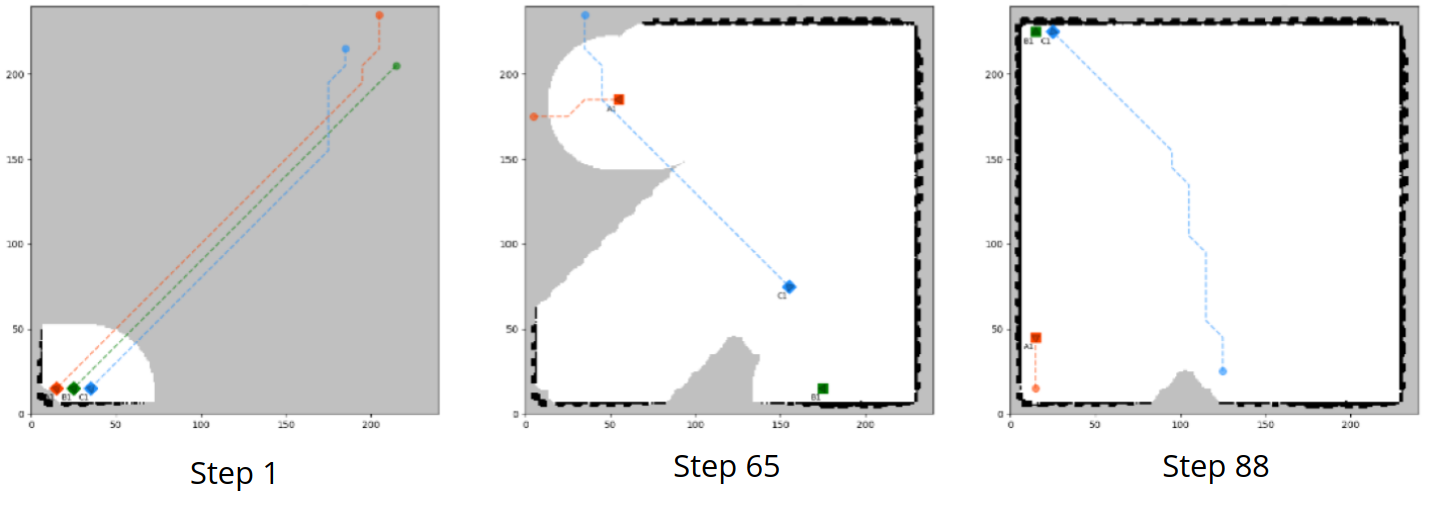}
         \caption{Environnement \textit{Empty Map}}
         \vspace{3mm}%
     \end{subfigure}
    \begin{subfigure}[b]{\textwidth}
         \centering
         \includegraphics[width=\textwidth]{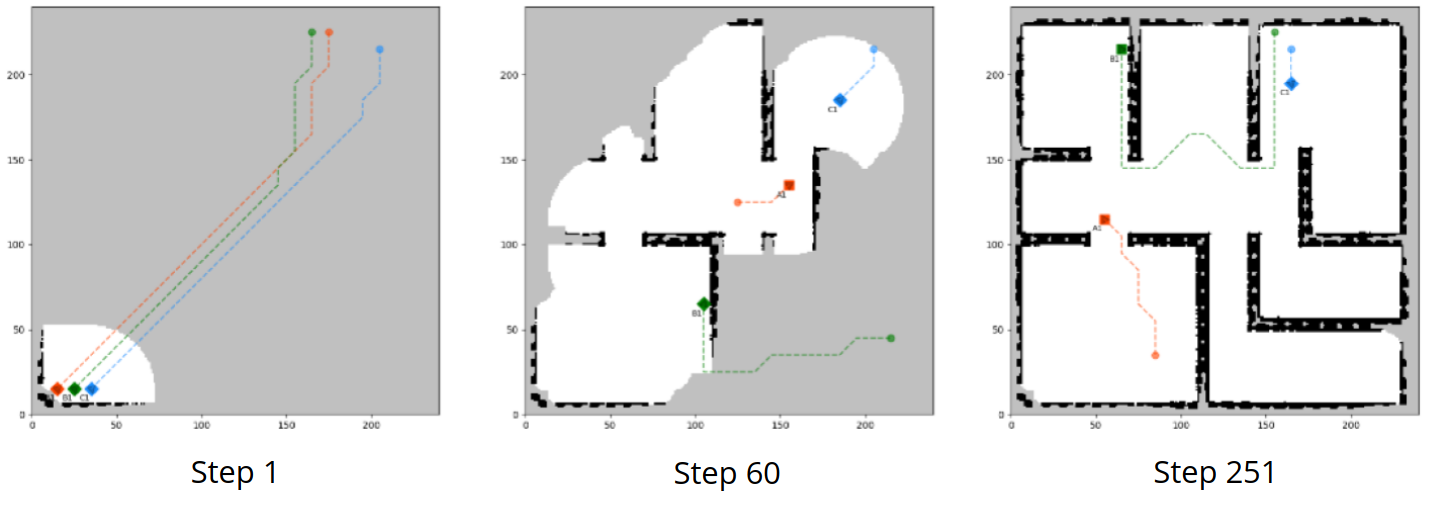}
         \caption{Environnement \textit{House Map}}
     \end{subfigure}
     
    \caption{Progression visuelle du scenario multirobots en utilisant la méthode xBOA avec 3 robots}
    \label{fig:c4.14}
\end{figure*}

La série d’expériences présentée dans cette section vise à valider l’adaptabilité de l’approche dans un scénario multirobots. Selon ce que l’on peut observer sur la figure \ref{fig:c4.14}, les robots déployés à partir de la même position de départ ont pu se disperser dans l’environnement pour explorer la surface entière sans aucun changement nécessaire dans l’algorithme pour coordonner leurs déplacements. Ceci montre la flexibilité de la modélisation proposée pour s’adapter aux scénarios mono et multirobots.

En effet, chaque robot possède sa propre population de solutions candidates et essaie de maximiser sa propre valeur de fitness indépendamment des autres robots. Ces robots n’échangent pas de messages entre eux, mais collaborent passivement en modifiant une carte globale qui est stockée dans une mémoire centrale. Lorsqu'un robot se déplace, il met à jour cette carte partagée afin d'y insérer les obstacles détectés et marquer la zone observée comme étant visitée. Les autres robots vont donc automatiquement éviter cette zone lorsqu'ils planifient les prochaines trajectoires puisque la fonction fitness pénalisera ces régions déjà visitées.


\begin{figure*}[b!]
    \vspace{1mm}%
    \centering
    \setlength{\abovecaptionskip}{0.3cm} 

    \centering
    \includegraphics[width=0.85\textwidth]{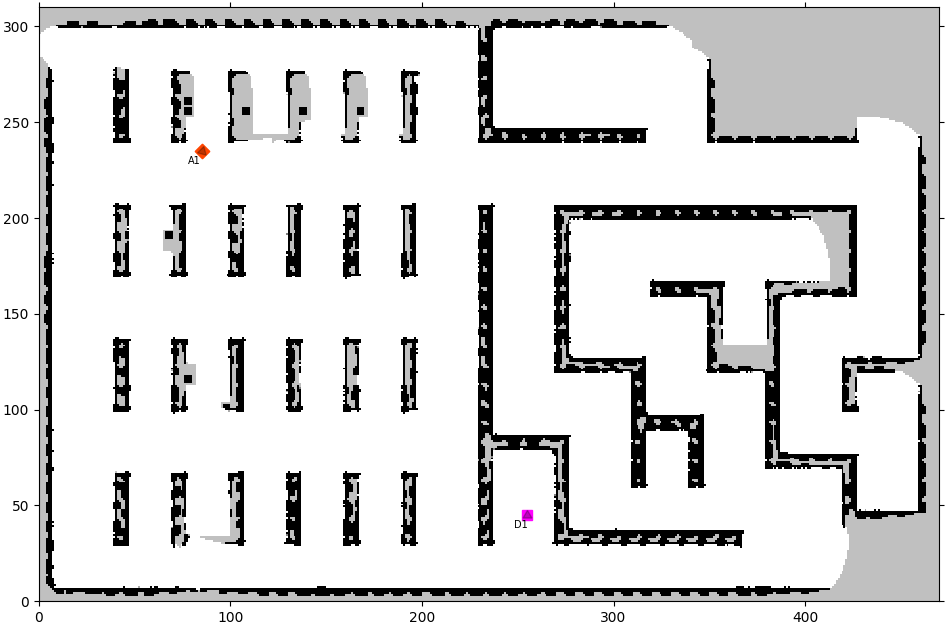}

    \caption{Progression visuelle d'un scénario à grande échelle en utilisant la méthode xBOA avec 2 robots}
    \label{fig:c4.15}
\end{figure*}

Étant donné que les robots ne sont pas conscients de la présence des autres robots, ils n'échangent pas de messages entre eux par rapport à leurs chemins planifiés ou des régions prévues à être visitées dans le futur. Tout ce qu'ils échangent c'est la position des obstacles détectés et les zones déjà visitées auparavant. De ce fait, il arrive que plusieurs robots décident de se diriger vers la même direction puisque cela maximise leur valeur de fitness. Ce comportement est observé au départ de l'expérience lorsqu'ils choisissent tous le chemin diagonal maximisant la zone à observer, ou parfois au milieu de l'expérience tel qu'on peut le voir sur la figure \ref{fig:c4.14}. Cependant, nous observons que dans l'ensemble, le système arrive quand même à faire disperser les robots dans des régions séparées, ce qui a pour effet d'accélérer le temps d'exploration de la zone comparé aux scénarios où on n'utilise qu'un seul robot.

Afin de montrer l'adaptabilité de l'approche à des environnements beaucoup plus larges, nous avons effectué une autre expérience en utilisant une carte inspirée d'une usine ayant une superficie égale à 30x50 mètres. La figure \ref{fig:c4.15} montre un résultat intermédiaire enregistré pendant l'exécution. Cette expérience a été répétée plusieurs fois en changeant le nombre de robots, elle a montré la capacité de l'approche à les pousser à explorer la zone entière même en cas de forte densité d'obstacles ou de l'augmentation de la superficie de l'environnement.

Par ailleurs, cette série d'expériences a permis aussi d'apprécier la facilité d'utilisation de notre simulateur, puisqu'il s'adapte automatiquement à l'ajout de nouveaux scénarios et au changement du nombre de robots. L'utilisateur n'aura qu'à effectuer un paramétrage de l'expérience sans se soucier des mécanismes internes de détection des obstacles, de mise à jour de la carte globale, ou de l'évitement des collisions entre les robots. Ce genre de tâches sont souvent un frein à beaucoup de chercheurs venant de domaines différents puisqu'elles nécessitent d'avoir une bonne compréhension des problématiques de navigation robotique et du fonctionnement des grilles d'occupation.

\section{Expérience 2: Comparaison entre les stratégies d'exploration à court terme et à long terme}

Cette expérience vise à comparer deux stratégies d'exploration. La première consiste à planifier la mission à long terme en sélectionnant plusieurs points de destination au départ de l'exécution ; calculer un chemin pour visiter ces points selon leur ordre de sélection ; puis répéter l'opération après avoir visité tous ces points. Quant à la deuxième stratégie, elle consiste à faire une planification à court terme en ne sélectionnant qu'un seul point à la fois ; le robot devra donc attendre jusqu'à avoir visité sa destination actuelle avant de choisir la prochaine.

\begin{figure*}[ht!]
    \centering

     \begin{subfigure}[b]{\textwidth}
         \centering
         \includegraphics[width=0.24\textwidth]{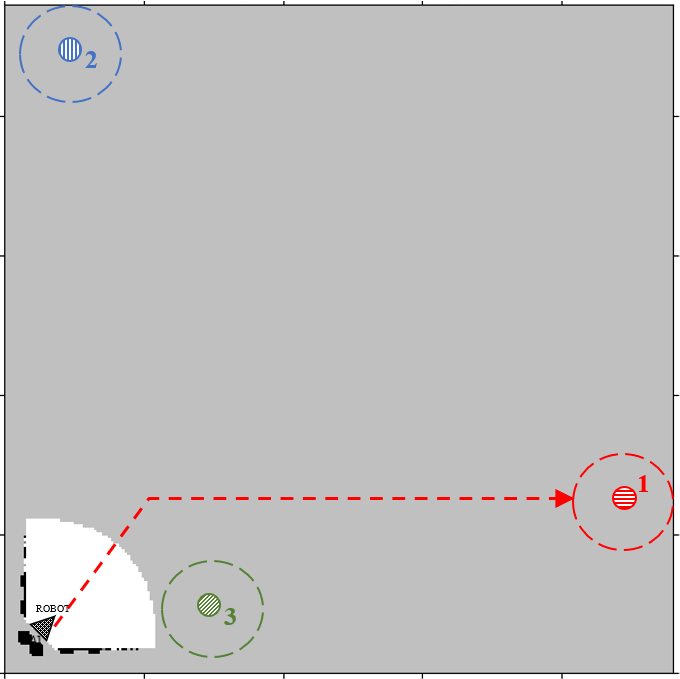}
         \includegraphics[width=0.24\textwidth]{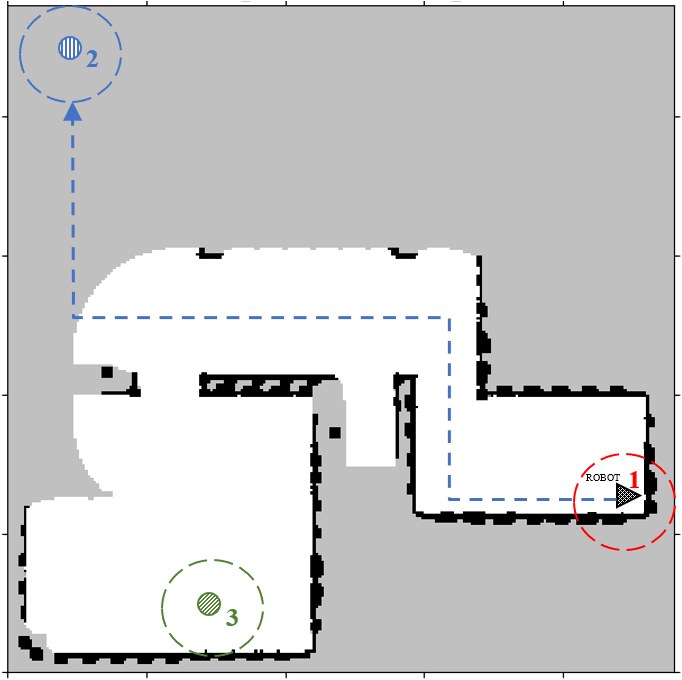}
         \includegraphics[width=0.24\textwidth]{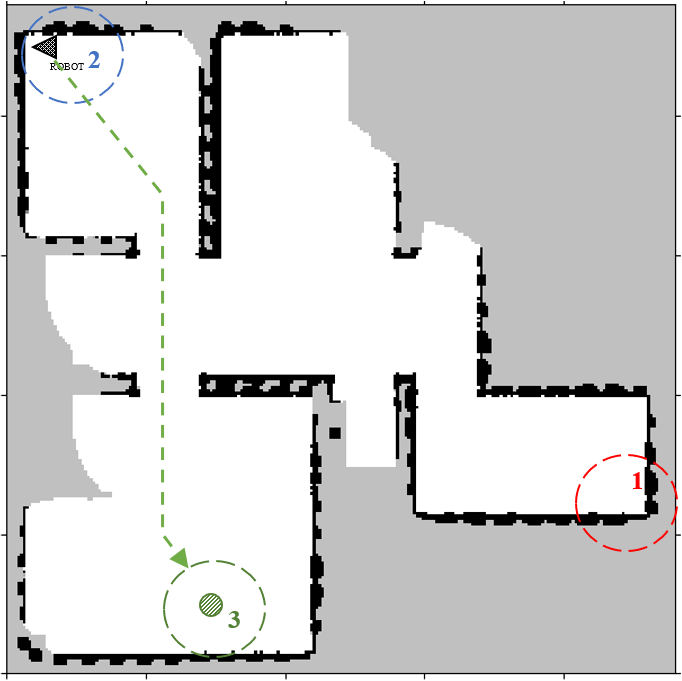}
         \includegraphics[width=0.24\textwidth]{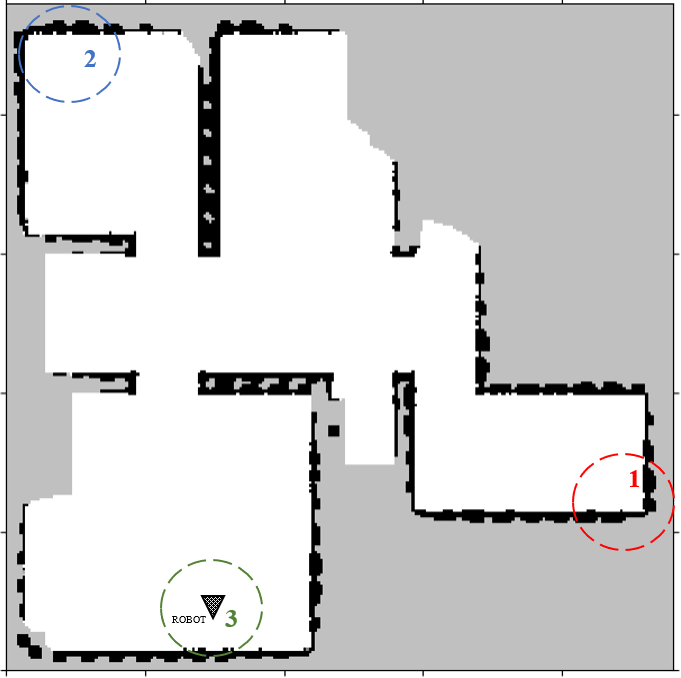}
         \caption{État d'avancement de la mission d'exploration en utilisant la stratégie à long terme (surface explorée 64\%)}
     \end{subfigure}
     \vspace{7mm}
     \begin{subfigure}[b]{\textwidth}
         \centering
         \includegraphics[width=0.24\textwidth]{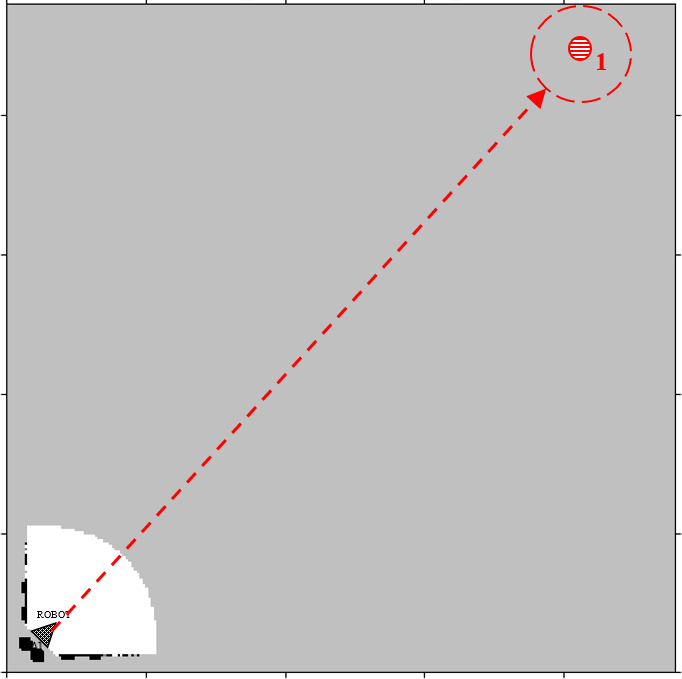}
         \includegraphics[width=0.24\textwidth]{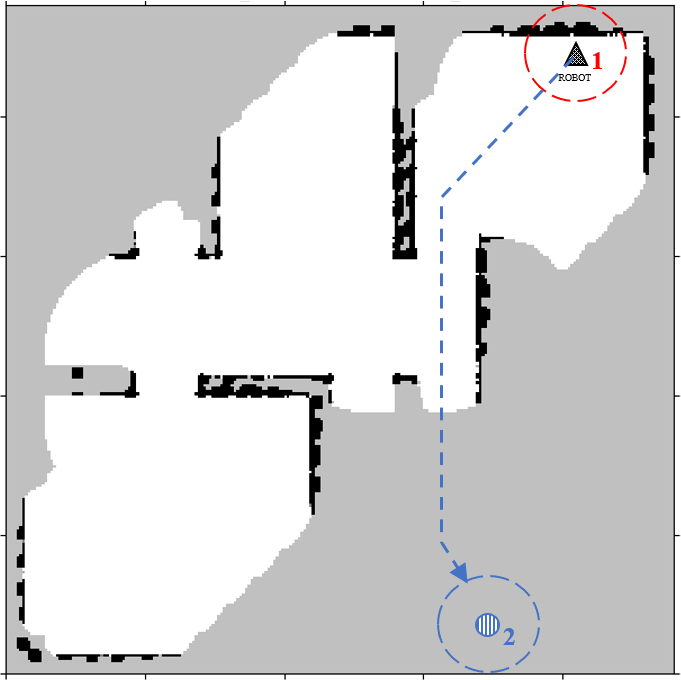}
         \includegraphics[width=0.24\textwidth]{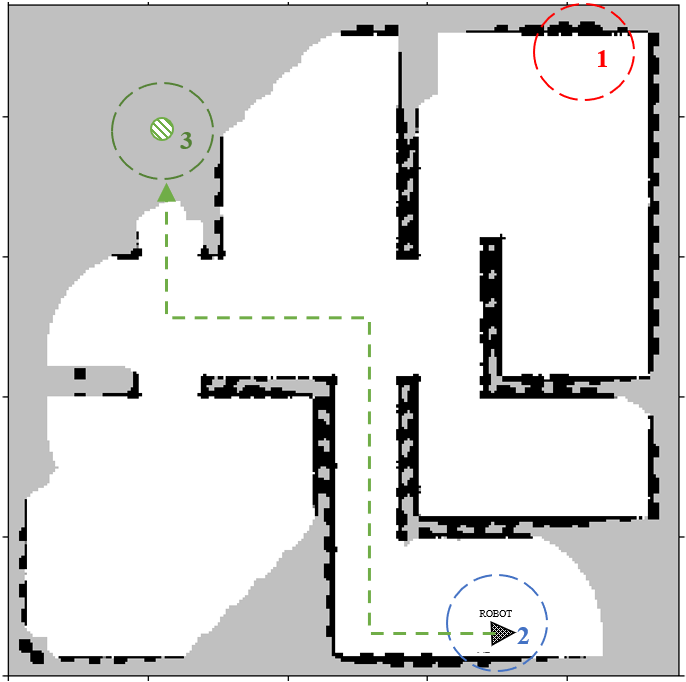}
         \includegraphics[width=0.24\textwidth]{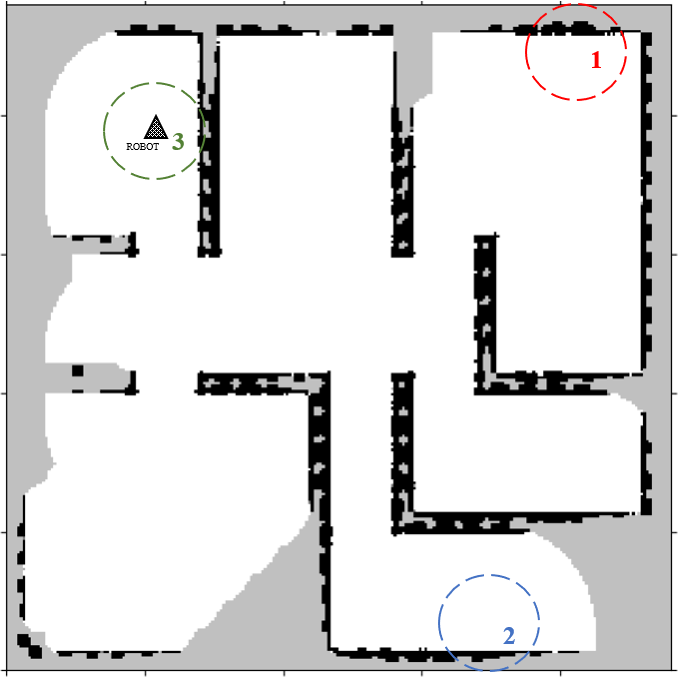}
         \caption{État d'avancement de la mission d'exploration en utilisant la stratégie à court terme (surface explorée 91\%)}
     \end{subfigure}
     \caption{Comparaison entre les deux stratégies d'exploration en utilisant la méthode xBOA}
     \label{fig:c4.3}
\end{figure*}

La figure \ref{fig:c4.3} montre un exemple d'exécution des deux stratégies en utilisant la méthode xBOA. Cette visualisation étape par étape montre que la stratégie à long terme est moins efficace que la stratégie à court terme lorsque l'environnement est inconnu à l'avance. En effet, la planification à long terme de plusieurs points de destination au début de la mission se base sur des informations très limitées puisque le robot n'a aucune connaissance à priori de l'emplacement des murs et des obstacles avant de les avoir détectés grâce à son LIDAR. Le processus d'optimisation choisit donc ces destinations en ne tenant compte que de l'information de distance à vue d'oiseau par rapport au robot. Cependant, lorsque celui-ci commence à se déplacer pour visiter ces points, il met à jour sa carte en collectant les informations des positions des obstacles et des différents chemins possibles. Toutefois, il n'utilisera ces informations que lors de la prochaine phase de sélection après avoir visité tous les points planifiés. En d'autres termes: le robot continuera donc à visiter les destinations prévues initialement sans profiter de l'apport des nouvelles informations récoltées en chemin, et tombera souvent dans une redondance poussant le robot à revisiter une zone déjà explorée auparavant. 

Pour illustrer ce problème de redondance, la figure \ref{fig:c4.3} montre un exemple où la visite du 3ème point en utilisant la stratégie à long terme n'a apporté aucun gain, et ceci à cause du manque de visibilité lors de la phase de sélection. Alors que la stratégie à court terme a permis d’éviter cette redondance en utilisant les informations récoltées lors de la visite du 1er et 2ème point pour pousser le robot à visiter une nouvelle zone.

Dans la stratégie à court terme, le robot ne sélectionne qu’un seul point de destination. Il se déplacera pour le visiter tout en mettant à jour la carte de l’environnement puis resélectionnera un nouveau point en utilisant les nouvelles informations récoltées en chemin. La qualité du choix est donc meilleure puisqu’il élimine les destinations n’apportant aucun gain supplémentaire ou qui sont trop coûteuses par rapport aux gains qu’ils pourront apporter. Cette stratégie est aussi avantageuse dans le cas des environnements dynamiques où les positions des obstacles changent à cause de portes ouvertes et fermées par exemple, ou à cause d’objets déplacés. Le robot pourra prendre en compte les derniers changements de la carte à chaque fois qu’une destination est atteinte.

L’inconvénient de cette stratégie cependant est la nécessité de répéter le processus de sélection des points de destination plus fréquemment. Il est donc nécessaire que le temps d'exécution de ce processus soit court puisque le robot restera en état d’attente le temps de finir ses calculs. 

Aussi nous remarquons que cette stratégie à réussi à explorer presque la totalité de la zone dans l'exemple montré par la figure \ref{fig:c4.3} en visitant 3 points seulement. Elle a pu atteindre un taux d'exploration de 91\% contrairement à la stratégie à long terme qui n'a réussi à explorer que 64\% de la surface de la zone, soit un résultat inférieur de 27\% comparé à la stratégie à court terme.

\section{Expérience 3: Recherche des meilleurs hyperparamètres}

Les métaheuristiques sont sensibles aux choix des paramètres initiaux. Une pratique courante dans la littérature consiste à comparer les performances des méthodes en utilisant les paramètres originaux utilisés par les auteurs dans leurs premières publications \cite{arora19}. Ceci peut être un bon choix si nous comparons ces méthodes en utilisant les fonctions de benchmarking standards ; toutefois, le problème d’exploration de zones inconnues est par nature un problème incrémental et partiellement observable, ce qui peut rendre les paramètres par défaut moins optimaux. 

De ce fait, nous avons effectué une expérience préliminaire afin de rechercher les meilleurs paramètres de chaque méthode pour ce problème en particulier. Ceci nous permutera de comparer les résultats de ces métaheuristiques dans leurs performances optimales et éviter qu’elles soient piégées dans des optimums locaux à cause de mauvaises initialisations.

Il n'existe pas de formule mathématique précise pour trouver les meilleurs hyperparamètres d'une méthode, ceci doit se faire expérimentalement en essayant plusieurs valeurs et comparer leurs résultats. Bien que beaucoup de chercheurs utilisent la méthode manuelle, nous avons préféré utiliser la bibliothèque \textit{Hyperopt} \cite{bergstra13}. Cette bibliothèque permet d’optimiser les paramètres d’entrée d’un algorithme en effectuant une recherche sélective sur une plage de valeurs. Elle offre plusieurs stratégies de recherche dont principalement la méthode de recherche aléatoire (\textit{Random Search} \cite{bergstra12}) et recherche par l'algorithme TPE (\textit{Tree-structured Parzen Estimator} \cite{bergstra13}) qui montrent des résultats meilleurs comparés aux  méthodes classiques telles que la stratégie de recherche par grille (\textit{Grid Search}) \cite{bergstra12}.

Le fonctionnement de base de la bibliothèque \textit{Hyperopt} consiste à exécuter plusieurs fois l'algorithme qu'on veut optimiser en utilisant des valeurs aléatoires pour l'initialisation des hyperparamètres, puis d'adapter ces valeurs selon les résultats obtenus après plusieurs essais. En répétant cette opération un nombre suffisant de fois, la bibliothèque arrive à réduire la plage de paramétrage afin de choisir une combinaison qui donne de bons résultats.

En profitant de la puissance de calcul du cluster de calcul intensif de l'\textit{USTOMB}, nous avons lancé l'optimisation des hyperparamètres de toutes les métaheuristiques incluses dans notre simulateur à raison de 30 essais par méthode en répétant chaque essai 3 fois, ce qui donne un total de 90 exécutions chacune. Le Tableau \ref{tab:4.1} liste les meilleurs paramètres trouvés durant cette expérience, nous nous baserons sur ces valeurs pour effectuer les prochaines expériences.

Certaines méthodes telles que ABC, GWO et CMAES ne requièrent pas de paramétrage manuel. Pour les autres méthodes, les paramètres suivants ont été sélectionnés : 

\begin{itemize}

    \item GA : nous avons utilisé une stratégie de sélection par tournoi de taille 2 et une mutation polynomiale avec un index de distribution de 76. 
    \item PSO : nous avons utilisé une taille de voisinage de 4 avec un facteur d’accélération cognitif supérieur au facteur d’accélération social, ce qui attire chaque particule vers la meilleure position dans son voisinage au lieu de retourner à sa meilleure position trouvée précédemment. 
    \item Variantes de BOA : ces paramètres sont expliqués en détail dans la section \ref{section:BOA_variants}. Les valeurs des meilleurs paramètres sont présentées dans le tableau \ref{tab:4.1}.

\end{itemize}

\begin{table*}[t!]
    \footnotesize
    \centering
    \caption{Valeurs des meilleurs hyperparamètres trouvés après 30 essais}
    \label{tab:4.1}
    
    \begin{tabular}{|c|c|c|c|c|c|}
    \hline
    \rowcolor{headerColor} 
    \textbf{Methods}         & \textbf{Hyperparameters}          & \textbf{Values}          & \textbf{Methods}            & \textbf{Hyperparameters} & \textbf{Values} \\ \hline
    
    \textbf{\begin{tabular}[c]{@{}c@{}}BOA\\ (pop size 20)\end{tabular}} &
      \begin{tabular}[c]{@{}c@{}}Power exponent\\ Sensor modality\\ Switch probability\end{tabular} &
      \begin{tabular}[c]{@{}c@{}}0.547\\ 0.602\\ 0.395\end{tabular} &
      \textbf{\begin{tabular}[c]{@{}c@{}}BOA\\ (pop size 5)\end{tabular}} &
      \begin{tabular}[c]{@{}c@{}}Power exponent\\ Sensor modality\\ Switch probability\end{tabular} &
      \begin{tabular}[c]{@{}c@{}}0.73\\ 0.577\\ 0.331\end{tabular} \\ \hline
    \textbf{\begin{tabular}[c]{@{}c@{}}xBOA\\ (pop size 20)\end{tabular}} &
      \begin{tabular}[c]{@{}c@{}}Power exponent\\ Sensor modality\\ Crossover probability \end{tabular} &
      \begin{tabular}[c]{@{}c@{}}0.905\\ 0.257\\ 0.593\end{tabular} &
      \textbf{\begin{tabular}[c]{@{}c@{}}xBOA\\ (pop size 5)\end{tabular}} &
      \begin{tabular}[c]{@{}c@{}}Power exponent\\ Sensor modality\\ Crossover probability\end{tabular} &
      \begin{tabular}[c]{@{}c@{}}0.994\\ 0.518\\ 0.583\end{tabular} \\ \hline
    \textbf{GA} &
      \begin{tabular}[c]{@{}c@{}}Crossover probability\\ Mutation probability\\ Mutation distribution index\end{tabular} &
      \begin{tabular}[c]{@{}c@{}}0.11\\ 0.215\\ 76.026\end{tabular} &
      \textbf{\begin{tabular}[c]{@{}c@{}}mBOA\\ (pop size 5)\end{tabular}} &
      \begin{tabular}[c]{@{}c@{}}Power exponent\\ Sensor modality\\ Switch probability\end{tabular} &
      \begin{tabular}[c]{@{}c@{}}0.61\\ 0.356\\ 0.762\end{tabular} \\ \hline
    \textbf{PSO} &
      \begin{tabular}[c]{@{}c@{}}Social component coef.\\ Cognitive component coef.\\ Max velocity\\ Inertia weight\end{tabular} &
      \begin{tabular}[c]{@{}c@{}}1.506\\ 3.379\\ 0.329\\ 0.449\end{tabular} &
      \textbf{\begin{tabular}[c]{@{}c@{}}ABOA\\ (pop size 5)\end{tabular}} &
      \begin{tabular}[c]{@{}c@{}}Power exponent\\ Sensor modality\\ Switch probability\\ $\mu$ \end{tabular} &
      \begin{tabular}[c]{@{}c@{}}0.992\\ 0.98\\ 0.983\\ 1.356\end{tabular} \\ \hline
    \textbf{\begin{tabular}[c]{@{}c@{}}ABC, GWO, \\ CMAES\end{tabular}} &
      \multicolumn{2}{c|}{\begin{tabular}[c]{@{}c@{}}No parameters to optimize\\ (auto-tuning)\end{tabular}} &
      \textbf{\begin{tabular}[c]{@{}c@{}}SABOA \\ (pop size 5)\end{tabular}} &
      Switch probability &
      0.237 \\ \hline
    \end{tabular}
\end{table*}

\section{Expérience 4: Comparaison de l'algorithme xBOA avec les métaheuristiques populaires}

Dans cette série d’expériences, nous allons comparer les performances des différentes métaheuristiques incluses dans notre simulateur pour la résolution du problème d'exploration de zones inconnues. 

Selon ce qu’on peut voir sur les Figures \ref{fig:c4.1} et \ref{fig:c4.2} ainsi que le Tableau \ref{tab:4.2}, les méthodes xBOA, PSO, GA et CMAES ont présenté des résultats presque similaires sur la plupart des scénarios, surtout dans le cas de l’exploration à court terme. Ceci signifie que le processus d’optimisation a convergé avec succès vers la solution optimale. Toutefois, nous observons une dégradation de la convergence de BOA et GWO à certaines périodes provoquées par leur blocage dans un optimum local poussant le robot à revisiter une zone déjà explorée pendant un certain intervalle de temps. Cependant, ces méthodes finissent par échapper à ce minimum local après un certain nombre de pas et rattraper leur retard. 

Il est important de noter que la métrique du « nombre de pas » (\textit{step number}) correspond au nombre de mouvements et de rotations effectués par le robot. Il est fortement corrélé avec la durée de la mission, mais de manière non linéaire puisque le temps nécessaire pour effectuer une rotation à 45° est différent de celui requis pour se déplacer un mètre en avant.

Nous avons choisi d’analyser le nombre de pas au lieu du temps d’exploration à cause de l’importance du facteur énergie dans le domaine de la robotique ; le nombre de mouvements qu’un robot peut effectuer est limité par la capacité de sa batterie. Si dans notre cas le robot avait une capacité de batterie pour effectuer seulement 100 pas par exemple, il ne pourra explorer que 80\% de la première scène (\textit{Empty map}) en utilisant l’algorithme BOA, au lieu de 85\% s’il utilisait xBOA, PSO ou CMAES. L’algorithme BOA apparaît donc moins efficace que les autres méthodes en considérant ce critère d’évaluation.

\begin{table*}[ht!]
    \footnotesize
    \centering
    \caption{\label{table:3}Taux d'exploration obtenus à la fin de la mission d'exploration}
    \label{tab:4.2}
    
    \begin{tabular}{|r|c|c|c|c|c|c|}
    \hline
    \rowcolor{headerColor}
      \multicolumn{7}{|c|}{\textit{Short-term exploration}} \\ \hline
    \multicolumn{1}{|l|}{} &
      \multicolumn{3}{c|}{\cellcolor[HTML]{C5E0B3}\textit{Empty map}} &
      \multicolumn{3}{|c|}{\cellcolor[HTML]{C5E0B3}\textit{House map}} \\ \hline
    \textit{\textbf{Method}} &
      \cellcolor[HTML]{FFCE93}\textit{Average} &
      \cellcolor[HTML]{FFCE93}\textit{Min} &
      \cellcolor[HTML]{FFCE93}\textit{Max} &
      \multicolumn{1}{|c|}{\cellcolor[HTML]{FFCE93}\textit{Average}} &
      \cellcolor[HTML]{FFCE93}\textit{Min} &
      \cellcolor[HTML]{FFCE93}\textit{Max} \\ \hline
    \textit{ABC} &
      98.12 &
      97.22 &
      \textbf{$\geq$ 99} &
      \multicolumn{1}{|c|}{90.1} &
      88.36 &
      92.7 \\ \hline
    \textit{BOA} &
      \cellcolor[HTML]{EFEFEF}96 &
      \cellcolor[HTML]{EFEFEF}94.79 &
      \cellcolor[HTML]{EFEFEF}\textbf{$\geq$ 99} &
      \multicolumn{1}{|c|}{\cellcolor[HTML]{EFEFEF}\textbf{93.4}} &
      \cellcolor[HTML]{EFEFEF}91.84 &
      \cellcolor[HTML]{EFEFEF}95.13 \\ \hline
    \textit{CMAES} &
      \textbf{98.49} &
      95.66 &
      \textbf{$\geq$ 99} &
      \multicolumn{1}{|c|}{93.29} &
      89.4 &
      \textbf{96.18} \\ \hline
    GA &
      \cellcolor[HTML]{EFEFEF}97.38 &
      \cellcolor[HTML]{EFEFEF}82.81 &
      \cellcolor[HTML]{EFEFEF}\textbf{$\geq$ 99} &
      \multicolumn{1}{|c|}{\cellcolor[HTML]{EFEFEF}93} &
      \cellcolor[HTML]{EFEFEF}\textbf{92.01} &
      \cellcolor[HTML]{EFEFEF}94.96 \\ \hline
    \textit{GWO} &
      93.03 &
      92.88 &
      93.05 &
      \multicolumn{1}{|c|}{67.7} &
      67.7 &
      67.7 \\ \hline
    \textit{PSO} &
      \cellcolor[HTML]{EFEFEF}98.16 &
      \cellcolor[HTML]{EFEFEF}\textbf{97.39} &
      \cellcolor[HTML]{EFEFEF}\textbf{$\geq$ 99} &
      \multicolumn{1}{|c|}{\cellcolor[HTML]{EFEFEF}93.19} &
      \cellcolor[HTML]{EFEFEF}88.71 &
      \cellcolor[HTML]{EFEFEF}94.79 \\ \hline
    \textit{xBOA} &
      98.23 &
      93.92 &
      \textbf{$\geq$ 99} &
      \multicolumn{1}{|c|}{91.63} &
      82.63 &
      94.44 \\ \hline
    \rowcolor{headerColor}
      \multicolumn{7}{|c|}{\textit{Long-term exploration}} \\ \hline
    \multicolumn{1}{|l|}{\textbf{Method}} &
      \cellcolor[HTML]{FFCE93}\textit{Average} &
      \cellcolor[HTML]{FFCE93}\textit{Min} &
      \cellcolor[HTML]{FFCE93}\textit{Max} & \multicolumn{1}{|c|}{\cellcolor[HTML]{FFCE93}\textit{Average}}
       &
      \cellcolor[HTML]{FFCE93}\textit{Min} &
      \cellcolor[HTML]{FFCE93}\textit{Max} \\ \hline
    \textit{ABC} &
      \textbf{98.02} &
      \textbf{96.52} &
      \textbf{$\geq$ 99} &
      \multicolumn{1}{|c|}{86.38} &
      80.38 &
      92.18 \\ \hline
    \textit{BOA} &
      \cellcolor[HTML]{EFEFEF}93.87 &
      \cellcolor[HTML]{EFEFEF}91.49 &
      \cellcolor[HTML]{EFEFEF}97.74 &
      \multicolumn{1}{|c|}{\cellcolor[HTML]{EFEFEF}84.61} &
      \cellcolor[HTML]{EFEFEF}73.26 &
      \cellcolor[HTML]{EFEFEF}92.36 \\ \hline
    \textit{CMAES} &
      94.16 &
      83.85 &
      \textbf{$\geq$ 99} &
      \multicolumn{1}{|c|}{82.23} &
      77.43 &
      86.45 \\ \hline
    \textit{GA} &
      \cellcolor[HTML]{EFEFEF}97.63 &
      \cellcolor[HTML]{EFEFEF}\textbf{96.52} &
      \cellcolor[HTML]{EFEFEF}\textbf{$\geq$ 99} &
      \multicolumn{1}{|c|}{\cellcolor[HTML]{EFEFEF}87.67} &
      \cellcolor[HTML]{EFEFEF}79.86 &
      \cellcolor[HTML]{EFEFEF}\textbf{95.48} \\ \hline
    \textit{GWO} &
      93.4 &
      93.4 &
      93.4 &
      \multicolumn{1}{|c|}{83.68} &
      83.68 &
      83.68 \\ \hline
    \textit{PSO} &
      \cellcolor[HTML]{EFEFEF}96.44 &
      \cellcolor[HTML]{EFEFEF}92.88 &
      \cellcolor[HTML]{EFEFEF}\textbf{$\geq$ 99} &
      \multicolumn{1}{|c|}{\cellcolor[HTML]{EFEFEF}87.06} &
      \cellcolor[HTML]{EFEFEF}\textbf{84.2} &
      \cellcolor[HTML]{EFEFEF}90.97 \\ \hline
    \textit{xBOA} &
      96.37 &
      93.57 &
      98.61 &
      \multicolumn{1}{|c|}{\textbf{89.9}} &
      79.16 &
      94.1 \\ \hline
    \end{tabular}
\end{table*}

\afterpage{

\begin{figure*}[pht!]
    \centering
    \setlength{\abovecaptionskip}{0.3cm} 

    \vspace{8mm}%
     \begin{subfigure}[b]{0.78\textwidth}
         \centering
         \includegraphics[width=\textwidth]{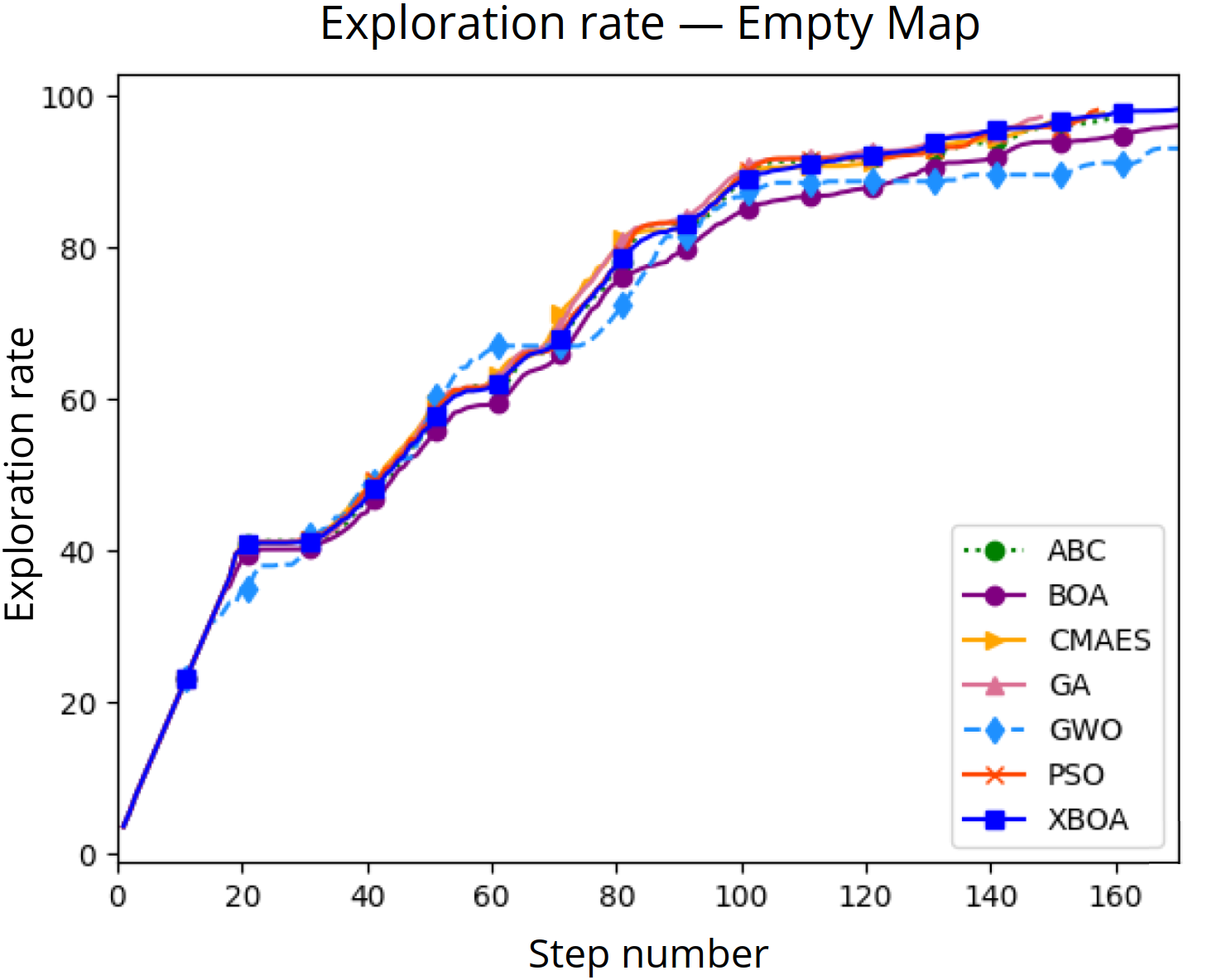}
         \hspace{10mm}
         \vspace{2mm}%
     \end{subfigure}
     
    \begin{subfigure}[b]{0.74\textwidth}
         \centering
         \includegraphics[width=\textwidth]{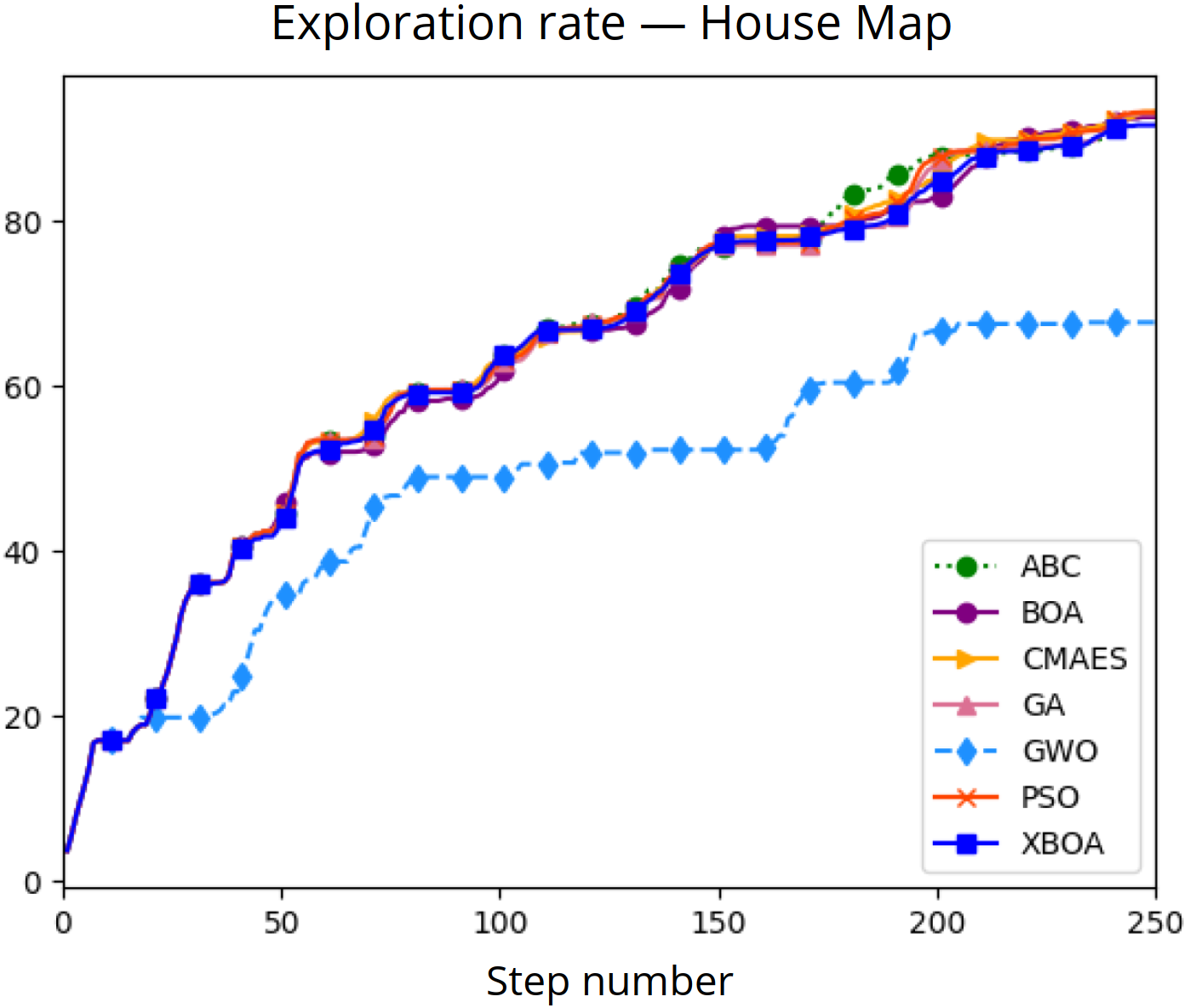}
     \end{subfigure}

     \vspace{1mm}%
     \textit{*Les résultats présentent les valeurs moyennes de 10 exécutions}
     
    \caption{Résultats de la simulation de la mission d'exploration  en utilisant la stratégie à court terme}
    \label{fig:c4.1}
\end{figure*}

\begin{figure*}[pht!]
    \centering
    \setlength{\abovecaptionskip}{0.3cm} 

     \begin{subfigure}[b]{0.75\textwidth}
         \centering
         \includegraphics[width=\textwidth]{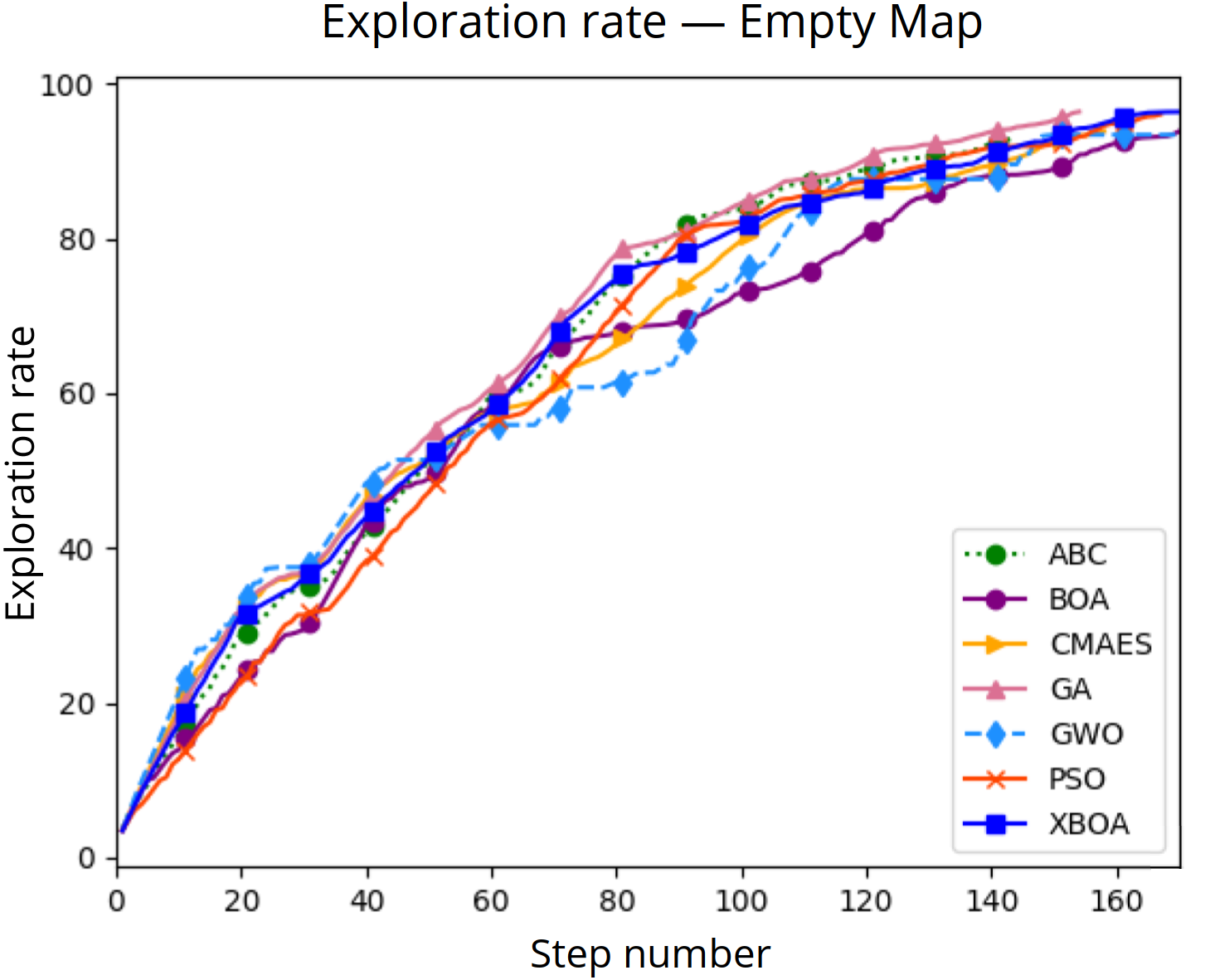}
         \vspace{2mm}%
     \end{subfigure}
    \hspace{5mm}%
    \begin{subfigure}[b]{0.715\textwidth}
         \centering
         \includegraphics[width=\textwidth]{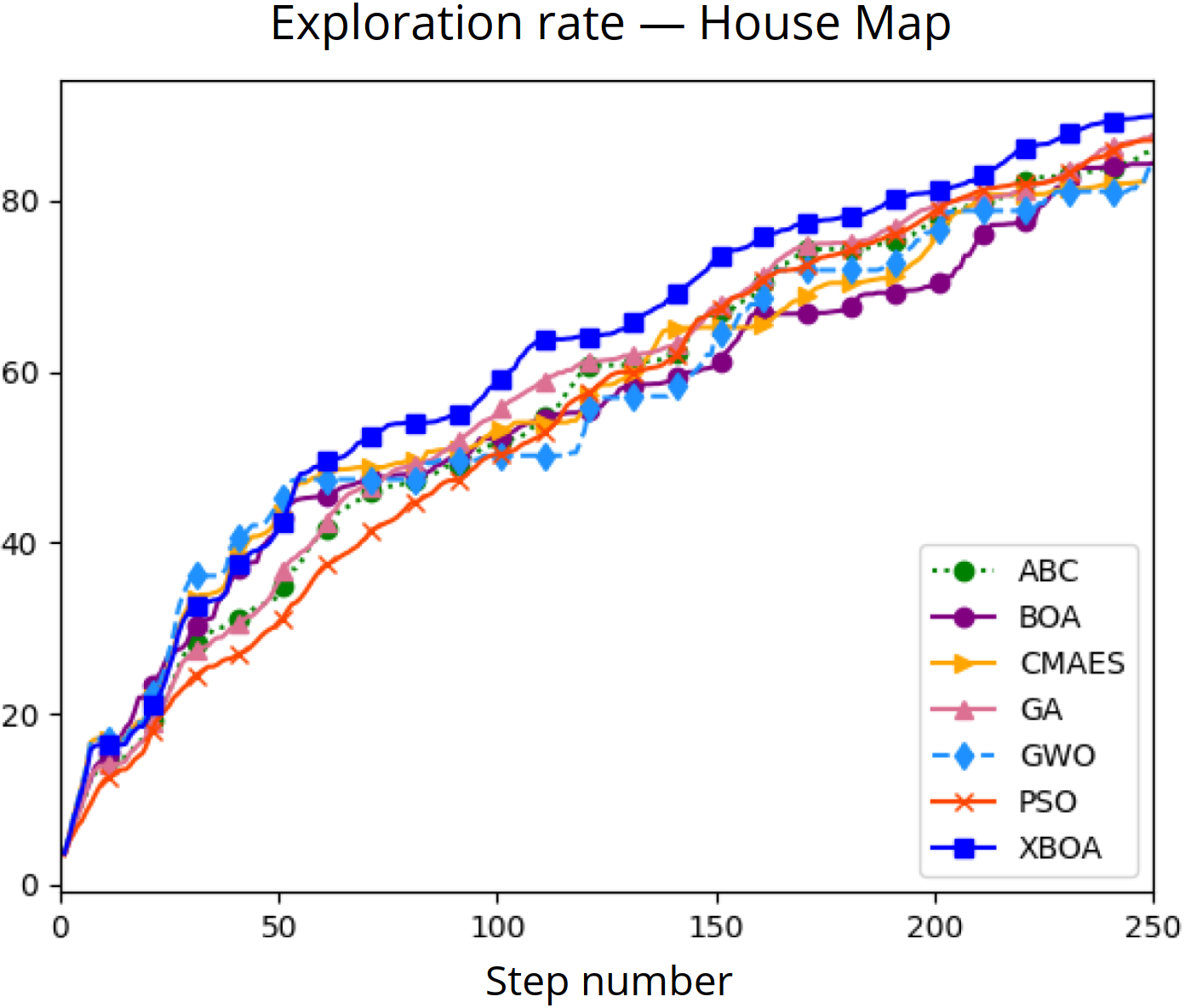}
     \end{subfigure}

     \vspace{1mm}%
     \textit{*Les résultats présentent les valeurs moyennes de 10 exécutions}
     
    \caption{Résultats de la simulation de la mission d'exploration en utilisant la stratégie à long terme}
    \label{fig:c4.2}
\end{figure*}

}

\begin{figure*}[pht!]
    \centering
    \setlength{\abovecaptionskip}{0.3cm} 

     \begin{subfigure}[b]{0.75\textwidth}
         \centering
         \includegraphics[width=\textwidth]{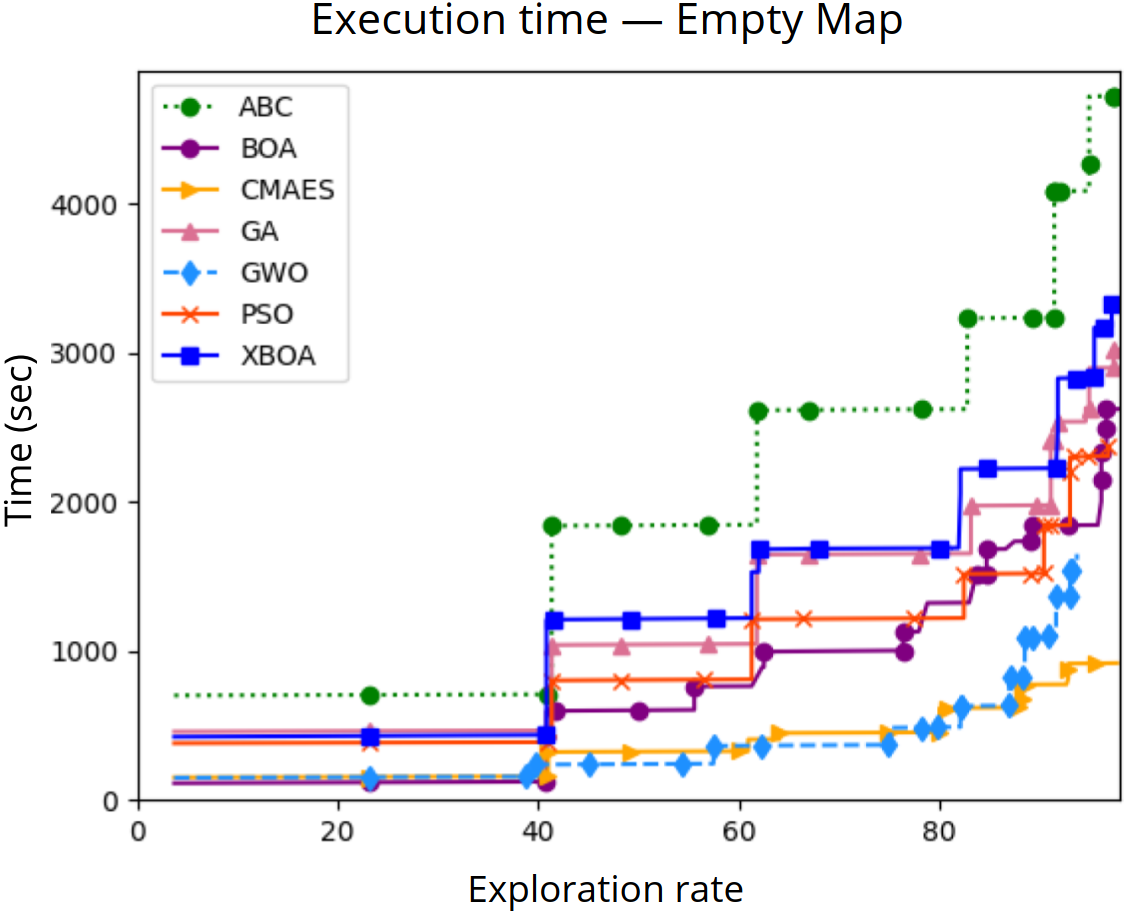}
         \vspace{2mm}%
     \end{subfigure}
     \hspace{5mm}%
    \begin{subfigure}[b]{0.715\textwidth}
         \centering
         \includegraphics[width=\textwidth]{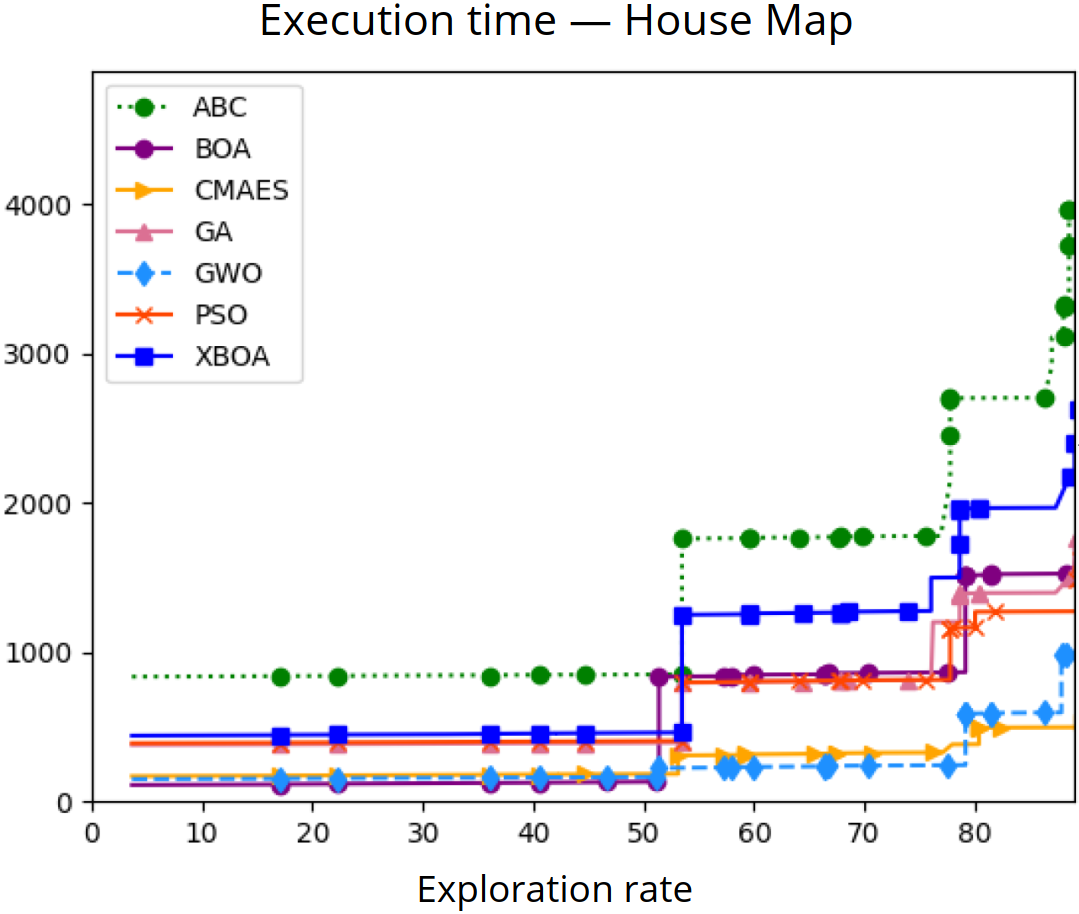}
     \end{subfigure}

     \vspace{1mm}%
     \textit{*Les résultats présentent les valeurs moyennes de 3 exécutions}
     
    \caption{Comparaison de la durée totale de la mission d'exploration pour la stratégie à court terme}
    \label{fig:c4.4}
\end{figure*}

\begin{figure*}[pht!]
    \centering
    \setlength{\abovecaptionskip}{0.3cm} 

     \begin{subfigure}[b]{0.75\textwidth}
         \centering
         \includegraphics[width=\textwidth]{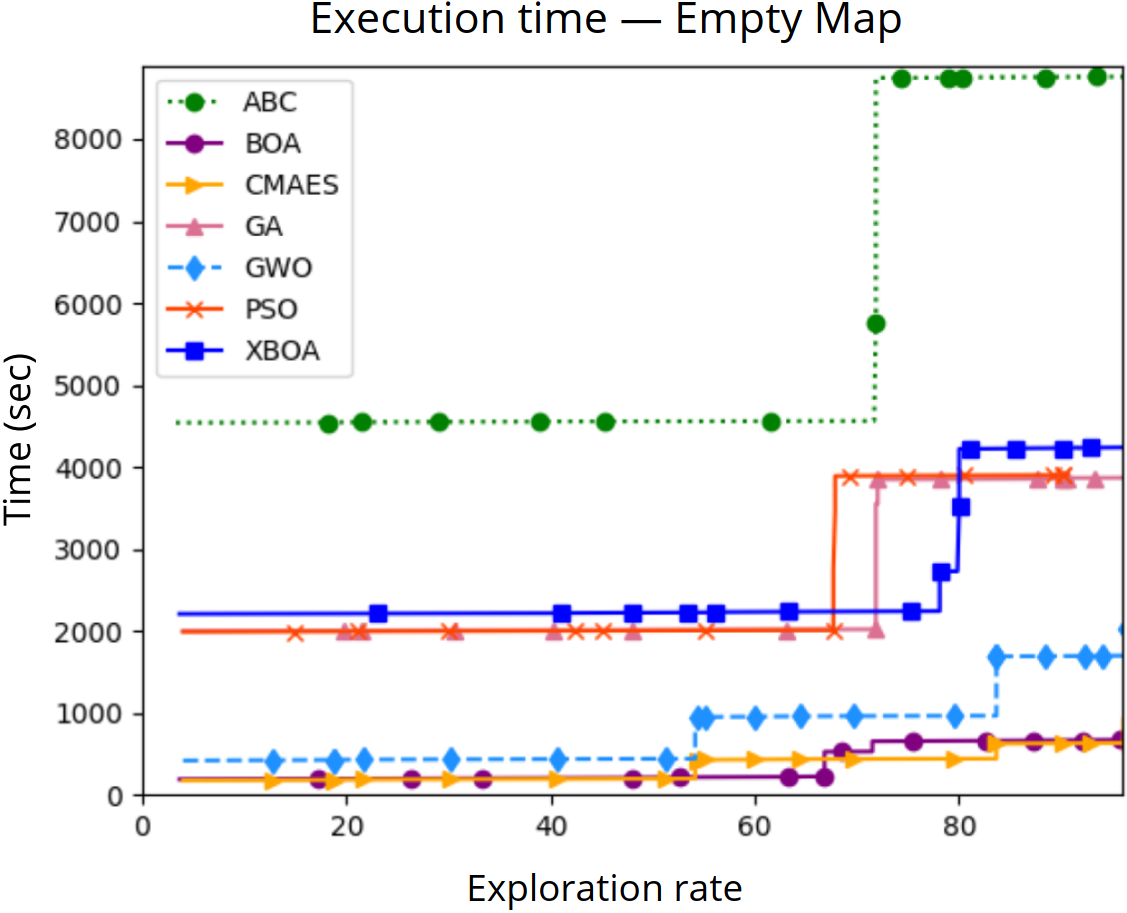}
         \vspace{2mm}%
     \end{subfigure}
     \hspace{5mm}%
    \begin{subfigure}[b]{0.715\textwidth}
         \centering
         \includegraphics[width=\textwidth]{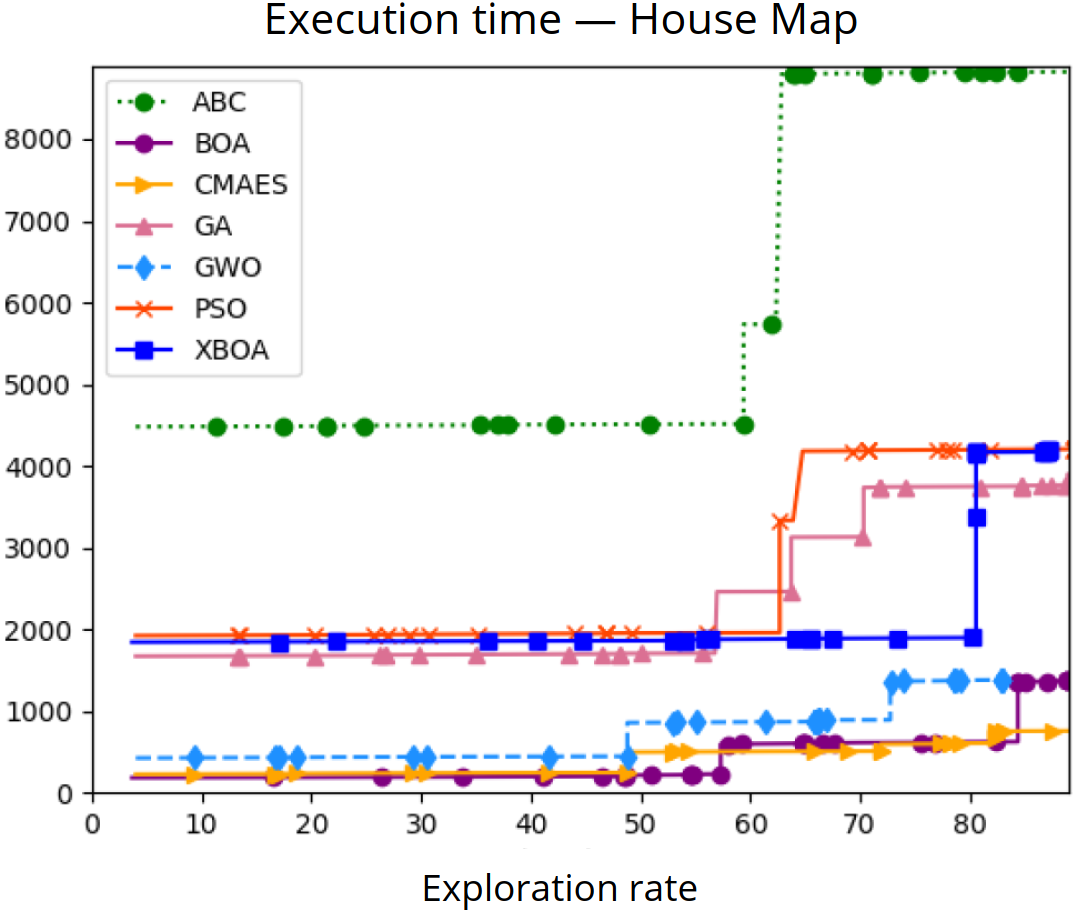}
     \end{subfigure}
     
     \vspace{1mm}%
     \textit{*Les résultats présentent les valeurs moyennes de 3 exécutions}
     
    \caption{Comparaison de la durée totale de la mission d'exploration pour la stratégie à long terme}
    \label{fig:c4.5}
\end{figure*}

Les Figures \ref{fig:c4.4} et \ref{fig:c4.5} nous permettent d’analyser le temps de calcul total pour chaque méthode durant la mission d’exploration. Étant donné que les processeurs sont une ressource coûteuse en termes de prix de fabrication et d’énergie consommée, la tendance dans le domaine des systèmes multirobots est de réduire au maximum le temps de calcul requis pour accomplir la mission afin de permettre l’utilisation de processeurs plus petits et moins gourmands. La méthode idéale devrait donner une qualité d’exploration maximale tout en nécessitant un minimum de temps de calcul. Toutefois, il est rarement le cas d’obtenir un tel résultat, ce qui nous contraint souvent à trouver un compris entre ces deux critères de comparaisons. 

Nous remarquons dans les résultats présentés par ces figures la présence de plusieurs pics sur l’axe du temps de calcul. Ces pics correspondent au temps requis par le robot pour calculer les prochains points à visiter en utilisant l’une des métaheuristiques, sachant que le robot est figé pendant ce temps puisqu’il n’a pas encore de point de destination pour pouvoir planifier un chemin de déplacement. Nous remarquons aussi que le temps de calcul pendant le déplacement du robot est relativement négligeable, ceci s’explique par le fait que le robot ne nécessite pas d’effectuer des opérations de calculs compliquées pendant le déplacement puisque cette phase se limite à mesurer la distance avec l’obstacle le plus proche pour éviter les collisions ainsi que la mise à jour des probabilités d’occupation dans la carte.

\begin{figure*}[pht!]
    \centering
    \setlength{\abovecaptionskip}{0.3cm} 

     \begin{subfigure}[b]{0.75\textwidth}
         \centering
         \includegraphics[width=\textwidth]{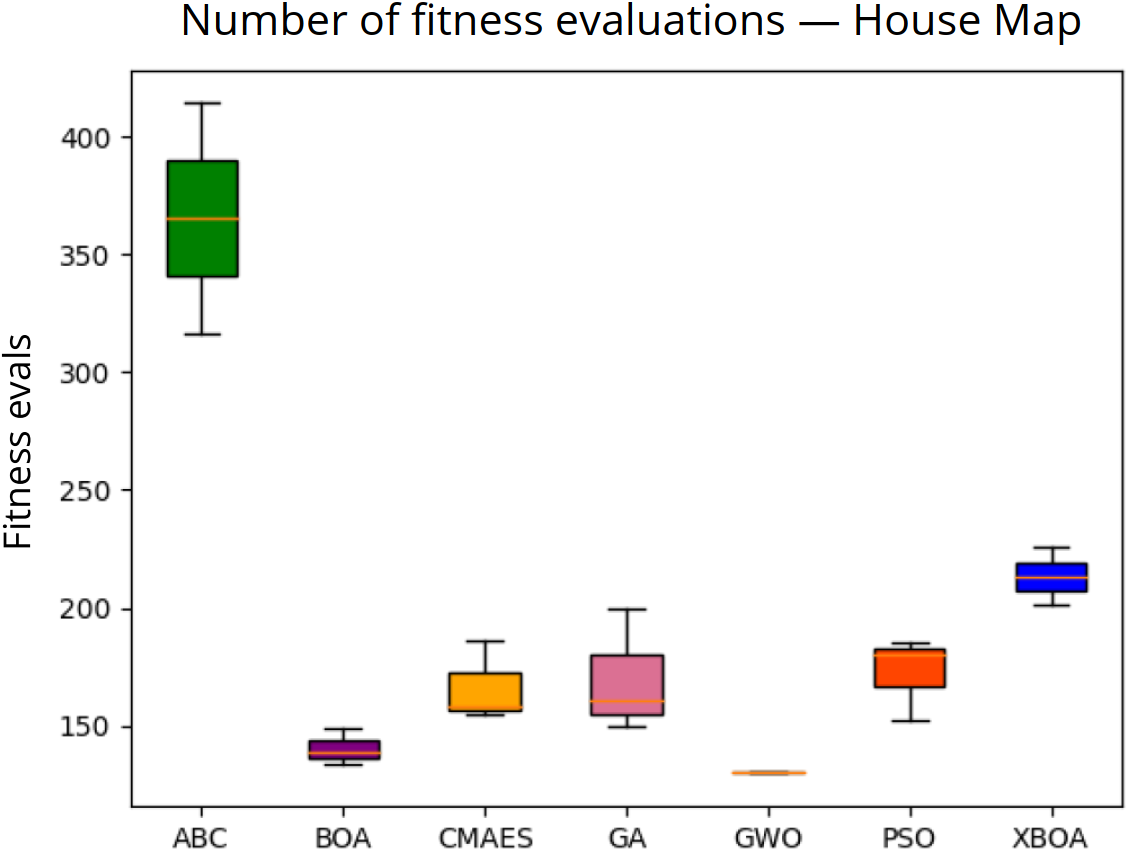}
         \vspace{2mm}%
         \caption{Nombre d'évaluations de la fonction de fitness}
         \vspace{10mm}%
     \end{subfigure}
     \vspace{10mm}%
    \begin{subfigure}[b]{0.75\textwidth}
         \centering
         \includegraphics[width=\textwidth]{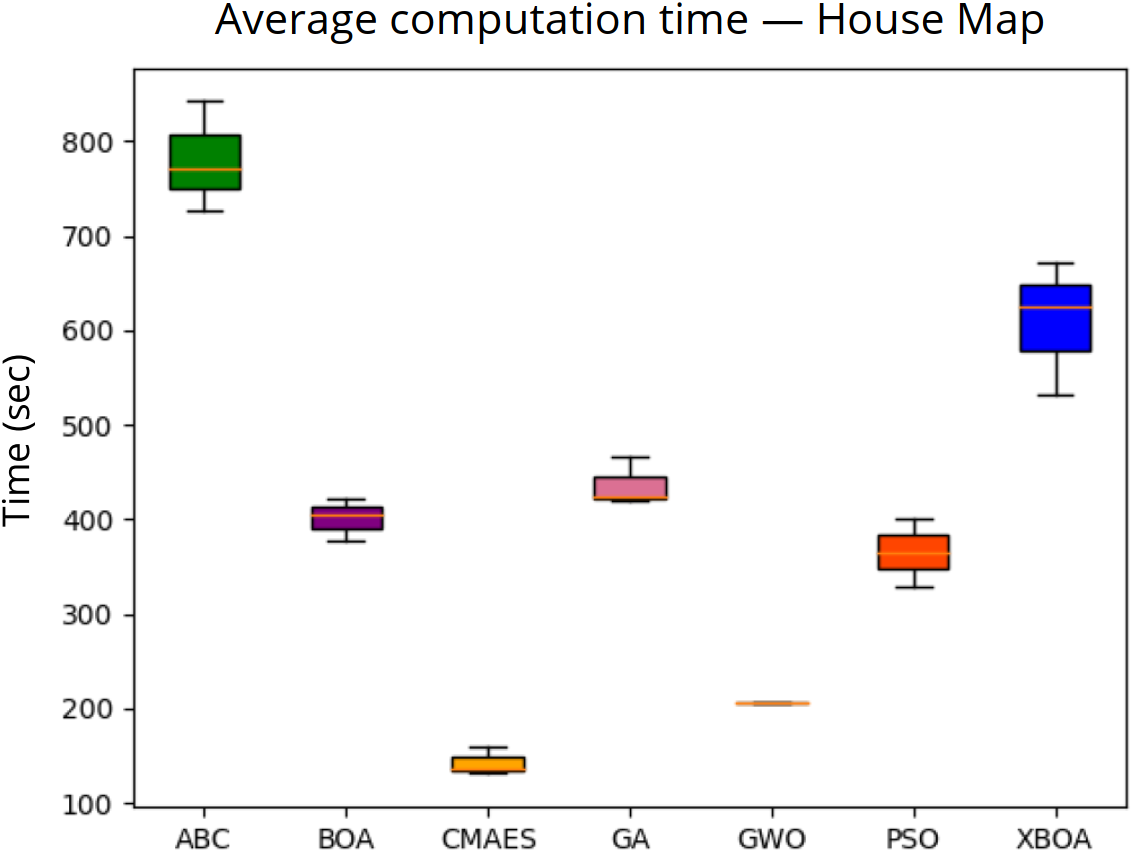}
         \vspace{2mm}%
         \caption{Temps moyen de calcul pour chaque métaheuristique}
     \end{subfigure}

     \vspace{1mm}%
     \textit{*Les résultats présentent les valeurs moyennes de 3 exécutions}
     
    \caption{Nombre d'évaluations de la fonction de fitness et temps moyen de calcul pour chaque métaheuristique}
    \label{fig:c4.6}
\end{figure*}

Bien que les méthodes xBOA, GA et ABC donnent de meilleurs résultats en termes de taux d’exploration comparés à BOA, elles nécessitent un temps de calcul plus long. Ceci s’explique par la simplicité des opérations de l’algorithme BOA et sa complexité réduite comparée aux autres méthodes. Les résultats présentés dans la Figure \ref{fig:c4.6} renforcent cette explication ; nous observons clairement que le temps de calcul de la méthode BOA est inférieur à celui de xBOA, GA et ABC, et que le nombre d’appels à la fonction de fitness (\textit{fitness evaluations}) est inférieur. 

Nous observons aussi que la méthode ABC requiert un temps de calcul considérable dans toutes les expériences, qui est supérieur au double du temps requis par les autres méthodes. Ceci est causé par la nature de l'algorithme ABC qui est divisé en 3 phases dont chacune nécessite la réévaluation des individus, ce qui conduit à un grand nombre d’exécutions de la fonction objectif. Cette lenteur le rend inadapté aux scénarios en situation réelle, car les robots ne doivent pas rester immobilisés pendant une longue période. 

Par ailleurs, nous remarquons que la méthode CMAES domine les autres méthodes sur le critère du temps de calcul, mais elle est parfois dominée par la méthode BOA en termes de taux d’exploration, alors que cette dernière est elle-même dominée par xBOA  dans tous les scénarios sur ce même critère. La méthode GWO est elle aussi dominée par xBOA sur ce critère, mais elle enregistre un temps de calcul plus rapide. 


Il est important de noter à partir de la figure \ref{fig:c4.6} que le temps de calcul moyen pour sélectionner le prochain point de destination est relativement long ; moyennant 150 à 450 secondes durant lesquelles le robot est à l’arrêt en attendant le résultat du processus d’optimisation. Ceci n’est pas recommandé pour des scénarios où le temps est un facteur critique tel que les missions de sauvetage par exemple. Deux potentielles solutions sont possibles dans ce cas. La première consiste à profiter des fonctionnalités de parallélisme offertes par les processeurs modernes pour évaluer plusieurs individus de la population en même temps. Des résultats préliminaires nous ont permis de réduire de moitié le temps d'exécution en utilisant cette technique. 

La deuxième stratégie consiste à réduire la taille de la population, ce qui revient donc à réduire le nombre de solutions candidates évaluées par la fonction objectif. Dans la série d’expériences suivante, nous allons analyser l’impact de cette deuxième stratégie sur la qualité des solutions générées.

\section{Expérience 5: Evaluation de la robustesse de l'algorithme xBOA face à la réduction de taille de la population}

\begin{table}[b]
    \small
    \centering
    \caption{\label{tab:4.3}Evolution du taux d'exploration selon la taille de la population en utilisant l'algorithme BOA}
    
    \begin{tabular}{|p{20mm}|p{15mm}|p{15mm}|p{15mm}|}
    \hline
    \rowcolor{headerColor}
    \multicolumn{4}{|c|}{\textit{Short-term exploration - House map}} \\ \hline
    \multicolumn{1}{|c|}{\textbf{\textit{Pop} size}} & \multicolumn{1}{c|}{\cellcolor[HTML]{FFCE93}\textit{Average}} & \multicolumn{1}{c|}{\cellcolor[HTML]{FFCE93}\textit{Min}} & \multicolumn{1}{c|}{\cellcolor[HTML]{FFCE93}\textit{Max}} \\ \hline
    
    \multicolumn{1}{|r|}{\textit{05 butterflies}} & \multicolumn{1}{c|}{86.11} & \multicolumn{1}{c|}{81.77} & \multicolumn{1}{c|}{\textbf{95.13}} \\ \hline
    
    \multicolumn{1}{|r|}{\textit{10 butterflies}} & \multicolumn{1}{c|}{\cellcolor[HTML]{EFEFEF}88.66}  & \multicolumn{1}{c|}{\cellcolor[HTML]{EFEFEF}79.68} & \multicolumn{1}{c|}{\cellcolor[HTML]{EFEFEF}\textbf{95.13}}  \\ \hline
    
    \multicolumn{1}{|l|}{\textit{20 butterflies}} & \multicolumn{1}{c|}{\textbf{92.84}} & \multicolumn{1}{c|}{\textbf{89.75}} & \multicolumn{1}{c|}{94.27}  \\ \hline
\end{tabular}
\end{table}

\begin{figure*}[b!] 
    \vspace{1mm}%
    \centering
    \setlength{\abovecaptionskip}{0.3cm} 

     \begin{subfigure}[b]{0.49\textwidth}
         \centering
         \includegraphics[width=\textwidth]{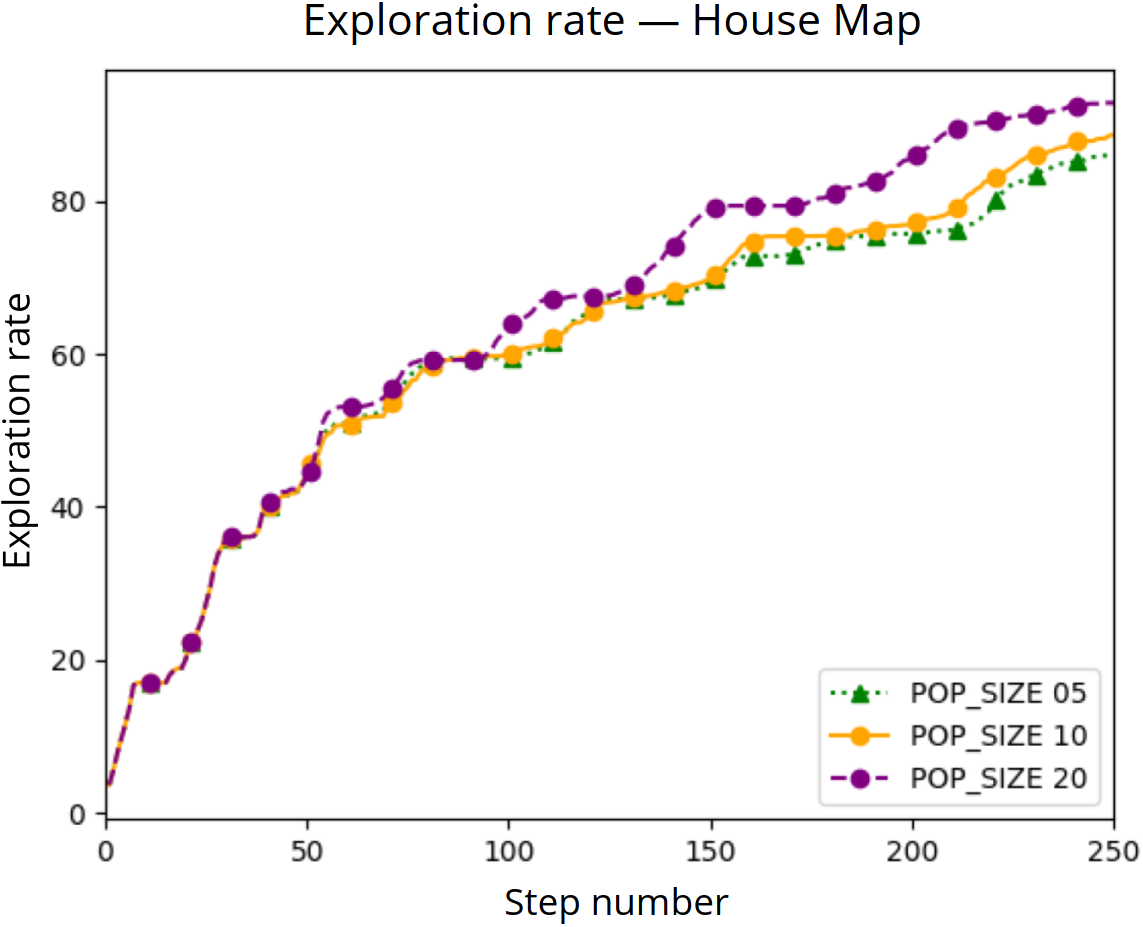}
     \end{subfigure}
    \begin{subfigure}[b]{0.49\textwidth}
         \centering
         \includegraphics[width=\textwidth]{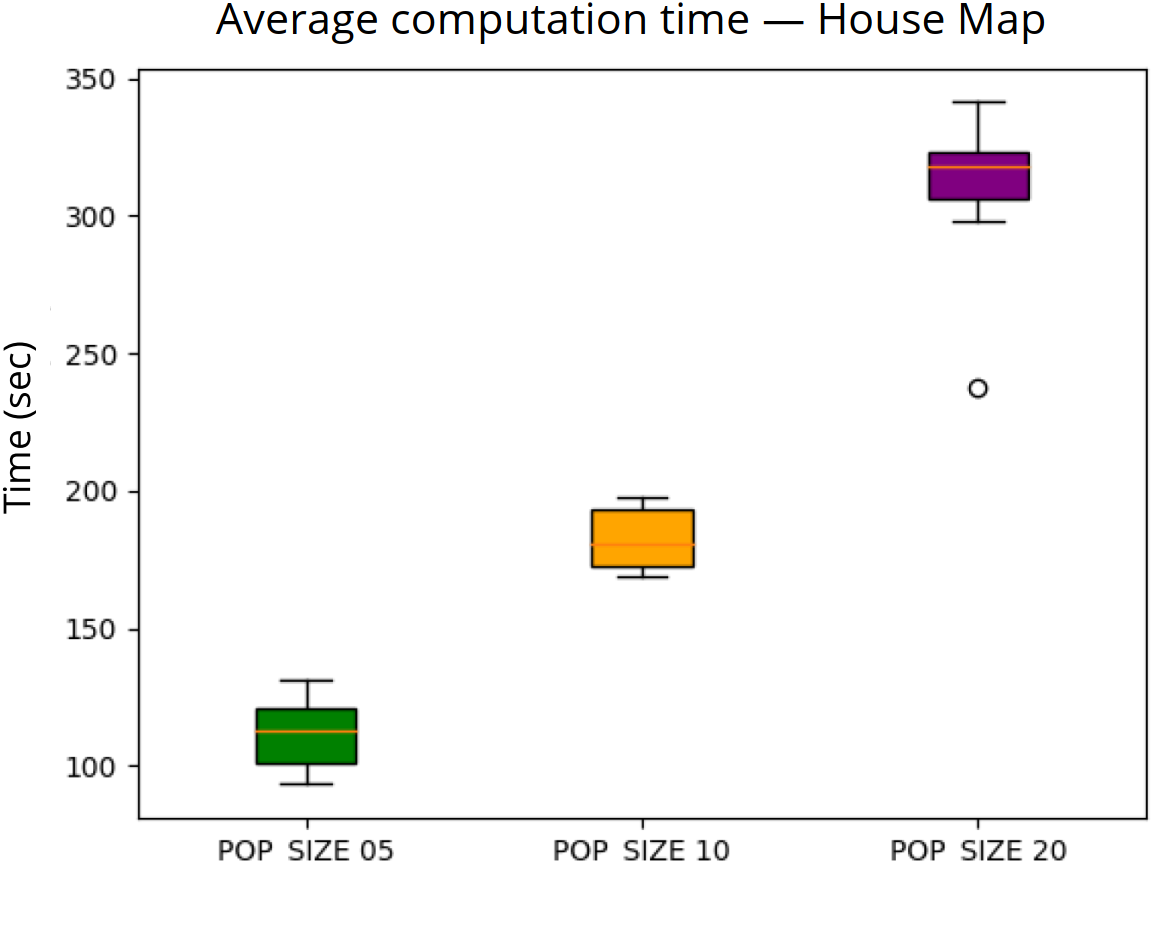}
     \end{subfigure}

     \vspace{1mm}%
     \textit{*Les résultats présentent les valeurs moyennes de 3 exécutions}
     
    \caption{Comparaison des résultats en utilisant l'algorithme BOA avec différentes tailles de population sur l'environnement \textit{House Map}}
    \label{fig:c4.7}
\end{figure*}

Afin d'étudier la possibilité d'accélérer les algorithmes BOA et xBOA en réduisant la taille de la population, nous avons effectué une série d'expériences afin de mesurer l’influence de la taille de la population sur ces algorithmes et la robustesse de ces derniers.

Dans un premier temps, nous avons effectué une expérience sur l'algorithme BOA original en variant la taille de la population de 20 à 5 papillons tout en gardant les autres paramètres fixes. 

Les résultats présentés sur la figure \ref{fig:c4.7} montrent que la réduction du nombre de solutions candidates (papillons) réduit considérablement le temps d'exécution de la métaheuristique. En effet, ce temps a diminué de 320 secondes à 110 secondes en divisant la taille de la population en quatre. Toutefois, ceci affecte aussi la qualité de l’exploration puisque le taux total de la surface explorée a baissé de 6.73\%, tel que nous pouvons le voir sur le tableau \ref{tab:4.3}.

Bien que cette baisse du taux d'exploration est importante, le gain apporté de pouvoir calculer le prochain point de destination en moins de 110 secondes apporte son lot d'avantages puisque ce temps d'attente est acceptable pour beaucoup de scénarios de robotique dans le monde réel par rapport au temps initial qui était trois fois plus long.

\begin{figure*}[pht!]
    \centering
    \setlength{\abovecaptionskip}{0.3cm} 

    \hspace{5mm}%
     \begin{subfigure}[b]{0.75\textwidth}
         \centering
         \includegraphics[width=\textwidth]{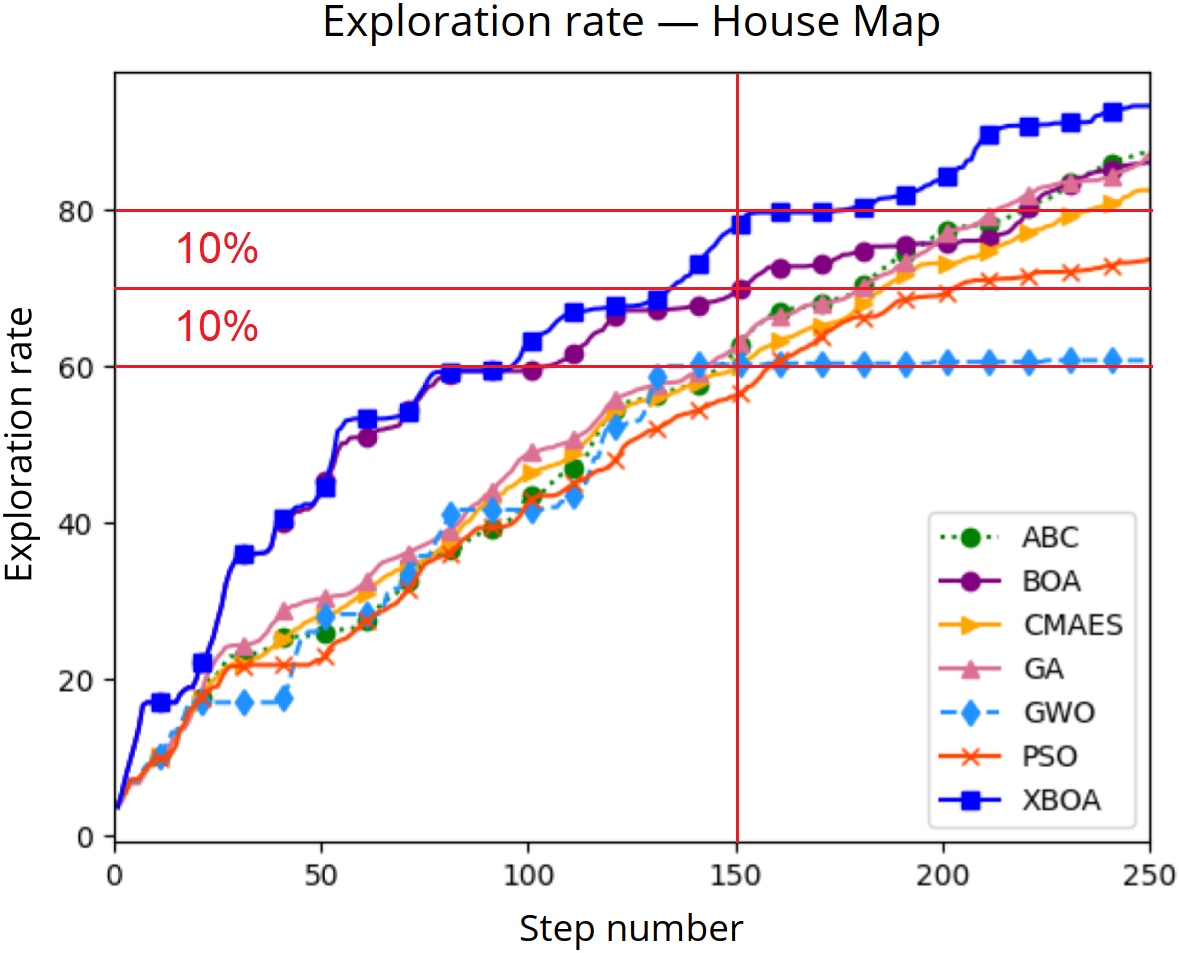}
         \vspace{2mm}%
     \end{subfigure}
    \begin{subfigure}[b]{0.75\textwidth}
         \centering
         \includegraphics[width=\textwidth]{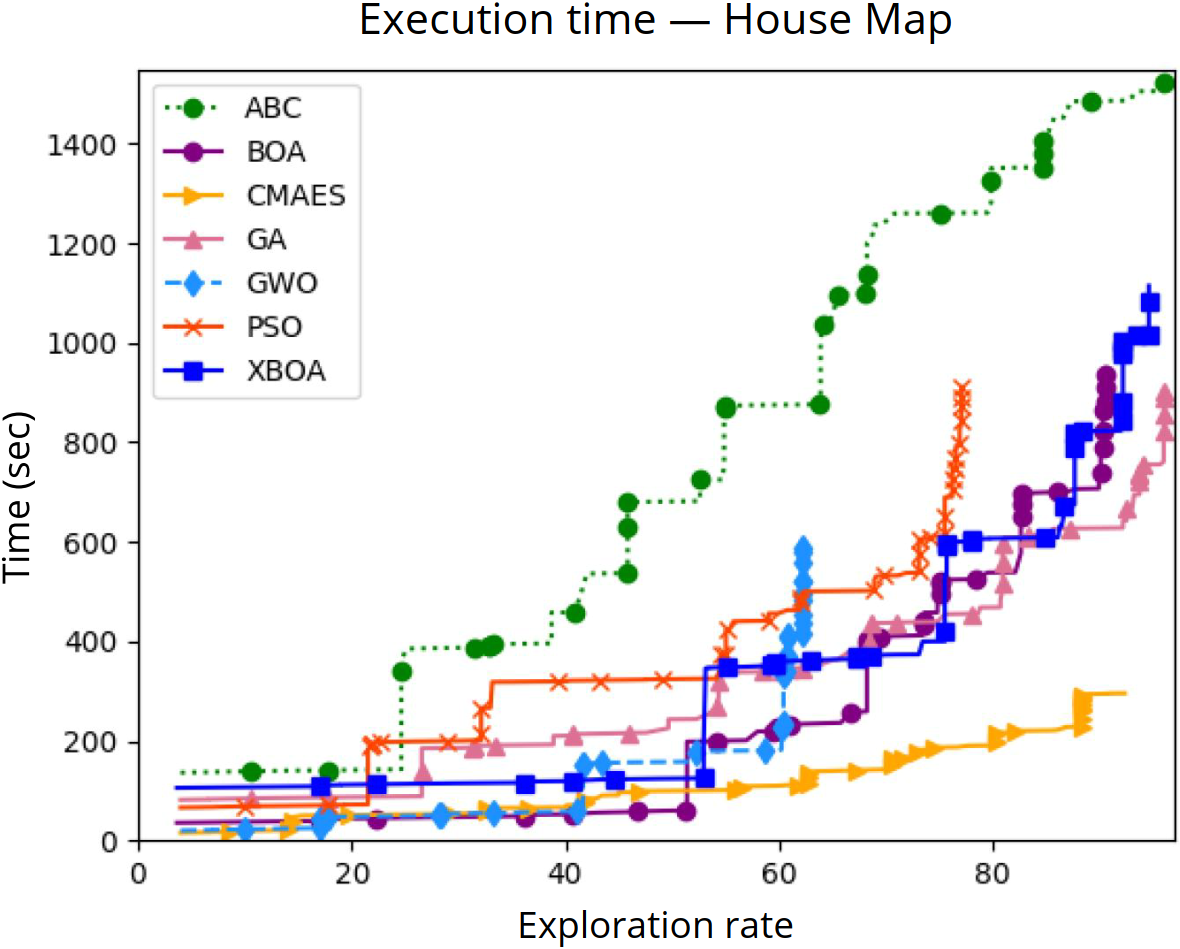}
     \end{subfigure}

     \vspace{1mm}%
     \textit{*Les résultats présentent les valeurs moyennes de 3 exécutions}
     
    \caption{
    Comparaison du taux d'exploration et durée de la mission en utilisant la stratégie à court-terme et une population de taille 5}
    \label{fig:c4.8}
\end{figure*}

\begin{figure*}[pht!]
    \centering
    \setlength{\abovecaptionskip}{0.3cm} 

     \begin{subfigure}[b]{0.75\textwidth}
         \centering
         \includegraphics[width=\textwidth]{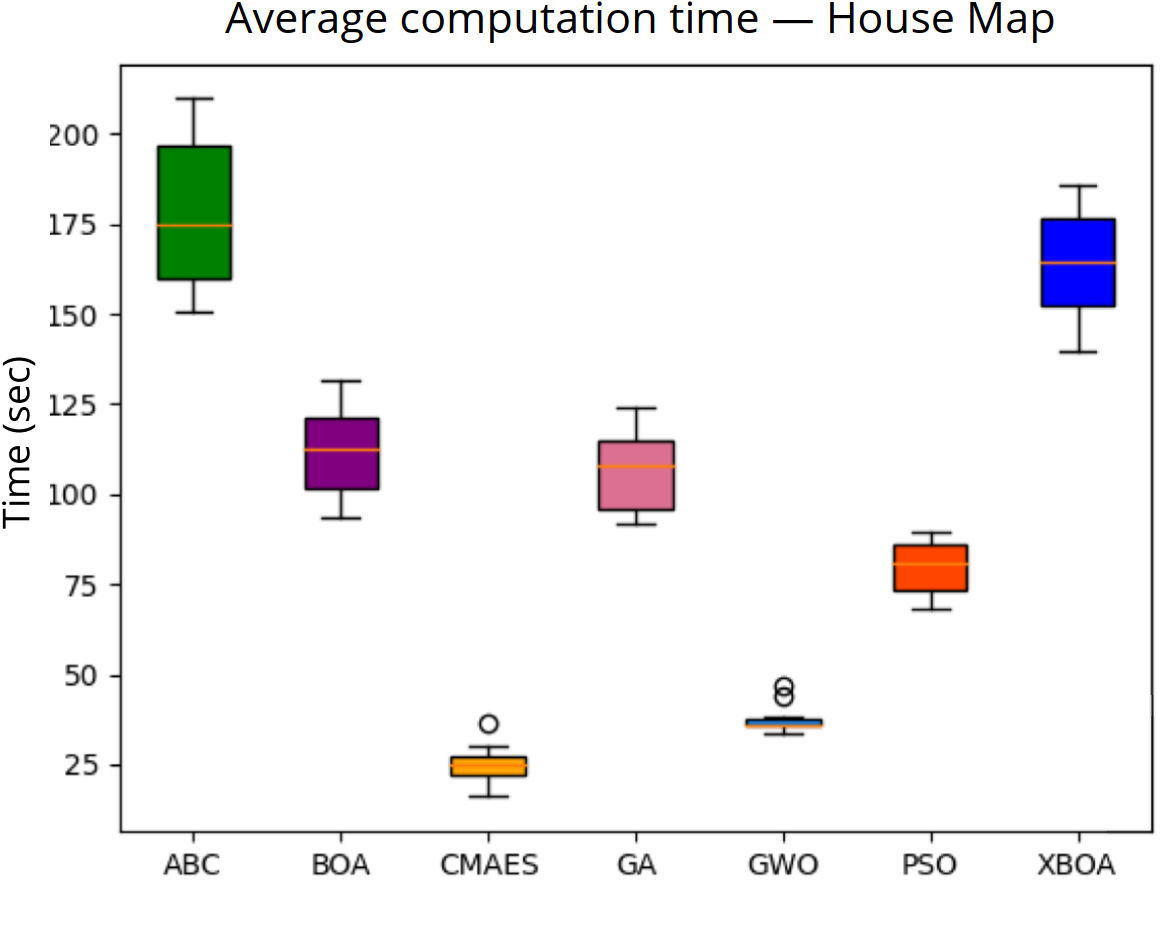}
         \vspace{2mm}%
     \end{subfigure}
    \begin{subfigure}[b]{0.75\textwidth}
         \centering
         \includegraphics[width=\textwidth]{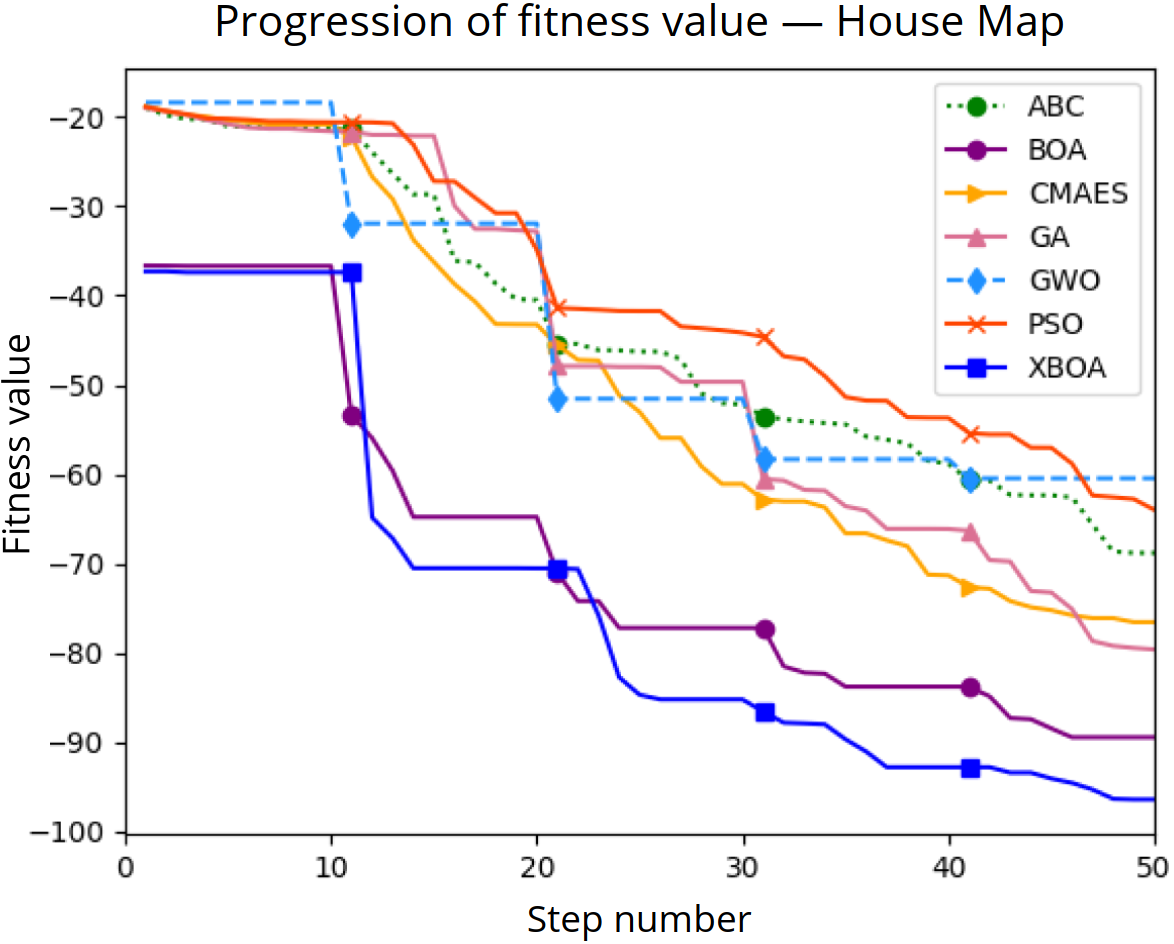}
     \end{subfigure}

     \vspace{1mm}%
     \textit{*Les résultats présentent les valeurs moyennes de 3 exécutions}
     
    \caption{Comparaison du temps moyen de calcul et convergence de la valeur de fitness en utilisant la stratégie à court-terme et une population de taille 5}
    \label{fig:c4.9}
\end{figure*}

En prenant ces résultats comme ligne de base, nous avons effectué d'autres expériences pour mesurer les performances de notre méthode ainsi que celles des autres métaheuristiques lorsque la taille de la population est réduite à 5 solutions. Les résultats sont présentés sur le tableau  \ref{tab:4.3} et les figures \ref{fig:c4.8} et \ref{fig:c4.9}.


La première remarque notable que nous pouvons constater en observant les graphes se trouve au niveau du taux d'exploration des méthodes PSO et GWO dont l'amélioration s'est arrêtée après avoir atteint les valeurs de 75\% et 62\% respectivement, malgré qu’elles démontraient des résultats proches de ceux des autres méthodes lorsque la taille de la population était plus grande.

La deuxième observation notable c'est que la surface de la zone visitée à la fin de la mission n'a pas pu dépasser le taux de 97\% pour l'environnement \textit{Empty Map} et 93.35\% pour l'environnement \textit{House Map}, et ceci, même si le niveau d'énergie du robot n'a pas encore atteint 0. Après la visualisation des trajectoires des robots dans le simulateur, nous avons remarqué que ceci est dû au blocage des robots dans un minima local où ils revisitent des régions déjà explorées. Ceci s'explique par le manque de diversification dans les solutions puisque le nombre très réduit de la population a engendré une convergence prématurée vers une solution unique. C'est-à-dire que tous les cinq individus de la population avaient (presque) la même valeur.

\begin{table*}[b!]
    \small
    \centering
    \caption{Taux d'exploration obtenus à la fin de la mission en utilisant l'algorithme BOA avec une population de taille 5}
    \label{tab:4.4}
    
    \begin{tabular}{|r|c|c|c|c|c|c|}
    \hline
    \rowcolor{headerColor}
      \multicolumn{7}{|c|}{\textit{Short-term exploration}} \\ \hline
    \multicolumn{1}{|l|}{} &
      \multicolumn{3}{c|}{\cellcolor[HTML]{C5E0B3}\textit{Empty map}} &
      \multicolumn{3}{|c|}{\cellcolor[HTML]{C5E0B3}\textit{House map}} \\ \hline
    \textit{\textbf{Method}} &
      \cellcolor[HTML]{FFCE93}\textit{Average} &
      \cellcolor[HTML]{FFCE93}\textit{Min} &
      \cellcolor[HTML]{FFCE93}\textit{Max} &
      \multicolumn{1}{|c|}{\cellcolor[HTML]{FFCE93}\textit{Average}} &
      \cellcolor[HTML]{FFCE93}\textit{Min} &
      \cellcolor[HTML]{FFCE93}\textit{Max} \\ \hline
      
        ABC &
        \textbf{97.32} & \textbf{95.83} & 98.61 & 
        \multicolumn{1}{|c|}{90.59} &
         84.72 & 96 \\ \hline
        BOA & 
        \multicolumn{1}{c|}{\cellcolor[HTML]{EFEFEF}96.09} &
        \multicolumn{1}{c|}{\cellcolor[HTML]{EFEFEF}92.01} &
        \multicolumn{1}{c|}{\cellcolor[HTML]{EFEFEF}$\geq$99} &
        \multicolumn{1}{|c|}{\cellcolor[HTML]{EFEFEF}87.13} &
        \multicolumn{1}{c|}{\cellcolor[HTML]{EFEFEF}83.68} &
        \multicolumn{1}{c|}{\cellcolor[HTML]{EFEFEF}95.13}
        \\ \hline
        CMAES &
        94.46 & 89.4 & 98.61 & 
        \multicolumn{1}{|c|}{82.8} &
         76.73 & 93.22 \\ \hline
        GA &
        \multicolumn{1}{c|}{\cellcolor[HTML]{EFEFEF}96.09} &
        \multicolumn{1}{c|}{\cellcolor[HTML]{EFEFEF}87.5} &
        \multicolumn{1}{c|}{\cellcolor[HTML]{EFEFEF}$\geq$99} &
        \multicolumn{1}{|c|}{\cellcolor[HTML]{EFEFEF}92.36} &
        \multicolumn{1}{c|}{\cellcolor[HTML]{EFEFEF}89.23} &
        \multicolumn{1}{c|}{\cellcolor[HTML]{EFEFEF}96}
        \\ \hline
        GWO &
        89.61 & 84.37 & 92.53 & 
        \multicolumn{1}{|c|}{62.03} &
         60.76 & 62.32 \\ \hline
        PSO &
        \multicolumn{1}{c|}{\cellcolor[HTML]{EFEFEF}93.03} &
        \multicolumn{1}{c|}{\cellcolor[HTML]{EFEFEF}84.37} &
        \multicolumn{1}{c|}{\cellcolor[HTML]{EFEFEF}98.26} &
        \multicolumn{1}{|c|}{\cellcolor[HTML]{EFEFEF}74.9} &
        \multicolumn{1}{c|}{\cellcolor[HTML]{EFEFEF}70.83} &
        \multicolumn{1}{c|}{\cellcolor[HTML]{EFEFEF}77.6}
        \\ \hline
        xBOA &
        96.14 & 92.88 & 98.26 & 
        \multicolumn{1}{|c|}{\textbf{93.35}} & \textbf{92.18} & \textbf{93.92} \\ \hline
    \end{tabular}
\end{table*}

\begin{figure*}[b!] 
    \vspace{1mm}%
    \centering
    \setlength{\abovecaptionskip}{0.3cm} 

     \begin{subfigure}[b]{0.46\textwidth}
         \centering
         \includegraphics[width=\textwidth]{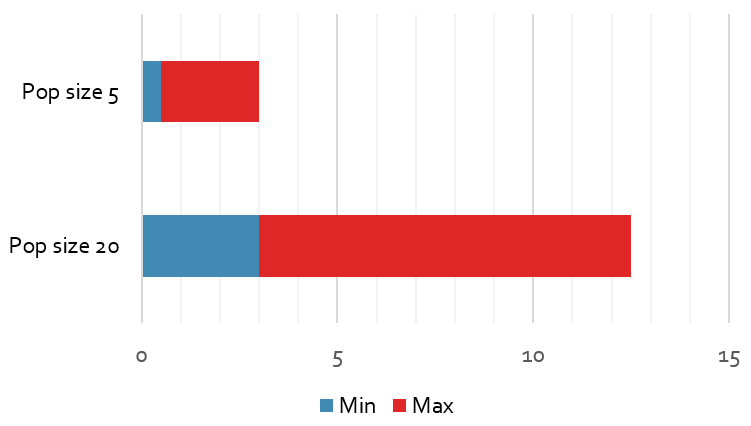}
         \caption{Temps de calcul moyen (en minutes)}
     \end{subfigure}
    \begin{subfigure}[b]{0.53\textwidth}
         \centering
         \includegraphics[width=\textwidth]{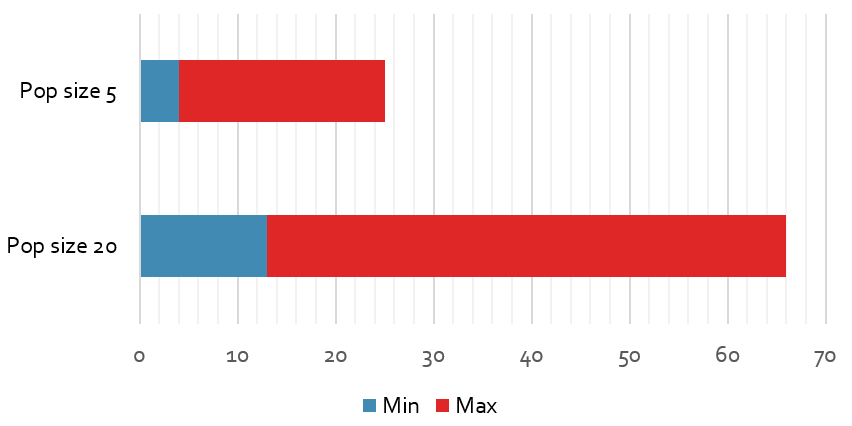}
         \caption{Temps total de la mission (en minutes)}
     \end{subfigure}

     \vspace{1mm}%
     
    \caption{Impact de la réduction de la taille de population sur le temps de calcul et temps de la mission en utilisant les différentes métaheuristiques}
    \label{fig:c4.16}
\end{figure*}

Une potentielle solution pour éviter ce problème serait d’augmenter l’espace de recherche graduellement en augmentant la taille de la population lorsque le taux de la surface explorée dépasse un certain seuil (par exemple 80\%). En d'autres termes: commencer la recherche en utilisant une taille de population réduite afin d'accélérer le temps de calcul, puis augmenter le nombre d'individus à la fin de la mission pour diversifier la population et échapper aux minimas locaux. Ceci devrait être une meilleure stratégie pour profiter des avantages des deux paramétrages, sans sacrifier la qualité du résultat obtenu à la fin de la mission.

Nous observons également que la méthode xBOA domine toutes les autres méthodes sur les critères du taux d’exploration et de la convergence de la valeur de fitness, mais elle souffre d'un temps d'exécution élevé. De l'autre côté, la méthode CMAES est dominante sur le critère du temps moyen de calcul qui avoisine les 25 secondes, mais elle souffre d'un taux d'exploration inférieur à xBOA de 10\%. La méthode ABC est la plus lente de toutes les méthodes, elle nécessite 175 secondes de temps de calcul pour arriver à un résultat presque similaire à celui de xBOA. Tandis que GA offre un bon compromis entre qualité et vitesse d'exécution. La figure \ref{fig:c4.16} montre la différence de ces résultats avec ceux obtenus lors de la série d'expériences précédente qui utilisait une taille de population de 20 individus.

Étant donné que la méthode xBOA a pu garder un taux d'exploration supérieur aux autres méthodes pendant toute la durée de la mission (voir la figure \ref{fig:c4.8}), et étant donné que les autres méthodes n'ont pu approcher le taux de xBOA que vers la fin de la mission, nous pouvons conclure que cette méthode est la plus robuste d'entre eux face à la réduction de la taille de la population et qu'elle serait plus avantageuse à utiliser dans beaucoup des situations. Par exemple, Si la batterie était limitée, xBOA permettrait d’explorer 10\% à 20\% plus de surface que les autres métaheuristiques en général.

Une solution alternative serait d'utiliser la méthode GA qui offre un résultats proches de xBOA vers la fin de la mission avec un temps d'exécution similaire à l'algorithme BOA classique.

Toutefois, dans le cas où le temps d'exécution est critique, nous pouvons envisager d'utiliser la méthode CMAES qui donne des résultats très rapides en sacrifiant la qualité des solutions. 

\begin{figure*}[ht!]
    \centering
    \setlength{\abovecaptionskip}{0.3cm} 

     \includegraphics[width=0.8\textwidth]{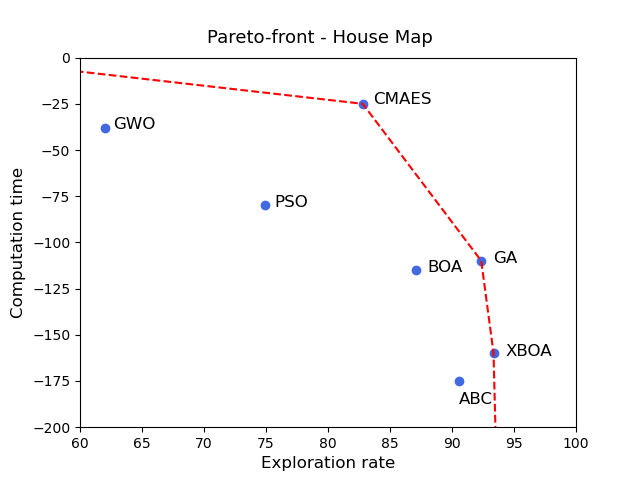}


     \vspace{1mm}%
     \textit{*Toutes les solutions sous la ligne rouge sont dominées}
     
    \caption{Visualisation du front des solutions dominantes (Pareto-front) pour les différentes métaheuristiques en utilisant une population de taille 5}
    \label{fig:c4.10}
\end{figure*}

Ces trois méthodes appartiennent au front des solutions non dominées (\textit{Pareto-front}) et sont donc simultanément optimales. Toutes les autres méthodes sont dominées et appartiennent à la liste des solutions non optimales. La figure \ref{fig:c4.10} illustre cette classification.

\section{Expérience 6: Comparaison de xBOA avec les autres variantes de l'algorithme BOA}

\begin{table}[t!]
    \small
    \centering
    \caption{Comparatif du taux d'exploration en utilisation les variantes de l'algorithme BOA avec une  population de taille 5}
    \label{tab:4.5}
    
    \begin{tabular}{|r|c|c|c|}
    \hline
    \multicolumn{4}{|c|}{\cellcolor[HTML]{C5E0B3}\textit{Short-term exploration - House map}} \\ \hline
    \multicolumn{1}{|c|}{\textbf{\textit{Method}}} & \multicolumn{1}{c|}{\cellcolor[HTML]{FFCE93}\textit{Average}} & \multicolumn{1}{c|}{\cellcolor[HTML]{FFCE93}\textit{Min}} & \multicolumn{1}{c|}{\cellcolor[HTML]{FFCE93}\textit{Max}} \\ \hline
    
    \multicolumn{1}{|r|}{\textit{BOA}} & \multicolumn{1}{c|}{\cellcolor[HTML]{EFEFEF}89.11} & \multicolumn{1}{c|}{\cellcolor[HTML]{EFEFEF}81.77} & \multicolumn{1}{c|}{\cellcolor[HTML]{EFEFEF}\textbf{95.13}}  \\ \hline    
    
    \multicolumn{1}{|r|}{\textit{ABOA}} & \multicolumn{1}{c|}{90.38} & \multicolumn{1}{c|}{81.77} & \multicolumn{1}{c|}{\textbf{95.13}} \\ \hline
    
    \multicolumn{1}{|r|}{\textit{mBOA}} &
    \multicolumn{1}{c|}{\cellcolor[HTML]{EFEFEF}89.14} &
    \multicolumn{1}{c|}{\cellcolor[HTML]{EFEFEF}81.07} &
    \multicolumn{1}{c|}{\cellcolor[HTML]{EFEFEF}\textbf{95.13}}  \\ \hline
    
    \multicolumn{1}{|r|}{\textit{SABOA}} &
    \multicolumn{1}{c|}{85.13} &
    \multicolumn{1}{c|}{82.98} &
    \multicolumn{1}{c|}{85.93}  \\ \hline
    
    \multicolumn{1}{|r|}{\textit{xBOA}} &
    \multicolumn{1}{c|}{\cellcolor[HTML]{EFEFEF}\textbf{93.35}} &
    \multicolumn{1}{c|}{\cellcolor[HTML]{EFEFEF}\textbf{92.18}} &
    \multicolumn{1}{c|}{\cellcolor[HTML]{EFEFEF}93.92}  \\ \hline
    
    \end{tabular}
    
\end{table}

Ayant constaté les avantages et les inconvénients de l'algorithme xBOA par rapport aux autres métaheuristiques, nous avons effectué une dernière série d'expériences dont le but est de comparer les performances de cet algorithme face aux autres variantes du BOA citées auparavant dans l'état de l'art (voir section \ref{section:BOA_variants}). Nous avons gardé la même taille de population que la série d'expériences précédente, à savoir 5 papillons.

\begin{figure*}[pht!]
    \vspace{1mm}%
    \centering
    \setlength{\abovecaptionskip}{0.3cm} 

     \begin{subfigure}[b]{0.75\textwidth}
         \centering
         \includegraphics[width=\textwidth]{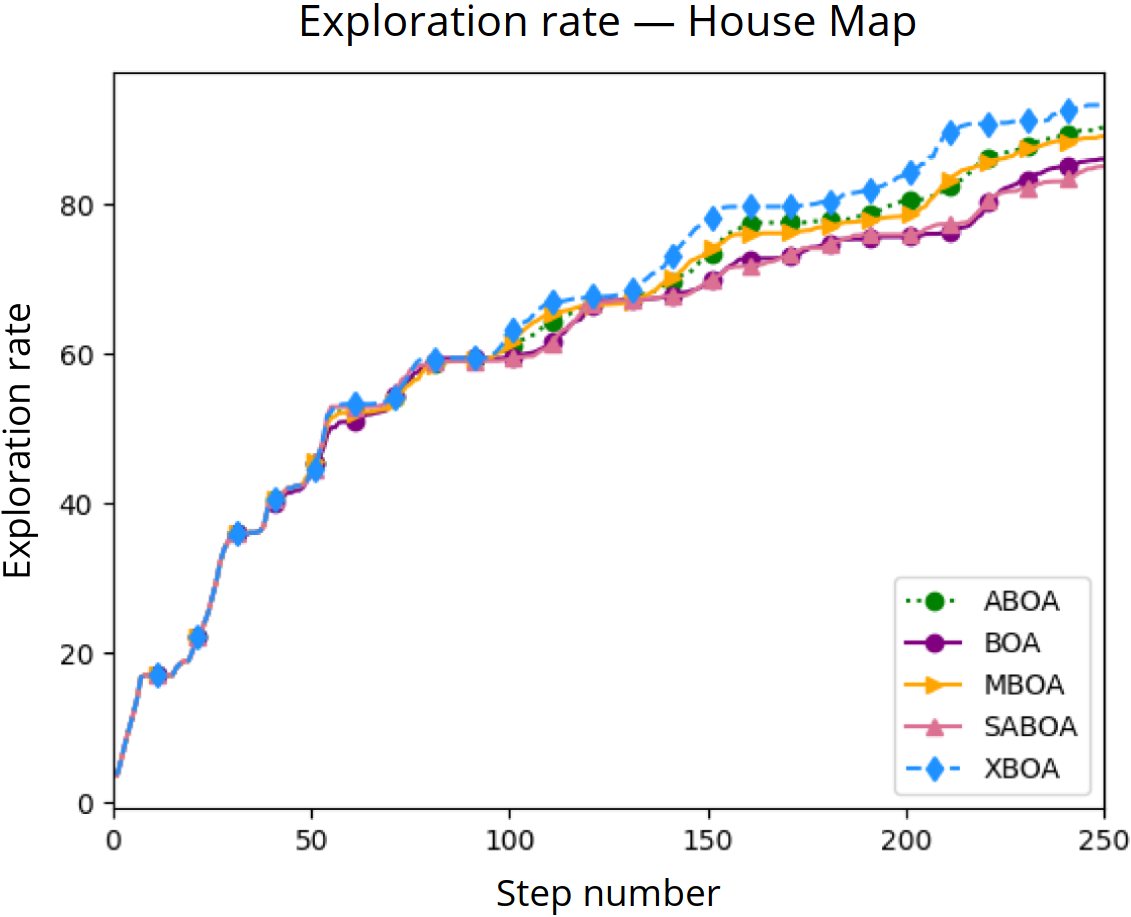}
         \vspace{2mm}%
     \end{subfigure}
    \begin{subfigure}[b]{0.75\textwidth}
         \centering
         \includegraphics[width=\textwidth]{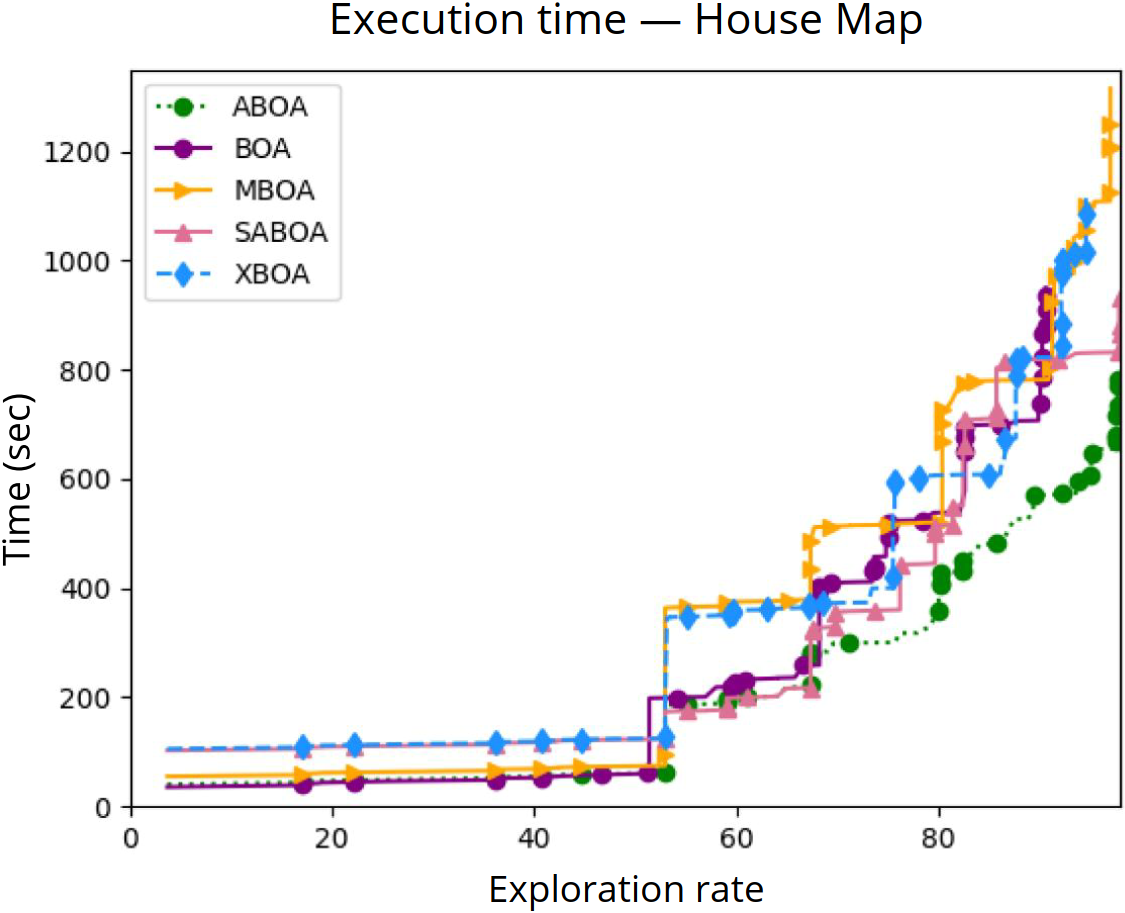}
     \end{subfigure}

     \vspace{1mm}%
     \textit{*Les résultats présentent les valeurs moyennes de 10 exécutions}
     
    \caption{Comparaison des variantes de l'algorithme BOA en utilisant la stratégie à court terme et une population de taille 5}
    \label{fig:c4.11}
\end{figure*}

\begin{figure*}[pht!]
    \vspace{1mm}%
    \centering
    \setlength{\abovecaptionskip}{0.3cm} 

     \begin{subfigure}[b]{0.75\textwidth}
         \centering
         \includegraphics[width=\textwidth]{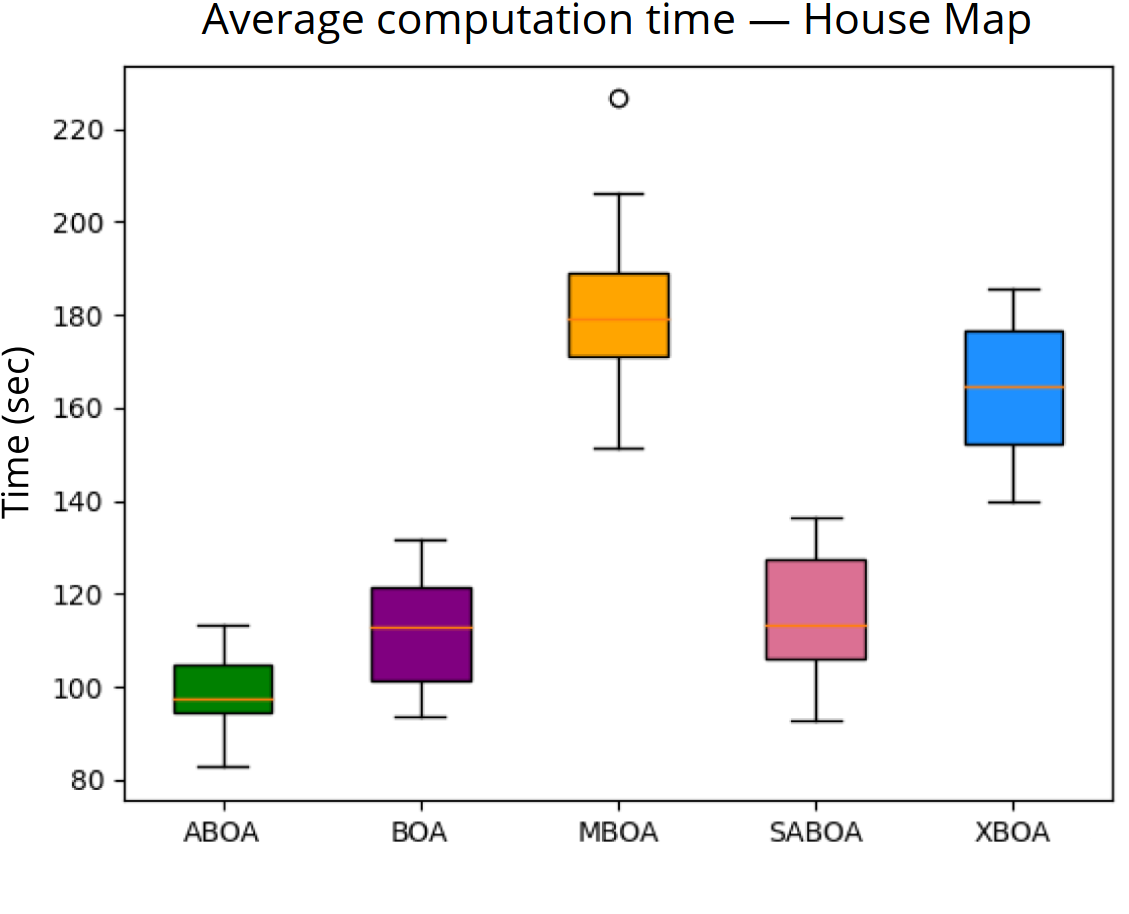}
         \vspace{2mm}%
     \end{subfigure}
    \begin{subfigure}[b]{0.75\textwidth}
         \centering
         \includegraphics[width=\textwidth]{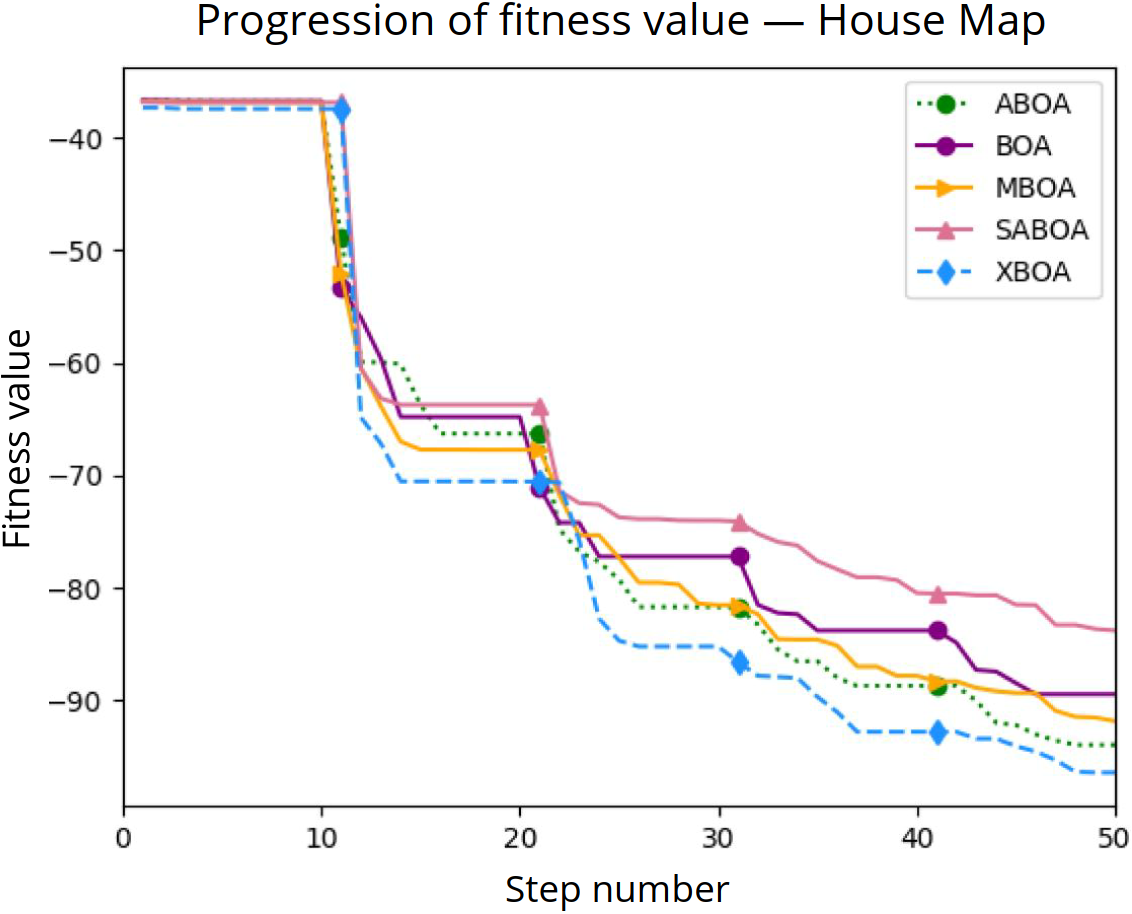}
     \end{subfigure}

     \vspace{1mm}%
     \textit{*Les résultats présentent les valeurs moyennes de 10 exécutions}
     
    \caption{Suite de la comparaison des variantes de l'algorithme BOA en utilisant la stratégie à court terme et une population de taille 5}
    \label{fig:c4.12}
\end{figure*}

Selon ce que l’on peut voir sur la figure \ref{fig:c4.11} et \ref{fig:c4.12} ainsi que le tableau \ref{tab:4.5}, la méthode xBOA domine les autres variantes dans le critère du taux d'exploration durant toute la durée de la mission. Elle atteint un résultat supérieur à celui de l'algorithme BOA original de 4.24\%. Elle les domine également dans le critère de la convergence de la valeur de fitness.

Cette amélioration est due à l'utilisation de l'opérateur de croisement qui augmente la diversité de la population. En effet, en remplaçant les solutions existantes par de nouvelles solutions, nous encourageons l'algorithme à explorer diverses zones dans l'espace de recherche au lieu d'intensifier la recherche au niveau local. Ceci permet à l’algorithme d’échapper au piège de converger prématurément vers un minima local, ce qui augmente les chances de trouver l'optimum global.

Toutefois, chaque nouvel individu créé doit être évalué, ce qui engendre l'augmentation du nombre d’évaluations de la fonction fitness. Par conséquent, le temps de calcul de l'algorithme augmente aussi. Étant donné que le paramètre de \textit{probabilité de croisement} était fixé à 0.583, il y a donc ~58\% plus d'individus à évaluer dans l'algorithme xBOA comparé au BOA classique.

L'algorithme mBOA souffre aussi d'un temps de calcul élevé, ceci est causé par les calculs supplémentaires ajoutés à cette variante dans la phase d'exploitation intensive. Bien que cette nouvelle phase permet d'accélérer la convergence de l'algorithme par rapport à la méthode BOA classique, elle reste tout de même dominée par l'algorithme xBOA que ce soit dans les critères du temps, de la convergence ou du taux d'exploration de la zone.

ABOA domine les autres variantes dans le critère du temps d'exécution, mais n'est pas parvenue à dépasser à un taux d’exploration de 90.38\%, alors que xBOA a pu atteindre un résultat supérieur. Quant à l'algorithme SABOA, il a donné le résultat le moins performant avec un taux d'exploration de 85.13\%, mais possède néanmoins l’avantage de n’avoir qu’un seul hyperparamètre en entrée, ce qui réduit la complexité du paramétrage.

Les deux solutions dominantes sont donc :
\begin{itemize}
    \item ABOA : qui possède le meilleur temps de calcul; et
    \item xBOA : qui obtient le meilleur taux d'exploration. 
\end{itemize}


Il faut donc considérer le choix d’un bon compris entre la maximisation de la surface explorée et le temps de calcul nécessaire. Ce choix dépendra de la nature de la mission : dans un contexte d'une mission de sauvetage par exemple, le facteur « temps » est très important, ce qui rend envisageable la décision de privilégier la réduction de la durée d'exécution au profit de la qualité d'exploration. En revanche, dans un contexte de déminage, de nettoyage ou de surveillance, il serait judicieux de privilégier la maximisation de la zone explorée au dépourvu du temps de calcul, jusqu’à une certaine mesure.

Un autre facteur a prendre en considération aussi est la stabilité des résultats. En effet, après avoir répété l'expérience 10 fois, le xBOA a enregistré un taux de stabilité élevé avec un résultat presque similaire à chaque exécution (\textasciitilde1 à 2\% d'écart). Alors que l'algorithme ABOA vacillait dans une marge de 15\% de différence entre les résultats.

\begin{figure*}[h!]
    \centering
    \setlength{\abovecaptionskip}{0.3cm} 

     \includegraphics[width=0.77\textwidth]{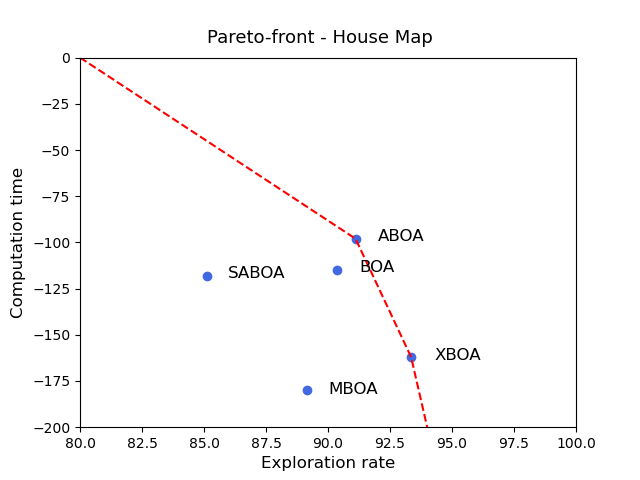}


     \vspace{1mm}%
     \textit{*Toutes les solutions sous la ligne rouge sont dominées}
     
    \caption{Visualisation du front des solutions dominante (Pareto-front) pour les variantes de BOA en utilisant une population de taille 5}
    \label{fig:c4.13}
\end{figure*}

\section{Expérience 7: Test en utilisant un robot réel}

Afin de valider l'adaptabilité de notre approche aux expériences du monde réel, nous avons effectué des tests sur le robot P3DX du laboratoire LARESI situé dans le département d'électronique de USTOMB. 
Pour cela, nous avons séparé le système en deux parties. D'un côté, un ordinateur portable connecté au réseau local du laboratoire via une connexion wifi joue le rôle de serveur. Il exécute le processus d'optimisation pour générer le prochain point de destination à visiter ainsi que l'algorithme de planification de chemin pour calculer les positions intermédiaires, qui sont ensuite transformées en un ensemble de commandes de mouvements puis envoyées au robot via une socket TCP. 

Une fois que le robot reçoit une commande de mouvement (move, turn, stop), il la transforme en une commande de vitesse pour le contrôle des moteurs en utilisant les routines du système ROS (\textit{Robot Operating System}), celui-ci communiquera directement avec le microcontrôleur du robot via l'interface ROSARIA pour exécuter les commandes. En parallèle à cela, un autre programme basé sur ROS s'occupe de collecter les informations mesurées par le Lidar à une fréquence de 10Hz puis les envoyer au serveur pour mettre à jour la carte de l'environnement.

Un problème rencontré durant les expériences était l'accumulation de l'erreur de localisation causée par le glissement du robot sur le sol. En effet, notre simulateur supposait que la localisation du robot était connue à chaque instant $t$, toutefois, dans le monde réel ceci ne pouvait se faire sans l'intégration d'un dispositif électronique de positionnement. La méthode que nous avons utilisée sur le robot P3DX se base sur l'odométrie, qui est une technique peu chère se basant sur le calcul du nombre de rotations des roues pour estimer la distance parcourue par le robot et son angle. Cependant, nous avons remarqué des mini-glissements du robot à la fin de chaque mouvement qui le fait dévier de 1 ou 2 centimètres de sa trajectoire. Bien que cette déviation n'est pas importante, son accumulation après plusieurs mouvements s'accroit et engendre une localisation erronée.

Afin de résoudre ce problème sans avoir besoin d'utiliser des dispositifs de localisation plus complexes, nous avons intégré un mécanisme de contrôle en boucle fermée (contrôlleur PID) pour réduire la vitesse de rotation des roues lorsque le robot se rapproche de l'emplacement prévu, avec une possibilité de revenir légèrement en arrière pour corriger sa position si nécessaire.

\begin{figure*}[b!] 
    \vspace{1mm}%
    \centering
    \setlength{\abovecaptionskip}{0.3cm} 

     \begin{subfigure}[b]{0.45\textwidth}
         \centering
         \includegraphics[width=\textwidth]{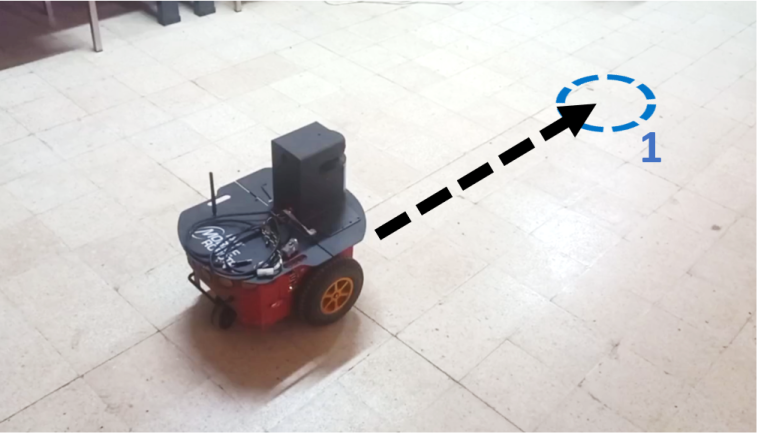}
     \end{subfigure}
    \begin{subfigure}[b]{0.47\textwidth}
         \centering
         \includegraphics[width=\textwidth]{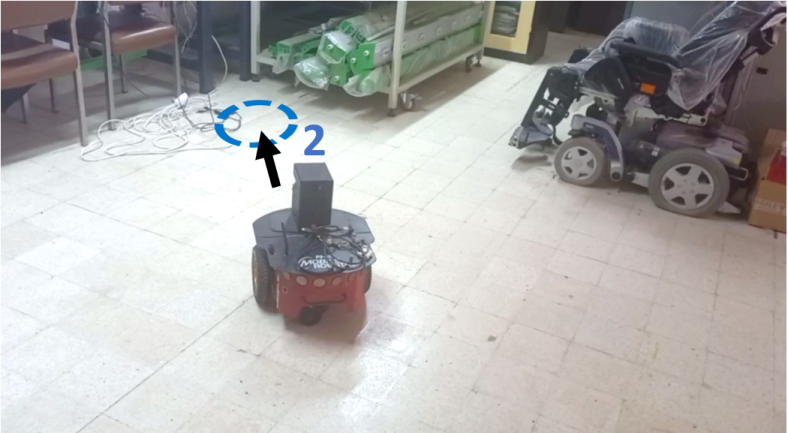}
     \end{subfigure}
    \begin{subfigure}[b]{0.45\textwidth}
         \centering
         \includegraphics[width=\textwidth]{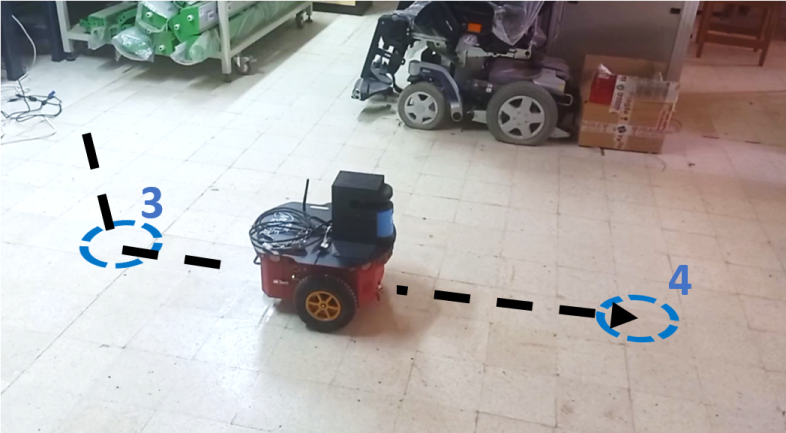}
     \end{subfigure}

     \vspace{1mm}%
     
    \caption{Expérience sur le robot réel}
    \label{fig:c4.17}
\end{figure*}

Les codes sources de ce contrôleur PID basé sur le système ROS ainsi que le programme pour le contrôle du robot en utilisant les commandes distantes par TCP sont disponible en open source sur le lien cité en bas de page \footnote{https://github.com/amineHorseman/pioneer\_bci}. La figure \ref{fig:c4.17} montre des photos prises durant les expériences dans le laboratoire.

Dans les travaux futurs, nous allons optimiser le système en intégrant directement les calculs des points de destination et de la planification de trajectoires dans l'ordinateur de bord du robot, et ceci afin qu'il puisse être autonome et continuer sa mission même si la connexion avec le serveur est interrompue. Étant donné les faibles capacités de calcul de cet ordinateur de bord, il serait judicieux de profiter des résultats obtenus durant l'expérience précédente par rapport à la réduction de la taille de population afin d'alléger les calculs. L'utilisation du xBOA dans ce contexte serait donc appropriée étant donné ses capacités de diversification des solutions dans les populations à petite taille, ainsi que sa stabilité à atteindre un taux d'exploration élevé lors de chaque exécution.



\section{Conclusion}

Nous avons présenté dans ce chapitre une série d'expériences effectuées dans le but d'évaluer les performances de l'algorithme xBOA. Nous l'avons comparé à plusieurs autres métaheuristiques, ainsi que les autres variantes de BOA récemment introduites dans la littérature.

L’évaluation des méthodes s’est basée sur 5 critères de comparaison. Les résultats montrèrent que l’algorithme xBOA surpasse BOA et ses autres variantes, mais requiert un temps de calcul élevé. L’algorithme proposé surpasse aussi les autres métaheuristiques telles que PSO, GA, GWO et ABC dans certains scénarios. 

Nous avons aussi effectué une expérience pour évaluer l’adaptabilité de notre approche dans un contexte multirobots. Les résultats nous ont permis de valider notre modélisation du mécanisme de synchronisation implicite et observer l'émergence d'un comportement collectif au sein des robots leur permettant de se disperser dans l'environnement et minimiser la redondance même pour des zones à large échelle. Nous avons également effectué une expérience sur un robot réel de type P3DX au sein du laboratoire LARESI basé sur le middleware ROS et une communication avec le serveur via une connexion wifi en utilisant des sockets TCP.


Les résultats obtenus durant ces expériences ont été satisfaisants sur certains critères, mais ont démontré aussi les limites de l'algorithme xBOA. Nous présenterons dans la conclusion générale quelques perspectives d'amélioration de cette méthode.
    
    \chapter*{Conclusion générale et perspectives}
\addcontentsline{toc}{chapter}{Conclusion générale et perspectives}
\lhead{Conclusion générale et perspectives}
\setstretch{1.5}
Nous avons présenté dans cette thèse nos travaux sur l’optimisation du comportement collectif d’un groupe de robots autonomes en utilisant les métaheuristiques.

Ce travail était divisé en deux parties principales. La première était consacrée à l’étude des fondements théoriques des systèmes multirobots et des métaheuristiques pour l’optimisation globale. Tandis que la deuxième présentait notre modélisation du problème d’exploration de zones inconnues avec des contraintes d’énergie, ainsi que la série d’expériences réalisées pour la valider.

Cette thèse a introduit deux contributions principales. D’un côté le développement d’une nouvelle plateforme de simulation appelée PyRoboticsLab pour l’évaluation et le benchmarking d’algorithmes dédiés à la résolution des problèmes de navigation robotique, planification de trajectoires, exploration et surveillance. Et de l’autre côté l’introduction d’un nouvel algorithme xBOA que nous avons proposé comme nouvelle variante de la métaheuristique BOA. 

Nous avons utilisé notre plateforme de benchmarking afin de comparer les performances de notre nouvel algorithme à plusieurs autres métaheuristiques utilisées dans la littérature. L’évaluation des méthodes s’est basée sur cinq critères de comparaison. Les résultats ont montré que l’algorithme xBOA surpasse BOA et ses autres variantes, mais requiert un temps de calcul plus long. L’algorithme proposé surpasse aussi les autres métaheuristiques telles que PSO, GA, GWO et ABC dans certains scénarios.

Nous avons procédé aussi à une série d’expériences pour évaluer l’adaptabilité de notre approche dans un contexte multirobots, et optimiser la coordination entre les robots en introduisant des mécanismes d’échange d’informations pour permettre aux robots de se disperser efficacement et minimiser la redondance.

D’autres améliorations d’ordre programmatique ont aussi été introduites telles que l’utilisation du parallélisme dans le processus d’évolution pour permettre d’évaluer plusieurs individus en même temps et accélérer le temps global de calcul des points de destinations.

Nous avons également effectué une expérience réelle sur un robot de type P3DX dont le but était de vérifier le bon fonctionnement des outils proposés. Cette expérience nous a permis de détecter et résoudre les problèmes de localisation basée sur l'odométrie et d’adapter notre approche au middleware ROS.

Dans les travaux futurs, nous nous consacrerons à améliorer la plateforme de simulation afin d’y intégrer plus d’algorithmes, et ce afin d’offrir aux chercheurs de ce domaine une suite complète de benchmarking et un outil uniformisé pour positionner leurs travaux. L’outil pourra aussi intégrer plus de scénarios et des environnements variés avec l’intégration d’obstacles dynamiques.

Le simulateur pourra aussi être utilisé à des fins pédagogiques afin de faire découvrir aux étudiants en informatique le domaine de l'intelligence artificielle appliquée aux problématiques de la robotique collective, sans qu'ils aient à se soucier des détails théoriques relatifs à la modélisation des capteurs laser, ou du calcul des probabilités d'occupation.

Du côté de l’algorithme xBOA, nous projetons de le tester sur d’autres types de problématiques d’optimisation combinatoire. Nous pourrons considérer la piste d’améliorer cet algorithme en introduisant un mécanisme plus adaptatif pour équilibrer les phases d’exploration et d’exploitation, ou de tester différents opérateurs de croisement pour analyser leur effet sur la convergence de l’algorithme.

\setstretch{1}


\pagebreak
\lhead{Bibliographie}
\printbibliography
\addcontentsline{toc}{chapter}{Bibliographie}

\end{document}